%% file: main.tex
\documentclass[10pt,letterpaper]{long-horizon-survey}

\usepackage{marvosym} 
\usepackage[numbers,sort&compress]{natbib}
\usepackage{booktabs}
\usepackage{multirow}
\usepackage{makecell}
\usepackage{array}
\usepackage{tabularx}
\usepackage{graphicx}
\usepackage{caption}
\usepackage{afterpage}
\usepackage{tikz}
\usetikzlibrary{arrows.meta}
\usepackage{longtable}
\usepackage{multirow}
\usetikzlibrary{calc,backgrounds}
\usepackage{adjustbox}
\input{figures/taxonomy-fonts}

\newcolumntype{Y}{>{\raggedright\arraybackslash}X}
\definecolor{ScopeBlue}{HTML}{3F51B5}
\definecolor{ScopeBlueLight}{HTML}{E8EAF6}
\definecolor{ContribAmber}{HTML}{FF8F00}
\definecolor{ContribAmberLight}{HTML}{FFF3E0}

\tcbset{
  surveyscope/.style={
    colback=ScopeBlueLight,
    colframe=ScopeBlue,
    colbacktitle=ScopeBlue!20!white,
    coltitle=black,
    boxrule=0.9pt,
    arc=2mm,
    left=3mm,
    right=3mm,
    top=2mm,
    bottom=2mm,
    fonttitle=\bfseries,
    title=Survey Scope
  },
  surveycontrib/.style={
    colback=ContribAmberLight,
    colframe=ContribAmber,
    colbacktitle=ContribAmber!20!white,
    coltitle=black,
    boxrule=0.9pt,
    arc=2mm,
    left=3mm,
    right=3mm,
    top=2mm,
    bottom=2mm,
    fonttitle=\bfseries,
    title=Contributions
  }
}

\runningtitle{Video Generation meets Agent Harness}

\title{%
  \textbf{Video Generation Meets Agent Harness}
  \\
  {
  \fontsize{12}{16}
  \selectfont\sffamily\bfseries\color{black}
  Agentic Video Generation, Agentic World Modeling and Recursive Self-Improvement}
}

\author{%
  \normalfont
  \sffamily
  Liaoyuan Fan$^{*}$ \quad
  Yuhao Zheng$^{*}$ \quad
  Dongxing Mao$^{*}$\quad
  Hangyu Ran$^{*}$ \quad
  Siyuan Hu$^{*}$ \quad
  Weijia Mao \quad
  Rui Zhao\\
  Weijia Wu \quad
  Guian Fang \quad
  Yanzhe Chen \quad
  Zeyu Zhu \quad
  Yuxin Jiang \quad
  Yiren Song \quad
  Yuchao Gu \quad
  Junchao Yi \quad
  Qi Hu \quad\\
  Dan Kong \quad
  Alex Jinpeng Wang \quad
  Kevin Qinghong Lin$^\text{\Letter}$ \quad
  Mike Zheng Shou$^\text{\Letter}$ \quad
  Wei Liu$^\text{\Letter}$
  \\[5pt]
  {\normalfont\fontsize{8.5}{11}\selectfont
  \sffamily
  Video Rebirth\quad
  National University of Singapore\quad
  Nanyang Technological University\quad
  Hong Kong University of Science and Technology\quad
  University of Science and Technology of China\quad
  Central South University\quad
  University of Electronic Science and Technology of China
  }
}

\fancypagestyle{surveycovernotes}{%
  \fancyhf{}
  \fancyfoot[L]{\normalfont\fontsize{7}{8}\selectfont\mbox{%
    $^{*}$Equal contribution. \quad
    $\text{\Letter}$ Corresponding author: \texttt{kevin.qh.lin@gmail.com}, \texttt{mike.zheng.shou@gmail.com}, \texttt{wl2223@columbia.edu}
    }}

}

\date{\vspace{-3ex}}

\begin{document}

\newgeometry{left=2.5cm,right=2.5cm,top=1.5cm,bottom=1.8cm,
  headheight=0pt,headsep=0pt,footskip=22pt}
\begingroup
\setlength{\parskip}{3pt}
\tcbset{titlebox/.append style={top=3mm,bottom=2mm}}
\makeatletter
\renewcommand{\maketitle}{%
  \begingroup\setlength{\parskip}{0pt}%
  \centering
  {\titlefont\@title\par}\vspace{5pt}%
  {\normalfont\rmfamily\fontsize{9.8}{12}\selectfont\@author\par}%
  \endgroup\vspace{5pt}\par
  \abscontent\vspace{2pt}\par
}
\makeatother

\begin{abstract}
\textbf{\large Abstract:}
Video generation is evolving from end-to-end synthesis toward agentic generation, in which planning, tool use, and feedback produce coherent multi-shot videos with greater control. Meanwhile, video world models provide action-conditioned observations and predicted futures and can serve as interactive environments for agents. We study this intersection through the \textit{agent harness}: the executable runtime that constructs an agent's context, exposes tools, maintains state, routes feedback, and enforces budgets around models such as video generators. The report is organized around three directions.
First, we study agentic video generation, where harnesses coordinate script and shot planning, asset management, shot production, and editing, and we discuss what the evaluation of such systems must separate. Second, we examine agentic world modeling, where harnesses mediate observation, control, execution, and state, including designs in which a game engine or a program written by a coding agent holds the world state while the video model renders it, and where agents and world models test and improve one another. Third, we examine recursive self-improvement (RSI) for video model development: we separate revisions of a current output from updates to harness state, generator parameters, and evaluators, and from changes to the procedure that produces later updates, and we identify the channels through which an improvement could re-enter development.
Across these directions, we analyze the agent capabilities that video imposes, from audiovisual perception, long-horizon planning, and memory to interaction through code and computer use, feedback-guided control, skill reuse, and multi-agent organization.
We conclude with open challenges, including autonomous video production, lifelong memory, programmable hybrid control, autonomous model development, provenance and cost accounting, and evaluation that links video quality and task performance to sustained improvement of models and harnesses.

\vspace{2mm}
\\
\textbf{Webpage}: \url{https://agentic-videogen.github.io/}\\
\textbf{GitHub}: \url{https://github.com/agentic-videogen/awesome-agentic-videogen}
\end{abstract}

\noindent\begin{minipage}{\linewidth}
  \setlength{\parskip}{0pt}
  \raggedright
  \includegraphics[height=8mm]{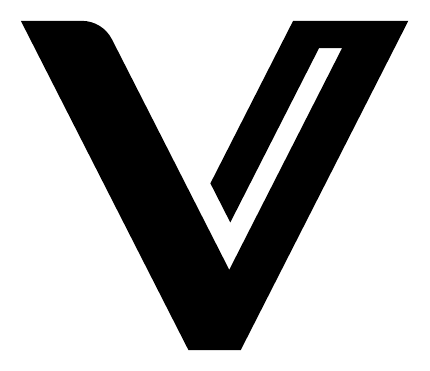}\par
  \vspace{2pt}
  {\color{black!18}\hrule height 0.4pt}
\end{minipage}\par

\begin{surveycover}
\maketitle
\end{surveycover}
\thispagestyle{surveycovernotes}

\vspace{2pt}
\noindent\begin{minipage}{\linewidth}
  \centering
  \setlength{\parskip}{0pt}
\includegraphics[width=\linewidth]{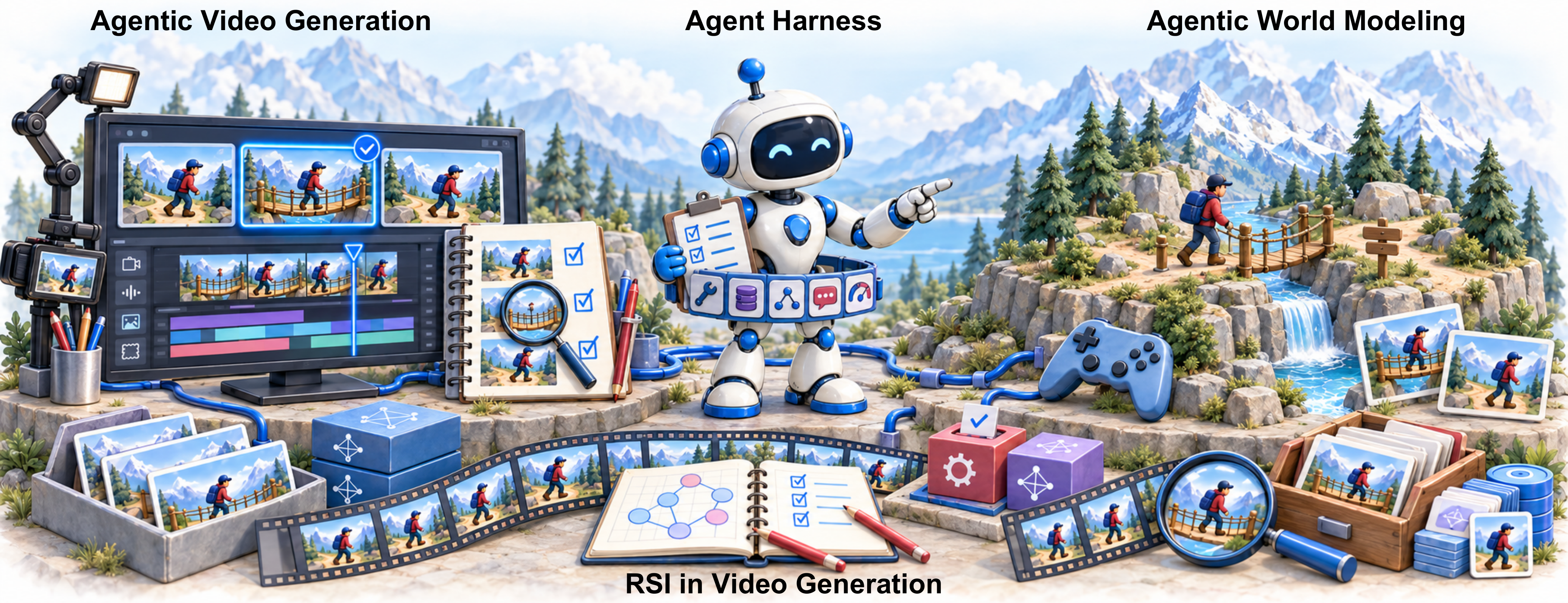}
  \captionsetup{type=figure,hypcap=false,skip=3pt,
    justification=raggedright,singlelinecheck=false}
  \captionof{figure}{\textbf{Video generation meets agent harness.}
    This report explores how agent harnesses revolutionize video generation across key areas: agentic video generation, agentic world modeling, and recursive self-improvement.}
  \label{fig:survey-teaser}
\end{minipage}
\par

\endgroup
\restoregeometry
\tableofcontents
\clearpage
\input{sections/01_introduction}

\clearpage
\pagestyle{fancy}

\input{sections/02_background}
\input{sections/04_production}
\begingroup
\setlength{\emergencystretch}{2em}
\widowpenalty=10000
\clubpenalty=10000
\captionsetup{justification=raggedright,singlelinecheck=false}
\input{sections/05_world_models}
\par
\endgroup
\input{sections/03_model_development}
\input{sections/06_agent_design}

\input{sections/07_future_challenges}

\clearpage
\bibliographystyle{unsrtnat}
\bibliography{main}

\end{document}

%% file: sections/01_introduction.tex
\section{Introduction}
\label{sec:introduction}

\subsection{Motivation}
\label{sec:introduction-background}

Video generators produce increasingly realistic, controllable, and temporally
coherent clips~\citep{kong2024hunyuanvideo}. A generator alone, however, does
not organize the decisions needed to complete a video task. A creative brief
must become coherent shots;
references must persist across scenes; takes must be inspected; edits must
preserve what is correct~\citep{huang2026vimax,vista2025}. Interactive
video must also support continuing actions grounded in observations, not
just a finished clip
\citep{gao2026lingbotworld2}.
These requirements motivate a transition from video generation
to agentic video generation: a goal-directed process in which generated results
inform subsequent decisions. An agent can revise a plan, retrieve a reference,
select a tool, or localize a failure in response to actual outcomes
\citep{li2026videogenagent}.
Agentic video generation denotes a system-level perspective, not a new generator
architecture: the video model provides synthesis or predicted transitions,
while the agent makes consequential choices around those capabilities.
Video couples these choices over time. References affect later shots
\citep{chen2026cinecrew},
actions change subsequent observations, and local repairs can disrupt
temporal consistency elsewhere. Plausible footage may still violate a
narrative requirement or fail to realize an intended action, so a quality
score alone rarely specifies what to do next
\citep{liu2024videogenerationevaluation,ding2026playworld}. Effective control
requires connecting observations to localized decisions while maintaining
project or interaction state and accounting for generation cost.

The development from large language models (LLMs) to agents offers a
foundation for this control. Rather than only generating a response, an
LLM can operate within a cycle of reasoning, tool use, and observation,
as demonstrated by ReAct~\citep{yao2023react}. Planning and memory extend
this cycle to tasks requiring multiple decisions~\citep{weng2023agents}.
Making these capabilities operational requires an agent harness:
the executable runtime around the model that constructs its context,
exposes tools, maintains state, routes feedback, and enforces permissions
and budgets~\citep{weng2026harness,ning2026code}
(Section~\ref{sec:background-harness}).
For video generation, the harness is where a creative brief or an
interactive goal becomes executable operations on generators, editors,
and world models. The central question shifts from generating a good
clip to reliably completing a video
task through feedback. Agentic video generation and video world-model interaction are two
concrete settings. A further question is whether experience from these tasks
can improve the system that will perform future tasks.

\begin{figure}[!htb]
  \centering
  \includegraphics[width=\linewidth]{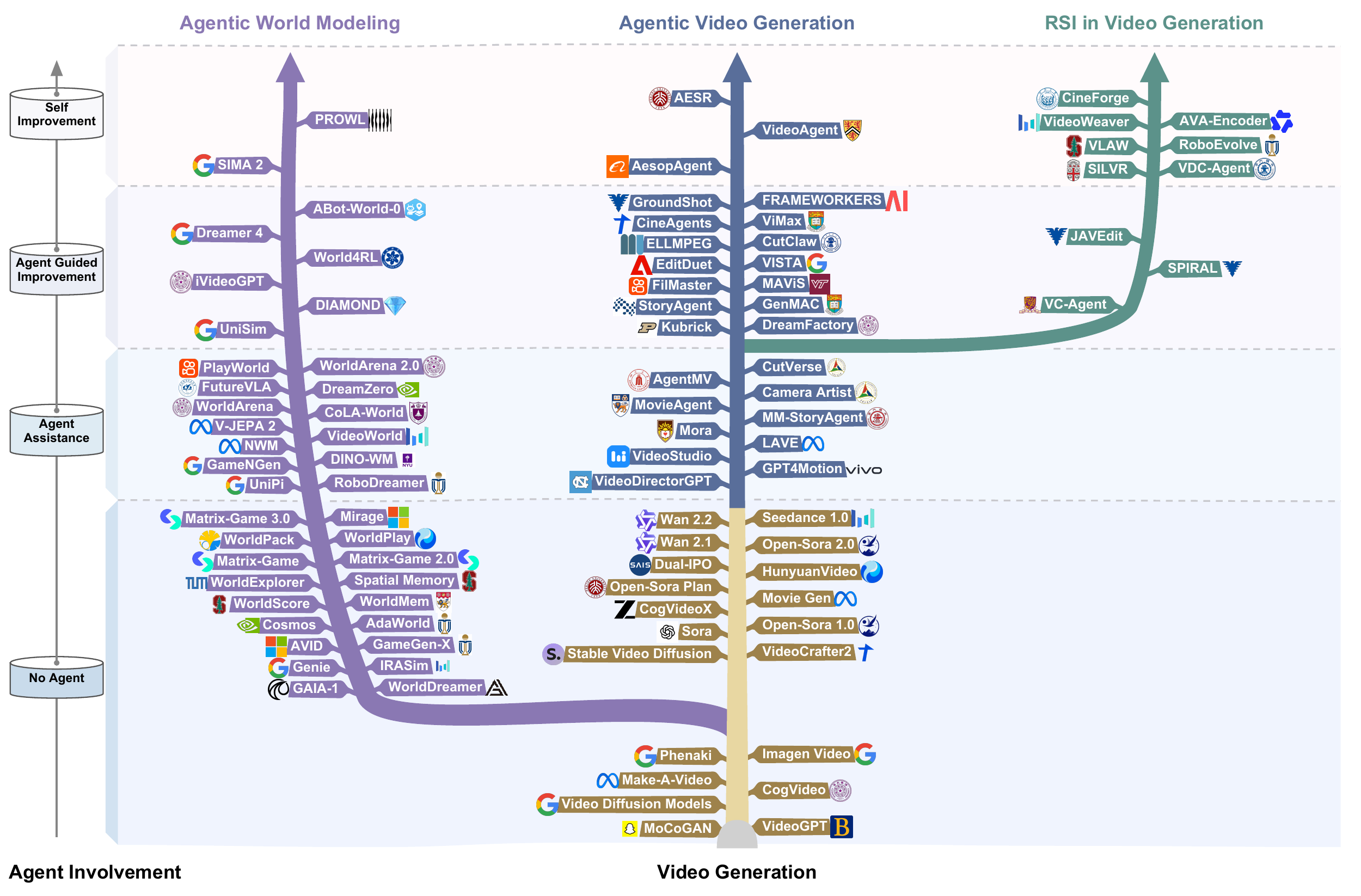}
  \captionsetup{skip=4pt,
    justification=raggedright,singlelinecheck=false}
  \caption{\textbf{Agent involvement across three directions of video generation.}
  Representative works branch from a shared video generation foundation into
  agentic world modeling, agentic video generation, and development toward
  recursive self-improvement (RSI). These directions connect video models to
  sustained world interaction, coordinated production, and the improvement of
  models and the agent harness. Vertical position reflects increasing agent
  involvement, from No Agent through Agent Assistance and Agent Guided
  Improvement to Self Improvement.}
  \label{fig:survey-evolution}
\end{figure}

Feedback changes different objects at different timescales. Prompt revision
or take replacement targets the current artifact~\citep{vista2025};
retaining a skill targets subsequent tasks~\citep{wei2026videoweaver};
updating a dataset, critic, or generator targets later model
versions~\citep{liu2025videoalign}. These outcomes
should not be conflated: effective repair of a current video does not by
itself establish improvement in future capability.
Beyond repairing individual outputs, context engineering considers
how instructions, retrieved evidence, memory, and execution records are
selected and organized for subsequent decisions~\citep{weng2026harness}.
Agentic Context Engineering (ACE), for example, accumulates and revises a
structured context from execution feedback rather than updating model
weights~\citep{zhang2025ace}. Such work shifts attention from a current
response toward reusable knowledge, while leaving open whether the
mechanisms that construct and update that knowledge can themselves improve.

Recent work on automated research extends these ideas to model
development. The AI Scientist links idea generation, coding, experiments,
and reporting~\citep{lu2024aiscientist}; \texttt{autoresearch} runs a bounded
loop of training code modification, experimentation, and validation-based
acceptance~\citep{karpathy2026autoresearch}. These workflows connect proposals
to empirical feedback, but automating a fixed experiment loop is not the
same as improving the loop itself. Research on self-improving agents,
exemplified by the Darwin G\"odel Machine and Regularized Recursive
Self-Improvement (RRSI), explores changes to the
agent's own code or harness~\citep{zhang2025dgm,xia2026rrsi}.
Such work motivates examining the processes that generate and validate
subsequent improvements.

Recursive self-improvement (RSI) asks whether a system can improve
the procedures that produce further improvements~\citep{chen2026surveyr04,duan2026surveyn21}.
Video-specific targets include failure diagnosis, data collection,
supervision construction, update
selection, and candidate validation. Repeating a fixed collection-and-training
pipeline is iterative improvement. The recursive direction additionally
allows evidence to revise that pipeline, changing how a later round discovers
and addresses weaknesses.
An agent harness provides an executable surface for this direction.
Its context construction, tool orchestration, memory, testing
workflow, and update-selection rules can be inspected and, where authorized,
revised \citep{weng2026harness}. This does not make every harness
self-improving: a revision must be retained and independently evaluated on
later tasks or improvement rounds under stated resource constraints
\citep{xia2026rrsi}.

Existing video research supplies parts of this connection: feedback can
revise reusable production skills~\citep{wei2026videoweaver}, uncertainty can
guide data acquisition for adaptation~\citep{savov2025autoexplore}, and
validation can govern critic renewal for later training
\citep{yang2026dualipo}. These are
building blocks rather than evidence that one system has completed the full
RSI loop. Our survey therefore connects three questions: how harnesses
organize agentic video generation, how they support video world-model interaction,
and how they may drive video-generation development toward RSI. The first
two establish task execution and sources of experience; the third examines
persistent capability and the procedures used to improve it.
Figure~\ref{fig:survey-evolution} maps representative work from a shared video
generation foundation into the three directions corresponding to the questions
below.

\subsection{Contributions and Positioning}
\label{sec:introduction-questions}
The survey addresses three questions about the influence of the agent harness
on video generation (Figure~\ref{fig:survey-teaser}), each corresponding to
one core chapter.

\textbf{How does the agent harness enable agentic video generation?}
Agentic Video Generation (Section~\ref{sec:production}) studies
how harnesses coordinate planning, execution, tool use, memory, editing, and
watermarking around shared project state, and what the evaluation of such
systems must separate. We examine how feedback changes
generation and editing decisions and how those changes affect video quality, consistency,
and efficiency.

\textbf{How can the agent harness be integrated into video
world models?}
Agentic World Modeling
(Section~\ref{sec:world-models}) locates where a harness can intervene in
observation, control, execution, event organization, and state management,
and examines designs in which an engine holds the world state
while the video model renders it.
We examine how these interventions support sustained interaction, control
fidelity, state continuity, and agent-driven evaluation. The focus of this
question is how a harness enables reliable interaction with a video world
model.

\textbf{How can video generation systems move toward recursive self-improvement?}
Toward RSI in Video Generation
(Section~\ref{sec:model-development}) examines how advances in video
generators and the agent harness can feed back into the process that produces
subsequent model improvements. We review foundation model improvement,
agent harness improvement, and evaluation, and identify the channels through
which gains in these components can strengthen failure diagnosis, data
curation, training, and validation in later development rounds. The goal is to identify mechanisms
and evidence needed for recursively improving development processes that
can support next-generation video generation models.

These questions connect through feedback: agentic video generation and world interaction
can expose failures, while development can turn selected evidence into
changes deployed in later tasks. Section~\ref{sec:agent-design} compares the shared
agent capabilities that support all three settings.

\textbf{Differentiating Our Work from Existing Surveys}

Existing surveys provide complementary accounts of video generation.
Work on spatiotemporal consistency examines generation models,
representations, training strategies, and evaluation, emphasizing the
mechanisms that maintain coherent video sequences~\citep{yin2025spatiotemporal}.
Post-training and alignment surveys organize methods for adapting video
generators to human intent, including supervised fine-tuning, distillation,
preference optimization, and inference-time adaptation~\citep{li2026videoposttraining}.
Evaluation surveys examine perceptual quality, semantic fidelity, and
alignment with user instructions~\citep{liu2024videogenerationevaluation}.
Together, these accounts explain how video models are built, improved,
and assessed.

More directly related to our system perspective, agentic visual generation
surveys organize the conditioning, execution, and adaptation decisions
available to a controller across visual modalities~\citep{huang2026agenticvisual},
or study autonomous multimodal content creation and the construction of
stateful artifacts~\citep{xu2026ags,wang2026artifactcreation}. Wu et al.\
place agentic generation and world-modeling generation on a single
capability ladder of visual generation~\citep{wu2026visualnewera}.
World model surveys characterize predictive and simulation capabilities
across domains, focus on action conditioning, memory, and responsiveness
in sustained interaction~\citep{chu2026agenticworldmodel,liu2026interactiveworldmodel},
or treat world models as interactive proxies that supply feedback to
continually improving agents~\citep{yang2026quovadis}.
Harness surveys analyze the execution layer around text and coding
agents~\citep{ning2026code,guo2026harnessdesign}.
Surveys of self-evolving agents and RSI examine evolutionary targets,
adaptation stages, and feedback loops across general AI
systems~\citep{gao2025surveyr02,chen2026surveyr04, wu2026rsisurvey}.
These perspectives share concepts such as feedback, memory, and agent
control, but none of them takes the harness around a video generator as the
unit of analysis across production, interaction, and model development.

Table~\ref{tab:survey-comparison-overview} compares related surveys and
adjacent studies by scope, thematic coverage (including RSI), and agent design.
Our survey complements these perspectives with a framework centered on
video generation that covers three settings in which a harness meets a video
model: as a synthesizer in production, as an interactive environment, and as
the object of development. The settings are linked through the harness:
agents organize generation and refinement, while video models provide
interactive experience that can inform agent decisions and further model
development. Where capability ladders treat agentic
generation and world modeling as successive stages of the
generator~\citep{wu2026visualnewera}, we take the agent harness around the
generator as the unit of analysis, so that one framework covers production,
interaction, and development.

Our contributions are threefold.
\begin{itemize}[leftmargin=*]
  \item A harness-centered framework. We define a common agent--harness
        execution loop (Section~\ref{sec:background}) and apply it to
        production, organized by workflow function
        (Section~\ref{sec:production}), and to world-model interaction,
        organized by the interfaces through which a harness intervenes,
        including designs in which an engine or a program holds the world
        state (Section~\ref{sec:world-models}).
  \item An account of improvement. We separate revisions of the current
        output from updates to harness state, generator parameters, and
        evaluators, and from changes to the development procedure. We identify four channels
        through which an improvement could re-enter video development.
        No surveyed system yet revises the procedure that trains and
        validates the generator (Section~\ref{sec:model-development}).
  \item Evaluation requirements and open problems. We state how the agent harness
        supports active testing and reciprocal evaluation of agents and
        world models (Section~\ref{sec:iwm-evaluation}), what the
        evaluation of agentic production must separate
        (Section~\ref{sec:rsi-evaluation}), 
        which capabilities video imposes on agents
        (Section~\ref{sec:agent-design}), and which challenges remain
        (Section~\ref{sec:challenges}).
\end{itemize}

\input{tables/survey_comparison_overview}


%% file: tables/survey_comparison_overview.tex
\begin{table}[!htbp]
\centering
\begingroup
\color{black}
\hypersetup{allcolors=black,citecolor=SurveyLinkBlue}
\SurveyTableSetup
\definecolor{scOverviewFamily}{HTML}{F0F5FB}
\definecolor{scOverviewOurs}{HTML}{C9DDF2}
\definecolor{scOverviewDivider}{HTML}{777777}
\definecolor{scOverviewGreen}{HTML}{00845A}
\definecolor{scOverviewRed}{HTML}{B92338}
\definecolor{scOverviewPartialGray}{HTML}{929292}
\DeclareRobustCommand{\scOverviewMark}[2]{\makebox[9pt][c]{%
  \tikz[x=0.7pt,y=0.7pt,baseline=-1.9pt]{%
    \path[use as bounding box] (-5,-4.8) rectangle (5,4.8);
    \begin{scope}[draw=#1,line width=1.15pt,line cap=round,line join=round]
      #2
    \end{scope}}}}
\DeclareRobustCommand{\scOverviewYes}{\scOverviewMark{scOverviewGreen}{%
  \draw (-3.3,-0.3) .. controls (-2.9,-1.2) and (-2.2,-2.5) .. (-1.7,-3.0)
    .. controls (-0.1,0.0) and (1.9,2.4) .. (3.5,3.8);}}
\DeclareRobustCommand{\scOverviewNo}{\scOverviewMark{scOverviewRed}{%
  \draw (-2.1,3.4) .. controls (-1.2,1.8) and (0.4,-1.5) .. (2.2,-3.3)
    (3.1,4.2) .. controls (1.2,2.0) and (-1.6,-1.2) .. (-3,-4);}}
\DeclareRobustCommand{\scOverviewPartial}{\scOverviewMark{scOverviewPartialGray}{%
  \draw (-3.3,-0.3) .. controls (-2.9,-1.2) and (-2.2,-2.5) .. (-1.7,-3.0)
    .. controls (-0.1,0.0) and (1.9,2.4) .. (3.5,3.8);}}
\newcommand{\scOverviewHeader}[1]{\makebox[\linewidth][c]{%
  \fontsize{7.0}{7.7}\selectfont\bfseries\makecell[c]{#1}}}
\captionsetup{position=top,labelfont={bf,color=black},
  textfont={rm,color=black},labelsep=colon,justification=raggedright,
  singlelinecheck=false,skip=7pt}
\caption[Comparison with related surveys and studies]{%
{\bfseries Comparison with related surveys and studies.}
Studies are grouped by primary research focus and compared in terms of
scope, thematic coverage, and agent design.
Our survey connects agentic video generation, agent--world interaction,
and video model development toward Recursive Self-Improvement (RSI)
through a unified agent-harness framework.
Green checks (\scOverviewYes), gray checks (\scOverviewPartial), and red crosses
(\scOverviewNo) indicate systematic coverage, partial coverage, and topics
outside a study's scope, respectively.}
\label{tab:survey-comparison-overview}
\fontsize{7.5}{8.35}\selectfont
\setlength{\tabcolsep}{1.7pt}
\setlength{\parskip}{0pt}
\renewcommand{\arraystretch}{1.00}
\arrayrulecolor{black}
\pgfmathsetlengthmacro{\scStudyWidth}{0.40*\linewidth+2*\tabcolsep}
\pgfmathsetlengthmacro{\scYearWidth}{0.055*\linewidth+2*\tabcolsep}
\pgfmathsetlengthmacro{\scRsiWidth}{0.035*\linewidth+2*\tabcolsep}
\pgfmathsetlengthmacro{\scStateWidth}{(\linewidth-\scStudyWidth-\scYearWidth-\scRsiWidth)/6}
\newcommand{\scOverviewHeaderPanel}{%
  \begin{tikzpicture}[x=1pt,y=1pt,baseline=0pt,
    every node/.style={inner sep=0pt,outer sep=0pt,align=center,
      font=\rmfamily\bfseries\fontsize{7.5}{8.35}\selectfont}]
    \path[use as bounding box] (0,0) rectangle (\linewidth,33pt);
    \draw[black,line width=0.8pt] (0,33pt)--(\linewidth,33pt);
    \node[anchor=west] at (\tabcolsep,16.5pt) {Survey / Study};
    \node at ({\scStudyWidth+0.5*\scYearWidth},16.5pt) {Year};
    \node at ({\scStudyWidth+\scYearWidth+\scStateWidth},27.1pt) {Scope};
    \node at ({\scStudyWidth+\scYearWidth+3.5*\scStateWidth+0.5*\scRsiWidth},27.1pt) {Thematic Coverage};
    \draw[scOverviewDivider,line width=0.3pt]
      ({\scStudyWidth+\scYearWidth+4pt},21.8pt)--
      ({\scStudyWidth+\scYearWidth+2*\scStateWidth-4pt},21.8pt);
    \draw[scOverviewDivider,line width=0.3pt]
      ({\scStudyWidth+\scYearWidth+2*\scStateWidth+4pt},21.8pt)--
      ({\scStudyWidth+\scYearWidth+5*\scStateWidth+\scRsiWidth-4pt},21.8pt);
    \begin{scope}[every node/.append style={font=\rmfamily\bfseries\fontsize{7}{7.7}\selectfont}]
      \node at ({\scStudyWidth+\scYearWidth+0.5*\scStateWidth},10.6pt) {Video\\Generation};
      \node at ({\scStudyWidth+\scYearWidth+1.5*\scStateWidth},10.6pt) {World\\Models};
      \node at ({\scStudyWidth+\scYearWidth+2.5*\scStateWidth},10.6pt) {Model\\Development};
      \node at ({\scStudyWidth+\scYearWidth+5*\scStateWidth+0.5*\scRsiWidth},10.6pt) {RSI};
      \node at ({\scStudyWidth+\scYearWidth+3.5*\scStateWidth},10.6pt) {Agentic\\Video Gen.};
      \node at ({\scStudyWidth+\scYearWidth+4.5*\scStateWidth},10.6pt) {Agent--World\\Interaction};
      \node at ({\scStudyWidth+\scYearWidth+5.5*\scStateWidth+\scRsiWidth},16.5pt) {Agent\\Design};
    \end{scope}
  \end{tikzpicture}}
\begin{tabularx}{\linewidth}{>{\raggedright\arraybackslash}p{0.40\linewidth}>{\centering\arraybackslash}p{0.055\linewidth}*{5}{>{\centering\arraybackslash}X}>{\centering\arraybackslash}p{0.035\linewidth}>{\centering\arraybackslash}X}
\multicolumn{9}{@{}c@{}}{\scOverviewHeaderPanel}\\[-2.5pt]
\midrule
\rowcolor{scOverviewFamily}[0pt][0pt]
\multicolumn{9}{@{}c@{}}{\rule[-2.7pt]{0pt}{10.6pt}{\bfseries\itshape Visual Generative Modeling}} \\
Adversarial Text-to-Image Synthesis~\citep{frolov2021surveyn01} & 2021 & \scOverviewNo & \scOverviewNo & \scOverviewYes & \scOverviewNo & \scOverviewNo & \scOverviewNo & \scOverviewNo \\
Diffusion Models: Methods and Applications~\citep{yang2022surveyn08} & 2022 & \scOverviewYes & \scOverviewNo & \scOverviewYes & \scOverviewPartial & \scOverviewNo & \scOverviewNo & \scOverviewPartial \\
RenAIssance: Text-to-Image Generation~\citep{bie2023surveyn09} & 2023 & \scOverviewPartial & \scOverviewNo & \scOverviewYes & \scOverviewNo & \scOverviewNo & \scOverviewNo & \scOverviewNo \\
Text-to-Image Diffusion Models~\citep{zhang2023surveyn02} & 2023 & \scOverviewYes & \scOverviewNo & \scOverviewYes & \scOverviewNo & \scOverviewNo & \scOverviewNo & \scOverviewNo \\
From Sora: Text-to-Video Generation~\citep{sun2024sorasurvey} & 2024 & \scOverviewYes & \scOverviewPartial & \scOverviewYes & \scOverviewYes & \scOverviewPartial & \scOverviewNo & \scOverviewPartial \\
Autoregressive Models in Vision~\citep{xiong2024surveyn06} & 2024 & \scOverviewYes & \scOverviewYes & \scOverviewYes & \scOverviewNo & \scOverviewYes & \scOverviewNo & \scOverviewPartial \\
Video Diffusion Models~\citep{melnik2024videodiffusionsurvey} & 2024 & \scOverviewYes & \scOverviewYes & \scOverviewYes & \scOverviewNo & \scOverviewPartial & \scOverviewNo & \scOverviewPartial \\
\rowcolor{scOverviewFamily}[0pt][0pt]
\multicolumn{9}{@{}c@{}}{\rule[-2.7pt]{0pt}{10.6pt}{\bfseries\itshape Visual Editing and Control}} \\
Diffusion-Based Image Editing~\citep{huang2024surveyn03} & 2024 & \scOverviewNo & \scOverviewNo & \scOverviewYes & \scOverviewNo & \scOverviewNo & \scOverviewNo & \scOverviewNo \\
Controllable Text-to-Image Generation~\citep{cao2024surveyn04} & 2024 & \scOverviewPartial & \scOverviewPartial & \scOverviewYes & \scOverviewNo & \scOverviewNo & \scOverviewNo & \scOverviewNo \\
Diffusion-Based Video Editing~\citep{sun2024videoeditingsurvey} & 2024 & \scOverviewYes & \scOverviewNo & \scOverviewYes & \scOverviewNo & \scOverviewNo & \scOverviewNo & \scOverviewNo \\
\rowcolor{scOverviewFamily}[0pt][0pt]
\multicolumn{9}{@{}c@{}}{\rule[-2.7pt]{0pt}{10.6pt}{\bfseries\itshape Long Video Generation and Memory}} \\
Long Video Generation~\citep{li2024longvideosurvey} & 2024 & \scOverviewYes & \scOverviewNo & \scOverviewYes & \scOverviewNo & \scOverviewNo & \scOverviewNo & \scOverviewNo \\
Long Video Generation: Latest Trends~\citep{waseem2024surveyn07} & 2024 & \scOverviewYes & \scOverviewNo & \scOverviewYes & \scOverviewYes & \scOverviewNo & \scOverviewNo & \scOverviewYes \\
Spatiotemporal Consistency~\citep{yin2025spatiotemporal} & 2025 & \scOverviewYes & \scOverviewPartial & \scOverviewYes & \scOverviewPartial & \scOverviewPartial & \scOverviewNo & \scOverviewPartial \\
The Past Frames the Future: Video Memory~\citep{chen2026surveyn14} & 2026 & \scOverviewYes & \scOverviewYes & \scOverviewYes & \scOverviewNo & \scOverviewPartial & \scOverviewNo & \scOverviewNo \\
\rowcolor{scOverviewFamily}[0pt][0pt]
\multicolumn{9}{@{}c@{}}{\rule[-2.7pt]{0pt}{10.6pt}{\bfseries\itshape Specialized Video Generation}} \\
Human Video Generation~\citep{lei2024humanvideosurvey} & 2024 & \scOverviewYes & \scOverviewNo & \scOverviewYes & \scOverviewNo & \scOverviewNo & \scOverviewNo & \scOverviewNo \\
Controllable Video Generation~\citep{ma2025controllablevideo} & 2025 & \scOverviewYes & \scOverviewPartial & \scOverviewYes & \scOverviewPartial & \scOverviewPartial & \scOverviewNo & \scOverviewPartial \\
Physics Cognition in Video Generation~\citep{lin2025physicscognitionsurvey} & 2025 & \scOverviewYes & \scOverviewYes & \scOverviewYes & \scOverviewYes & \scOverviewYes & \scOverviewNo & \scOverviewYes \\
\rowcolor{scOverviewFamily}[0pt][0pt]
\multicolumn{9}{@{}c@{}}{\rule[-2.7pt]{0pt}{10.6pt}{\bfseries\itshape Generation Evaluation and Alignment}} \\
Diffusion Model Alignment~\citep{liu2024surveyn11} & 2024 & \scOverviewYes & \scOverviewNo & \scOverviewYes & \scOverviewNo & \scOverviewNo & \scOverviewNo & \scOverviewNo \\
AI-Generated Video Evaluation~\citep{liu2024videogenerationevaluation} & 2024 & \scOverviewYes & \scOverviewNo & \scOverviewYes & \scOverviewNo & \scOverviewNo & \scOverviewNo & \scOverviewPartial \\
Preference Alignment for Image Generation~\citep{wu2025surveyn05} & 2025 & \scOverviewPartial & \scOverviewPartial & \scOverviewYes & \scOverviewNo & \scOverviewPartial & \scOverviewNo & \scOverviewPartial \\
Video Post-Training and Alignment~\citep{li2026videoposttraining} & 2026 & \scOverviewYes & \scOverviewPartial & \scOverviewYes & \scOverviewYes & \scOverviewPartial & \scOverviewNo & \scOverviewYes \\
\rowcolor{scOverviewFamily}[0pt][0pt]
\multicolumn{9}{@{}c@{}}{\rule[-2.7pt]{0pt}{10.6pt}{\bfseries\itshape Visual Production Workflows}} \\
AI-Empowered Persuasive Video Generation~\citep{liu2021persuasivevideosurvey} & 2021 & \scOverviewYes & \scOverviewNo & \scOverviewYes & \scOverviewNo & \scOverviewNo & \scOverviewNo & \scOverviewNo \\
LLMs Meet Multimodal Generation and Editing~\citep{he2024llmsgenerationediting} & 2024 & \scOverviewYes & \scOverviewNo & \scOverviewYes & \scOverviewYes & \scOverviewNo & \scOverviewNo & \scOverviewYes \\
Visual Generation in the New Era~\citep{wu2026visualnewera} & 2026 & \scOverviewYes & \scOverviewYes & \scOverviewYes & \scOverviewPartial & \scOverviewYes & \scOverviewNo & \scOverviewYes \\
Agentic Generative Systems~\citep{xu2026ags} & 2026 & \scOverviewYes & \scOverviewPartial & \scOverviewPartial & \scOverviewYes & \scOverviewPartial & \scOverviewNo & \scOverviewYes \\
Agentic Artifact Creation~\citep{wang2026artifactcreation} & 2026 & \scOverviewYes & \scOverviewYes & \scOverviewPartial & \scOverviewYes & \scOverviewYes & \scOverviewNo & \scOverviewYes \\
Beyond Visual Generation: Agentic Creation~\citep{mei2026beyond} & 2026 & \scOverviewYes & \scOverviewNo & \scOverviewYes & \scOverviewYes & \scOverviewNo & \scOverviewNo & \scOverviewYes \\
Agentic Visual Generation: Models to Control~\citep{huang2026agenticvisual} & 2026 & \scOverviewYes & \scOverviewPartial & \scOverviewYes & \scOverviewYes & \scOverviewPartial & \scOverviewYes & \scOverviewYes \\
Agentic Visual Generation~\citep{xing2026agenticvisualsurvey} & 2026 & \scOverviewYes & \scOverviewYes & \scOverviewYes & \scOverviewYes & \scOverviewPartial & \scOverviewPartial & \scOverviewYes \\
\rowcolor{scOverviewFamily}[0pt][0pt]
\multicolumn{9}{@{}c@{}}{\rule[-2.7pt]{0pt}{10.6pt}{\bfseries\itshape Video Understanding Agents}} \\
Agentic Video Understanding~\citep{deng2026agenticvideounderstanding} & 2026 & \scOverviewNo & \scOverviewPartial & \scOverviewYes & \scOverviewNo & \scOverviewPartial & \scOverviewNo & \scOverviewYes \\
Video-RSI: Harness Evolution (method)~\citep{luo2026videorsi} & 2026 & \scOverviewNo & \scOverviewNo & \scOverviewPartial & \scOverviewNo & \scOverviewNo & \scOverviewNo & \scOverviewYes \\
\rowcolor{scOverviewFamily}[0pt][0pt]
\multicolumn{9}{@{}c@{}}{\rule[-2.7pt]{0pt}{10.6pt}{\bfseries\itshape General and Interactive World Models}} \\
Sora as a World Model?~\citep{puspitasari2024soraworldsurvey} & 2024 & \scOverviewYes & \scOverviewYes & \scOverviewYes & \scOverviewPartial & \scOverviewYes & \scOverviewNo & \scOverviewYes \\
Understanding World or Predicting Future?~\citep{ding2024worldmodels} & 2024 & \scOverviewYes & \scOverviewYes & \scOverviewYes & \scOverviewNo & \scOverviewYes & \scOverviewNo & \scOverviewYes \\
Interactive Generative Video~\citep{yu2025surveyn12} & 2025 & \scOverviewYes & \scOverviewYes & \scOverviewYes & \scOverviewNo & \scOverviewYes & \scOverviewNo & \scOverviewYes \\
Visual World Simulation: A Roadmap~\citep{yue2025surveyn13} & 2025 & \scOverviewYes & \scOverviewYes & \scOverviewYes & \scOverviewNo & \scOverviewYes & \scOverviewNo & \scOverviewYes \\
Agentic World Modeling~\citep{chu2026agenticworldmodel} & 2026 & \scOverviewYes & \scOverviewYes & \scOverviewYes & \scOverviewNo & \scOverviewYes & \scOverviewPartial & \scOverviewYes \\
Quo Vadis, World Modeling?~\citep{yang2026quovadis} & 2026 & \scOverviewNo & \scOverviewYes & \scOverviewYes & \scOverviewNo & \scOverviewYes & \scOverviewNo & \scOverviewYes \\
\rowcolor{scOverviewFamily}[0pt][0pt]
\multicolumn{9}{@{}c@{}}{\rule[-2.7pt]{0pt}{10.6pt}{\bfseries\itshape Embodied World Models}} \\
World Models for Embodied AI~\citep{li2025embodiedworldmodels} & 2025 & \scOverviewYes & \scOverviewYes & \scOverviewYes & \scOverviewPartial & \scOverviewYes & \scOverviewNo & \scOverviewYes \\
A Step Toward World Models: Manipulation~\citep{zhang2025surveyn18} & 2025 & \scOverviewYes & \scOverviewYes & \scOverviewYes & \scOverviewNo & \scOverviewYes & \scOverviewNo & \scOverviewYes \\
Robotic Video World Models~\citep{mei2026surveyn19} & 2026 & \scOverviewYes & \scOverviewYes & \scOverviewYes & \scOverviewNo & \scOverviewYes & \scOverviewNo & \scOverviewYes \\
World Models for Robot Learning~\citep{hou2026robotworldmodels} & 2026 & \scOverviewYes & \scOverviewYes & \scOverviewYes & \scOverviewNo & \scOverviewYes & \scOverviewNo & \scOverviewYes \\
World Models for Robotic Manipulation~\citep{wang2026surveyn20} & 2026 & \scOverviewYes & \scOverviewYes & \scOverviewYes & \scOverviewNo & \scOverviewYes & \scOverviewNo & \scOverviewYes \\
World-Action Models for Robot Learning~\citep{lu2026surveyn17} & 2026 & \scOverviewYes & \scOverviewYes & \scOverviewYes & \scOverviewNo & \scOverviewYes & \scOverviewNo & \scOverviewYes \\
Embodied World Models: Plausible to Actionable~\citep{yao2026surveyn16} & 2026 & \scOverviewYes & \scOverviewYes & \scOverviewYes & \scOverviewNo & \scOverviewYes & \scOverviewNo & \scOverviewYes \\
\rowcolor{scOverviewFamily}[0pt][0pt]
\multicolumn{9}{@{}c@{}}{\rule[-2.7pt]{0pt}{10.6pt}{\bfseries\itshape Agent Harness}} \\
Code as Agent Harness~\citep{ning2026code} & 2026 & \scOverviewNo & \scOverviewYes & \scOverviewPartial & \scOverviewNo & \scOverviewYes & \scOverviewNo & \scOverviewYes \\
Agent System and Harness Design~\citep{guo2026harnessdesign} & 2026 & \scOverviewNo & \scOverviewNo & \scOverviewPartial & \scOverviewNo & \scOverviewNo & \scOverviewPartial & \scOverviewYes \\
\rowcolor{scOverviewFamily}[0pt][0pt]
\multicolumn{9}{@{}c@{}}{\rule[-2.7pt]{0pt}{10.6pt}{\bfseries\itshape Self-Evolving Systems}} \\
Self-Evolution of Large Language Models~\citep{tao2024surveyr01} & 2024 & \scOverviewNo & \scOverviewNo & \scOverviewYes & \scOverviewNo & \scOverviewNo & \scOverviewPartial & \scOverviewYes \\
Self-Evolving Agents~\citep{gao2025surveyr02} & 2025 & \scOverviewNo & \scOverviewPartial & \scOverviewYes & \scOverviewNo & \scOverviewPartial & \scOverviewPartial & \scOverviewYes \\
RSI: Bounded Refinement to Research Loops~\citep{chen2026surveyr04} & 2026 & \scOverviewPartial & \scOverviewNo & \scOverviewYes & \scOverviewNo & \scOverviewNo & \scOverviewYes & \scOverviewYes \\
Recursive Self-Improvement in AI~\citep{wu2026rsisurvey} & 2026 & \scOverviewNo & \scOverviewNo & \scOverviewYes & \scOverviewNo & \scOverviewNo & \scOverviewYes & \scOverviewYes \\
Self-Improvements in Modern Agentic Systems~\citep{ren2026surveyr03} & 2026 & \scOverviewPartial & \scOverviewYes & \scOverviewYes & \scOverviewNo & \scOverviewYes & \scOverviewYes & \scOverviewYes \\
The Last AI Built by Humans: Genuine RSI~\citep{duan2026surveyn21} & 2026 & \scOverviewPartial & \scOverviewYes & \scOverviewYes & \scOverviewNo & \scOverviewYes & \scOverviewYes & \scOverviewYes \\
\midrule
\rowcolor{scOverviewOurs}
{\bfseries Ours} & 2026 & \scOverviewYes & \scOverviewYes & \scOverviewYes & \scOverviewYes & \scOverviewYes & \scOverviewYes & \scOverviewYes \\
\bottomrule
\end{tabularx}
\endgroup
\end{table}

%% file: sections/02_background.tex
\section{Foundations, Preliminaries, and Framework}
\label{sec:background}

This section establishes the vocabulary needed to compare methods that were
developed in different communities.  We first describe video generation as a
conditional spatiotemporal modeling problem, abstracting away from a
particular generator family.  We then separate the agent that makes
decisions from the harness that executes them and from the models and
tools on which it acts. We distinguish the roles of feedback-producing
components, then formulate harness intervention across agentic video generation,
world-model interaction, and model development toward RSI.

\subsection{Video Generation Foundations}
\label{sec:background-video}

Let a video be a sequence
$\mathbf{x}=(x_1,\ldots,x_T)$, optionally accompanied by synchronized audio.
A conditional video generator $G_\theta$ models
\begin{equation}
  \mathbf{x} \sim p_\theta(\mathbf{x}\mid \mathbf{c},z),
  \label{eq:video-generation}
\end{equation}
where $\mathbf{c}$ denotes one or more conditions and $z$ captures stochastic
variation.  The condition may contain text, reference images or clips,
keyframes, masks, depth or pose signals, trajectories, camera parameters,
audio, or actions.  This formulation covers text-to-video, image-to-video,
video continuation, video-to-video translation, inpainting, editing, and
action-conditioned prediction.  In an agentic system, $\mathbf{c}$ is not
necessarily supplied once by a user: it can be constructed, decomposed, or
revised by a controller over multiple attempts.

We first introduce diffusion and flow matching as generative formulations,
then autoregressive video generation and its streaming execution, and
finally video world models as a functional role in
interaction. These distinctions are not mutually exclusive model categories.

\textbf{Diffusion and score-based models} learn to reverse a corruption
process, typically through repeated denoising steps.  Video Diffusion Models
extended this construction jointly across spatial and temporal dimensions
\citep{ho2022videodiffusion}; later systems explored full-duration space-time
generation and large-scale latent video diffusion
\citep{bartal2024lumiere,kong2024hunyuanvideo}.  Most large systems operate in
a compressed latent space to reduce the cost of high-resolution synthesis
\citep{rombach2021latentdiffusion}, and transformer denoisers provide a
scalable alternative to U-Net backbones \citep{peebles2022dit}.
\textbf{Flow matching} learns a time-dependent vector field,
\begin{equation}
  \frac{d\mathbf{x}_\lambda}{d\lambda}
  = v_\theta(\mathbf{x}_\lambda,\lambda,\mathbf{c}),
  \label{eq:flow-video}
\end{equation}
that transports samples from a simple distribution to the data distribution;
flow matching provides a simulation-free objective for learning this field
\citep{lipman2022flowmatching}. Here $\lambda$ is continuous sampling time,
distinct from the discrete interaction index $t$ used below.

\textbf{Autoregressive video generation} factorizes generation into successive
conditional predictions. Token-based models tokenize video, text, images,
and sometimes audio, and write
\begin{equation}
  p_\theta(\mathbf{u}\mid\mathbf{c})
  = \prod_{i=1}^{N} p_\theta(u_i\mid u_{<i},\mathbf{c}),
  \label{eq:autoregressive-video}
\end{equation}
where $\mathbf{u}$ is a discrete or continuous token sequence. Phenaki uses
compressed causal video tokens for variable-length generation, while
VideoPoet casts multiple video, image, text, and audio tasks into a unified
autoregressive interface
\citep{villegas2022phenaki,kondratyuk2023videopoet}. Such interfaces are
attractive to a harness because heterogeneous inputs and outputs can be
expressed through a common sequence model, although long sequences make
planning and temporal state preservation difficult. Autoregressive temporal
generation can also be combined with diffusion within each generated frame
or chunk, as illustrated by Self Forcing \citep{huang2025selfforcing}.

For incremental execution, streaming generation produces and exposes successive frames or
chunks before the complete video is available, rather than requiring a
full-clip output before execution can continue. It is an execution mode,
not a generative formulation mutually exclusive with diffusion, flow
matching, or autoregression. Self Forcing demonstrates streaming
autoregressive diffusion, and Matrix-Game 2.0 combines few-step
autoregressive diffusion with action-conditioned streaming generation
\citep{huang2025selfforcing,he2025matrixgame2}. For a harness, incremental
outputs provide opportunities to observe results and, where supported,
update controls between generation intervals. This makes latency, temporal
state retention, and the interval at which a new control can take effect
important interface properties; streaming output alone does not establish
interactive control.

\textbf{Video world models.} For this survey, a video world model is a
learned model of visual environmental dynamics that predicts future video
observations from visual history and supported actions or controls
\citep{bruce2024genie,ding2026playworld}. Using the same parameter notation
$\theta$, a prediction over $\Delta$ interaction steps can be written as
\begin{equation}
  \mathbf{x}_{t+1:t+\Delta}
  \sim p_\theta\bigl(\cdot\mid\mathbf{x}_{\leq t},
    \boldsymbol{\alpha}_{<t+\Delta},\mathbf{c}\bigr),
  \label{eq:video-world-model}
\end{equation}
where $t$ indexes interaction time, $\boldsymbol{\alpha}_{<t+\Delta}$ contains
past controls and the proposed controls over the prediction horizon, and
$\mathbf{c}$ specifies additional conditions. Controls may be keyboard
inputs, camera trajectories, robot commands, or learned latent actions.
The world-model role extends synthesis to action-conditioned prediction or
continuing interaction: returned frames can inform the next control or a
planning decision. A long, plausible video alone does not establish this
role. Neither an explicit 3D scene representation nor access to hidden state
is required; control fidelity and continuity of interaction consequences
must instead be verified empirically. Section~\ref{sec:world-models}
examines how a harness organizes these capabilities into an interactive system.

Across these formulations and execution modes, agenticity is determined by
the decisions wrapped around or embedded within the generator, not by the
choice of sampler alone. Video changes the design of an agentic loop in
ways that image generation does not. Quality depends on both frame-level
fidelity and cross-frame behavior, including identity preservation, motion
plausibility, camera continuity, object permanence, causal ordering, audio
synchronization, and long-horizon narrative state. Sampling and evaluation
are expensive, a single scalar score rarely identifies where a failure
occurs, and a local repair can introduce downstream inconsistencies. These
properties motivate hierarchical plans, persistent state, temporally
localized critics, and budget-aware decisions about whether to accept,
edit, extend, or regenerate a candidate
\citep{yin2025spatiotemporal,liu2024videogenerationevaluation}.

Table~\ref{tab:video-notation} summarizes the notation for generation and
world-model prediction. Sampling time, token position, and interaction time
index different processes; bold $\mathbf{c}$ denotes generator conditions.
\input{tables/video_generation_notation}

\subsection{Agents and Harness Engineering}
\label{sec:background-harness}

An \textbf{agent} is a goal-conditioned decision process that selects
consequential actions from observations and uses their outcomes to guide
subsequent decisions. Let $t$ index decision steps, $g$ denote a goal,
$o_t$ the observation available before decision $t$, $a_t$ the selected
action, $y_t$ its execution outcome, and $f_t$ the feedback derived from
that outcome. The available interaction history is
$h_t=(o_0,a_0,y_0,f_0,\ldots,o_t)$. A controller policy $\pi_\phi$
specifies
\begin{equation}
  a_t \sim \pi_\phi(\cdot\mid g,h_t,\mathcal{A}_t),
  \qquad a_t\in\mathcal{A}_t,
  \label{eq:agent-policy}
\end{equation}
where $\mathcal{A}_t$ is the admissible action set, including termination
when permitted. The policy can be learned, programmed, or hybrid; using
an LLM or VLM is an implementation choice, not the definition of agency.
Training a model for reasoning or tool use develops its capabilities,
but does not by itself provide the execution process that turns its
outputs into actions \citep{dubois2025llmagents,yao2023react}.

\textbf{Agent Components.} We use the functional decomposition
\begin{equation}
  \mathfrak{A}_\phi
  =(\mathcal{O},\pi_\phi,\mathcal{M},\mathcal{T},\mathcal{F}),
  \label{eq:agent-components}
\end{equation}
covering observation, planning and decision-making, memory, tool-mediated
action, and feedback. This is an organizing abstraction rather than a
mandatory module inventory: planning may be implicit in the policy, memory
may be limited to the current history, and feedback may consist of returned
observations rather than a separate learned critic
\citep{weng2023agents,shinn2023reflexion}. A plan structures future actions;
the policy determines whether to follow or revise it as evidence arrives.
The generator $G_\theta$ from Section~\ref{sec:background-video} can be one
tool in $\mathcal{T}$; its parameters $\theta$ are distinct from the
controller parameters $\phi$ when a learned controller is used.

\textbf{Harness Engineering.} A functional agent description says what
the system must do, but not how its decisions become reliable operations.
For video generation, this requires coordinating observations, tool calls,
and retained project state across successive decisions.
An \textbf{agent harness} is the executable runtime that constructs
controller context, exposes tool interfaces, dispatches and schedules
operations, maintains state, routes feedback, and enforces permissions,
budgets, and stopping rules \citep{weng2026harness,ning2026code}.
It is more than a collection of prompts or a tool list. It specifies,
for example, which frames a controller sees, how a requested edit is
applied, which artifact version is committed, and how a failed operation
returns to the decision loop. Delegation and background jobs, when
supported, are further execution operations rather than prerequisites
for being an agent.

For this survey, we represent these responsibilities as
\begin{equation}
  \mathcal{H}
  =(\mathcal{O},\mathcal{C},\mathcal{T},\mathcal{X},
    \mathcal{M},\mathcal{F},\mathcal{R}),
  \label{eq:harness-components}
\end{equation}
where $\mathcal{C}$ constructs the controller context, $\mathcal{X}$
executes actions, and $\mathcal{R}$ determines admissible operations and
termination conditions under task, resource, and permission constraints.
$\mathcal{O}$, $\mathcal{M}$, $\mathcal{T}$, and $\mathcal{F}$ realize the
corresponding functional agent roles, with $\mathcal{M}$ also committing
updates to execution state. The agent and harness are thus not disjoint
component inventories: the agent denotes the goal-directed decision
process, while the harness implements and constrains its operation.

Let $s_t=(q_t,m_t,b_t,e_t)$ be the harness-managed state before decision
$t$. Here $q_t$ contains current task specifications, plans, artifacts,
and their versions; $m_t$ contains stored evidence and experience, with
its within-task or cross-task retention scope specified explicitly;
$b_t$ records the remaining resource budget; and $e_t$ contains tool
and runtime records, including execution status, pending operations,
timestamps, and newly available inputs. Interaction history may be
retained in these records or accessed through memory. This state need
not reveal the environment's hidden state.
Writing subscripts $\mathcal{H}$ for operators configured by a particular
harness, one decision cycle is
\begin{equation}
  \begin{aligned}
    o_t &= \mathcal{O}_{\mathcal{H}}(g,s_t), &
    c_t &= \mathcal{C}_{\mathcal{H}}(g,o_t,s_t), \\
    \mathcal{A}_t &= \mathcal{R}_{\mathcal{H}}(g,s_t;\mathcal{T}), &
    a_t &\sim \pi_\phi(\cdot\mid c_t,\mathcal{A}_t), \\
    y_t &\sim \mathcal{X}_{\mathcal{H}}
      (\cdot\mid s_t,a_t;\mathcal{T}), &
    f_t &= \mathcal{F}_{\mathcal{H}}(g,s_t,a_t,y_t), \\
    s_{t+1} &= \mathcal{M}_{\mathcal{H}}(s_t,a_t,y_t,f_t). &&
  \end{aligned}
  \label{eq:agent-harness-loop}
\end{equation}
Here $c_t$ is the selected context presented to the controller, rather
than necessarily the complete history $h_t$ used in the abstraction
of Equation~\ref{eq:agent-policy}. It may contain task instructions,
observations, retrieved memory, plans, and execution constraints.
The operator $\mathcal{M}_{\mathcal{H}}$ commits memory together with
task, budget, and execution-state updates. The next observation is
obtained from the updated records and newly accessible evidence.
This loop makes the distinction concrete: the controller proposes
what to do, the harness governs what can actually run and what persists,
and the tools or environment produce the consequences.

A decision cycle need not correspond to a video frame or a complete
generation call. In asynchronous or streaming execution, $y_t$ may
contain an acknowledgement or an available video chunk; $e_t$ tracks
pending operations and associates later results with their originating
requests, making those results available to subsequent observations.

Table~\ref{tab:component-boundaries} summarizes the agent and harness,
their functional components, and the quantities in one execution cycle.
Calligraphic symbols denote components or sets; the subscript $t$ indexes
a decision cycle. The history $h_t$ records prior interaction, the state
$s_t$ stores harness-managed information, and the context $c_t$ contains
the information selected for the current controller invocation.
In particular, $c_t$ is controller context, whereas $\mathbf{c}$ in
Section~\ref{sec:background-video} is a generator condition;
$a_t$ is a harness-level action, which may construct generator conditions,
issue one or more world-model controls $\boldsymbol{\alpha}$,
or invoke another tool.
\input{tables/agent_harness_notation}

\textbf{Feedback roles and boundaries.}
\label{sec:background-feedback}
An internal \textbf{critic} or \textbf{verifier} supplies diagnosis or
checks through $\mathcal{F}_{\mathcal{H}}$ before termination, so the policy
can change its next action. An \textbf{external evaluator} reports evidence
to researchers or auditors without controlling that run. A \textbf{reward
model} produces a scalar or preference signal; it may support ranking,
search, or offline learning, but does not itself specify the next action.
An \textbf{editor} belongs to $\mathcal{T}$ because it changes an artifact,
whereas a critic produces evidence about one. The same model can occupy
different roles depending on where its output is routed.

These boundaries also matter for attribution. Changing $G_\theta$,
$\pi_\phi$, $\mathcal{H}$, or the budget can each improve end-to-end results
for different reasons. Runtime feedback may revise context, a plan, or
memory while leaving both $\theta$ and $\phi$ fixed.
The harness $\mathcal{H}$ specifies executable components and rules,
whereas $s_t$ records their current contents and execution state;
updating $s_t$ does not by itself establish persistent improvement
across tasks. Human participation or a reward signal alone does not
imply agentic control or parameter learning \citep{liu2025videoalign}.
Section~\ref{sec:background-formulation} uses this shared notation to
distinguish agentic video generation, world interaction, and persistent
development updates.

\subsection{A Unified View of Harness Intervention}
\label{sec:background-formulation}
\label{sec:framework}

The three questions in Section~\ref{sec:introduction-questions} concern
different objects of harness intervention: an agentic video generation workflow, an
interactive world, and the development of future video-generation capability.
We compare these settings through the decisions a harness makes executable
and the changes retained from feedback. A system may connect several
settings, with each connection requiring an identifiable transfer of evidence
or an accepted update.

Using the state, context, and action notation of
Section~\ref{sec:background-harness}, we summarize the execution cycle as
\begin{equation}
  a_t \sim \pi_\phi(\cdot\mid c_t,\mathcal{A}_t), \qquad
  s_{t+1}\sim\mathcal{K}_{\mathcal{H}}
    (\cdot\mid s_t,a_t;\mathcal{T}).
  \label{eq:harness-intervention}
\end{equation}
Here $\mathcal{K}_{\mathcal{H}}$ is the aggregate transition induced by
tool execution, feedback routing, and state commitment in
Equation~\ref{eq:agent-harness-loop}, not a transition model for the world's
hidden state. The tool set $\mathcal{T}$ may include generators, critics,
editors, interaction interfaces, and development tools. The recorded trace
$\tau=(s_0,c_0,a_0,y_0,f_0,\ldots,s_{t_{\mathrm{end}}})$ supplies evidence for subsequent
decisions and possible development updates.

\begin{samepage}
\textbf{Agentic video generation} instantiates $q_t$ as a project containing
requirements, plans, assets, and candidate videos. Actions change the
production process or its artifacts: schedule a shot, select a generator,
inspect a temporal defect, or apply a local edit. Critics and human feedback
help the controller decide which intervention to attempt next. Section~\ref{sec:production}
examines how these interventions affect output quality, consistency, and
execution efficiency, and which skills can be reused across requests.
\par\end{samepage}

\textbf{Video world-model interaction} instantiates $q_t$ through generated
observations, interaction history, and exposed or estimated scene state.
The generator supplies action-conditioned transitions, while the harness
organizes controls, events, execution timing, and relevant memory. An action
therefore changes the conditions of a continuing interaction, whose outcomes
become observations for later decisions. Section~\ref{sec:world-models}
examines where the harness can intervene and how control fidelity, state
continuity, and grounded evaluation establish its effect.

\textbf{Video model development toward RSI} instantiates $q_t$ as a
development task involving data, supervision, checkpoints, and evaluation
records. Actions select learning experience, revise supervision, launch
training, test a candidate, or retain a validated change. Section~\ref{sec:model-development}
studies how these decisions connect data engineering, training, and evaluation.
The RSI target additionally makes the procedures governing future development
modifiable. Let $\xi_k$ collect persistent data, model, critic, and reusable
harness state at round $k$, and let $\rho_k$ describe its testing,
data-selection, and update procedures. The production harness belongs to
$\xi_k$: a change to its prompts, memory, skills, or controller counts as a
change to $\rho_k$ only when it alters how later updates are proposed,
selected, or validated. A prospective development process can
then be represented as
\begin{equation}
  (\xi_{k+1},\rho_{k+1})
  =\mathcal{U}(\xi_k,\rho_k;\mathcal{L}_k,F_k,B_k),
  \label{eq:development-update}
\end{equation}
where $\mathcal{L}_k=\{\tau^{(j)}\}$ contains the traces collected in round $k$,
$F_k$ is development feedback, and $B_k$ is the round's resource budget.
The operator $\mathcal{U}$ is fixed; it applies whatever procedure $\rho_k$
specifies, so revising $\rho_k$ does not require revising $\mathcal{U}$. Holding $\rho_k$ fixed describes
repeated execution of an improvement procedure. RSI concerns validated
changes to that procedure that make later improvement rounds more effective.
Independent assessment must establish this benefit beyond the feedback used
to select the changes.

Agentic video generation and world interaction can supply failure evidence for development,
while accepted model and harness changes can improve later generation and
interaction. Such transfers require selecting useful traces, grounding
correction targets, and validating the resulting updates. Online and offline
operation are execution choices within these settings. Across chapters, we
compare observations, action authority, intervention targets, feedback
consumers, retained state, and independently assessed improvement.

%% file: tables/video_generation_notation.tex
\begingroup
\SurveyTableSetup
\small
\setlength{\tabcolsep}{4pt}
\renewcommand{\arraystretch}{1.08}
\begin{longtable}{@{}
  >{\raggedright\arraybackslash}p{0.28\textwidth}
  >{\raggedright\arraybackslash}p{\dimexpr0.72\textwidth-2\tabcolsep\relax}@{}}
\caption{\textbf{Symbols in video generation and video world modeling.}
The notation distinguishes sampling time, token position, and interaction
time; a world model is a functional role rather than a separate sampler family.}
\label{tab:video-notation}\\
\toprule
Symbol & Definition \\
\midrule
\endfirsthead
\multicolumn{2}{@{}l}{\textbf{Table \thetable\ (continued)}}\\
\toprule
Symbol & Definition \\
\midrule
\endhead
\midrule
\multicolumn{2}{r@{}}{\footnotesize Continued on the next page}\\
\endfoot
\bottomrule
\endlastfoot
$\mathbf{x}=(x_1,\ldots,x_T)$, $T$
  & Video sequence, individual frames, and sequence length \\
$G_\theta$, $\theta$
  & Conditional generator and its parameters \\
$p_\theta(\cdot\mid\cdot)$
  & Model's conditional probability law \\
$\mathbf{c}$
  & Generator conditions \\
$z$
  & Stochastic variation \\
$\lambda$, $\mathbf{x}_\lambda$
  & Continuous sampling time and sample state along the flow \\
$v_\theta(\mathbf{x}_\lambda,\lambda,\mathbf{c})$
  & Learned time-dependent vector field \\
$\mathbf{u}$, $u_i$, $u_{<i}$, $N$
  & Token sequence, current token, preceding tokens, and token count \\
$t$, $\Delta$
  & Discrete interaction index and prediction horizon \\
$x_t$
  & Visual observation at interaction step $t$, a frame or returned video chunk \\
$\mathbf{x}_{\leq t}$
  & Visual observation history through step $t$ \\
$\mathbf{x}_{t+1:t+\Delta}$
  & Predicted future observations \\
$\alpha_t$, $\boldsymbol{\alpha}_{<t}$, $\boldsymbol{\alpha}_{\leq t}$
  & Current control; past controls; control history including the current control \\
$\boldsymbol{\alpha}_{<t+\Delta}$
  & Past controls and proposed controls over the horizon \\
$\boldsymbol{\alpha}_{t:t+\Delta-1}$
  & Future control sequence, supplied in conditional prediction or jointly generated with video \\
$\mathbf{s}_t^{\mathrm{w}}$
  & Explicit or latent predictive state of the world model, distinct from harness-managed $s_t$ \\
\end{longtable}
\endgroup

%% file: tables/agent_harness_notation.tex
\begingroup
\SurveyTableSetup
\small
\setlength{\tabcolsep}{4pt}
\renewcommand{\arraystretch}{1.08}
\begin{longtable}{@{}
  >{\raggedright\arraybackslash}p{0.28\textwidth}
  >{\raggedright\arraybackslash}p{\dimexpr0.72\textwidth-2\tabcolsep\relax}@{}}
\caption{\textbf{Symbols and functional roles of agents and the agent harness.}
Components need not be separate networks or software modules.
The subscript $\mathcal{H}$ denotes an operator configured by the harness;
lower-case variables describe one execution cycle.}
\label{tab:component-boundaries}\\
\toprule
Symbol & Definition \\
\midrule
\endfirsthead
\multicolumn{2}{@{}l}{\textbf{Table \thetable\ (continued)}}\\
\toprule
Symbol & Definition \\
\midrule
\endhead
\midrule
\multicolumn{2}{r@{}}{\footnotesize Continued on the next page}\\
\endfoot
\bottomrule
\endlastfoot

$\mathfrak{A}_\phi$
  & Functional agent decomposition \\

$\pi_\phi$, $\phi$
  & Controller policy and its parameters, when learned;
    distinct from generator parameters $\theta$ \\

$\mathcal{H}$
  & Executable agent harness \\

$\mathcal{O}$
  & Observation component \\

$\mathcal{C}$
  & Context-construction component \\

$\mathcal{T}$
  & Available tools and their interfaces \\

$\mathcal{X}$
  & Action-execution component \\

$\mathcal{M}$
  & Memory and state-update component \\

$\mathcal{F}$
  & Feedback component \\

$\mathcal{R}$
  & Admissibility and termination component \\

$t$
  & Discrete decision-cycle index, not necessarily
    a video-frame index \\

$g$
  & Goal \\

$o_t$, $h_t$
  & Observation available before decision $t$ and available
    interaction history $h_t=(o_0,a_0,y_0,f_0,\ldots,o_t)$ \\

$c_t$
  & Selected controller context, distinct from the complete
    history $h_t$ and generator conditions $\mathbf{c}$ \\

$a_t$, $\mathcal{A}_t$
  & Selected harness-level action and currently admissible
    action set \\

$y_t$, $f_t$
  & Execution outcome and feedback derived from that outcome \\

$s_t=(q_t,m_t,b_t,e_t)$
  & Harness-managed task, memory, budget, and runtime state
    before decision $t$ \\

$q_t$, $m_t$
  & Current task specifications, plans, artifacts, and versions;
    stored evidence and experience with a specified retention scope \\

$b_t$, $e_t$
  & Remaining resource budget; tool and runtime records,
    including execution status, pending operations, timestamps,
    and newly available inputs \\

\end{longtable}
\endgroup

%% file: sections/04_production.tex
\clearpage
\begingroup
\setlength{\emergencystretch}{2em}
\section{Agentic Video Generation}
\label{sec:production}

In agentic video generation, a creative brief is developed into an edited audiovisual work through planning, shot production, and audiovisual editing (Figure~\ref{fig:production-workflow}). Scripts and shot specifications define the narrative and visual requirements; reusable assets guide the generation of candidate takes; and an editing timeline assembles the selected footage with narration and music. Decisions propagate across these artifacts: changing a character reference may affect several shots, replacing a take may alter the cut, and retiming the cut may require adjustments to the audio. Agents can participate at different stages by making production decisions, invoking tools, and inspecting intermediate results. The agent harness connects these activities and maintains the project state needed to revise decisions and coordinate changes across stages.

\par\addvspace{4pt}
\noindent\begin{minipage}{\linewidth}
\setlength{\parskip}{0pt}
\centering
\includegraphics[width=\linewidth]{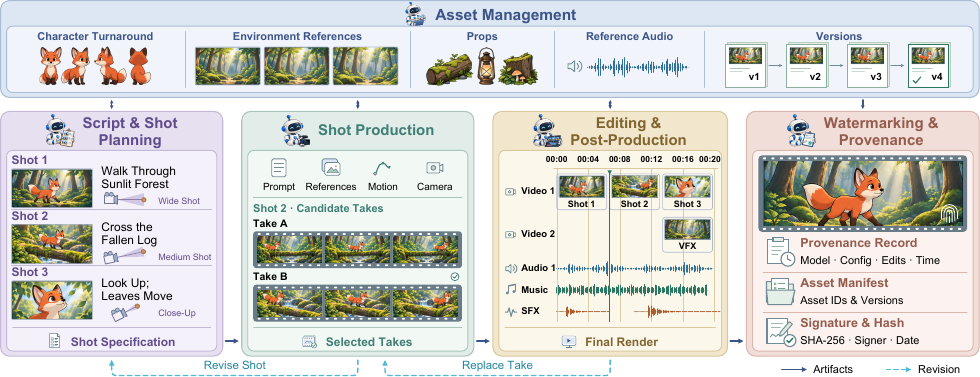}
\captionof{figure}{\textbf{Artifacts and revision in agentic video generation.} Shot specifications lead to candidate takes and an edited audiovisual sequence. Shared assets support references, continuity, and version tracking; dashed arrows request revision. The final panel shows watermarking and provenance at delivery. The example is illustrative and not taken from a surveyed system.}
\label{fig:production-workflow}
\end{minipage}\par\addvspace{4pt}

We organize the literature by the workflow function a system mainly serves: \emph{script and shot planning} decides what will be filmed (Section~\ref{sec:agentic_video_planning}); \emph{asset management} keeps characters, scenes, and other references consistent across shots (Section~\ref{sec:production_assets}); \emph{shot production} creates the footage (Section~\ref{sec:agentic_video_execution}); and \emph{editing and post-production} assembles and modifies it (Section~\ref{sec:agentic_video_editing}) and Section~\ref{sec:agentic_video_watermarking} covers watermarking and provenance at delivery. Asset management runs alongside the other three functions, which all read and update its records. Within each function, subcategories are separated by the object a system mainly changes, and a system covering several functions is placed by its main contribution.

\subsection{Script and Shot Planning}
\label{sec:agentic_video_planning}

Planning turns a brief into decisions that downstream tools can execute, at three levels of detail. \emph{Script development} decides what happens: events, characters, and dialogue. \emph{Shot and task planning} breaks the script into a shot list and the tasks needed to realize it. \emph{Camera and staging} decides how a single shot is filmed: camera position, shot size, angle, and movement, and where actors stand and move (blocking). All three produce plans, recorded as text, trajectories, or engine camera objects; low-level camera control that conditions the generator belongs to shot production. A revision at a higher level affects more: changing an event can invalidate several scenes, whereas changing the camera keeps the story and assets intact.

\noindent\textbf{Script Development.} MovieAgent~\citep{wu2025movieagent} decomposes a film into scenes and shots through hierarchical reasoning by production roles, keeping character, subtitle, and audio requirements attached to each shot. ScriptAgent~\citep{scriptallneed2026} makes this intermediate representation explicit: a scripter translates dialogue into a cinematic script, and a director uses it to coordinate generation with frame anchoring. MAViS~\citep{mavis2025} extends the plan across script, shots, characters, keyframes, animation, and audio through an Explore--Examine--Enhance process. Co-Director~\citep{codirector2026} couples global exploration of creative directions with local multimodal self-refinement, and One Sentence, One Drama~\citep{onesentence2026} combines story debate, spatially grounded references, and reviewers that target particular script or visual failures. Both systems choose which narrative to realize before generating it. Selecting a better take and rewriting the story should be evaluated separately, because the latter changes the target against which later footage is judged.

\noindent\textbf{Shot and Task Planning.} AniMaker~\citep{animaker2025} connects a storyboard to search over generated clip candidates; its photography and reviewer components evaluate each candidate in its story context. FilMaster~\citep{filmaster2025} connects reference retrieval and cinematic planning to a rough-cut-to-fine-cut process driven by simulated audience feedback. FRAMEWORKERS~\citep{frameworkers2026} plans production tasks rather than shot content: a Director selects tasks and sub-agents from an editable Task Stack, and an Assistant executes them in a shared Workspace that stores the resulting artifacts. Whether the plan is a storyboard, a retrieved shot plan, or a task stack, writing it down lets a failure be traced to the specification, the source material, or the generated candidate.

\noindent\textbf{Camera and Staging.} FilmAgent~\citep{filmagent2024} plans actor positions, actions, and camera setups in a virtual set: director--screenwriter critique corrects the script, and cinematographer debate followed by director judgment selects the camera setup before filming. Mind-of-Director~\citep{mindofdirector2026} extends these decisions to 3D scene construction and character blocking and exposes the previsualization in an editable engine timeline. In both systems, a staging decision becomes a change to a production environment that can be inspected before final rendering. Camera Artist~\citep{cameraartist2026} adds a cinematography agent that specifies shot size, angle, and movement across shots; CamPilot~\citep{campilot2026} learns camera-work planning across consecutive shots; and AutoMV for real estate~\citep{automvrealestate2025} plans camera trajectories alongside material, subtitles, and music. A textual instruction such as ``slow dolly-in'' may be interpreted loosely by a generator, whereas a geometric trajectory or engine camera is executed exactly, so camera quality should be assessed on the form that is executed.

\subsection{Asset Management}
\label{sec:production_assets}
\label{sec:agentic_video_memory}

Asset management links reusable material to the decisions that consume it. We distinguish \emph{asset creation}, which produces new references; \emph{identity consistency}, which keeps an entity recognizable across shots; and \emph{cross-shot continuity}, which tracks the state that later shots must respect. Useful records include entity identity, scene membership, the time span in which an asset is valid, generation conditions, and dependencies on other assets. The main distinction is between a stable reference and a changing state: a character's identity stays fixed while pose, location, or clothing changes, and each generation request needs the right combination.

\noindent\textbf{Asset Creation.} Anim-Director~\citep{animdirector2024} creates character and setting references, evaluates candidates, and uses the selected images to guide animation. MM-StoryAgent~\citep{mmstoryagent2025} prepares visual assets, narration, music, and sound effects with modality-specific agents that exchange structured messages. Replacing a visual event can then require a different sound effect, narration length, or musical cue. Creating assets thus involves choosing among candidates, and each audio asset must be tied to the scenes where it applies.

\noindent\textbf{Identity Consistency.} AniME~\citep{anime2025} places a director's global memory above specialized agents and selects control conditions through MCP interfaces. MUSE~\citep{muse2026} translates narrative intent into identity, spatial, and temporal constraints and uses multimodal verification to drive targeted revision. AutoMV for music videos~\citep{automv2025} maintains character references, routes shots to narrative or singing-video tools, and regenerates or falls back when a verifier fails a shot. These systems differ in whether reference use is mediated by a memory interface or by explicit constraints, but all must link the intended entity to the condition supplied to a tool.

\noindent\textbf{Cross-shot Continuity.} GroundShot~\citep{groundshot2026} schedules shots by their usefulness as future references and verifies references extracted from generated footage before storing them, so an unreliable early shot does not become the basis for later ones; as a result, generation order can differ from narrative order. MVAgent~\citep{kong2026mvagent} separates persistent appearance references from the state observed at the end of a shot and composes typed conditions for the next shot through observer, transition, and orchestrator agents. ViMax~\citep{huang2026vimax} combines hierarchical narrative retrieval with dependency-aware tracking of character and environment states. AgentMV~\citep{agentmv2026} records references, segment dependencies, recurring motifs, costs, and execution history, and uses them to allocate a song-level generation budget. Reusing an accepted motif saves a generation call, but a stale state or misaligned recurrence can propagate an error to every segment that reuses it. Keeping appearance, recent state, and narrative facts in separate records makes it less likely that a visually similar but outdated reference is reused.

\noindent\textbf{Consistency Inside the Generator.} Multi-shot generators offer a different route to the same goal. Long Context Tuning~\citep{guo2025lct} extends the attention of a single-shot model to all shots in a scene, with interleaved 3D position embeddings and asynchronous noise, so that shots can be generated jointly or extended one at a time. HoloCine~\citep{holocine2026} generates a whole scene at once, localizing each shot's prompt through window cross-attention and keeping attention dense within shots but sparse between them, and reports persistent characters and scenes without external records. The two routes keep consistency in different places. In a multi-shot generator it lives in the model's context, so it needs no reference bookkeeping but is bounded by the scene the model can attend to. In a harness it lives in explicit records, which can be inspected, corrected, and reused across scenes, and which allow one shot to be regenerated without touching the others; the cost is that every record must be supplied to the generator correctly.

\subsection{Shot Production}
\label{sec:agentic_video_execution}
\label{sec:agentic_video_tools}

Shot production executes a shot specification to obtain candidate footage. We distinguish three routes: \emph{model-based generation} uses a learned video generator; \emph{code-based animation} writes programs, such as Manim scripts, that render the footage; and \emph{3D scene and simulation} builds a spatial scene whose simulation or rendering produces the footage or conditions its generation.

\noindent\textbf{Model-based Generation.} PhyT2V~\citep{phyt2v2025} refines prompts to improve physical plausibility, and GenMAC~\citep{huang2026genmac} coordinates design, verification, suggestion, and correction agents for compositional generation. SCMAPR~\citep{scmapr2026} verifies rewritten prompts against their constraints before generation. VISTA~\citep{vista2025} selects candidates by pairwise comparison and revises the next request from visual, audio, and contextual critiques, while VideoGen-Agent~\citep{li2026videogenagent} learns to coordinate augmentation, generation, and verification tools over multiple interactions. In these systems, the controller can change how evidence is obtained as well as what is generated, and each revision must keep the requirements it does not target, such as identity and timing. MotionAgent~\citep{motionagent2025} converts motion intent into object trajectories and camera parameters that condition the generator. Checking a prompt catches omitted requirements, but only inspecting the generated take reveals failures introduced during synthesis. Many generation pipelines run as ComfyUI node graphs that agents construct and execute; these systems are discussed in Section~\ref{sec:agent-design}.

\noindent\textbf{Code-based Animation.} TheoremExplainAgent~\citep{theoremexplainagent2025} turns explanation scenes into Manim animations and repairs code from execution errors. TeachMaster~\citep{teachmaster2026} adds debugging, synchronization, layout validation, and fallback after repeated rendering failures, and LASEV~\citep{beyondendtoend2026} coordinates solution, illustration, and narration agents through shared state. Code2Video~\citep{code2video2026} combines code repair with a visual critic for layout problems, EvoPresent~\citep{evopresent2026} refines presentation design with aesthetic feedback, and mPresenter~\citep{mpresenter2026} adds audience-oriented outline critique and compile--render--evaluate iteration. These checks catch different failures. Code that runs can still express an incorrect explanation, and correct narration can still be mistimed against the animation, so separate signals let the controller revise the responsible artifact.

\noindent\textbf{3D Scene and Simulation.} LiVER~\citep{liver2026} arranges assets, lighting, and cameras in a scene and uses the rendering as a condition for synthesis, whereas MoReGen~\citep{moregen2025} renders the result of physical simulation directly. FantasyHSI~\citep{fantasyhsi2026} uses navigation, action planning, and a critic to correct drift in human--scene interaction. Kubrick~\citep{he2024kubrick} iterates on Blender programs from reviewer observations, and GEST-Agent~\citep{gestagent2026} checks scene-building operations against simulator state before execution. These three systems validate at different points: after generated motion, after rendered execution, or before an operation is accepted. Cutscene Agent~\citep{cutsceneagent2026} gives specialists MCP access to Unreal Engine's cinematic tools, so a local camera or animation change can be applied without discarding the rest of the scene. The benefit depends on how faithfully the tool interface exposes the engine and on which scene properties the agent can observe. When the scene keeps running and every frame is rendered from it, the generator becomes a renderer of an explicit world state (Section~\ref{sec:iwm-explicit-state}).

\subsection{Editing and Post-production}
\label{sec:agentic_video_editing}

Editing works on existing footage at two levels. At the sequence level, \emph{story-driven cutting} decides which source intervals enter the cut and in what order, and \emph{timeline and tool execution} carries out such decisions on a timeline or media files. At the frame level, \emph{frame content editing} changes pixels inside a shot without moving it in the cut. We group semantic edits with restoration because both rewrite an existing shot and must leave everything outside the target unchanged. At both levels, the editing state must keep source timecodes and applied operations so that a revision touches only the intended interval.

\noindent\textbf{Story-driven Cutting.} CutClaw~\citep{zhao2026cutclaw} decomposes long source footage, anchors a narrative to the music, and refines the cut with editor and reviewer agents. DIRECT~\citep{direct2026} passes editing intent between screenwriter, director, and editor levels, and feedback can change the retrieval query and candidate shots. CineAgents~\citep{cineagents2026} builds hierarchical narrative memory from source footage and develops a creative blueprint into a compiled sequence. The three systems differ in where the narrative anchor comes from: the music in CutClaw, an intent passed from screenwriter to editor in DIRECT, and a memory built from the source footage in CineAgents.

\noindent\textbf{Timeline and Tool Execution.} LAVE~\citep{wang2024lave} translates user intentions into timeline operations within an interactive editor, while EditDuet~\citep{sandoval2025editduet} alternates editor actions with critic feedback in a non-linear editing environment. In LAVE a person guides the next action, and in EditDuet a critic evaluates the evolving sequence; both rely on an explicit timeline whose state can be inspected after each operation. ELLMPEG~\citep{ellmpeg2026} retrieves tool documentation and reflects on media-processing commands before executing them; a valid command can still target the wrong interval, so the rendered output still needs inspection. Crayotter~\citep{crayotter2026} stores blueprints, tool calls, intermediate renders, and exports as persistent artifacts, which supports failure diagnosis and resuming interrupted jobs. These records also let a later revision recover why a source interval was chosen and which downstream operations depend on it.

\noindent\textbf{Frame Content Editing.} Aurora~\citep{aurora2026} uses a tool-augmented agent to complete underspecified editing requests and construct conditions for a video editing model. MMLVE-Agent~\citep{mmlveagent2026} decomposes long footage and instructions by shot while tracking entities, so an edit reaches the intended occurrence of an entity. MA-VFR~\citep{mavfr2026} applies the same pattern to restoration: facial-detail and temporal-consistency agents propose toolchains that an evolving coordinator combines, and restoration results update the coordination memory for later tool choices. Both semantic edits and restoration must leave unaffected content intact, and the replacement must reach every timeline position that uses the shot.

\subsection{Watermarking and Provenance}
\label{sec:agentic_video_watermarking}

Watermarks embed recoverable evidence in the media signal, and provenance records describe how an asset was derived. Where a mark is inserted determines whether it covers a generated take, an edited clip, or the final export. \emph{Generation-time} methods need access to the generator: VideoShield~\citep{videoshield2025} encodes information in the generation noise and localizes tampering by inversion, SIGMark~\citep{sigmark2026} supports blind extraction robust to temporal disturbances, LVMark~\citep{lvmark2024} learns marking within a latent video diffusion model, and SPDMark~\citep{spdmark2026} learns key-indexed parameter shifts that recover frame order. \emph{Post-production} methods work on existing footage, including footage from closed generators: RBMark~\citep{rbmark2025} embeds marks in wavelet coefficients without a learned model, and SepVAMark~\citep{sepvamark2025} learns separable audiovisual marks for source tracing and deepfake detection. Detecting an alteration and locating it are separate capabilities, and robustness to compression says nothing about detecting unauthorized edits. C2PA Content Credentials~\citep{c2pa2026} bind signed assertions about origin, ingredients, and edits to an asset. In agentic video generation, a keyframe becomes an ingredient of a clip and a clip an ingredient of the cut, so a locally regenerated shot needs its own identity and marking record.
\par
\endgroup

%% file: sections/05_world_models.tex
\section{Agentic World Modeling}
\label{sec:iwm-harness}
\label{sec:world-models}
\label{sec:video-world-models}

This chapter examines how an agent harness connects video generation to
sustained world interaction. We first compare video generation models with
interactive world models and describe their components
(Section~\ref{sec:iwm-interface}); then review semantic, state, and feedback
mechanisms for agentic world modeling (Section~\ref{sec:harness-for-iwm}),
and systems in which an engine or a program holds the world state while a
video model renders it (Section~\ref{sec:iwm-explicit-state}).
Finally, we examine interaction reliability, active testing supported by the
agent harness, and reciprocal evaluation of agents and world models
(Section~\ref{sec:iwm-evaluation}).

\subsection{From Video Generation Models to Interactive World Models}
\label{sec:iwm-overview}
\label{sec:iwm-interface}

This chapter focuses on interactive world models built around video generation models as their predictive core. Both model visual content and its temporal evolution, and can use similar diffusion or autoregressive architectures. Conventional video generation usually produces a visual sequence conditioned on text, images, or existing video, with emphasis on visual quality, condition adherence, and temporal consistency. Interactive world models use this capability for sustained interaction: they receive current observations and action conditions, predict the resulting changes, and return generated observations for subsequent interaction. The distinction lies primarily in how generation is organized and what it is used for. In an interactive world model, the consequences of an action need to remain reflected in subsequent steps, and generated observations must provide usable environmental information for the next action.

A direct path from video generation to interactive world modeling is to extend video continuation into ongoing prediction conditioned on actions. An initial image or video establishes the starting point, while a user or agent supplies control signals during execution. The model generates subsequent frames or video segments under these controls and continues prediction from the updated context. This transition requires introducing action conditions, learning the relationship between actions and visual changes~\citep{jiang2026olafworld}, and supporting continued generation. Genie~\citep{bruce2024genie}, for example, learns latent actions to turn video prediction into a generative environment with frame by frame control. GameNGen~\citep{valevski2025gamengen} predicts the next frame from past frames and player actions, demonstrating how a video diffusion model can support interactive game simulation. These systems turn generation for describing or continuing visual content into environmental evolution driven by actions.

Using the notation in Table~\ref{tab:video-notation}, let $\alpha_t$ be the
control applied at interaction step $t$ and $x_{t+1}$ the returned visual
observation. The one step case of Eq.~\eqref{eq:video-world-model} is
\begin{equation}
  \begin{aligned}
    x_{t+1}
      &\sim p_\theta\bigl(\cdot\mid\mathbf{x}_{\leq t},
        \boldsymbol{\alpha}_{\leq t},\mathbf{c}\bigr), \\
    \mathbf{x}_{\leq t+1}
      &= (\mathbf{x}_{\leq t},x_{t+1}),
      \qquad
      \boldsymbol{\alpha}_{\leq t}
      = (\boldsymbol{\alpha}_{<t},\alpha_t).
  \end{aligned}
  \label{eq:iwm-observation-update}
\end{equation}
An observation may be a frame or a returned video chunk; $t$ counts
interaction steps, rather than sampling steps or token positions. The
returned observation extends the history available for the next control.

Functionally, these world models connect current observations and state representations, action or instruction conditions, and the generation of subsequent visual observations. Current observations may be a single image, recent video, or multiple views that represent the scene at the current interaction step. Control conditions may include keystrokes, camera poses, motion trajectories, robot actions, or natural language instructions interpreted by the model or a control module. The predictive module combines these conditions with historical context to generate subsequent visual content, allowing the user or agent to assess the consequences of an action. State may be represented explicitly through objects, poses, and scene structure, or implicitly through video frames and latent features. Many video world models directly predict subsequent observations without separately exposing a complete physical state. Predicting state evolution and generating visual observations are therefore often performed by the same model.

For implementations that summarize the relevant history in an explicit or
latent predictive state $\mathbf{s}_t^{\mathrm{w}}$, transition and
observation generation can be represented as
\begin{equation}
  \begin{aligned}
    \mathbf{s}_{t+1}^{\mathrm{w}}
      &\sim p_\theta\bigl(\cdot\mid\mathbf{s}_t^{\mathrm{w}},
        \alpha_t,\mathbf{c}\bigr), \\
    x_{t+1}
      &\sim p_\theta\bigl(\cdot\mid\mathbf{s}_{t+1}^{\mathrm{w}},
        \mathbf{c}\bigr).
  \end{aligned}
  \label{eq:iwm-predictive-state}
\end{equation}
The superscript $\mathrm{w}$ distinguishes this predictive state from the
harness-managed state $s_t$ in Section~\ref{sec:background-harness}. This is
a functional decomposition, not a requirement for separate networks or an
accessible physical state.

Sustained interaction also requires retaining information beyond the current view and a finite context window. If an object has been moved, for example, that change should remain evident when the same region is revisited after leaving the field of view. Memory can supplement recent video context by storing and retrieving past scenes, spatial relations, and interaction outcomes to support subsequent prediction. WorldMem~\citep{xiao2025worldmem} stores historical frames with state information such as poses and timestamps in a memory bank, then retrieves relevant content to support scene reconstruction. WorldPlay~\citep{sun2026worldplay} reconstructs historical context so that earlier frames that remain geometrically relevant can continue to inform generation. These mechanisms extend the information available for interaction from recent video context to longer histories, supporting the maintenance of world state. Implementations may use explicit memory banks, structured state, or latent representations; a separate memory module is not required in every model.

In embodied settings, video world models further predict the consequences of robot actions in physical environments, connecting visual generation to planning, policy learning, and control. UniSim~\citep{yang2024unisim} simulates the visual changes caused by high level instructions and low level controls, providing interactive experience for robot planning and policy learning. Cosmos~\citep{nvidia2025cosmos} builds world foundation models through large scale video pretraining and adapts them through post training to physical tasks such as robotic manipulation. DreamZero~\citep{ye2026dreamzero} builds on a pretrained video diffusion model to jointly predict future video and robot actions, forming a world action model. These works illustrate different paths from predicting the visual consequences of actions to jointly modeling future observations and executable actions. Physical world models and world action models are thus closely related but not identical: the former emphasize predicting the evolution of physical environments, while the latter additionally incorporate action generation into the modeling process.

Joint prediction of future video and actions can instead be written as
\begin{equation}
  \bigl(\mathbf{x}_{t+1:t+\Delta},
    \boldsymbol{\alpha}_{t:t+\Delta-1}\bigr)
  \sim p_\theta\bigl(\cdot\mid\mathbf{x}_{\leq t},
    \boldsymbol{\alpha}_{<t},\mathbf{c}\bigr).
  \label{eq:iwm-joint-video-action}
\end{equation}
Here the future control sequence $\boldsymbol{\alpha}_{t:t+\Delta-1}$ is
an output, whereas the proposed future controls are supplied as conditions
in Eq.~\eqref{eq:video-world-model}. This separates simulation under
supplied controls from the joint generation of observations and controls.

\begin{figure}[!htb]
  \centering
  \includegraphics[width=\linewidth]{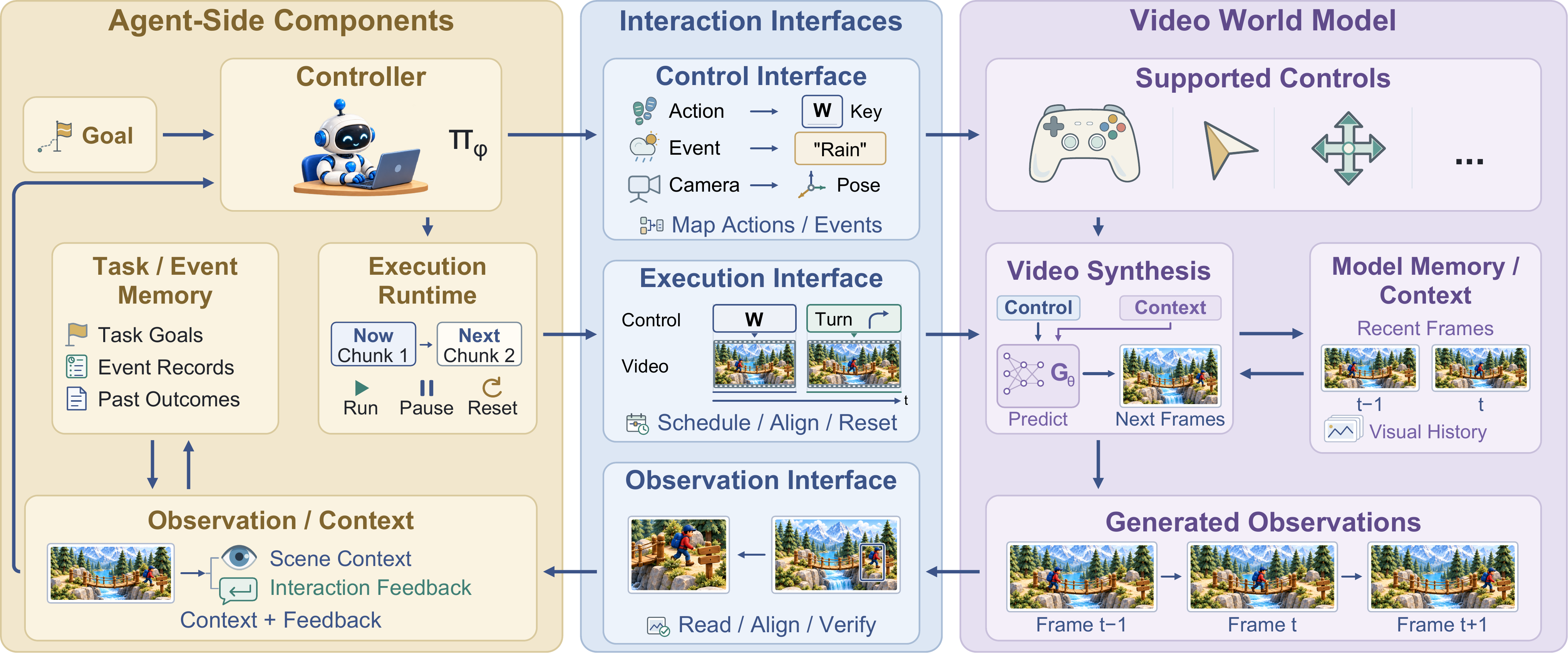}
  \caption{\textbf{From video generation to harness-mediated world interaction.}
  Conditional visual generation remains the shared core. In a video world model,
  one-shot conditioning becomes sustained control, temporal context must support
  continuing state, and output frames become observations for the next decision.
  Around this core, the harness observes, decides, translates controls, schedules
  generation, and incorporates returned feedback and relevant history.
  The central interfaces translate controls, coordinate execution, and interpret
  returned observations.
  External memory and tracking are distinct from model-side state or context.
  The diagram summarizes shared, adapted, and added components; a particular
  system need not implement every component. Scene thumbnails illustrate visual
  observations; they are not experimental results.}
  \label{fig:world-model-runtime}
\end{figure}

These capabilities provide the basis for connecting an agent harness to a world model. The world model supplies visual predictions and interaction feedback conditioned on actions, while the agent harness organizes action selection, context management, and subsequent interaction around task goals. World models based on video generation can thereby serve as environments in which agents continually observe, act, and revise their plans. Figure~\ref{fig:world-model-runtime} illustrates the shared generative core and its connection to the surrounding agent harness.

\subsection{Harness Interventions}
\label{sec:harness-for-iwm}

Beyond relaying inputs and outputs, a harness can organize the semantics, state information, and feedback on which continuing generation depends. We group these interventions into two families: semantic/event mechanisms organize how subsequent changes are expressed; state/feedback mechanisms determine which history should continue to constrain generation and how deviations are handled. Both can be analyzed through the same sequence: world-model limitation $\rightarrow$ intervention point $\rightarrow$ mechanism $\rightarrow$ evidenced system capability. The evidence for this layer is still thin: most mechanisms below are demonstrated in a single system or proposed as designs, and few studies report a gain attributable to the harness alone.

\textbf{Semantic and event interventions} address the gap between high-level intent and executable generation conditions. A request may describe a desired interaction without identifying its target, preconditions, or event order. GameGen-X~\citep{che2025gamegenx} provides a model-side precursor through instruction tuning that connects user inputs to character behavior and scene content. LingBot-World~2.0~\citep{gao2026lingbotworld2} adds a VLM that interprets the current scene and user interactions before proposing subsequent events. More generally, a harness can combine the current scene, user intent, and interaction history into event conditions with explicit objects and temporal relations. This changes the conditions received by the generator; rewriting a prompt does not itself update the model's dynamics.

LingBot-World~2.0~\citep{gao2026lingbotworld2} provides a concrete example through its Director--Pilot framework. A VLM interprets the scene and user interaction and proposes events, while the video generator realizes subsequent visual changes. The report distinguishes character-level semantic interaction, object interaction assisted by SAM-based tracking across video chunks, and textual world interventions that modify the environment or introduce entities. These mechanisms connect high-level intent to conditions for sustained generation. The interaction demonstrations support this system design; an isolated quantitative gain attributable to the harness is not reported.

The corresponding questions concern whether control becomes more explicit and reliable: can a target be specified consistently, does an event occur at an appropriate time, do its effects respect preceding interactions, and are actor actions kept distinct from world interventions? A plausible VLM-generated event description is evidence about condition organization. Whether the video model realizes the intended transition must be judged from the resulting interaction. Semantic proposal quality and execution reliability are therefore separate properties.

\textbf{State and feedback interventions} address the loss of historical constraints during long-horizon generation. WorldMem~\citep{xiao2025worldmem} motivates memory through the need to reconstruct scenes after long temporal or viewpoint gaps; WorldPlay~\citep{sun2026worldplay} addresses the attenuation of geometrically important historical context. The current frame may omit an off-screen object, an earlier scene change, or an unfinished event. At the harness level, a possible design maintains object records, scene summaries, and event logs; retrieves history relevant to the current interaction; and returns it through supported reference-frame, text, or state interfaces. Memory selects and transmits constraints. A stored record provides effective state support only when the model can consume it and its subsequent generation is affected.

External state management should be distinguished from model-side memory. WorldMem~\citep{xiao2025worldmem} stores historical frames with poses and timestamps, retrieves relevant memory units, and incorporates them into prediction through state-aware attention. WorldPlay~\citep{sun2026worldplay} uses Reconstituted Context Memory to combine recent temporal context with spatially relevant historical video chunks, with temporal reframing preserving the influence of distant information. These are memory designs integrated with the generator, not generic external agent logs. At the harness level, tracking associates targets across observations, event logs record completed interactions, and retrieval selects information to return to the generator. These mechanisms may complement model-side memory, but cannot replace the model's use of that information. LingBot-World~2.0~\citep{gao2026lingbotworld2} likewise notes that long-term visual stability is not equivalent to true memory when revisiting regions outside the retained context.

Feedback connects stored state to the next intervention. EVA~\citep{chi2025eva} supplies a concrete prediction-side example: its Reflection of Generation connects intermediate VLM reasoning with video prediction. Reflection about a predicted trajectory should nevertheless be distinguished from selecting and executing an action in an online environment. More generally, a harness can compare new frames with prior records, executed controls, and expected events to identify lost object references, unrealized events, or overwritten history. Depending on the available interfaces, it may retrieve additional history, narrow the next control conditions, request further observations, or re-execute from a recoverable state. These are closed-loop design options, not a uniform set of features already implemented by every surveyed system. Their benefit should appear as fewer subsequent deviations, more reliable recovery, or more persistent state, rather than merely an explanation produced by a checker.

Table~\ref{tab:world-model-mechanisms} compares representative video world models through their video backbones, control interfaces, and external harness roles. Interaction, memory, and evaluation roles describe the listed runtime configurations; model-internal memory and training-only collection are distinguished from external harnesses. Active testing and reciprocal evaluation are discussed in Section~\ref{sec:iwm-evaluation}.

\input{tables/world_model_mechanisms}

The level at which an improvement occurs matters. If model weights remain fixed while event organization, retrieval, or feedback correction improves interaction success, the evidence supports better inference-time system performance. SCOPE~\citep{jiang2026scope} makes this distinction explicit by adapting typed external controls through bounded changes supported by development evidence while keeping the video world model frozen. A claim that the world model itself improves must identify the updated parameters or persistent components and evaluate the resulting capability under matched harnesses, controls, and resource budgets. VLAW~\citep{guo2026vlaw}, by contrast, fine-tunes the world model on real policy rollouts. Better performance with more informative conditions and more reliable state transitions learned through model updates are distinct contributions.

\subsection{Explicit World State and Generative Rendering}
\label{sec:iwm-explicit-state}

\begin{figure}[!htb]
  \centering
  \includegraphics[width=\linewidth]{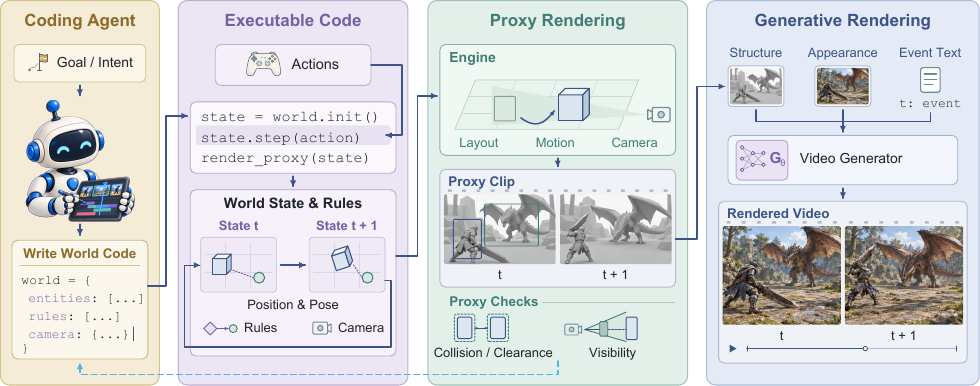}
  \caption{\textbf{Explicit world state and generative rendering.}
  A coding agent authors or revises a world program that maintains state and
  applies interaction rules. Actions update program state, which an engine
  converts into proxy frames for the video generator. An appearance anchor
  and event text complement these structural conditions; proxy inspection can
  guide code revision (dashed). This division of labor is illustrated by
  EdenLive~\citep{vast2026edenlive}, SolarRenderer~\citep{fang2026solarrenderer},
  and Code World Model~\citep{chen2026cwm}. Game scenes are conceptual
  illustrations, not outputs of these systems.}
  \label{fig:explicit-state-rendering}
\end{figure}

The interventions above leave world dynamics to the video model and improve the conditions it receives, so their effect depends on whether the model uses what it is given. A different division of labor removes state and dynamics from the generator. A game engine, a simulator, or a program written by a coding agent holds the world state and applies its rules, and a video model only turns the resulting coarse structure into realistic frames. Matrix-Game 4.0~\citep{zhang2026matrixgame4} calls this design an executable game harness: a lightweight game engine maintains the world state and executes interaction rules, and an autoregressive video model renders the low-poly frames the engine produces. EdenLive~\citep{vast2026edenlive} describes the same split, in which the upstream program executes the rules and maintains the world state, whereas the video model renders the resulting reference video. The generator is then a renderer rather than a simulator, and the program that holds the state acts as an environment whose state the harness can read and edit directly instead of estimating it from frames. Figure~\ref{fig:explicit-state-rendering} summarizes this division of labor.

\textbf{What the structure fixes.} The exported representation decides how much of each frame the engine specifies and how much the renderer must supply. Sparse layouts fix where things are but not what they look like: StreamRender-H3~\citep{nvidia2026streamrender} conditions on flat-color semantic maps from a browser game engine, and the Programmable World Model~\citep{huang2026programmable} on 3D oriented bounding boxes compiled into pixel-aligned conditions. Geometric proxies add shape, as in the white-box frames that Magpie~\citep{zhan2026magpie} receives from Unreal Engine, the low-poly frames of Matrix-Game 4.0, the coarse 3D simulations behind Coarse-to-Real~\citep{gomeznogales2026coarsetoreal}, and the depth or surface normals used by AgentGarten~\citep{chi2026agentgarten}. G-buffers also carry material channels and leave mainly lighting and fine appearance to the model~\citep{huang2026worldrenderer,zeng2026rgbx,liang2025diffusionrenderer}. Most of these renderers are trained for the representation their engine exports, so a new engine or a new kind of proxy requires a new renderer. Two exceptions loosen this coupling: Code Video Model~\citep{liang2026codevideomodel} conditions a frozen generator on proxy renders without training, and SolarRenderer~\citep{fang2026solarrenderer} passes untyped structural videos through the reference pathway of a frozen backbone with a single control LoRA and evaluates on representations absent from its training data.

\textbf{Where the agent acts.} Placing an agent upstream of the renderer raises the question of when it is consulted. In Code World Model~\citep{chen2026cwm}, a coding agent reasons about events and writes the code that maintains the world state, after which deterministic per-step updates run without invoking the agent; its reasoning is spent on authoring the rules, not on producing every frame. The Programmable World Model instead translates natural-language instructions into programs for a lightweight engine, and Matrix-Game 4.0 uses a multimodal coding agent to turn a player's ideas into game code, so that scenes and gameplay can be created and modified in natural language. AgentGarten, EdenLive, and Code Video Model use an agent to build the world or scene before rendering, and the World Narrative Model~\citep{chen2026wnm} constructs an editable physical world with layouts, motion, cameras, and lighting before synthesis rather than during continuous interaction. In all of these designs the interfaces of Section~\ref{sec:iwm-interface} change form. An intervention becomes an edit to program state or rules rather than a prompt or an action code, the observation interface can read state from the program instead of estimating it from frames, and objects outside the view persist regardless of how much context the generator retains.

\textbf{What the agent produces.} An agent upstream of the renderer produces three inputs rather than one. The first is a world program that fixes geometry, motion, and camera. Code World Model starts from existing game-engine code and composes reusable proxy primitives whose position, pose, layout, camera, and state markers are addressable parameters. The Programmable World Model recovers entities from a reference image with a 3D detector and writes their states, relations, interaction rules, event triggers, and objectives. EdenLive builds standalone Three.js games in stages that define the mechanics, the static scene, interactive behavior, and the interface, and Code Video Model writes one Three.js program per shot, computing mathematically defined dynamics such as a double pendulum directly in code. The second input is an appearance anchor. Instead of describing appearance in code, most systems have an image model redraw the first proxy frame into a realistic first frame, as in Code World Model, Code Video Model, EdenLive, and StreamRender-H3. The third is text that places events in time: the agent of Code World Model writes timed action scripts, the coding agent of Matrix-Game 4.0 writes dynamic scene captions that condition the renderer together with memory references, the engine of EdenLive emits a caption for the salient events of each chunk, and an LLM agent annotates the training clips of StreamRender-H3 with timestamped narration. Structure, appearance, and event timing thus come from code, an image model, and language, and the renderer has to reconcile them.

\textbf{Checking the program before rendering.} Because the world is a program, many errors can be caught before any frame is synthesized. EdenLive runs each game with real input and checks the resulting state, repairing failed checks within a fixed budget. AgentGarten evaluates test rollouts from several probe cameras, queries the engine for contact, collision clearance, and occlusion, and repairs faulty poses or unsupported structures. The World Narrative Model checks script syntax, runtime exceptions, collisions, and boundary conformance, and compares rendered depth and silhouettes against references, while Code Video Model renders keyframes into a contact sheet and revises the program against the scene description. These checks inspect exact state through simulation or draft renders instead of judging generated video. They do not cover the final step, in which the renderer turns verified structure into pixels.

\textbf{Speed and supervision.} With the engine holding the dynamics, the video model sets the frame rate, and the systems that report real-time rendering reach it through few-step distillation. Magpie reports a compute-side rate of about 32 frames per second on one H100, AgentGarten over 35 frames per second on one H100 with four denoising steps, and EdenLive at least 24 frames per second on a consumer GPU; AlayaRenderer-Flash~\citep{alaya2026renderer} raises a G-buffer renderer from 0.56 to 31.54 frames per second, whereas Streaming Code World Model~\citep{liao2026streamingcwm} describes its four-step version as not yet real time. Frame rate does not determine responsiveness. Magpie reports an end-to-end response of about 1.55 seconds alongside its frame rate, whereas Matrix-Game 4.0, which renders four-frame chunks with one or two denoising steps, reports 60 frames per second at 480p on one RTX 5090 with 112 milliseconds from key press to display. Training pairs involve a trade-off between alignment and coverage. Recordings from an engine align every structural frame with its target, as in Magpie's 300 hours of interactive Unreal Engine play, the aligned low-poly and high-fidelity video pairs of Matrix-Game 4.0, and the synchronized RGB and G-buffer frames that Generative World Renderer captures from commercial games, but they cover only the engines they come from. Structure estimated from real video, as for part of AgentGarten's depth and normal conditions or DiffusionRenderer's G-buffers, covers more content but carries estimator errors and can differ from what an engine exports. Proxy2World~\citep{xu2026proxy2world} avoids authored proxy--video pairs by learning from posed RGB-D video, and Code Video Model conditions a frozen generator on proxy renders without additional training. Agent-built worlds also begin to supply data: EdenLive synthesizes coarse-to-fine training pairs from agent-built proxy videos, and the evaluation sets of the Programmable World Model, Code Video Model, and SolarRenderer are built from worlds or programs written by agents.

\textbf{What explicit state does not solve.} Moving dynamics into a program relocates several difficulties. The renderer still invents whatever the structure leaves open, such as materials, lighting, cloth, and fluids, and nothing in the geometry keeps that invented appearance the same when a region is revisited. Coarse structure can be ambiguous, as Magpie notes for the depth of white-box frames. The output can also depart from the structure it was given, which makes adherence a quantity to measure; SolarRenderer, for example, compares point tracks and edges of the rendered video against its structural input. Because the intended state is available in the program, the harness has a reference against which adherence can be estimated at every rendering step, which a generator that keeps its state internally does not provide. The world, finally, can express only what the program represents, and an event outside its rules requires new code or a return to learned dynamics. How to combine programmed and learned dynamics within one world remains open (Section~\ref{sec:challenges_hybrid_control}).

\subsection{Evaluation in Agentic World Modeling}
\label{sec:iwm-evaluation}
\label{sec:iwm-coevolution}

\textbf{From video quality to interaction reliability.} Video world models inherit the requirements of video generation for visual quality, semantic alignment, and temporal coherence, but interaction additionally requires generated states to support subsequent actions. A visually plausible video may still ignore input actions, forget prior states, or produce consequences that violate environmental rules. Evaluation must therefore extend beyond video clips under given conditions to state evolution under successive actions and the support these predictions provide for agent decisions. Following the evaluation framework of agentic world modeling~\citep{chu2026agenticworldmodel}, we organize these requirements along three dimensions: long horizon coherence, intervention sensitivity, and constraint consistency. Long horizon coherence concerns whether states and the consequences of past actions are maintained during sustained interaction. Intervention sensitivity asks whether changing an action induces a directionally meaningful change in the outcome. Constraint consistency examines whether generated trajectories obey environmental rules. Together, these requirements shift attention from convincing images to reliable interaction outcomes.

In video world models, exploration, revisits, and controlled interactions make these requirements concrete. Orbia~\citep{lu2026orbia}, for example, uses controlled exploration and revisits to separately assess input preservation, the persistence of generated content, and whether different views depict a single 3D scene. Such tests reveal failures that short video clips may conceal: a model can generate plausible images from the current viewpoint yet alter objects, layouts, or spatial relations when returning to a previously visited region. Intervention sensitivity can be tested by changing an action at a single step under matched initial conditions and checking whether subsequent states respond according to its intended semantics. Constraint checks further assess whether these responses respect collision constraints, support relations, object properties, or task rules. Different images under different actions establish responsiveness, but correctness must also be judged against the expected consequences.

\textbf{Active evaluation supported by the agent harness.} An agent harness can turn these requirements into goal directed active tests. With the world model held fixed, an agent selects actions according to the test objective, inspects returned observations, and adjusts subsequent controls to reach states that expose model failures. WorldScore~\citep{duan2025worldscore} evaluates controllability, quality, and dynamics through sequences of next scene generation tasks, providing a unified basis for evaluating world generation. PlayWorld~\citep{ding2026playworld} introduces multimodal Agent Players that pursue long horizon test objectives through sustained interaction, covering geometric consistency, interaction fidelity, and state evolution within and beyond the current field of view. Its motivation is that different models may require different action sequences to accomplish the same test objective. Adjusting actions based on observations, rather than fixing a control sequence in advance, allows the test to accommodate each model's interface and actual generated outcomes.

In this process, the agent harness organizes test objectives, action execution, historical records, and outcome verification. Evaluation records should retain requested actions, executed controls, returned observations, and verification evidence to distinguish world state errors from interface limitations or a tester's failure to perform the intended operation. Comparisons across models require matched test objectives, controller configurations, and resource budgets, with necessary interface adaptations documented. The tester's control authority must also be constrained. For example, external world edits must not directly manufacture outcomes that should arise through interaction with the model. If development feedback is used to adjust an external control policy, final evaluation also requires independent held out tasks. SCOPE~\citep{jiang2026scope} provides a concrete example by separating development feedback from held out assessment and freezing the selected control policy before the latter begins.

\textbf{Reciprocal evaluation of agents and world models.} The tests above treat the world model as the evaluation target, while the world model can conversely provide an environment for testing agents. This reciprocal role is already evident in embodied settings: agents probe the reliability of generated worlds through interaction, while generated worlds test agents' perception, planning, and action capabilities across different scenes, objectives, and interaction conditions. SIMA~2~\citep{sima2025sima2} conducts qualitative evaluations in held out environments generated by Genie~3, examining whether an agent trained in game environments can generalize to newly generated naturalistic and urban scenes. This illustrates the use of world models as environments for agent evaluation.

Such reciprocal evaluation must distinguish world model reliability from agent capability. A weaker agent may fail to reach states that would reveal model defects, while incorrect state transitions may make ineffective actions appear successful. Task success in generated worlds therefore needs to be related to outcomes in independent environments. WorldArena~\citep{shang2026worldarena} tests policies of different capabilities in world models controlled through actions and compares their task success rates with outcomes in the RoboTwin simulator, assessing the validity of world models as policy evaluators.

This motivates paired comparisons: hold the agent and agent harness fixed while comparing different world models, then compare different agents under matched world model conditions. Reports should separately present state transition fidelity, task completion rates, and agreement in policy rankings, with independent simulators or real environments used for verification wherever possible. Such evaluation addresses two questions: whether generated worlds withstand interactive testing by agents, and whether the agent performance measured within them reflects actual task capability.

%% file: tables/world_model_mechanisms.tex
\begin{table}[!htb]
\centering
\SurveyTableSetup
\definecolor{wmHeader}{HTML}{E5EFFB}
\definecolor{wmLatent}{HTML}{70508C}
\definecolor{wmAction}{HTML}{A34559}
\definecolor{wmPose}{HTML}{326F99}
\definecolor{wmSemantic}{HTML}{2D7762}
\definecolor{wmText}{HTML}{97701F}
\definecolor{wmInteraction}{HTML}{B35B28}
\definecolor{wmMemory}{HTML}{505A68}
\definecolor{wmEvaluation}{HTML}{963D79}
\definecolor{wmAbsent}{HTML}{B92338}
\DeclareRobustCommand{\wmBadge}[2]{%
  \begingroup
  \setlength{\fboxsep}{0pt}%
  \colorbox{#1}{%
    \makebox[11.5pt][c]{\rule[-2.4pt]{0pt}{11.5pt}%
      \SurveyTableFont\bfseries\fontsize{9}{10}\selectfont
      \color{white}#2}}%
  \endgroup}
\newcommand{\wmNoHarness}{%
  \makebox[11.5pt][c]{\tikz[baseline=-2.5pt,x=0.9pt,y=0.9pt]{%
    \draw[wmAbsent,line width=1.55pt,line cap=round]
      (-2.1,3.4) .. controls (-1.2,1.8) and (0.4,-1.5) .. (2.2,-3.3)
      (3.1,4.2) .. controls (1.2,2.0) and (-1.6,-1.2) .. (-3,-4);}}}
\newcommand{\wmHeaderText}[1]{{\bfseries #1}}
\newcommand{\wmRoles}[1]{\begingroup
  \renewcommand{\arraystretch}{1}%
  \begin{tabular}[c]{@{}c@{}}#1\end{tabular}\endgroup}
\newcommand{\wmBackbone}[1]{{\fontsize{8.3}{10.5}\selectfont #1}}
\captionsetup{labelfont={bf,color=black},
  textfont={rm,color=black},labelsep=colon,skip=10pt}
\caption[Video world models, backbones, control interfaces, and the agent harness.]{%
{\bfseries Video world models, backbones, control interfaces, and the agent harness.}
The table compares video backbones, control interfaces
(\wmBadge{wmLatent}{L} latent actions, \wmBadge{wmAction}{A} explicit actions,
\wmBadge{wmPose}{P} camera poses/trajectories, \wmBadge{wmSemantic}{S} semantic/object interactions,
\wmBadge{wmText}{T} text prompts/events), and external runtime harness roles
(\wmBadge{wmInteraction}{I} interaction, \wmBadge{wmMemory}{M} memory,
\wmBadge{wmEvaluation}{E} evaluation).
Red crosses indicate no external harness in the listed runtime configuration; NR means not reported.
Model-internal memory and training-only collectors are excluded.
LingBot-World 2.0's memory is limited to short-term object tracking across chunks, not persistent world memory.}
\label{tab:world-model-mechanisms}
\fontsize{8.8}{10.8}\selectfont
\setlength{\tabcolsep}{5pt}
\renewcommand{\arraystretch}{2.1}
\setlength{\extrarowheight}{0pt}
\begin{tabular}{>{\raggedright\arraybackslash}m{110pt}
  >{\centering\arraybackslash}m{24pt}
  >{\raggedright\arraybackslash}m{112pt}
  >{\centering\arraybackslash}m{92pt}
  >{\centering\arraybackslash}m{\dimexpr\linewidth-388pt\relax}}
\toprule
\wmHeaderText{World Model} & \wmHeaderText{Year}
& \wmHeaderText{Video Backbone} & \wmHeaderText{Control Interface}
& \wmHeaderText{Agent Harness}\\
\midrule
Genie~\citep{bruce2024genie} & 2024
& \wmBackbone{ST-transformer / MaskGIT}
& \wmBadge{wmLatent}{L}
& \wmNoHarness\\
GameNGen~\citep{valevski2025gamengen} & 2024
& \wmBackbone{Stable Diffusion 1.4}
& \wmBadge{wmAction}{A}
& \wmNoHarness\\
WorldMem~\citep{xiao2025worldmem} & 2025
& \wmBackbone{Conditional DiT\newline(Diffusion Forcing)}
& \wmBadge{wmAction}{A}\,\wmBadge{wmPose}{P}
& \wmNoHarness\\[3pt]
Matrix-Game 2.0~\citep{he2025matrixgame2} & 2025
& \wmBackbone{SkyReels-V2-I2V-1.3B}
& \wmBadge{wmAction}{A}
& \wmNoHarness\\
ABot-World-0~\citep{jiang2026abotworld} & 2026
& \wmBackbone{Wan2.2}
& \wmBadge{wmAction}{A}
& \wmNoHarness\\
AdaWorld~\citep{gao2025adaworld} & 2025
& \wmBackbone{Stable Video Diffusion}
& \wmBadge{wmLatent}{L}\,\wmBadge{wmAction}{A}
& \wmBadge{wmInteraction}{I}
\\
Genie 3~\citep{parkerholder2025genie3} & 2025
& \wmBackbone{NR}
& \wmBadge{wmAction}{A}\,\wmBadge{wmText}{T}
& \wmRoles{\wmBadge{wmInteraction}{I}
  \\[2pt]
  \wmBadge{wmEvaluation}{E}
  }\\[6pt]
LingBot-World~\citep{gao2026lingbotworld} & 2026
& \wmBackbone{Wan2.2}
& \wmBadge{wmAction}{A}\,\wmBadge{wmPose}{P}\,\wmBadge{wmText}{T}
& \wmRoles{\wmBadge{wmInteraction}{I}
  \\[2pt]
  \wmBadge{wmEvaluation}{E}
  }\\[6pt]
LingBot-World 2.0~\citep{gao2026lingbotworld2} & 2026
& \wmBackbone{Causal video DiT}
& \wmBadge{wmAction}{A}\,\wmBadge{wmPose}{P}\,\wmBadge{wmSemantic}{S}\,\wmBadge{wmText}{T}
& \wmRoles{\wmBadge{wmInteraction}{I}\,\wmBadge{wmMemory}{M}\\[2pt]
  \wmBadge{wmEvaluation}{E}
  }\\[6pt]
WorldPlay~\citep{sun2026worldplay} & 2025
& \wmBackbone{HunyuanVideo-1.5}
& \wmBadge{wmAction}{A}\,\wmBadge{wmPose}{P}
& \wmBadge{wmEvaluation}{E}
\\
\bottomrule
\end{tabular}
\end{table}

%% file: sections/03_model_development.tex
\section{Toward Recursive Self-Improvement in Video Generation}
\label{sec:model-development}

Under the definition of Section~\ref{sec:framework}, recursive
self-improvement (RSI) requires a validated change to the development
procedure $\rho_k$ that makes later rounds more effective.
The video generation literature provides many mechanisms for improving
generators and the agent systems that use them, together with emerging
examples of persistent changes to parts of the development process.
However, the systems surveyed here do not yet demonstrate that a revised
development procedure improves subsequent rounds of generator development.

We organize the discussion around three perspectives:
(\emph{i}) Foundation Model Improvement examines data curation and model
training;
(\emph{ii}) Agent Harness Improvement examines the planning, tool use,
memory, and reusable procedures that organize generation and editing;
and (\emph{iii}) Evaluation examines the feedback used to propose updates
and the evidence needed to validate their effects.
Most existing methods update the development state $\xi_k$, including
model parameters and retained harness components, under a fixed procedure.
Reusing an improved component to produce further updates can create a
feedback loop, but establishing RSI additionally requires evidence that
the procedure producing those updates has itself improved.
We examine this distinction within each perspective.

\definecolor{DevelopmentBlue}{HTML}{2B6CB0}
\definecolor{DevelopmentBlueLight}{HTML}{E8F1FB}
\definecolor{DevelopmentGold}{HTML}{B7791F}
\definecolor{DevelopmentGoldLight}{HTML}{FFF6DF}
\tikzset{
  development stage/.style={
    draw=DevelopmentBlue,
    fill=DevelopmentBlueLight,
    rounded corners=2pt,
    align=center,
    line width=0.7pt,
    inner sep=6pt,
    font=\small
  },
  development harness/.style={
    draw=DevelopmentGold,
    fill=DevelopmentGoldLight,
    rounded corners=3pt,
    align=center,
    line width=0.7pt,
    inner sep=6pt,
    font=\small
  },
  development flow/.style={
    -{Latex[length=2mm]},
    thick,
    DevelopmentBlue
  },
  development control/.style={
    {Latex[length=1.6mm]}-{Latex[length=1.6mm]},
    dashed,
    thick,
    DevelopmentGold
  },
  development potential/.style={
    -{Latex[length=2mm]},
    densely dotted,
    thick,
    black!55
  }
}

\subsection{Foundation Model Improvement}
\label{sec:development-foundation-model}
\label{sec:development-data}

This subsection reviews approaches to improving video generation
foundation models through two coupled components of model
development: \emph{Data Curation} and \emph{Model Training}.
As summarized in Figure~\ref{fig:foundation-model-taxonomy},
data curation determines the training data and supervision
available to the model, while model training determines how
these signals are used for optimization.
We include established methods to characterize the interventions
available to an improving system, and examine emerging agentic
approaches according to which development decisions they
automate or adapt.
These decisions include selecting data, constructing supervision,
refining data pipelines, and using evaluative feedback for
model optimization.
The distinction relevant to RSI is whether these interventions only
change the data or model being developed, or also improve the procedures
used to construct and validate later updates.

\input{tables/foundation_model}

\subsubsection{Data Curation}

Data curation improves the data and supervision available for
training video generation models.
We organize existing approaches into three categories:
(\emph{i}) Data Acquisition \& Retrieval, which collects task-relevant data;
(\emph{ii}) Annotation \& Synthesis, which constructs training
supervision through annotation and synthetic data generation;
and (\emph{iii}) Data Refinement, which improves data quality
through filtering and correction.

\textbf{Data Acquisition \& Retrieval.}
Data acquisition and retrieval provide training samples that meet
the varying requirements of video generation models.
WebVid~\citep{bain2021frozen} collects millions of video--text pairs
from the web, providing a widely used data source for video generation.
HD-VG-130M~\citep{wang2023videofactory} constructs a larger collection
of high-definition videos for text-to-video generation.
VideoUFO~\citep{wang2025videoufo} identifies topics of interest from
real user prompts and retrieves corresponding videos from YouTube
to improve coverage of content requested by users.
Phantom-Data~\citep{chen2025phantomdata} retrieves reference images
of the same subject across different visual contexts to construct
cross-pair training data, reducing the copy-paste problem in
subject-to-video generation.
VC-Agent~\citep{zhang2025vcagent} uses feedback from users' accepted
samples and rejection reasons to update its filtering criteria
for subsequent video retrieval.
The resulting data are used to fine-tune
CogVideoX~\citep{yang2024cogvideox}.
These approaches connect acquisition to user demand and feedback.
Related exploration-driven environment learning uses uncertainty to
select which experience to acquire~\citep{savov2025autoexplore},
illustrating how observed capability gaps can guide data collection.

\textbf{Annotation \& Synthesis.}
Annotation and synthesis provide supervision for video generation
models by annotating existing videos and generating synthetic
training data.
Panda-70M~\citep{chen2024panda} generates candidate captions using
multiple cross-modality teachers and selects the best through
a trained video--text retrieval model.
ShareGPT4Video~\citep{chen2024sharegpt4video} compares consecutive
keyframes to describe changes in actions and camera motion,
producing detailed video captions for model training.
Vidu S2-Avatar~\citep{zhang2026vidus2} employs multiple agents
to generate and verify temporally ordered captions aligned
with individual events.
For synthetic data generation,
PhysInOne~\citep{zhou2026physinone} synthesizes videos covering
diverse physical phenomena to improve the physical plausibility
of video generation models.
Follow-Your-Instruction~\citep{feng2025followyourinstruction}
uses MLLM agents to construct 3D scenes, verify layouts across
views, and plan temporal evolution, producing synthetic data
for video model fine-tuning.

Agent-built worlds provide another source of synthetic supervision
(Section~\ref{sec:iwm-explicit-state}).
Their programs expose scene structure and intended state, providing
references for annotation and opportunities to target observed failures.
EdenLive~\citep{vast2026edenlive} synthesizes training pairs from
agent-built proxy videos.
When generated samples are repeatedly reused for training, however,
errors and omissions can accumulate across rounds.
Studies of recursive training show losses of distributional coverage,
quality, or diversity under the settings they examine
\citep{shumailov2023curse,alemohammad2024mad}.
The provenance and mixture of synthetic and real data therefore matter
when evaluating whether later rounds benefit from generated supervision.

\textbf{Data Refinement.}
Data refinement improves the quality of video training data
through filtering, correction, and quality assessment.
OpenVid-1M~\citep{nan2024openvid} improves video training data
through quality filtering and recaptioning.
Koala-36M~\citep{wang2025koala} introduces a Video Training
Suitability Score (VTSS) that combines multiple quality indicators
to select videos suitable for model training.
For interactive world models,
SolarWM~\citep{huang2026solarwm} converts 1.43 million clips
from ten datasets into a unified record of observations,
metric camera geometry, captions, quality metadata,
selection decisions, and provenance, and separates source
processing from mixture construction, so that a new training
mixture can be defined by changing filtering criteria or sampling
weights without reprocessing the sources.
Ditto~\citep{bai2025ditto} uses an agent to generate editing
instructions and filter synthetic video samples, together
with denoising enhancement to improve data quality.
JAVEdit~\citep{chen2026javedit} uses agents to diagnose failures
in data generation pipelines, modify code, prompt templates,
or pipeline settings, and retain validated fixes for subsequent tasks.

\subsubsection{Model Training}

Model training develops and improves video generation capabilities
through different learning and optimization strategies.
We organize existing approaches into four categories:
(\emph{i}) Pre-training, which establishes general visual and temporal
generation capabilities through large-scale training;
(\emph{ii}) Supervised Fine-Tuning (SFT), which adapts pretrained models
using task-specific supervision;
(\emph{iii}) Reinforcement Learning \& Reward Modeling, which leverages
evaluative feedback to guide model optimization;
and (\emph{iv}) Distillation, which improves generation efficiency
through knowledge transfer from teacher models.

\textbf{Pre-training.}
Pre-training establishes general visual and temporal generation
capabilities through large-scale training, where training efficiency
and data utilization are important considerations.
Make-A-Video~\citep{singer2022make} leverages text-to-image priors
and unpaired videos to learn video generation without paired
text--video data.
Stable Video Diffusion~\citep{blattmann2023stablevideo} adopts
a staged training approach, starting from an image model and
proceeding through video pre-training and high-quality fine-tuning.
CogVideoX~\citep{yang2024cogvideox} accommodates videos of varying
lengths and resolutions through progressive training and
diffusion timestep sampling.
HunyuanVideo~\citep{kong2024hunyuanvideo} uses scaling experiments
to guide model size and training token allocation.
Wan~\citep{wan2025wan} employs progressive image--video training
from low to high resolutions to improve the efficiency and
stability of large-scale video pre-training.

\textbf{Supervised Fine-Tuning.}
Supervised fine-tuning (SFT) adapts pretrained video generation
models using supervised data to improve video quality and
learn specific motion or control capabilities.
Tune-A-Video~\citep{wu2023tune} adapts a pretrained image diffusion
model for video generation using a single text--video pair.
AnimateDiff~\citep{guo2024animatediff} trains a reusable motion
module that can be integrated into personalized image
diffusion models.
MotionDirector~\citep{zhao2023motiondirector} separates appearance
and motion learning to customize motion from reference videos
without copying their appearance.
VideoCrafter2~\citep{chen2024videocrafter2} selectively fine-tunes
spatial modules with high-quality images to improve visual
quality without degrading motion generation.
VACE~\citep{vace2025} introduces a unified conditioning framework
for adapting video models to diverse creation and editing tasks.
DreamRunner~\citep{wang2024dreamrunner} uses LLM planning to
identify required motions, retrieves relevant reference videos,
and performs test-time fine-tuning to learn corresponding
motion priors.
MotionAgent~\citep{motionagent2025} combines motion fields
constructed by an agent with a fine-tuned optical flow adapter
to enable precise motion control.

\textbf{Reinforcement Learning \& Reward Modeling.}
Reinforcement learning (RL) improves video generation models
by optimizing their parameters using reward signals.
Reward modeling plays an important role in this process
by learning to assess generated videos and provide rewards
for optimization.
Constructing reliable rewards requires considering multiple
aspects of video quality, as well as the visual evidence
and scoring criteria used for assessment.
VideoScore~\citep{he2024videoscore} trains a video quality
assessment model on human annotations of visual quality,
temporal consistency, and other evaluation dimensions.
VideoReward~\citep{liu2025videoalign} trains a video reward
model on multidimensional human preferences.
RewardDance~\citep{wu2025rewarddance} shifts from regressive
to generative reward modeling, aligning reward prediction
with the VLM's next-token prediction objective, enabling
more effective scaling across model size and input context.
VR-Thinker~\citep{wang2025vrthinker} actively selects additional
video frames during reasoning to gather visual evidence.
RewardVerse~\citep{tang2026rewardverse} learns to generate
task-specific rubrics for assessing video quality.
The latter two improve reward modeling by refining the
visual evidence considered and the criteria used to assess
generated videos.

Beyond constructing reliable reward signals, another important
direction is how to leverage them to optimize video generation
models.
InstructVideo~\citep{yuan2024instructvideo} uses image reward
models to fine-tune video diffusion models, reformulating
reward optimization as editing to reduce training costs.
VideoDPO~\citep{liu2025videodpo} constructs preference pairs
from composite scores for offline preference optimization.
DanceGRPO~\citep{xue2025dancegrpo} incorporates rewards from
VideoReward into GRPO to optimize the video generator.
SPIRAL~\citep{yang2026spiral} incorporates agent feedback into
multi-step action video generation: a PlanAgent decomposes
action goals, a CriticAgent inspects generated video clips
and triggers revisions, and the video generator is further
optimized through GRPO using rewards from the critic.
Dual-IPO~\citep{yang2026dualipo} alternates improvements
to the reward model and the generator.
Such methods allow evaluators to shape the data and objectives
used for generator optimization, making the reliability of those
evaluators a development concern as well as an evaluation concern
(Section~\ref{sec:rsi-evaluation}).

\textbf{Distillation.}
Distillation transfers knowledge from teacher models to student
models to improve video generation efficiency without substantially
sacrificing generation quality.
T2V-Turbo~\citep{li2024t2vturbo,li2025t2vturbov2} incorporates
differentiable reward feedback into consistency distillation
to improve video quality with fewer sampling steps.
Motion Consistency Model~\citep{zhai2024motion} separates motion
distillation from appearance enhancement, using high-quality
image data to improve frame appearance without degrading
temporal dynamics.
Transition Matching Distillation (TMD)~\citep{nie2026transition}
approximates multi-step denoising trajectories using a small
number of probability transitions modeled by conditional flows.
PDD~\citep{shaul2026pdd} accelerates generation by predicting
multiple denoising steps within each model evaluation,
improving sampling efficiency and generation diversity.
AnyFlow~\citep{gu2026anyflow} learns flow map transitions over
arbitrary time intervals, allowing distilled models to improve
generation quality as more sampling steps become available.

Distillation can also be used to train autoregressive video
diffusion models, which generate video chunks sequentially
through diffusion conditioned on previously generated content.
CausVid~\citep{yin2025causvid} transfers knowledge from a
multi-step bidirectional diffusion teacher to a few-step
autoregressive student for faster video generation.
Building on CausVid, Self Forcing~\citep{huang2025selfforcing}
further trains the student on its own generated histories
through autoregressive rollouts and video-level distribution
matching to reduce the training--inference gap.
Causal Forcing~\citep{zhu2026causal} addresses the architectural
mismatch between bidirectional teachers and autoregressive
students by using an autoregressive teacher to provide generation
trajectories for student initialization, followed by distribution
matching with the bidirectional teacher.

Across data curation and model training, improved generators can
provide supervision, improved evaluators can change which samples are
selected, and production feedback can guide acquisition.
These mechanisms connect model use to further development.
They can nevertheless operate under fixed rules for collecting data,
optimizing parameters, and accepting candidates.

JAVEdit provides a more direct example of a persistent change to part of
the development procedure by retaining validated pipeline fixes.
The remaining question is whether such changes make later development
rounds more effective, including their effects on training and candidate
validation.
This requires measuring the quality and cost of subsequent updates,
beyond the quality of the data or model produced in the current round
(Section~\ref{sec:rsi-evaluation}).

\subsection{Agent Harness Improvement}
\label{sec:development-scaffold}

The agent harness (Section~\ref{sec:background-harness})
can play two roles in development.
A \emph{production harness} organizes generation and editing
for a request; a \emph{development harness} diagnoses failures,
curates data, launches training, and validates candidates.
The components reviewed below belong mostly to production
harnesses.
Many of them revise only the current output; those that retain
memory, rules, skills, or trained controller weights update $\xi_k$.
These mechanisms also provide ways to organize development decisions.
Their relevance to RSI depends on whether retained changes improve the
procedures used to diagnose failures, construct updates, and validate
later candidates.

We review four functional components:
(\emph{i}) Planning \& Orchestration,
(\emph{ii}) Tool Use \& Control,
(\emph{iii}) Memory \& Experience, and
(\emph{iv}) Skills \& Reusable Procedures.
Together, they cover production organization, individual
operations, and the retention of evidence and methods
for future use.
Adaptation and training mechanisms are discussed separately
at the end.

\input{tables/rsi_scaffold}

\subsubsection{Planning \& Orchestration}

Video production involves deciding what to produce,
organizing execution, and responding to failures.
Following these responsibilities, we review Production
Planning, Execution Coordination, and Review \& Recovery.

\noindent\textbf{Production Planning.}
Production planning translates a request into creative
specifications and structured production steps.
MovieAgent~\citep{wu2025movieagent} decomposes stories into
scene and shot plans.
Co-Director~\citep{codirector2026} explores creative
configurations while revising storylines and keyframes.
MultiMedia-Agent~\citep{multimediaagent2026} addresses
executable plans for multimedia production, while
AgenticGen~\citep{agenticgen2026} connects advertising
strategy selection to executable drafts.
These approaches span planning granularities from
overall creative intent to individual production steps.

\noindent\textbf{Execution Coordination.}
Execution coordination manages how tools and agents
carry out a plan, including action order and dependencies.
VideoGen-Agent~\citep{li2026videogenagent} organizes
multi-turn tool use into production trajectories.
NEWTON~\citep{feng2026newton} uses a planner to organize
actions involving a fixed generator, verifier, and toolset.
Recommendation as Generation~\citep{cheng2026rag}
connects instruction and video-agent policies within
recommendation-oriented production under quality constraints.
The inputs and controls for individual tool calls
are discussed under Tool Use \& Control.

\noindent\textbf{Review \& Recovery.}
Review and recovery determines where feedback should
be applied and which stages should be revisited when
intermediate results are unsatisfactory.
CHIEF~\citep{chief2026} converts audience and film-critic
feedback into revision requests, while
CogenPVG~\citep{cogenpvg2026} distributes generator--critic
pairs across argument, storyboard, asset, and editing stages.
VideoAgent~\citep{zhou2026videoagentediting} changes the
execution structure by correcting edges, adding missing
nodes, and revising the order of an editing-agent DAG.
CutClaw~\citep{zhao2026cutclaw} backtracks to alternative
footage intervals after review.
These mechanisms direct revision within the workflow;
the corrective operations themselves belong to
the tool-use and editing interface.

\subsubsection{Tool Use \& Control}

Production plans must be translated into operations
supported by available tools.
We review Tool Selection \& Conditioning for guiding
generation, followed by Editing \& Repair for correcting
existing results.

\noindent\textbf{Tool Selection \& Conditioning.}
Tool selection and conditioning matches a production
requirement to a generator and expresses it through
prompts, references, and other control inputs.
SPAgent~\citep{spagent2024} evaluates newly introduced
generators and incorporates their capability profiles
into model selection.
VISTA~\citep{vista2025} combines candidate tournaments
with visual, audio, and contextual critique to refine
prompts, while VQQA~\citep{song2026vqqa} uses diagnostic
questions to guide refinement.
SCMAPR~\citep{scmapr2026} verifies rewritten prompts
against atomic requirements before generation.
GenMAC~\citep{huang2026genmac} extends revision to prompts,
bounding boxes, and guidance through specialist feedback.
Beyond Trial-and-Error~\citep{beyondtrial2026} combines
semantic prompt revision with search over seed and guidance.

Specialized controllers address particular conditioning
decisions.
MVAgent~\citep{kong2026mvagent} constructs conditions
across multi-shot trajectories, while
CamPilot~\citep{campilot2026} focuses on camera-work control.
PhyPrompt~\citep{phyprompt2026} produces physics-aware
prompts, and DPP-GRPO~\citep{dppgrpo2025} targets
diversity and relevance in prompt rewriting.

\noindent\textbf{Editing \& Repair.}
Editing and repair specifies changes to generated content
or its underlying representation while seeking to preserve
satisfactory parts of the result.
VideoRepair~\citep{lee2026videorepair} refines misaligned
regions while protecting correctly generated content.
PhysAgent~\citep{physagent2026} repairs executable
scene-state and event programs after reconstruction
and simulation checks.
CutCraft~\citep{zeng2026cutcraft} adjusts cut points,
audio offsets, and transitions, with reported trade-offs
between transition quality and within-shot synchronization.
Crayotter~\citep{yan2026grpb} addresses sequences of
semantic editing decisions, while
Aurora~\citep{aurora2026} plans editing conditions
for a fixed video editor.
These interfaces expose different correction units,
with different preservation constraints and costs.

\subsubsection{Memory \& Experience}

Production decisions draw on information accumulated
within a project and across previous attempts.
We distinguish Project Context from Retained Experience
according to the role of the information being preserved.

\noindent\textbf{Project Context.}
Project context maintains the characters, references,
and intermediate states with which subsequent production
decisions must remain consistent.
DreamFactory~\citep{xie2024dreamfactory} carries character
descriptions and preceding visual references through
multi-scene production.
CineAgents~\citep{cineagents2026} verifies a production
blueprint against hierarchical source-film memory.
Both make earlier information available for later
planning and consistency checking.

\noindent\textbf{Retained Experience.}
Retained experience records previous actions and outcomes
so that related decisions can draw on evidence of
successful choices and recurring failures.
AesopAgent~\citep{wang2024aesopagent} combines a knowledge
store with Experience-RAG, preserving feedback entries
for workflow, prompt, and utility decisions.
A$^2$RD~\citep{song2026a2rd} stores successful and
unsuccessful prompt rewrites with rubric scores,
combining retrieval with hierarchical segment repair.
EditDuet~\citep{sandoval2025editduet} retains scored and
repaired communication examples for its Editor and Critic.
Its collection-specific in-context examples support
coordination, while cross-collection accumulation
remains untested.

\subsubsection{Skills \& Reusable Procedures}

Production knowledge can be reused as shared guidance
or as routines assembled for different requests.
We examine these forms through Prompts \& Production
Rules, followed by Skill Construction \& Composition.

\noindent\textbf{Prompts \& Production Rules.}
Prompts and production rules encode shared instructions
and constraints governing how stages are performed,
reviewed, and retried.
AESR~\citep{aesr2026} maintains strategies, templates,
and pitfalls in a prompting playbook alongside
current-video repair.
CineForge~\citep{cineforge2026} represents production
guidance through stage contracts, prompt rules,
retry budgets, and review routing.
AVA-Encoder~\citep{li2026avaencoder} uses shared shot
and keyframe encoding prompts for film reconstruction,
separating reusable instructions from repairs
to the current film knowledge graph.

\noindent\textbf{Skill Construction \& Composition.}
Skill construction and composition packages production
methods into reusable routines and combines them
into workflows for different requests.
SkillPE~\citep{skillpe2026} constructs a cinematic skill
library from variants of expert seed skills inspired
by film references.
The selected library remains fixed during deployment.
VideoWeaver~\citep{wei2026videoweaver} combines fixed
foundation skills with category-level composition skills
and creator skills for constructing further workflows.
Its evaluation covers evolved composition skills
and skill creators applied to unseen categories,
extending reuse from production routines to the
procedures that build them.

\medskip
\noindent\textbf{Adaptation and Learning Across Components.}
Within-task adaptation refines the current production
attempt through mechanisms such as prompt revision
in VISTA~\citep{vista2025} and workflow revision
in VideoAgent~\citep{zhou2026videoagentediting}.
Cross-task learning instead retains updates for later use.
Non-parametric approaches revise production rules
in CineForge~\citep{cineforge2026}, shared encoding prompts
in AVA-Encoder~\citep{li2026avaencoder}, and skill libraries
in SkillPE~\citep{skillpe2026}.
Retention still requires independent validation:
AESR~\citep{aesr2026} accumulates its playbook using
test-set feedback.

Parametric approaches optimize controller policies.
VideoGen-Agent~\citep{li2026videogenagent} combines
supervised trajectories with GRPO, while
NEWTON~\citep{feng2026newton} uses trajectory-level Flow-GRPO.
MultiMedia-Agent~\citep{multimediaagent2026} applies staged
supervised fine-tuning and preference optimization;
Aurora~\citep{aurora2026} combines supervised fine-tuning
with offline DPO.
Specialized supervision includes shot comparisons
in MVAgent~\citep{kong2026mvagent}, film-derived labels
in CamPilot~\citep{campilot2026}, video feedback
in PhyPrompt~\citep{phyprompt2026}, and text-space rewards
in DPP-GRPO~\citep{dppgrpo2025}.
Crayotter~\citep{yan2026grpb} addresses credit assignment
across editing stages.
Deployment further supplies click-through preferences
in AgenticGen~\citep{agenticgen2026} and interaction feedback
in Recommendation as Generation~\citep{cheng2026rag}.

Production experience can also inform development beyond the current
request.
The feedback collected by production systems becomes useful for model
or harness improvement when it identifies which capability failed,
under which conditions, and which intervention could address it.
Retaining traces or preferences supplies evidence, but the effect on
later development depends on how that evidence changes data collection,
training objectives, or update selection.

Persistent revision of the development harness extends this use of
experience to the procedures that produce updates.
VideoWeaver's creator skills illustrate this direction within skill
construction.
Related work outside video generation includes
Video-RSI~\citep{luo2026videorsi}, which evolves the harness of video
understanding agents with frozen model weights, and the Darwin G\"odel
Machine~\citep{zhang2025dgm} and RRSI~\citep{xia2026rrsi}, which revise
agent code or harnesses across generations.
For video generation, the mechanisms reviewed here support parts of
such a process, but do not yet establish that revisions to the
development harness make successive rounds of generator development
more effective.
Section~\ref{sec:rsi-evaluation} examines the evaluation needed to assess
both retained capability gains and changes to the improvement procedure.

\subsection{Evaluation in Self-Improvement}
\label{sec:rsi-evaluation}

Evaluation supports progress toward RSI by providing
feedback for proposing updates and independent evidence
for assessing their effects.
We distinguish two complementary perspectives:
(\emph{i}) Video Output Evaluation, which assesses
video quality, task fulfillment, and observable failures;
and (\emph{ii}) Agent Harness Evaluation, which examines
how the production system contributes to these outcomes,
diagnoses failures, and carries out corrective interventions.
For agentic video generation, identifying an unsatisfactory
result is only the first step: useful feedback must help
connect visual evidence to decisions or representations
that the system can revise.
Across both perspectives, development feedback should
remain distinct from evidence used to validate improvements.
Evaluation should also distinguish improvements to the current output,
retained capabilities that benefit later tasks, and changes to the
procedure that produces subsequent updates.

\subsubsection{Video Output Evaluation}

Video output evaluation measures the properties of generated
videos and identifies where they fail to meet requirements.
We review four complementary elements:
(\emph{i}) Metrics and Benchmarks;
(\emph{ii}) Model-Based Judging;
(\emph{iii}) Human Assessment and Evaluator Reliability;
and (\emph{iv}) Agentic Evaluation Pipelines.
Together, they specify what is tested, how it is measured,
how judgments are validated, and how evaluation is executed.

\textbf{Metrics and Benchmarks.}
Metrics and benchmarks translate video capabilities
into test cases and quantitative measurements.
Fr\'echet Video Distance (FVD)~\cite{unterthiner2018fvd}
measures distributional similarity between real and
generated videos, but its sensitivity to visual content
can obscure motion defects~\cite{ge2024contentbias}.
FVMD~\cite{liu2024fvmd} instead measures motion through
tracked points.
VBench~\cite{huang2024vbench} and
EvalCrafter~\cite{liu2024evalcrafter} extend assessment
to visual quality, temporal consistency, motion,
and text--video alignment.
More targeted benchmarks operationalize specific capabilities:
T2V-CompBench~\cite{sun2025t2vcompbench} evaluates
compositional instruction following;
VBench-2.0~\cite{zheng2025vbench2} covers physical
and commonsense plausibility;
VideoPhy~\cite{bansal2024videophy} and
VideoPhy-2~\cite{bansal2025videophy2} test physical interactions.
StateBench~\cite{miao2026statebench},
V2V-Bench~\cite{liu2026v2vbench}, and
AVGen-Bench~\cite{zhou2026avgenbench} extend coverage
to state consistency, source-preserving editing,
and audiovisual generation, respectively. DirectorBench~\cite{chen2026directorbench} extends evaluation to
long-form agentic production through structured checkpoints for
script, visual, audio, cross-modal, and stability criteria.
These measurements capture different properties:
distributional fidelity, temporal consistency, and
successful execution of a requested action should
not be treated as interchangeable.

Explicit world representations can provide additional reference
information (Section~\ref{sec:iwm-explicit-state}).
Agent-built worlds supply reference-backed evaluation sets for the
Programmable World Model~\cite{huang2026programmable},
Code Video Model~\cite{liang2026codevideomodel}, and
SolarRenderer~\cite{fang2026solarrenderer}.
Specified objects, states, and events can support checks of whether
generated content follows the intended structure.
These references strengthen evaluation only to the extent that adherence
is checked against the rendered video; the correctness of a program
alone does not establish the correctness of its visual realization.

\textbf{Model-Based Judging.}
Learned scoring models and LLM/VLM judges assess requirements
that fixed summary metrics capture poorly.
LLMs can decompose instructions into criteria,
while VLMs inspect visual evidence.
VideoScore~\cite{he2024videoscore} and
VideoReward~\cite{liu2025videoalign} learn scoring functions
from human ratings or preferences.
Video-Bench~\cite{han2025videobench} uses structured
questions, while VideoScore2~\cite{he2025videoscore2}
and AIGVE-MACS~\cite{liu2025aigvemacs} combine scores
with explanatory feedback.
Rubric-guided evaluation makes criteria and scoring
rules explicit:
VideoArgus~\cite{zeng2026videoargus} constructs output-blind
rubrics from the input,
RewardVerse~\cite{tang2026rewardverse} learns rubric-based
scoring, and RAVEN-Eval~\cite{jia2026raveneval} uses
order-reversed pairwise comparisons to reduce position bias.
Criterion-level judgments can identify unmet requirements,
but an explanation of a visual defect does not by itself
establish its cause in the production process.

\textbf{Human Assessment and Evaluator Reliability.}
Human assessment provides supervision and a reference for validating
automated judgments.
SafeSora~\cite{dai2024safesora} separates helpfulness and harmlessness,
EvalVerse~\cite{yang2026evalverse} incorporates filmmaking expertise,
and PhyGround~\cite{lin2026phyground} uses observable physical criteria.
Evaluator reliability concerns agreement with humans and sensitivity
to errors: UVE~\cite{liu2025uve} examines scoring and frame-sampling
choices, Spotlight~\cite{chinchure2025spotlight} tests error detection
and temporal localization, and SLVMEval~\cite{matsuda2026slvmeval}
introduces controlled defects into long videos.

When judges also supply learning feedback, independent evaluation
helps distinguish quality gains from reward overoptimization
\cite{gao2023overoptimization}.
VideoGen-Agent~\cite{li2026videogenagent} reuses its training judge
and rubrics for automated testing, making its human comparisons
important additional evidence.
Don't Read the Log~\cite{xu2026dontreadlog} identifies execution-log
bias in the tested Qwen-VL judges, underscoring the need to inspect
generated videos directly and specify the evidence available to
each judge.

\textbf{Agentic Evaluation Pipelines.}
Agentic evaluation pipelines organize instruction analysis,
evidence collection, and scoring into executable procedures.
VideoGen-Eval~\cite{yang2025videogeneval} and
VideoArgus~\cite{zeng2026videoargus} combine instruction
analysis with specialized inspection tools,
while VQQA~\cite{song2026vqqa} generates video-specific questions.
Evaluation Agent~\cite{zhang2025evaluationagent} and
Open Evaluation Agent~\cite{tian2026openevaluationagent}
adaptively select prompts and tools.
RAVEN-Eval~\cite{jia2026raveneval} reduces comparison costs
through selected reference generators.
TEAR~\cite{he2026tear} and
SceneSplit~\cite{lee2026scenesplit} extend testing
to temporally structured safety failures.
Prompt coverage, sampling policies, tool configurations,
judge reliability, and evaluation cost jointly determine
whether measurements are reproducible and informative.
Active Testing~\cite{kossen2021activetesting} further
highlights that adaptive test selection requires care
when estimating overall performance;
failure rates on deliberately adversarial prompts
should be distinguished from those on representative requests.

These measurements identify observable defects,
but do not by themselves establish which production
decision caused a failure or how it should be corrected.
The next perspective combines output evidence with plans,
intermediate states, and execution traces to evaluate
diagnosis and repair.
Because repeated optimization can overfit particular
tasks or judges, resulting gains should be checked
on held-out requests with independent evaluation.

\subsubsection{Agent Harness Evaluation}

Agent harness evaluation examines how prompts, memory,
skills, tool interfaces, and control logic support
video production and its improvement.
We organize this assessment around four dimensions:
(\emph{i}) Effectiveness and Efficiency, which assesses
performance gains and their costs;
(\emph{ii}) Diagnosability and Repairability, which examines
whether failures lead to accurate diagnoses and effective
corrective actions;
(\emph{iii}) Generalization, which tests whether gains
extend beyond development tasks;
and (\emph{iv}) Continual Adaptation, which tracks progress
and regressions across successive updates.

The evaluation setting should specify what changes,
what persists, and which feedback is available.
Rethinking the Evaluation of Harness Evolution for
Agents~\cite{wang2026harnessevaluation} compares
four approaches:
\emph{parallel sampling} explores independent trajectories
under a fixed harness;
\emph{sequential refinement} revises trajectories using
previous attempts while keeping the harness fixed;
\emph{harness evolution} updates a shared harness across tasks;
and \emph{harness scaling} revises the harness for
an individual task.
These settings differ in their update targets and scope
of reuse.
In video generation, VISTA~\cite{vista2025} and
VQQA~\cite{song2026vqqa} refine individual requests.
Evaluating persistent harness changes additionally
requires examining their effects on subsequent tasks.

\textbf{Effectiveness and Efficiency.}
Effectiveness concerns task completion, instruction fulfillment,
and video quality; efficiency concerns the resources needed to
develop and operate the harness.
An agentic pipeline can improve on a single generation call by
sampling more candidates, revising the plan, or supplying better
generation conditions.
To attribute gains to harness revision, comparisons should control
model and tool versions, match feedback access and resource budgets,
and include fixed-harness sampling or refinement baselines.
Ablations of plan revision and conditioning should use matched
sampling budgets to distinguish their contributions from
additional generation attempts.
When a script or plan is rewritten, the resulting video should be
evaluated against the original user brief so that plan revision
does not redefine task success.
Controller training and generator upgrades should be evaluated
separately where applicable.

Resource reporting should include generator, critic, and controller
calls, wall-clock time, and computational or monetary cost.
Update-development costs should be reported separately from
subsequent production costs.

AFlow~\cite{zhang2024aflow} reports performance--cost comparisons,
while SICA~\cite{robeyns2025sica} evaluates accuracy, runtime,
and monetary cost.
Evo-Bench~\cite{huang2026evobench} records development-side
progress and final held-out performance.
Meta-Harness~\cite{lee2026metaharness} uses prior harness code,
scores, and execution traces to inform subsequent proposals,
illustrating how aggregate results can be supplemented with
process evidence.
In video generation, AgentMV~\cite{agentmv2026} allocates generation
resources under a global budget constraint and evaluates the
resulting quality--cost trade-offs.
VideoWeaver~\cite{wei2026videoweaver} combines process and output
assessment with human rankings and resource-use measurements.
VideoGen-Agent~\cite{li2026videogenagent} includes fixed-tool
training ablations and human comparisons; its generator-upgrade
experiment separately measures the benefit of changing
generation tools.
Such comparisons help identify which intervention contributes
to a gain and at what cost.

\textbf{Diagnosability and Repairability.}
Evaluation should assess whether a harness identifies actionable
failure causes and corrects them while preserving satisfactory content.
VideoRepair~\cite{lee2026videorepair} illustrates regional correction,
while PhysAgent~\cite{physagent2026} exposes scene-state and event
programs as repair targets.
Using shared failed outputs or controlled faults, comparisons with
unguided retries and full regeneration should hold generators,
tools, and resource budgets fixed, reporting diagnostic accuracy,
repair success and locality, preservation of correct content,
and recovery cost.
Proposed causes should be checked through targeted interventions,
since visual localization alone does not establish a causal diagnosis.

\textbf{Generalization.}
Generalization assesses whether a selected harness's
benefits extend beyond the tasks used to develop it.
New examples from the development distribution test
in-distribution generalization, while different task
categories or domains test out-of-distribution generalization.
The selected harness and its shared cross-task state
remain fixed during this assessment.
Prescribed within-task reasoning and repair may still
operate, but test feedback should not drive candidate
selection or persistent updates.

GEPA~\cite{agrawal2025gepa} separates candidate selection
on validation tasks from final testing,
while ADAS~\cite{hu2024adas} examines transfer across
tasks and models.
RRSI~\cite{xia2026rrsi} evaluates evolved harnesses
on other benchmarks.
In video generation, VideoWeaver~\cite{wei2026videoweaver}
tests frozen evolved skills on new cases and task categories,
while VideoGen-Agent~\cite{li2026videogenagent} evaluates
an unseen task combination.
For diagnostic and repair mechanisms, generalization
also concerns unfamiliar failure types and production
configurations.
Development-time validation can guide whether a rule
or skill should be broadly retained or restricted
to a category; final testing should remain independent
of those decisions.

\textbf{Continual Adaptation.}
Continual adaptation assesses how harness performance
changes over successive updates.
Evaluation tracks whether additional experience improves
performance, how quickly gains emerge, and whether
later updates degrade previously supported capabilities.
The task distribution may remain stable or change
with production requirements and available tools.
ACE~\cite{zhang2025ace} and
Evo-Harness~\cite{wei2026evoharness} use completed tasks
to inform subsequent updates.
EvoHarnessBench~\cite{ke2026evoharnessbench} distinguishes
externally supplied harness expansion from experience-driven
adaptation, showing that additional tools, skills,
or agents can impair earlier-task performance even
with fixed model weights.
For video harnesses, evaluation should track quality,
completion rate, and cumulative cost across updates,
testing current and earlier task families for gains
and regressions.
Repeated diagnosis-and-repair tests can additionally
reveal whether the harness becomes more effective
at correcting recurring failures.
Stagewise adaptation should use data separate from
evaluation; online evaluation should score each task
before its feedback is used for subsequent updates.

Together, these perspectives assess video quality, retained harness
capabilities, and the ability to diagnose failures and verify repairs.
Establishing RSI requires an additional comparison at the level of the
development procedure $\rho_k$.
A fixed-procedure baseline and a system allowed to revise its procedure
should start from matched components and receive comparable feedback
access and resource budgets.
The retained procedural changes should then be used in subsequent
development rounds, with independent tasks assessing the quality and
cost of the updates those rounds produce.
Evaluation should track whether the revised procedure produces better
updates for a given budget, reaches comparable quality at lower cost,
and preserves earlier capabilities across repeated rounds.

%% file: tables/foundation_model.tex
\begin{figure*}[t]
\centering
\input{figures/foundation-model-taxonomy}
\captionsetup{position=bottom}
\caption{\textbf{Taxonomy of foundation model improvement for video generation.}
Representative works are organized by development area and category.
Model training covers pre-training and three post-training categories.}
\label{fig:foundation-model-taxonomy}
\end{figure*}
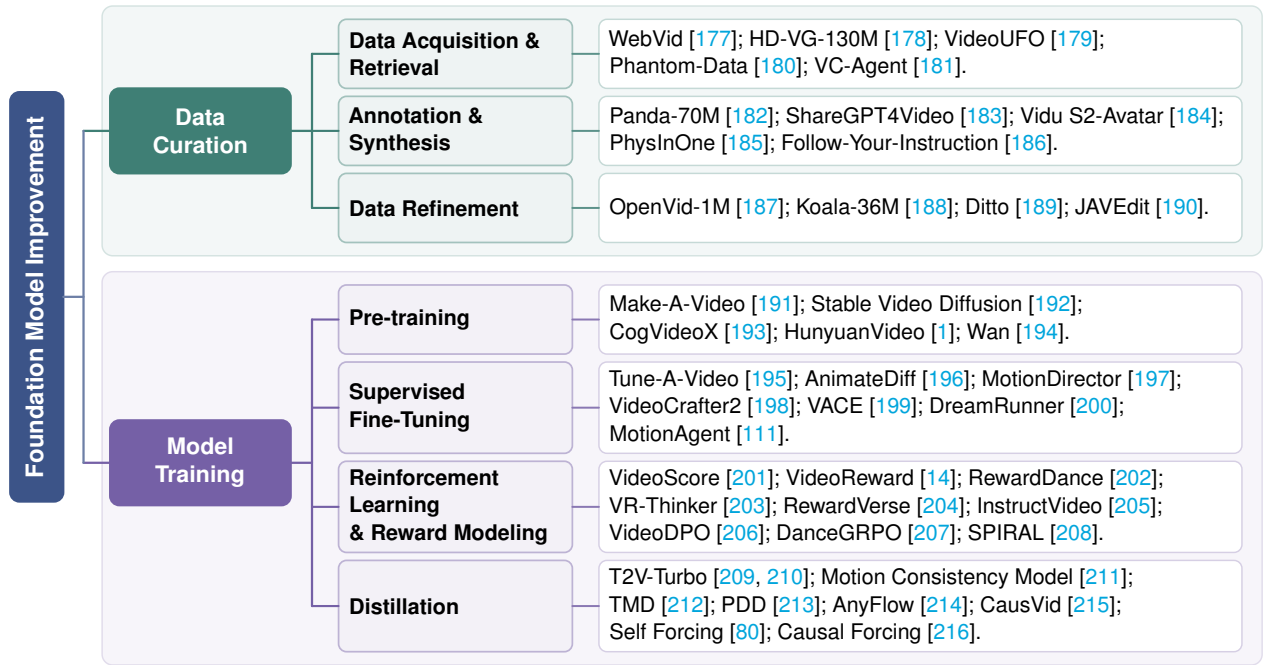

%% file: figures/foundation-model-taxonomy.tex
\input{figures/taxonomy-style}
\begingroup
\SurveyTaxonomyLandscape
\newcommand{\fmWork}[2]{\SurveyTaxonomyWork{#1}{#2}}
\SurveyTaxonomyStart{Foundation Model Improvement}
\SurveyTaxonomyGroup{taxTeal}{Data\\Curation}
\SurveyTaxonomyRow{Data Acquisition \& Retrieval}{%
  \fmWork{WebVid}{bain2021frozen};
  \fmWork{HD-VG-130M}{wang2023videofactory};
  \fmWork{VideoUFO}{wang2025videoufo};
  \fmWork{Phantom-Data}{chen2025phantomdata};
  \fmWork{VC-Agent}{zhang2025vcagent}.}
\SurveyTaxonomyRow{Annotation \& Synthesis}{%
  \fmWork{Panda-70M}{chen2024panda};
  \fmWork{ShareGPT4Video}{chen2024sharegpt4video};
  \fmWork{Vidu S2-Avatar}{zhang2026vidus2};
  \fmWork{PhysInOne}{zhou2026physinone};
  \fmWork{Follow-Your-Instruction}{feng2025followyourinstruction}.}
\SurveyTaxonomyRow{Data Refinement}{%
  \fmWork{OpenVid-1M}{nan2024openvid};
  \fmWork{Koala-36M}{wang2025koala};
  \fmWork{Ditto}{bai2025ditto};
  \fmWork{JAVEdit}{chen2026javedit}.}
\SurveyTaxonomyEndGroup

\SurveyTaxonomyGroup{taxPurple}{Model\\Training}
\SurveyTaxonomyRow{Pre-training}{%
  \fmWork{Make-A-Video}{singer2022make};
  \fmWork{Stable Video Diffusion}{blattmann2023stablevideo};
  \fmWork{CogVideoX}{yang2024cogvideox};
  \fmWork{HunyuanVideo}{kong2024hunyuanvideo};
  \fmWork{Wan}{wan2025wan}.}

\SurveyTaxonomyRow{Supervised Fine-Tuning}{%
  \fmWork{Tune-A-Video}{wu2023tune};
  \fmWork{AnimateDiff}{guo2024animatediff};
  \fmWork{MotionDirector}{zhao2023motiondirector};
  \fmWork{VideoCrafter2}{chen2024videocrafter2};
  \fmWork{VACE}{vace2025};
  \fmWork{DreamRunner}{wang2024dreamrunner};
  \fmWork{MotionAgent}{motionagent2025}.}
\SurveyTaxonomyRow{Reinforcement Learning\\\& Reward Modeling}{%
  \fmWork{VideoScore}{he2024videoscore};
  \fmWork{VideoReward}{liu2025videoalign};
  \fmWork{RewardDance}{wu2025rewarddance};
  \fmWork{VR-Thinker}{wang2025vrthinker};
  \fmWork{RewardVerse}{tang2026rewardverse};
  \fmWork{InstructVideo}{yuan2024instructvideo};
  \fmWork{VideoDPO}{liu2025videodpo};
  \fmWork{DanceGRPO}{xue2025dancegrpo};
  \fmWork{SPIRAL}{yang2026spiral}.}
\SurveyTaxonomyRow{Distillation}{%
  \fmWork{T2V-Turbo}{li2024t2vturbo,li2025t2vturbov2};
  \fmWork{Motion Consistency Model}{zhai2024motion};
  \fmWork{TMD}{nie2026transition};
  \fmWork{PDD}{shaul2026pdd};
  \fmWork{AnyFlow}{gu2026anyflow};
  \fmWork{CausVid}{yin2025causvid};
  \fmWork{Self Forcing}{huang2025selfforcing};
  \fmWork{Causal Forcing}{zhu2026causal}.}
\SurveyTaxonomyEndGroup
\SurveyTaxonomyFinish
\endgroup

%% file: figures/taxonomy-style.tex
\ifdefined\SurveyTaxonomyStart\else
\definecolor{taxBlue}{HTML}{3C5488}
\definecolor{taxTeal}{HTML}{3F7F75}
\definecolor{taxPurple}{HTML}{7560A6}
\definecolor{taxAmber}{HTML}{9A6A19}
\definecolor{taxCoral}{HTML}{B84C42}
\definecolor{taxBerry}{HTML}{A23875}
\definecolor{taxOlive}{HTML}{6D7C33}
\definecolor{taxPerception}{HTML}{637B3E}
\definecolor{taxPlanning}{HTML}{A17A23}
\definecolor{taxMemory}{HTML}{985978}
\definecolor{taxCoding}{HTML}{AF6344}
\definecolor{taxComputer}{HTML}{487D69}
\definecolor{taxContext}{HTML}{827047}
\definecolor{taxCommunication}{HTML}{8B4C45}
\ifdefined\taxCategoryBox\else
\newsavebox{\taxCategoryBox}
\newsavebox{\taxWorksBox}
\newlength{\taxWidth}
\newlength{\taxRight}
\newlength{\taxWorksWidth}
\newlength{\taxY}
\newlength{\taxTop}
\newlength{\taxBottom}
\newlength{\taxHeight}
\newlength{\taxCenter}
\newlength{\taxFirstGroupY}
\newlength{\taxLastGroupY}
\newlength{\taxRootSpan}
\newcounter{taxRowNumber}
\newcounter{taxFirstRow}
\newcounter{taxMiddleRow}
\fi
\newcommand{\taxBodyFont}{\TaxonomyArial\mdseries\fontsize{9}{11.4}\selectfont
  \hyphenpenalty=10000\exhyphenpenalty=10000}
\newcommand{\taxCategoryFont}{\TaxonomyArial\bfseries\fontsize{9}{11.4}\selectfont
  \hyphenpenalty=10000\exhyphenpenalty=10000}
\newcommand{\taxGroupFont}{\TaxonomyArial\bfseries\fontsize{9.5}{11.8}\selectfont
  \hyphenpenalty=10000\exhyphenpenalty=10000}
\newcommand{\taxCategoryWidth}{84pt}
\newcommand{\taxCategoryLeft}{132pt}
\newcommand{\taxWorksLeft}{238pt}
\newcommand{\taxGroupCenter}{76pt}
\newcommand{\taxGroupWidth}{64pt}
\newcommand{\taxGroupXPad}{6pt}
\newcommand{\taxBranchX}{122pt}
\newcommand{\taxXPad}{6pt}
\newcommand{\taxRowPad}{5pt}
\newcommand{\taxMinRowHeight}{30pt}
\newcommand{\taxRowGap}{4pt}
\newcommand{\taxGroupPad}{8pt}
\newcommand{\taxGroupGap}{9pt}
\newcommand{\taxRootBranchX}{28pt}
\newcommand{\taxRegionLeft}{35pt}
\newcommand{\taxMinGroupHeight}{36pt}
\newcommand{\taxAlignMiddleChild}{0}
\newcommand{\taxReferenceLineWidth}{.75pt}
\newcommand{\SurveyTaxonomyLandscape}{%
  \renewcommand{\taxBodyFont}{\TaxonomyArial\mdseries\fontsize{9}{10.3}\selectfont
    \hyphenpenalty=10000\exhyphenpenalty=10000}%
  \renewcommand{\taxCategoryFont}{\TaxonomyArial\bfseries\fontsize{9}{10.3}\selectfont
    \hyphenpenalty=10000\exhyphenpenalty=10000}%
  \renewcommand{\taxGroupFont}{\TaxonomyArial\bfseries\fontsize{9.5}{10.6}\selectfont
    \hyphenpenalty=10000\exhyphenpenalty=10000}%
  \renewcommand{\taxCategoryWidth}{80pt}%
  \renewcommand{\taxCategoryLeft}{124pt}%
  \renewcommand{\taxWorksLeft}{222pt}%
  \renewcommand{\taxGroupCenter}{72pt}%
  \renewcommand{\taxGroupWidth}{60pt}%
  \renewcommand{\taxGroupXPad}{4pt}%
  \renewcommand{\taxBranchX}{114pt}%
  \renewcommand{\taxXPad}{4pt}%
  \renewcommand{\taxRowPad}{3pt}%
  \renewcommand{\taxMinRowHeight}{26pt}%
  \renewcommand{\taxRowGap}{3pt}%
  \renewcommand{\taxGroupPad}{5pt}%
  \renewcommand{\taxGroupGap}{6pt}%
  \renewcommand{\taxMinGroupHeight}{32pt}%
}
\newcommand{\SurveyTaxonomyCompact}{%
  \renewcommand{\taxAlignMiddleChild}{1}%
  \renewcommand{\taxReferenceLineWidth}{.8pt}%
  \renewcommand{\taxBodyFont}{\TaxonomyArial\mdseries\fontsize{7}{7.6}\selectfont
    \hyphenpenalty=10000\exhyphenpenalty=10000}%
  \renewcommand{\taxCategoryFont}{\TaxonomyArial\bfseries\fontsize{7.4}{8.3}\selectfont
    \hyphenpenalty=10000\exhyphenpenalty=10000}%
  \renewcommand{\taxGroupFont}{\TaxonomyArial\bfseries\fontsize{8.2}{9.6}\selectfont
    \hyphenpenalty=10000\exhyphenpenalty=10000}%
  \renewcommand{\taxCategoryWidth}{60pt}%
  \renewcommand{\taxCategoryLeft}{112pt}%
  \renewcommand{\taxWorksLeft}{188pt}%
  \renewcommand{\taxGroupCenter}{67pt}%
  \renewcommand{\taxGroupWidth}{58pt}%
  \renewcommand{\taxGroupXPad}{3pt}%
  \renewcommand{\taxBranchX}{104pt}%
  \renewcommand{\taxXPad}{4pt}%
  \renewcommand{\taxRowPad}{1.5pt}%
  \renewcommand{\taxMinRowHeight}{18pt}%
  \renewcommand{\taxRowGap}{1.3pt}%
  \renewcommand{\taxGroupPad}{2pt}%
  \renewcommand{\taxGroupGap}{3.5pt}%
  \renewcommand{\taxMinGroupHeight}{26pt}%
}
\newcommand{\SurveyTaxonomyWork}[2]{\mbox{#1~\citep{#2}}}

\newcommand{\SurveyTaxonomyStart}[1]{%
  \setlength{\parindent}{0pt}\setlength{\parskip}{0pt}%
  \def\taxRootTitle{#1}%
  \setlength{\taxWidth}{\linewidth}%
  \setlength{\taxRight}{\dimexpr\taxWidth-8pt\relax}%
  \setlength{\taxWorksWidth}{\dimexpr\taxRight-\taxWorksLeft-\taxXPad-\taxXPad\relax}%
  \setcounter{taxRowNumber}{0}%
  \setlength{\taxY}{0pt}%
  \begin{tikzpicture}[text=black,line cap=round,line join=round]
}

\newcommand{\SurveyTaxonomyGroup}[2]{%
  \def\taxColor{#1}\def\taxGroupTitle{#2}%
  \setlength{\taxTop}{\taxY}%
  \addtolength{\taxY}{\taxGroupPad}%
  \setcounter{taxFirstRow}{\value{taxRowNumber}}\stepcounter{taxFirstRow}%
}

\newcommand{\SurveyTaxonomyRow}[2]{%
  \stepcounter{taxRowNumber}%
  \begin{pgfinterruptpicture}%
    \global\setbox\taxCategoryBox=\hbox{\parbox{\taxCategoryWidth}{%
      \raggedright\taxCategoryFont #1\par}}%
    \global\setbox\taxWorksBox=\hbox{\parbox{\taxWorksWidth}{%
      \raggedright\taxBodyFont #2\par}}%
    \global\taxHeight=\dimexpr\ht\taxCategoryBox+\dp\taxCategoryBox\relax\relax
    \ifdim\dimexpr\ht\taxWorksBox+\dp\taxWorksBox\relax>\taxHeight
      \global\taxHeight=\dimexpr\ht\taxWorksBox+\dp\taxWorksBox\relax\relax
    \fi
  \end{pgfinterruptpicture}%
  \addtolength{\taxHeight}{\dimexpr\taxRowPad+\taxRowPad\relax}%
  \ifdim\taxHeight<\taxMinRowHeight\setlength{\taxHeight}{\taxMinRowHeight}\fi
  \setlength{\taxCenter}{\dimexpr\taxY+\taxHeight/2\relax}%
  \node[draw=\taxColor!48,fill=\taxColor!9,rounded corners=3pt,
    line width=.65pt,inner xsep=\taxXPad,inner ysep=\taxRowPad,outer sep=0pt,
    minimum height=\taxHeight,anchor=west,align=left]
    (taxSub\thetaxRowNumber) at (\taxCategoryLeft,-\taxCenter) {\usebox{\taxCategoryBox}};
  \node[draw=\taxColor!30,fill=white,rounded corners=3pt,
    line width=.55pt,inner xsep=\taxXPad,inner ysep=\taxRowPad,outer sep=0pt,
    minimum height=\taxHeight,anchor=west,align=left]
    (taxRef\thetaxRowNumber) at (\taxWorksLeft,-\taxCenter) {\usebox{\taxWorksBox}};
  \draw[\taxColor!85,line width=\taxReferenceLineWidth]
    (taxSub\thetaxRowNumber.east)--(taxRef\thetaxRowNumber.west);
  \addtolength{\taxY}{\taxHeight}\addtolength{\taxY}{\taxRowGap}%
}

\newcommand{\SurveyTaxonomyEndGroup}{%
  \setlength{\taxBottom}{\dimexpr\taxY-\taxRowGap+\taxGroupPad\relax}%
  \ifnum\value{taxFirstRow}=1
    \edef\taxFirstGroup{\thetaxRowNumber}%
  \fi
  \ifnum\taxAlignMiddleChild=1
    \setcounter{taxMiddleRow}{\numexpr(\value{taxFirstRow}+\value{taxRowNumber})/2\relax}%
    \coordinate (taxGroupMid\thetaxRowNumber) at (taxSub\thetaxMiddleRow.center);
  \else
    \coordinate (taxGroupMid\thetaxRowNumber) at
      ($(taxSub\thetaxFirstRow.center)!0.5!(taxSub\thetaxRowNumber.center)$);
  \fi
  \node[draw=\taxColor,fill=\taxColor,text=white,rounded corners=3.5pt,
    line width=.8pt,inner xsep=\taxGroupXPad,inner ysep=6pt,outer sep=0pt,
    minimum height=\taxMinGroupHeight,text width=\taxGroupWidth,align=center,
    font=\taxGroupFont]
    (taxCap\thetaxRowNumber) at (\taxGroupCenter,0pt |- taxGroupMid\thetaxRowNumber)
    {\taxGroupTitle};
  \draw[\taxColor,line width=.85pt]
    (\taxRootBranchX,0pt |- taxCap\thetaxRowNumber.west)--(taxCap\thetaxRowNumber.west);
  \ifnum\taxAlignMiddleChild=1
    \draw[\taxColor,line width=.8pt]
      (taxCap\thetaxRowNumber.east)--(taxSub\thetaxMiddleRow.west);
    \draw[\taxColor,line width=.8pt]
      (\taxBranchX,0pt |- taxSub\thetaxFirstRow.west)--(\taxBranchX,0pt |- taxSub\thetaxRowNumber.west);
    \foreach \r in {\thetaxFirstRow,...,\thetaxRowNumber}{%
      \ifnum\r=\value{taxMiddleRow}\else
        \draw[\taxColor,line width=.8pt]
          (\taxBranchX,0pt |- taxSub\r.west)--(taxSub\r.west);
      \fi}
    \path let \p1=(taxCap\thetaxRowNumber.east),
      \p2=(taxSub\thetaxMiddleRow.west) in
      \pgfextra{\pgfmathsetmacro{\taxStemDelta}{\y1-\y2}%
        \typeout{TAXONOMY STEM group-end=\thetaxRowNumber, middle-row=\thetaxMiddleRow, delta=\taxStemDelta}};
  \else
    \draw[\taxColor,line width=.85pt]
      (taxCap\thetaxRowNumber.east)--(\taxBranchX,0pt |- taxCap\thetaxRowNumber.east);
    \draw[\taxColor,line width=.85pt]
      (\taxBranchX,0pt |- taxSub\thetaxFirstRow.west)--(\taxBranchX,0pt |- taxSub\thetaxRowNumber.west);
    \foreach \r in {\thetaxFirstRow,...,\thetaxRowNumber}{%
      \draw[\taxColor,line width=.8pt]
        (\taxBranchX,0pt |- taxSub\r.west)--(taxSub\r.west);}
  \fi
  \begin{pgfonlayer}{background}
    \path[fill=\taxColor!6,draw=\taxColor!22,line width=.55pt,rounded corners=4pt]
      (\taxRegionLeft,-\taxTop) rectangle (\dimexpr\taxWidth-.4pt\relax,-\taxBottom);
  \end{pgfonlayer}%
  \setlength{\taxY}{\dimexpr\taxBottom+\taxGroupGap\relax}%
}

\newcommand{\SurveyTaxonomyFinish}{%
  \coordinate (taxRootMidpoint) at
    ($(taxCap\taxFirstGroup.center)!0.5!(taxCap\thetaxRowNumber.center)$);
  \coordinate (taxRootJunction) at (\taxRootBranchX,0pt |- taxRootMidpoint);
  \pgfextracty{\taxFirstGroupY}{\pgfpointanchor{taxCap\taxFirstGroup}{center}}%
  \pgfextracty{\taxLastGroupY}{\pgfpointanchor{taxCap\thetaxRowNumber}{center}}%
  \setlength{\taxRootSpan}{\dimexpr\taxFirstGroupY-\taxLastGroupY\relax}%
  \node[fill=taxBlue,draw=taxBlue,text=white,rounded corners=3.5pt,
    rotate=90,line width=.8pt,inner xsep=9pt,inner ysep=4pt,outer sep=0pt,
    minimum width=\taxRootSpan,minimum height=20pt,
    font=\TaxonomySans\bfseries\fontsize{10}{12}\selectfont]
    (taxRoot) at (10.4pt,0pt |- taxRootMidpoint) {\taxRootTitle};
  \draw[taxBlue!75,line width=.85pt]
    (\taxRootBranchX,0pt |- taxCap\taxFirstGroup.west)--(\taxRootBranchX,0pt |- taxCap\thetaxRowNumber.west);
  \draw[taxBlue!75,line width=.85pt] (taxRoot.south)--(taxRootJunction);
  \typeout{TAXONOMY width=\the\taxWidth, bottom=\the\taxBottom, rows=\thetaxRowNumber}%
  \end{tikzpicture}%
}
\fi

%% file: tables/rsi_scaffold.tex

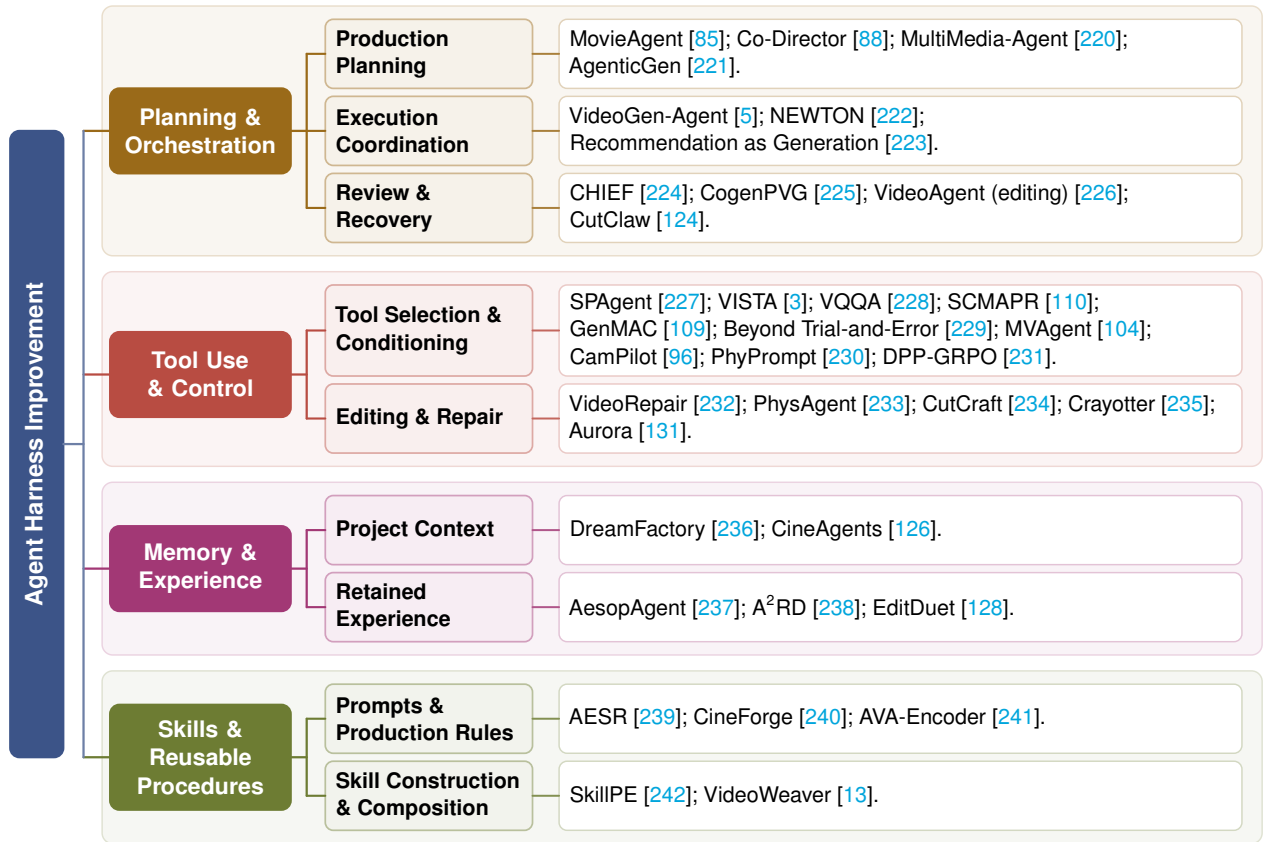
\begin{figure*}[t]
\centering
\input{figures/rsi-scaffold-taxonomy}
\captionsetup{position=bottom}
\caption{\textbf{Functional taxonomy of agent harness improvement
in video generation.} Methods are grouped by their primary
contribution to planning and orchestration, tool use and
control, memory and experience, or reusable procedures.}
\label{fig:rsi-scaffold-taxonomy}
\end{figure*}

%% file: figures/rsi-scaffold-taxonomy.tex
\input{figures/taxonomy-style}
\begingroup
\SurveyTaxonomyLandscape
\renewcommand{\taxGroupWidth}{63pt}
\renewcommand{\taxGroupXPad}{2.5pt}
\renewcommand{\taxCategoryWidth}{70pt}
\renewcommand{\taxCategoryLeft}{119pt}
\renewcommand{\taxBranchX}{112pt}
\renewcommand{\taxWorksLeft}{207pt}
\newcommand{\rscWork}[2]{\SurveyTaxonomyWork{#1}{#2}}
\SurveyTaxonomyStart{Agent Harness Improvement}
\SurveyTaxonomyGroup{taxAmber}{Planning \&\\Orchestration}
\SurveyTaxonomyRow{Production Planning}{%
  \rscWork{MovieAgent}{wu2025movieagent};
  \rscWork{Co-Director}{codirector2026};
  \rscWork{MultiMedia-Agent}{multimediaagent2026};
  \rscWork{AgenticGen}{agenticgen2026}.}
\SurveyTaxonomyRow{Execution Coordination}{%
  \rscWork{VideoGen-Agent}{li2026videogenagent};
  \rscWork{NEWTON}{feng2026newton};
  \rscWork{Recommendation as Generation}{cheng2026rag}.}
\SurveyTaxonomyRow{Review \& Recovery}{%
  \rscWork{CHIEF}{chief2026};
  \rscWork{CogenPVG}{cogenpvg2026};
  \rscWork{VideoAgent (editing)}{zhou2026videoagentediting};
  \rscWork{CutClaw}{zhao2026cutclaw}.}
\SurveyTaxonomyEndGroup

\SurveyTaxonomyGroup{taxCoral}{Tool Use\\\& Control}
\SurveyTaxonomyRow{Tool Selection \& Conditioning}{%
  \rscWork{SPAgent}{spagent2024};
  \rscWork{VISTA}{vista2025};
  \rscWork{VQQA}{song2026vqqa};
  \rscWork{SCMAPR}{scmapr2026};
  \rscWork{GenMAC}{huang2026genmac};
  \rscWork{Beyond Trial-and-Error}{beyondtrial2026};
  \rscWork{MVAgent}{kong2026mvagent};
  \rscWork{CamPilot}{campilot2026};
  \rscWork{PhyPrompt}{phyprompt2026};
  \rscWork{DPP-GRPO}{dppgrpo2025}.}
\SurveyTaxonomyRow{Editing \& Repair}{%
  \rscWork{VideoRepair}{lee2026videorepair};
  \rscWork{PhysAgent}{physagent2026};
  \rscWork{CutCraft}{zeng2026cutcraft};
  \rscWork{Crayotter}{yan2026grpb};
  \rscWork{Aurora}{aurora2026}.}
\SurveyTaxonomyEndGroup

\SurveyTaxonomyGroup{taxBerry}{Memory \&\\Experience}
\SurveyTaxonomyRow{Project Context}{%
  \rscWork{DreamFactory}{xie2024dreamfactory};
  \rscWork{CineAgents}{cineagents2026}.}
\SurveyTaxonomyRow{Retained Experience}{%
  \rscWork{AesopAgent}{wang2024aesopagent};
  \rscWork{A\textsuperscript{2}RD}{song2026a2rd};
  \rscWork{EditDuet}{sandoval2025editduet}.}
\SurveyTaxonomyEndGroup

\SurveyTaxonomyGroup{taxOlive}{Skills \&\\Reusable\\Procedures}
\SurveyTaxonomyRow{Prompts \& Production Rules}{%
  \rscWork{AESR}{aesr2026};
  \rscWork{CineForge}{cineforge2026};
  \rscWork{AVA-Encoder}{li2026avaencoder}.}
\SurveyTaxonomyRow{Skill Construction \& Composition}{%
  \rscWork{SkillPE}{skillpe2026};
  \rscWork{VideoWeaver}{wei2026videoweaver}.}
\SurveyTaxonomyEndGroup
\SurveyTaxonomyFinish
\endgroup

%% file: sections/06_agent_design.tex
\section{Agent Design for Video Generation}
\label{sec:agent-design}

\begin{figure}[htbp]
\centering
\includegraphics[width=\linewidth]{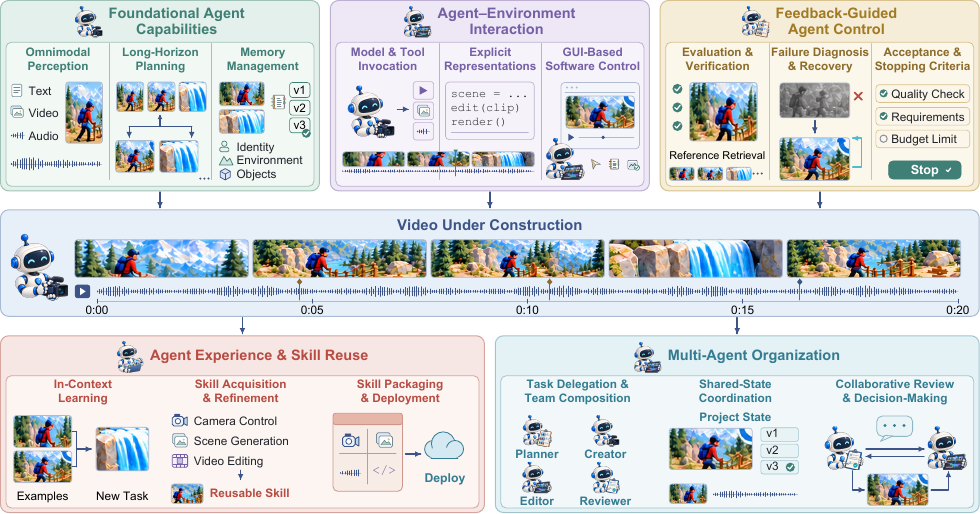}
\caption{\textbf{Agent design for video generation.}
Five design aspects surround a shared video under construction.
Foundational capabilities, environment interaction, and feedback-guided
control support audiovisual production. Experience and skill reuse support
adaptation, while multi-agent organization enables coordination.
Scenes and timing are conceptual illustrations.}
\label{fig:agent_design_teaser}
\end{figure}

Agent design for video generation connects perceptual and reasoning capabilities with generation interfaces, feedback, reusable experience, and collaboration (Figure~\ref{fig:agent_design_teaser}). The production, world-model, and improvement settings reviewed in above sections impose different requirements on this architecture. An isolated clip can often be produced through a small number of model calls, while a long narrative or continuing interaction can require persistent state, explicit controls, and recovery from partial failures. Reliable design allocates these mechanisms to the constraints that materially affect task success.

Figure~\ref{fig:agent_design_table} organizes these five design aspects into their constituent dimensions and representative works. Perception, memory, and planning establish the information available to the controller; model invocation, coding, and GUI interaction determine how decisions are realized. Feedback governs assessment, recovery, and acceptance, while skill reuse and multi-agent organization influence how experience and responsibility are distributed. These design choices should be assessed through their effects on output quality, controllability, cost, and transfer across complete workflows.

Together, these design aspects govern how a video agent pursues a goal $g$ through successive generation and revision decisions. At decision cycle $t$, the controller $\pi_\phi$ receives context $c_t$ assembled from the goal, current observations, and relevant production state. This state, $s_t=(q_t,m_t,b_t,e_t)$ in the formulation of Section~2, tracks task and artifact records $q_t$, memory $m_t$, remaining budget $b_t$, and runtime records $e_t$. When generation is invoked, the selected prompts, references, and controls form the conditions $\mathbf c$ supplied to the video generator $p_\theta$.

\input{figures/agent_design_map}

\subsection{Foundational Agent Capabilities}
\label{sec:agent-foundations}
A video agent requires perceptual evidence, persistent state, and reasoning over interdependent decisions. These capabilities must accommodate temporal change, audiovisual correspondence, and uncertain generation outcomes. Their required granularity depends on the task: isolated clips may need little history, whereas long narratives and continuing interactions require selective access to earlier constraints.

\textbf{Omnimodal perception.} Omnimodal perception interprets text, images, video, and audio to recover task-relevant entities, events, relationships, and constraints. Source material defines what should be generated, while intermediate outputs reveal what has been realized. Application-specific systems emphasize source fidelity and creative intent~\cite{animdirector2024,automvrealestate2025,jeon2026atvg,preacher2025,presentagent2025}; social and interactive settings additionally require interpreting preceding behavior and communicative context~\cite{zhang2026interacvid,peng2025svbench,chief2026}. Perceptual state should distinguish requested properties, observed content, and uncertain inferences so that later decisions do not treat an assumption as an established fact.

Cross-modal interpretation associates observations with shared entities and temporal intervals. Musical structure, lyrics, and beats can constrain shot duration and edit timing~\cite{automv2025,agentmv2026,mvweaver2026}, while audiovisual production connects speech, expressions, subtitles, and background sound~\cite{wu2025movieagent,mavis2025,cineagi2026}. Correspondence does not always require simultaneity: voice-over, reaction shots, and off-screen events can intentionally separate sound from visible action. Geometric state and tool reports provide supplementary observations whose provenance should remain distinct from rendered evidence.

The main design trade-off is perceptual coverage versus acquisition cost. Sparse frames and event summaries support many semantic checks, but short-lived motion defects and sound timing may require denser observations. Targeted questions, adaptive sampling, and criterion-specific inspection provide selective alternatives~\cite{song2026vqqa,liu2025aigvemacs,zeng2026videoargus}. Evidence acquisition is most useful when it resolves uncertainty that can change a generation, preservation, or acceptance decision. The controller should retain the source and temporal scope of each finding and revisit the media when compressed summaries cannot support the required judgment.

\textbf{Memory management.} Memory management represents, updates, and retrieves the information needed beyond the current observation. Visual references preserve appearance, while structured records capture entity state, shot dependencies, accepted versions, and production history~\cite{long2024videostudio,hu2024storyagent,xie2024dreamfactory,videomemory2026,filmworld2026,liu2026reca,huang2026vimax,soap2soap2026}. Stable identities and locations must remain distinguishable from changing poses, clothing, viewpoints, and interactions~\cite{mondal2026canvas,cotrisygen2026}. Multi-turn editing also requires retaining earlier changes and user preferences~\cite{lee2026memoryv2v,videocreator2026,liang2025univa}. Raw media offers detailed, traceable evidence; structured state and summaries reduce access cost but may omit information needed for a later decision.

Admission and updates require reliability checks because generated content can introduce errors into subsequent conditions. GroundShot~\cite{groundshot2026}, StoryMem~\cite{zhang2025storymem}, and Memento~\cite{wei2026memento} illustrate selective verification or retention of references. A practical memory should record whether an entry is a source requirement, an accepted output, a superseded version, or a rejected candidate. Changes to a reference should expose their downstream dependencies, allowing affected shots to be reconsidered while preserving unrelated accepted content.

Retrieval constructs the context available to the next decision. Entity-indexed, semantic, spatial, and bounded-history mechanisms offer different precision--cost trade-offs~\cite{vandersanden2026emvid,an2026onestory,yu2025contextmemory,li2025vmem,xiao2025worldmem,wu2025spatialmemory,chen2025longanimation,peng2026freemem,kim2026memrope,xue2026ringforcing}. Model-side retention~\cite{wu2025corgi,luo2026filmweaver} and sparse retrieval for video understanding~\cite{song2024moviechat,he2024malmm,fan2024videoagent,wang2025videotree} provide enabling mechanisms. Extending general agent memory~\cite{park2023generativeagents} to video therefore requires policies for selecting relevant evidence, compressing it without losing critical relations, and preserving provenance. Thus $c_t=\mathcal C_{\mathcal H}(g,o_t,s_t)$ combines the goal, current observation $o_t$, and selected state within the controller's context limit. Provenance and version records in $m_t$ and $q_t$ allow omitted evidence to be retrieved when later decisions require it.

\textbf{Long-horizon planning.} Long-horizon planning decomposes an objective into scenes, shots, segments, and supporting operations while preserving prerequisites and temporal constraints. Narrative order can differ from production order when a later shot requires a reliable reference that must be created first. Time-varying prompts, layouts, entity groups, and trajectories expose different levels of commitment~\cite{xu2023freebloom,lian2024llmgrounded,lin2024videodirectorgpt,motionagent2025}; camera and shot methods add composition, blocking, transitions, and montage constraints~\cite{cameraartist2026,campilot2026,zhou2026dramadirector,zeng2026cutcraft}. The plan should connect each task to its inputs, expected outputs, and dependent decisions.

A detailed global plan is useful when dependencies are known and consistency requirements are strong. Local planning with retrieved state is more flexible when stochastic outputs determine the conditions of subsequent shots. Candidate search is justified when alternative realizations differ materially and can be compared within the available budget. AniMaker~\cite{animaker2025} and Co-Director~\cite{codirector2026} illustrate exploration in video production; Language Agent Tree Search~\cite{zhou2024lats} supplies a general search foundation. Resource estimates should include generation, verification, and possible rework, directing additional trials toward decisions that strongly affect later content.

The controller must also allocate its own reasoning context. Memory-efficient attention reduces memory traffic~\cite{dao2022flashattention}, but a longer input does not guarantee reliable use of distant evidence~\cite{liu2023lostmiddle}. Retrieved constraints and explicit dependencies can support bounded planning calls while keeping the global objective accessible. The appropriate design balances the cost of constructing and maintaining a detailed plan against the risk of omissions, inconsistent state, and expensive downstream repair.

\subsection{Agent--Environment Interaction}
\label{sec:agent-interaction}
A video agent interacts with generators, reference stores, editors, renderers, and, in interactive settings, a world model. Interaction can take the form of invoking a task-level capability, constructing executable controls, or operating a visible software interface. Ordinary model calls remain central: an agent can choose conditions, obtain candidates, and revise generation without writing code or using a GUI. Interface design determines the control available to the agent and the evidence returned after execution.

\textbf{Model and tool invocation.} Task-level invocation selects an available capability and supplies its inputs: a text-to-video request, a reference-conditioned shot, an editing operation, an inspection query, or a world-model control. A single generator is appropriate when it exposes the needed controls; additional models and tools can address limitations in identity, motion, sound, or editing. VideoGen-Agent~\cite{li2026videogenagent} and MVAgent~\cite{kong2026mvagent} illustrate policies that choose augmentation, generation, and reference operations. A useful tool set exposes information gathering and candidate inspection alongside synthesis, enabling uncertainty to be resolved before an expensive commitment.

Conditioning interfaces determine how much of the intended behavior can be communicated. Reference videos or image pairs can convey motion, physical behavior, perspective, and appearance transformations~\cite{zhang2024vidit,liu2024aicl,bian2025videoasprompt,chen2026viper,song2025worldwander,qu2026vincienext}; context layout and efficient conditioning affect the cost of providing these signals~\cite{fei2024videoincontext,he2025fulldit2}. The agent must match the reference to the requested relation and preserve target-specific content. Fine-grained parameters improve flexibility but increase selection errors, while higher-level tools simplify reasoning at the cost of hidden internal choices. ComfyMind~\cite{comfymind2025} illustrates composition through higher-level functional modules.

Editing systems expose operations at different levels of abstraction, including semantic timeline changes and executable processing commands~\cite{wang2024lave,sandoval2025editduet,zhou2026videoagentediting,zhao2026cutclaw,ellmpeg2026}. Reliable invocation requires validated arguments, identifiable input and output versions, and recoverable asynchronous job state. For continuing world interaction, a control must additionally be linked to the interval and state in which it takes effect. The interface should support checking both whether an operation completed and whether its output satisfies the task; the world-model settings are examined in Section~4.

\textbf{Coding and program execution.} Coding constructs or revises the executable logic through which content is produced. Programs and declarative specifications expose geometry, timing, animation, and data relationships as variables and operations that can be inspected and changed. ComfyUI~\cite{comfy2026nodes} provides an editable graph of executable nodes; scripting and graph construction offer different ways to compose such operations. The design choice is which properties require explicit control and which visual details can be delegated to neural synthesis. This is particularly consequential for precise event timing, physical layouts, and instructional content.

GPT4Motion~\cite{lv2024gpt4motion} and VideoCoCo~\cite{videococo2026} combine procedural dynamics with generative appearance synthesis. Editable blockouts in Scene2Scene~\cite{mei2026scene2scene}, the fal workflow~\cite{odin2026clayrender}, and Seedance's clay-reference interface~\cite{bytedance2026seedance25} illustrate related uses of coarse spatial control with different degrees of human involvement. Scene and physics methods expose inspectable conditions at the cost of construction and simulation~\cite{cai2026liver,chen2026wnm,feng2026newton,nguyen2026physplan,he2024kubrick,moregen2025}. Executable teaching, data, and browser graphics support precise visual structure and timing~\cite{manimator2025,chen2025code2video,teachmaster2026,beyondendtoend2026,datamagic2026,tseng2024keyframer,shirakawa2025mggen,shen2024datastory}, with adjacent foundations in visual programming and graphics decomposition~\cite{gupta2023visprog,suris2023vipergpt,multiagentcad2026,blenderagentic2025}.

A hybrid workflow can retain precise structure in an executable scene program $P$ and delegate appearance synthesis to a neural generator:
\begin{equation}
\mathbf x \sim p_\theta\bigl(\cdot\mid (\mathbf c^{\mathrm{base}},\operatorname{Render}(P)),z\bigr).
\label{eq:agent-design-hybrid-execution}
\end{equation}
Here $\operatorname{Render}(P)$ supplies a draft video or structural guide, while $\mathbf c^{\mathrm{base}}$ contains the remaining conditions; together they instantiate $\mathbf c$ from Section~2, with $z$ controlling sampling randomness. Neural refinement may alter geometry or timing, so the resulting video $\mathbf x$ must be checked against the constraints encoded in $P$.

Execution should preserve the correspondence between program controls, runtime state, and rendered intervals. Errors and simulation traces expose process failures, while inspection establishes whether a valid program achieved the intended result. Access to both the specification and final video supports targeted diagnosis, local editing, and reusable parameterized procedures. These benefits must justify the effort of constructing, executing, and checking the explicit representation.

\textbf{GUI-based software control.} GUI control is valuable when required production operations are unavailable through a stable programmatic interface or depend on inspecting the application's visible state. The agent observes previews, timelines, tracks, selections, and parameter panels and acts through mouse, keyboard, and shortcuts. General interface-control research provides relevant foundations~\cite{niu2024screenagent,xie2024osworld,cheng2024videogui,openai2026gpt6astra,sun2026agentslastexam}, while CutVerse~\cite{hu2026cutverse} evaluates video-production applications. Temporal targeting is especially demanding: an apparently successful interaction can affect the wrong clip, track, or interval.

GUI operations trade broad application coverage for sensitivity to layout, hidden state, and interaction history. Scrubbing and playback may be necessary to verify effects that a static screenshot cannot establish. AssistEditor~\cite{gao2024assisteditor} illustrates combining visual and programmatic interaction. Hybrid execution can use direct tools for repeatable processing and the GUI for application-specific edits, provided asset versions, timecodes, and project state remain consistent across interfaces. Recovery should start from an observed state rather than assume that a click or shortcut had the intended effect.

\subsection{Feedback-Guided Agent Control}
\label{sec:agent-feedback-control}
Feedback controls whether an agent accepts a result, gathers more evidence, corrects an artifact, or changes the remaining workflow. The measurements reviewed in Section~5 become operationally useful when they are attached to identifiable constraints and available interventions. Effective control distinguishes assessing a result, diagnosing its failure, and deciding whether further effort is warranted.

\textbf{Evaluation and verification.} Evaluation estimates perceptual quality and task fulfillment, while verification checks explicit requirements and execution conditions. Compiler or tool success cannot establish correct geometry, narrative fulfillment, or audiovisual timing. Fine-grained assessment and temporal localization provide evidence about content defects~\cite{zhang2026vqinsight,chinchure2025spotlight,chen2026directorbench}; Evaluation Agent~\cite{zhang2025evaluationagent} illustrates adaptive evidence collection. Simulation state and executable predicates provide complementary checks for physics and animation~\cite{physagent2026,simworlds2026,ma2025mover}, and rendering feedback can guide code revision~\cite{silva2026manimagent}.

The inspection procedure should match the requirement being tested. Cheap validity checks can prune unusable candidates before expensive multimodal review, while denser temporal inspection is justified for transient or relational errors. Findings should identify the relevant entities, intervals, criteria, and uncertainty. Learned critics can disagree or inherit model biases, so subjective goals and ambiguous cases may require human assessment. Feedback quality is best judged by whether it supports reliable acceptance or correction decisions, alongside its agreement with independent evaluation.

\textbf{Failure diagnosis and recovery.} Diagnosis connects an observed discrepancy to a plausible source: the request, reference, plan, program, generator configuration, or execution operation. SCMAPR~\cite{scmapr2026} checks rewritten prompts before generation, whereas searching seeds and guidance settings can address failures not resolved by prompt revision~\cite{beyondtrial2026}. Critics and revision loops connect different defects to subsequent interventions~\cite{vista2025,song2026vqqa,huang2026genmac,soap2soap2026,groundshot2026}. A useful diagnosis must support an operation that the system can perform and distinguish uncertainty about the cause from confirmed failure evidence.

Recovery selects the smallest intervention likely to correct the problem while preserving accepted content. Options include local editing, regeneration, reference replacement, parameter adjustment, or revision of an upstream plan. Feedback from environment execution~\cite{soni2024videoagentself}, rendered presentation artifacts~\cite{deeppresenter2026}, and creator preferences~\cite{chief2026} complements visual critique. ReAct~\cite{yao2023react} and Reflexion~\cite{shinn2023reflexion} provide general foundations for using observations and retained feedback in later decisions. Local repair limits immediate cost but may leave a faulty dependency intact; broader regeneration can address the cause while introducing new variation. A recoverable accepted state and checks for collateral changes are therefore essential to either strategy.

\textbf{Acceptance and stopping criteria.} Acceptance should be based on satisfied task constraints and sufficient quality evidence; stopping should also account for diminishing improvement, retry limits, latency, and remaining budget. Candidate tournaments and selective critique~\cite{vista2025} can reduce assessment cost, while global allocation across segments~\cite{agentmv2026} helps direct trials toward influential or difficult shots. Further inspection or generation is justified when its expected decision value exceeds its cost. Already accepted material should be reconsidered when a relevant dependency changes, not merely because more sampling is available.

Acceptance checks use the latest artifact and verification records in $q_t$; stopping also considers the remaining budget $b_t$ and execution history in $e_t$. Budget exhaustion does not establish successful completion: the system should record whether it accepted the artifact, stopped with unresolved constraints, or escalated for user judgment. Pending operations must be completed or explicitly cancelled before closure. Hard requirements support explicit checks, while aesthetic goals may need bounded comparisons or human selection. Repeated critique without an achievable intervention is a reason to change strategy or stop. Comparisons should report rejected candidates, incomplete tasks, and total generation and evaluation cost, so additional sampling is not attributed to a better controller.

\subsection{Agent Experience and Skill Reuse}
\label{sec:agent-experience}
\label{sec:agent-context-skills}
Experience can supply context for current inference, be consolidated into reusable skills, or support later policy updates. Building on the improvement mechanisms in Section~5, reliable reuse depends on what is retained, how the procedure is validated, and whether its assumptions remain valid in the next task. Skill acquisition changes available knowledge; packaging and deployment make that knowledge accessible to the agent and its tools.

\textbf{In-context learning.} In-context learning uses supplied examples to adapt inference while keeping controller parameters fixed~\cite{brown2020fewshot}. DreamRunner~\cite{wang2024dreamrunner} illustrates planning from an example, while A$^2$RD~\cite{song2026a2rd} contrasts successful and failed refinement cases. Contextual examples are useful when tasks change quickly, relevant precedents are available, or model weights cannot be updated. They are less dependable when retrieved cases omit critical conditions, conflict with the current request, or consume more context than the controller can effectively use.

Examples in $c_t$ should preserve the task, the decision or procedure, and its observed result, allowing the controller $\pi_\phi$ to adapt its decisions while $\phi$ remains fixed. Memory supplies candidate records; the controller determines which relationships apply to the current story, shot, or tool choice. Evaluation should compare downstream outcomes with and without relevant examples under matched tasks, initial observations, available tools, generator parameters $\theta$, and generation budgets. Length-matched irrelevant examples help distinguish useful experience from prompt-length effects. Following fixed instructions or replaying stored code alone does not establish adaptation from examples.

\textbf{Skill acquisition and refinement.} Recurring solutions can be abstracted into prompting rules, tool-use procedures, parameterized programs, or generation workflows. ExpeL~\cite{zhao2024expel}, Agent Workflow Memory~\cite{wang2025awm}, and Voyager~\cite{wang2024voyager} provide general precedents for extracting or retaining reusable knowledge. Video systems develop experience stores, model-specific playbooks, composition skills, and validated production updates~\cite{sam2026actionagent,wang2024aesopagent,aesr2026,wei2026videoweaver,cineforge2026}. Useful skills retain the input assumptions, asset requirements, operations, constraints, and failure cases that explain when the procedure works.

Refinement should address recurrent failure patterns and be evaluated on tasks distinct from those used to propose the change. Broad recipes transfer more easily but leave execution details unresolved; narrowly tuned procedures can achieve stronger control while depending on particular models or tool versions. When stable decision errors require a learned policy change, parameter training offers another mechanism~\cite{yang2026spiral,cheng2026rag,agenticgen2026,li2026videogenagent}. Its costs and benefits should be attributed separately from improved retrieval or reusable instructions. Retaining earlier capabilities after each update is as important as improving the latest task.

\textbf{Skill packaging and deployment.} A deployable skill combines procedural guidance with the resources and checks needed to execute it. The Agent Skills specification~\cite{agentskills2026spec} illustrates a package containing a \texttt{SKILL.md} file with metadata and instructions, plus optional scripts, references, and assets loaded as needed. This can express a shot-construction procedure, blockout-to-video workflow, or editing routine. Parameterized programs are useful when timing, geometry, and local-edit controls must remain explicit. The package should identify its dependencies, supported inputs, and expected outputs; its format alone does not establish learned competence or transfer.

Skills supply knowledge about when and how to use a capability, whereas tools expose executable operations. MCP~\cite{mcp2025tools} standardizes tool discovery and invocation through declared argument schemas. Comfy Skills~\cite{comfy2026skills} packages workflow guidance and commands, while Comfy MCP~\cite{comfy2026mcp} exposes workflow discovery, execution, and output retrieval. Deployment must reconcile the skill's assumptions with available nodes, models, assets, parameter semantics, and permissions. Tests on new tasks and after tool or generator changes should establish both portability and retention, including whether the resulting video still satisfies the skill's acceptance checks.

\subsection{Multi-Agent Organization}
\label{sec:agent-organization}
Multiple agents are useful when specialization, parallel work, or independent review compensates for the cost of routing information and synchronizing state. A single controller can remain preferable when dependencies are tight and the required tools fit one context. Team design must establish who is responsible for a task, what information collaborators share, and how competing assessments lead to an accepted result.

\textbf{Task delegation and team composition.} Delegation assigns a collaborator a task, relevant context, permitted changes, and an expected deliverable. Role-based video systems~\cite{yuan2024mora,wu2025movieagent,mavis2025} organize specialized production components, while FRAMEWORKERS~\cite{frameworkers2026} illustrates dynamic routing as tasks and available capabilities evolve. Fixed roles simplify accountability for stable workflows; dynamic teams accommodate heterogeneous requests but introduce routing and context-selection errors. Studies of agent scaling, routing, and pruning~\cite{li2024moreagents,yue2025masrouter,li2025agentdropout} motivate selecting team size according to task demands rather than maximizing the number of agents.

A delegation request can take the form $a_t=(\texttt{delegate},i,g_t^{(i)},c_t^{(i)})$, where $i$ identifies the collaborator, $g_t^{(i)}$ its subgoal, and $c_t^{(i)}$ its scoped context. The request must belong to the currently admissible action set $\mathcal A_t$ and respect its permissions and budget constraints. Runtime records $e_t$ link delayed results to their requests, preserving task ownership until the returned artifacts are verified and incorporated into shared state.

ComfyUI-based systems distinguish the team that constructs a workflow from the graph that executes it. ComfyAgent~\cite{comfybench2024} uses cooperating agents to construct workflows, ComfyGPT~\cite{comfygpt2025} specializes agents in reformulation, graph construction, refinement, and execution, and ComfyMind~\cite{comfymind2025} combines modular composition with planning and feedback. Their evidence concerns general generative-workflow construction; applying these organizations to video also requires temporal and audiovisual assessment. A computation node is not an independent decision-maker merely because it belongs to a graph.

Video-oriented projects also use the term \emph{agent swarm}. FilmSwarm~\cite{filmswarm2026} describes a director-led pre-production pipeline with dynamically instantiated specialists, shared references, and roundtable deliberation; its documented deliverables include a video production plan. SwarmZero's Livepeer video-generator example~\cite{swarmzero2026video} coordinates script, scene, image, video, editing, and upload agents through a manager. These project and documentation examples establish integration patterns rather than controlled evidence of decentralized intelligence or improved final-video quality. Swarm configurations need explicit delegation and communication policies, and should demonstrate gains over a single controller with comparable tools and budget.

\textbf{Shared-state coordination.} Collaborators need consistent asset identities, accepted shot versions, temporal constraints, and dependency records. Shared specifications and persistent state~\cite{chen2026cinecrew,agentmv2026,hollywoodtown2025} reduce repeated context reconstruction but require clear ownership and update propagation. Typed media references~\cite{kong2026mvagent} preserve details that prose can omit. Parallel slide, scene, or shot production~\cite{paper2video2025,datamagic2026,soap2soap2026} is most effective when outputs can be checked against stable shared conditions; coupled changes to the same state may require sequential execution. An accepted update should identify which downstream tasks must revisit their inputs.

Communication format should preserve the information needed by the recipient. Text and structured artifacts support inspection and broad compatibility. LatentMAS~\cite{zou2025latentmas} and Vision Wormhole~\cite{liu2026visionwormhole} explore latent message interfaces with compatibility or alignment requirements. Shared world representations in WorldWeaver~\cite{worldweaver2026} and OmniDrive~\cite{meng2026omnidrive} couple generated viewpoints rather than production-role decisions. Their relevance lies in how information is preserved and propagated; improved coordination must still be measured in the target workflow. More detailed messages help only when recipients can use them without excessive context and communication cost.

\textbf{Collaborative review and decision-making.} Collaborative review distributes the assessment of proposed plans or artifacts and returns actionable feedback to the responsible component~\cite{huang2026genmac,sandoval2025editduet,onesentence2026,hu2025filmagent}. Independent perspectives can expose overlooked errors, but correlated critics and repeated debate may increase cost without improving judgment. Reviewer assignments should specify the criterion, affected content, required evidence, and authority to recommend or accept a change. A clear decision rule is needed when agents disagree, particularly for subjective creative goals.

Useful review produces a verifiable revision or an explicit acceptance decision. Critiques should preserve the relation between the observed defect and proposed intervention, while the reviser reports what changed and what was retained. The shared state is updated after checking the revised result against the relevant constraints. Evaluation should compare independent review, manager arbitration, and other decision procedures at matched budgets, including latency, unresolved disagreement, and collateral changes. Escalation or termination rules are necessary when further discussion cannot resolve the task.

Effective video agents combine control precision with selective use of computation. Explicit representations are valuable where geometry, timing, or content must be inspectable; richer observations are valuable where uncertainty can change a decision; and reusable skills or collaboration are valuable where recurring procedures and separable responsibilities justify their overhead. The resulting architecture should make accepted state, failure evidence, resource use, and retained improvements traceable, allowing system-level benefits to be distinguished from stronger generators or additional sampling.

%% file: figures/agent_design_map.tex
\begin{figure*}[!htbp]
\centering
\input{figures/taxonomy-style}
\begingroup
\SurveyTaxonomyLandscape
\renewcommand{\taxAlignMiddleChild}{1}
\renewcommand{\taxBodyFont}{\TaxonomyArial\mdseries\fontsize{7.4}{8.2}\selectfont
  \hyphenpenalty=10000\exhyphenpenalty=10000}
\renewcommand{\taxCategoryFont}{\TaxonomyArial\bfseries\fontsize{7.6}{8.7}\selectfont
  \hyphenpenalty=10000\exhyphenpenalty=10000}
\renewcommand{\taxGroupFont}{\TaxonomyArial\bfseries\fontsize{8.2}{9.6}\selectfont
  \hyphenpenalty=10000\exhyphenpenalty=10000}
\renewcommand{\taxGroupCenter}{65pt}
\renewcommand{\taxGroupWidth}{56pt}
\renewcommand{\taxGroupXPad}{2.5pt}
\renewcommand{\taxCategoryWidth}{62pt}
\renewcommand{\taxCategoryLeft}{105pt}
\renewcommand{\taxBranchX}{98pt}
\renewcommand{\taxWorksLeft}{188pt}
\renewcommand{\taxRowPad}{2pt}
\renewcommand{\taxMinRowHeight}{22pt}
\renewcommand{\taxRowGap}{3.2pt}
\renewcommand{\taxGroupPad}{4pt}
\renewcommand{\taxGroupGap}{5pt}
\definecolor{sagmGroup1}{HTML}{3F7F75}
\definecolor{sagmGroup2}{HTML}{7560A6}
\definecolor{sagmGroup3}{HTML}{A17A23}
\definecolor{sagmGroup4}{HTML}{AF6344}
\definecolor{sagmGroup5}{HTML}{985978}
\SurveyTaxonomyStart{Agent Design for Video Generation}
\SurveyTaxonomyGroup{sagmGroup1}{Foundational\\Agent\\Capabilities}
\SurveyTaxonomyRow{Omnimodal Perception}{%
  Anim-Director~\cite{animdirector2024};
  AutoMV (real estate)~\cite{automvrealestate2025};
  ATVG~\cite{jeon2026atvg};
  Preacher~\cite{preacher2025};
  PresentAgent~\cite{presentagent2025};
  InteracVid~\cite{zhang2026interacvid};
  SVBench~\cite{peng2025svbench};
  CHIEF~\cite{chief2026};
  AutoMV (music)~\cite{automv2025};
  AgentMV~\cite{agentmv2026};
  MVWeaver~\cite{mvweaver2026};
  MovieAgent~\cite{wu2025movieagent};
  MAViS~\cite{mavis2025};
  CineAGI~\cite{cineagi2026};
  VQQA~\cite{song2026vqqa};
  AIGVE-MACS~\cite{liu2025aigvemacs};
  VideoArgus~\cite{zeng2026videoargus}.
}
\SurveyTaxonomyRow{Memory Management}{%
  VideoStudio~\cite{long2024videostudio};
  StoryAgent~\cite{hu2024storyagent};
  DreamFactory~\cite{xie2024dreamfactory};
  VideoMemory~\cite{videomemory2026};
  FilmWorld~\cite{filmworld2026};
  ReCA~\cite{liu2026reca};
  ViMax~\cite{huang2026vimax};
  Soap2Soap~\cite{soap2soap2026};
  CANVAS~\cite{mondal2026canvas};
  EvoState~\cite{cotrisygen2026};
  Memory-V2V~\cite{lee2026memoryv2v};
  VideoCreator~\cite{videocreator2026};
  UniVA~\cite{liang2025univa};
  GroundShot~\cite{groundshot2026};
  StoryMem~\cite{zhang2025storymem};
  Memento~\cite{wei2026memento};
  EM-Vid~\cite{vandersanden2026emvid};
  OneStory~\cite{an2026onestory};
  Context as Memory~\cite{yu2025contextmemory};
  VMem~\cite{li2025vmem};
  WorldMem~\cite{xiao2025worldmem};
  Video World Models with Long-term Spatial Memory~\cite{wu2025spatialmemory};
  LongAnimation~\cite{chen2025longanimation};
  FreeMem~\cite{peng2026freemem};
  MemRoPE~\cite{kim2026memrope};
  Ring Forcing~\cite{xue2026ringforcing};
  Corgi~\cite{wu2025corgi};
  FilmWeaver~\cite{luo2026filmweaver};
  MovieChat~\cite{song2024moviechat};
  MA-LMM~\cite{he2024malmm};
  VideoAgent (understanding)~\cite{fan2024videoagent};
  VideoTree~\cite{wang2025videotree};
  Generative Agents~\cite{park2023generativeagents}.
}
\SurveyTaxonomyRow{Long-Horizon Planning}{%
  Free-Bloom~\cite{xu2023freebloom};
  LLM-grounded Video Diffusion~\cite{lian2024llmgrounded};
  VideoDirectorGPT~\cite{lin2024videodirectorgpt};
  MotionAgent~\cite{motionagent2025};
  Camera Artist~\cite{cameraartist2026};
  CamPilot~\cite{campilot2026};
  DramaDirector~\cite{zhou2026dramadirector};
  CutCraft~\cite{zeng2026cutcraft};
  AniMaker~\cite{animaker2025};
  Co-Director~\cite{codirector2026};
  Language Agent Tree Search~\cite{zhou2024lats};
  FlashAttention~\cite{dao2022flashattention};
  Lost in the Middle~\cite{liu2023lostmiddle}.
}
\SurveyTaxonomyEndGroup

\SurveyTaxonomyGroup{sagmGroup2}{Agent--\\Environment\\Interaction}
\SurveyTaxonomyRow{Model \& Tool Invocation}{%
  VideoGen-Agent~\cite{li2026videogenagent};
  MVAgent~\cite{kong2026mvagent};
  VidIT~\cite{zhang2024vidit};
  AICL~\cite{liu2024aicl};
  Video-As-Prompt~\cite{bian2025videoasprompt};
  VIPER~\cite{chen2026viper};
  WorldWander~\cite{song2025worldwander};
  VINCIE-NExT~\cite{qu2026vincienext};
  Video Diffusion Transformers are In-Context Learners~\cite{fei2024videoincontext};
  FullDiT2~\cite{he2025fulldit2};
  ComfyMind~\cite{comfymind2025};
  LAVE~\cite{wang2024lave};
  EditDuet~\cite{sandoval2025editduet};
  VideoAgent (editing)~\cite{zhou2026videoagentediting};
  CutClaw~\cite{zhao2026cutclaw};
  ELLMPEG~\cite{ellmpeg2026}.
}
\SurveyTaxonomyRow{Coding \& Program Execution}{%
  ComfyUI~\cite{comfy2026nodes};
  GPT4Motion~\cite{lv2024gpt4motion};
  VideoCoCo~\cite{videococo2026};
  Scene2Scene~\cite{mei2026scene2scene};
  fal clay-render workflow~\cite{odin2026clayrender};
  Seedance 2.5~\cite{bytedance2026seedance25};
  LiVER~\cite{cai2026liver};
  World Narrative Model~\cite{chen2026wnm};
  NEWTON~\cite{feng2026newton};
  PhysPlan~\cite{nguyen2026physplan};
  Kubrick~\cite{he2024kubrick};
  MoReGen~\cite{moregen2025};
  Manimator~\cite{manimator2025};
  Code2Video~\cite{chen2025code2video};
  TeachMaster~\cite{teachmaster2026};
  Beyond End-to-End Video Models~\cite{beyondendtoend2026};
  DataMagic~\cite{datamagic2026};
  Keyframer~\cite{tseng2024keyframer};
  MG-Gen~\cite{shirakawa2025mggen};
  Data Director~\cite{shen2024datastory};
  VisProg~\cite{gupta2023visprog};
  ViperGPT~\cite{suris2023vipergpt};
  Multi-Agent CAD Code Generation~\cite{multiagentcad2026};
  Blender Agentic Animation~\cite{blenderagentic2025}.
}
\SurveyTaxonomyRow{GUI-Based Software Control}{%
  ScreenAgent~\cite{niu2024screenagent};
  OSWorld~\cite{xie2024osworld};
  VideoGUI~\cite{cheng2024videogui};
  GPT-6 Astra~\cite{openai2026gpt6astra};
  Agents' Last Exam~\cite{sun2026agentslastexam};
  CutVerse~\cite{hu2026cutverse};
  AssistEditor~\cite{gao2024assisteditor}.
}
\SurveyTaxonomyEndGroup

\SurveyTaxonomyGroup{sagmGroup3}{Feedback-\\Guided\\Agent Control}
\SurveyTaxonomyRow{Evaluation \& Verification}{%
  VQ-Insight~\cite{zhang2026vqinsight};
  Spotlight~\cite{chinchure2025spotlight};
  DirectorBench~\cite{chen2026directorbench};
  Evaluation Agent~\cite{zhang2025evaluationagent};
  PhysAgent~\cite{physagent2026};
  SimWorlds~\cite{simworlds2026};
  MoVer~\cite{ma2025mover};
  ManimAgent~\cite{silva2026manimagent}.
}
\SurveyTaxonomyRow{Failure Diagnosis \& Recovery}{%
  SCMAPR~\cite{scmapr2026};
  Beyond Trial-and-Error~\cite{beyondtrial2026};
  VISTA~\cite{vista2025};
  VQQA~\cite{song2026vqqa};
  GenMAC~\cite{huang2026genmac};
  Soap2Soap~\cite{soap2soap2026};
  GroundShot~\cite{groundshot2026};
  VideoAgent (self-improvement)~\cite{soni2024videoagentself};
  DeepPresenter~\cite{deeppresenter2026};
  CHIEF~\cite{chief2026};
  ReAct~\cite{yao2023react};
  Reflexion~\cite{shinn2023reflexion}.
}
\SurveyTaxonomyRow{Acceptance \& Stopping Criteria}{%
  VISTA~\cite{vista2025};
  AgentMV~\cite{agentmv2026}.
}
\SurveyTaxonomyEndGroup

\SurveyTaxonomyGroup{sagmGroup4}{Agent\\Experience\\\& Skill Reuse}
\SurveyTaxonomyRow{In-Context Learning}{%
  GPT-3~\cite{brown2020fewshot};
  DreamRunner~\cite{wang2024dreamrunner};
  A\textsuperscript{2}RD~\cite{song2026a2rd}.
}
\SurveyTaxonomyRow{Skill Acquisition \& Refinement}{%
  ExpeL~\cite{zhao2024expel};
  Agent Workflow Memory~\cite{wang2025awm};
  Voyager~\cite{wang2024voyager};
  Action Agent~\cite{sam2026actionagent};
  AesopAgent~\cite{wang2024aesopagent};
  AESR~\cite{aesr2026};
  VideoWeaver~\cite{wei2026videoweaver};
  CineForge~\cite{cineforge2026};
  SPIRAL~\cite{yang2026spiral};
  Recommendation as Generation~\cite{cheng2026rag};
  AgenticGen~\cite{agenticgen2026};
  VideoGen-Agent~\cite{li2026videogenagent}.
}
\SurveyTaxonomyRow{Skill Packaging \& Deployment}{%
  Agent Skills~\cite{agentskills2026spec};
  MCP~\cite{mcp2025tools};
  Comfy Skills~\cite{comfy2026skills};
  Comfy MCP~\cite{comfy2026mcp}.
}
\SurveyTaxonomyEndGroup

\SurveyTaxonomyGroup{sagmGroup5}{Multi-Agent\\Organization}
\SurveyTaxonomyRow{Task Delegation \& Team Composition}{%
  Mora~\cite{yuan2024mora};
  MovieAgent~\cite{wu2025movieagent};
  MAViS~\cite{mavis2025};
  FRAMEWORKERS~\cite{frameworkers2026};
  More Agents Is All You Need~\cite{li2024moreagents};
  MasRouter~\cite{yue2025masrouter};
  AgentDropout~\cite{li2025agentdropout};
  ComfyAgent / ComfyBench~\cite{comfybench2024};
  ComfyGPT~\cite{comfygpt2025};
  ComfyMind~\cite{comfymind2025};
  FilmSwarm~\cite{filmswarm2026};
  SwarmZero video swarm~\cite{swarmzero2026video}.
}
\SurveyTaxonomyRow{Shared-State Coordination}{%
  Better Call CineCrew~\cite{chen2026cinecrew};
  AgentMV~\cite{agentmv2026};
  Hollywood Town~\cite{hollywoodtown2025};
  MVAgent~\cite{kong2026mvagent};
  Paper2Video~\cite{paper2video2025};
  DataMagic~\cite{datamagic2026};
  Soap2Soap~\cite{soap2soap2026};
  LatentMAS~\cite{zou2025latentmas};
  Vision Wormhole~\cite{liu2026visionwormhole};
  WorldWeaver~\cite{worldweaver2026};
  OmniDrive~\cite{meng2026omnidrive}.
}
\SurveyTaxonomyRow{Collaborative Review \& Decision-Making}{%
  GenMAC~\cite{huang2026genmac};
  EditDuet~\cite{sandoval2025editduet};
  One Sentence, One Drama~\cite{onesentence2026};
  FilmAgent~\cite{hu2025filmagent}.
}
\SurveyTaxonomyEndGroup

\SurveyTaxonomyFinish
\endgroup
\caption{\textbf{Taxonomy and literature coverage of agent design for video generation.} Five complementary design aspects are organized into fifteen dimensions aligned with Section~\ref{sec:agent-design}. All references cited in the section are listed under their corresponding dimensions, including video-specific methods, general foundations, and relevant tools. Works appear under multiple dimensions when discussed in more than one part of the section.}
\label{fig:agent-design-literature-map}
\label{fig:agent_design_table}
\label{tab:agent-design-literature-map}
\end{figure*}
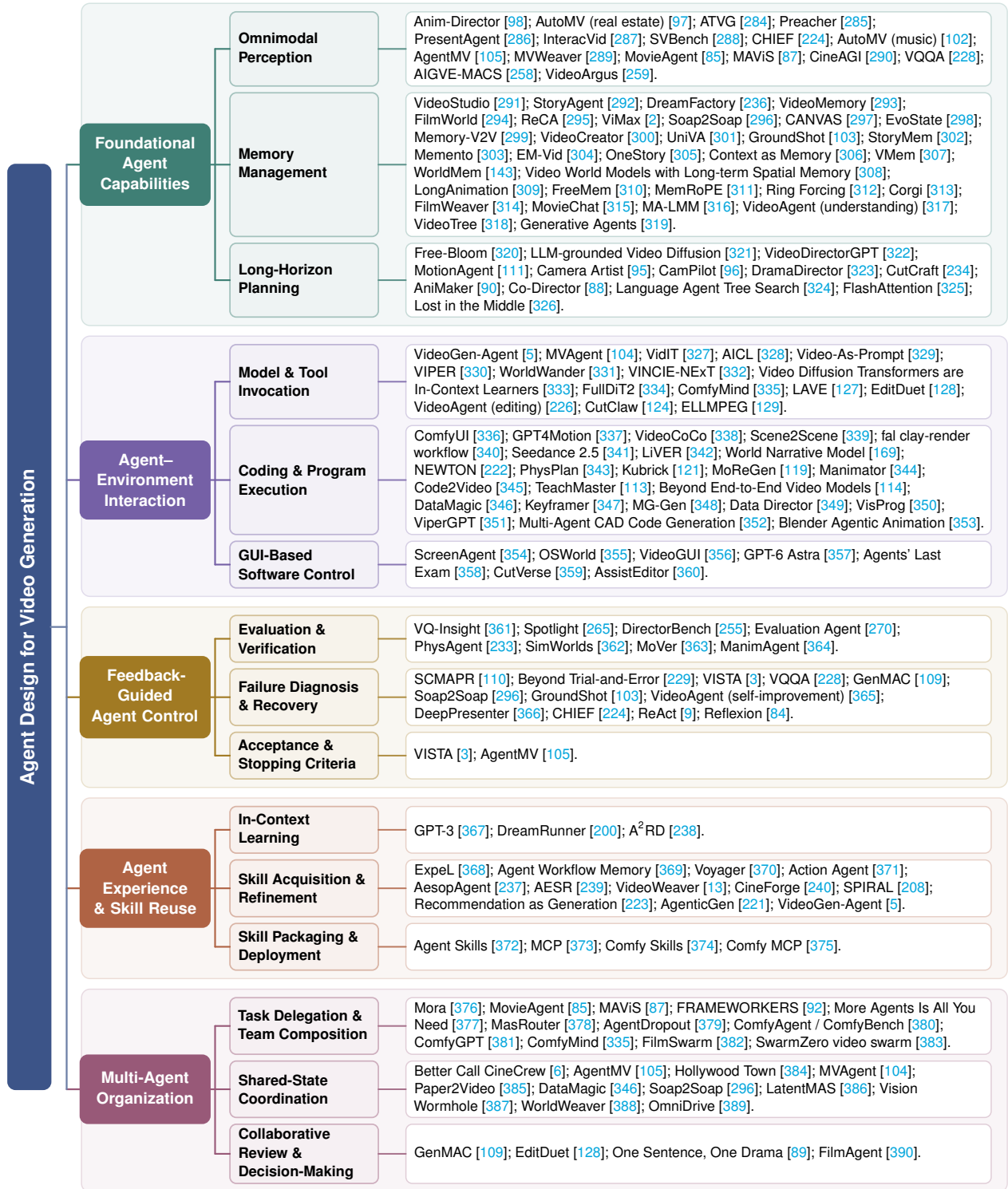

%% file: sections/07_future_challenges.tex
\section{Future Work and Challenges}
\label{sec:challenges}

\subsection{Fully End-to-end Video Production}
\label{sec:challenges_autonomous_production}
Fully end-to-end video production means autonomously turning an initial idea into a complete film or television episode through scriptwriting, audiovisual design, asset creation, generation, editing, and final delivery. The target is a finished work that meets commercial requirements for quality, continuity, and delivery, without relying on repeated human prompting and revision.
The agent harness must coordinate dependencies, allocate resources, and recover from tool failures across this pipeline. Critic feedback should link a specific failure to an actionable repair—for example, revising a plan, replacing a reference, regenerating a shot, or re-editing a sequence—while preserving accepted content.
DirectorBench~\citep{chen2026directorbench} finds transitions between units to be the lowest-scoring part of current long-form workflows (Section~\ref{sec:rsi-evaluation}), highlighting the need to assess continuity across unit boundaries. Acceptance and stopping criteria grounded in the original creative intent and available budget remain undeveloped, leaving the harness without a reliable basis for deciding when further revision is warranted and when the production is ready for delivery.

\subsection{Lifelong Memory}
\label{sec:challenges_lifelong_memory}
Autonomous agentic video generation and sustained world interaction require agents to maintain stable, persistent memory across shots and interaction sessions.
Generated frames alone do not necessarily provide a dependable record of character identity, object properties, scene changes, or completed events, especially when these facts leave the current field of view.
StoryMem~\citep{zhang2025storymem} conditions later shots on a dynamically updated visual memory, while WorldMem~\citep{xiao2025worldmem} retrieves memory frames together with associated states to reconstruct previously observed scenes.
These mechanisms provide foundations for continuity, but lifelong memory should also preserve useful experience beyond a single generated sequence or simulated episode.

A future agent harness should distinguish project specific records, such as accepted character references and event histories, from transferable experience, such as successful tool configurations, repair strategies, and reusable production procedures.
The challenge is not simply to store more context, but to retrieve relevant evidence, update records when the project changes, and discard obsolete or unreliable information without losing established constraints.
Memory entries should retain their provenance, temporal scope, and artifact versions so that the agent can distinguish verified outcomes from unconfirmed plans.
Evaluation should test whether retained experience improves new projects and revisited environments, while controlling for additional context and avoiding transfer of conflicting identities or outdated assumptions.

\subsection{Hybrid Code-Pixel Generation}
\label{sec:challenges_hybrid_control}
Hybrid code and pixel generation combines programs that represent scene state and controllable dynamics with video models that synthesize the resulting imagery.
Code supplies explicit object, camera, and interaction constraints, often through rendered proxy frames, while the learned generator adds appearance and motion detail, so programmable structure and pixel synthesis jointly determine the video.
Programmable scaffolds offer a promising foundation: the World Narrative Model~\citep{chen2026wnm} uses an editable physical representation to guide video synthesis, while PhysAgent~\citep{physagent2026} combines executable physical programs with simulation, verification, and repair.
These approaches make generation conditions inspectable and editable rather than expressing them only through prompts.
Code and rendered proxy videos already serve as persistent scaffolds shared by an agent and a generator in recent engine- and program-driven renderers (Section~\ref{sec:iwm-explicit-state}), but each is typically trained for one structural representation, only some run in real time, and their paired data come from a small number of engines.

The Genie 4 announcement~\citep{hadsell2026genie4} explicitly identifies hybrid world models as a future direction, combining structured reasoning by code and language models with pixel level simulation.
It describes closer integration of Gemini for scene structure and narrative planning with Genie for audiovisual simulation as an early step toward this goal.
Video generators and world models could further expose callable code interfaces for querying state, updating entities, specifying events, controlling cameras and actions, and advancing the simulation.
The agent would maintain a structured state, translate its updates into supported generation controls, and verify whether the resulting video realizes those updates.
A key challenge is keeping program state, agent memory, and generated observations aligned: executing a command or storing an object record does not guarantee that the generator preserves its consequences.
The interface must therefore support both explicit intervention and verification of its effects, allowing the agent harness to preserve accepted state and correct deviations over extended generation and interaction.

\subsection{Autonomous Human-free Development}
\label{sec:challenges_autonomous_development}
The same autonomy should extend from producing videos to developing the models that produce them.
A longer term goal is an agent harness that independently diagnoses model weaknesses, acquires or constructs relevant data, designs training experiments, executes updates, evaluates candidates, and retains or rejects changes without manual intervention in each iteration.
Failures from agentic video generation and trajectories from world interaction could supply evidence for this development loop, allowing the system to identify missing capabilities and generate targeted supervision.
Achieving RSI, however, requires more than automating a fixed sequence of data collection and training.
Improved capabilities must also strengthen the procedures used to discover failures, select data, design interventions, and validate subsequent updates.
The agent harness should therefore be able to revise development code, experiment strategies, and reusable improvement procedures using evidence from earlier rounds.
Major challenges include attributing gains to specific interventions, preventing errors in generated supervision from accumulating, coordinating updates to generators and critics, and preserving earlier capabilities. Among these, the one through which an evaluator selects data and accepts updates is the most exposed, because video development relies mainly on learned judges.
Independent validation and recoverable model versions are necessary to determine whether autonomous iterations produce durable progress rather than adaptation to the system's own evaluation signals.

\subsection{Efficiency}
\label{sec:challenges_efficiency}
Efficiency concerns the quality and scale of finished work that a system can deliver under a fixed time and resource budget.
Review loops add generator, critic, and controller calls, failed attempts, and wall clock time, yet these quantities are rarely reported under a common accounting protocol.
A common accounting, reported alongside quality, would make it possible to compare a multi-agent pipeline with a single long-context generator and to judge whether an improvement round was worth its cost.
The agent harness should allocate resources across generation, inspection, and repair, reuse accepted assets and intermediate results, and decide when further refinement no longer justifies another call.
Comparisons should account for all attempted operations, including rejected candidates and verification, rather than reporting only the final generation.
For model development, cumulative data, training, and validation costs must also be tracked across update rounds.

\subsection{Provenance and Safety}
\label{sec:challenges_provenance_safety}
\label{sec:challenges_provenance_cost}
Agentic production multiplies the operations applied to a single asset. A shot may be generated, locally regenerated, edited, re-timed, and composited by different tools before delivery, and a mark or credential attached at one step does not necessarily describe the result of later ones (Section~\ref{sec:agentic_video_watermarking}). A harness already records which tool produced which artifact; extending these records into signed provenance for every accepted version, as C2PA Content Credentials~\citep{c2pa2026} do for edits, would make the history of a delivered video auditable. Identity references raise a related question that current systems leave to the user: whether a person whose likeness is stored as an asset consented to its reuse across projects, and how a harness should enforce such limits.

Safety should cover both generated content and the operations used to produce it.
The agent harness should bind reuse permissions to identity assets, check generated and edited outputs against content restrictions, and enforce access and execution limits on production tools and generated code.
These constraints should follow assets across derived versions and projects, so that an edit, a tool change, or memory reuse cannot bypass an established restriction.

\subsection{Emerging Tasks}
\label{sec:challenges_emerging_tasks}
\label{sec:challenges_evaluation}
Agentic video generation could enable new tasks in which video creation, interactive experience, and iterative development are parts of the same workflow.
Automated game creation is one promising example.
GameNGen~\citep{valevski2025gamengen} demonstrates action conditioned simulation of an existing game, and GameGen-X~\citep{che2025gamegenx} supports generation and interactive control of game videos.
Building on such capabilities, an agent harness could coordinate narrative and level design, asset generation, gameplay code, audiovisual synthesis, and automated playtesting, using observed failures to revise the resulting experience.
However, generating visually playable trajectories is not equivalent to authoring a complete game.
The system must also establish explicit rules, persistent state, success and failure conditions, and reproducible interactions, and verify that visual outcomes agree with the intended mechanics.

Other prospective tasks include interactive storytelling that adapts to audience actions, personalized audiovisual learning environments, and scenario generation for testing agents in simulated worlds.
These tasks shift the objective from delivering a fixed sequence to maintaining a coherent experience whose content changes in response to participants.
They also connect the three directions of this survey: agentic video generation constructs the audiovisual content, video world modeling supports interaction, and model development could address limitations revealed by repeated use.

These tasks require benchmarks for complete video systems, not just isolated outputs.
Existing evaluations remain largely centered on generated video quality or the capabilities of individual video models.
Recent benchmarks cover parts of the problem. DirectorBench~\citep{chen2026directorbench} diagnoses long-form generation workflows, WorldScore~\citep{duan2025worldscore} and PlayWorld~\citep{ding2026playworld} test world models over sequences of generation or interaction, and harness-evolution studies~\citep{wang2026harnessevaluation,ke2026evoharnessbench} compare updated and fixed harnesses on general agent tasks. None of them yet measures, for video, whether the agent harness can autonomously complete an entire production workflow, maintain persistent state across extended generation and interaction, recover from failures, and improve system capabilities through successive development rounds.
Agent orchestration also creates opportunities to extend video duration and support richer narratives, more characters, and more complex event dependencies.
Benchmarks should therefore measure the generation horizon and content complexity that a complete system can sustain under explicit quality, consistency, and resource constraints, rather than evaluating only predefined clips.
Progress should be evaluated through task completion, controllability, persistent consistency, and development cost, in addition to visual quality.
For RSI, evaluation must additionally track gains, regressions, and cumulative costs across multiple updates, using independent tasks to test whether improvements persist and transfer.
Comparisons should establish whether an evolving system becomes better at producing subsequent improvements, rather than benefiting from additional sampling, stronger tools, or repeated training.

%% file: main.bib
@misc{liu2024videogenerationevaluation,
  title         = {A Survey of {AI}-Generated Video Evaluation},
  author        = {Liu, Xiao and Xiang, Xinhao and Li, Zizhong and Wang, Yongheng and Li, Zhuoheng and Liu, Zhuosheng and Zhang, Weidi and Ye, Weiqi and Zhang, Jiawei},
  year          = {2024},
  eprint        = {2410.19884},
  archivePrefix = {arXiv},
  url           = {https://arxiv.org/abs/2410.19884}
}

@misc{yin2025spatiotemporal,
  title         = {A Survey: Spatiotemporal Consistency in Video Generation},
  author        = {Yin, Zhiyu and Chen, Kehai and Bai, Xuefeng and Jiang, Ruili and Li, Juntao and Li, Hongdong and Liu, Jin and Xiang, Yang and Yu, Jun and Zhang, Min},
  year          = {2025},
  eprint        = {2502.17863},
  archivePrefix = {arXiv},
  url           = {https://arxiv.org/abs/2502.17863}
}

@misc{liu2026interactiveworldmodel,
  title         = {Towards Interactive Video World Modeling: Frontiers, Challenges, Benchmarks, and Future Trends},
  author        = {Liu, Jiuming and Ni, Chaojun and Liu, Mengmeng and Peng, Chensheng and Wang, Fangjinhua and Shen, Sitian and Pollefeys, Marc and Tomizuka, Masayoshi and Tewari, Ayush and Kristensson, Per Ola},
  year          = {2026},
  eprint        = {2606.01164},
  archivePrefix = {arXiv},
  url           = {https://arxiv.org/abs/2606.01164}
}

@misc{huang2026vimax,
  title         = {{ViMax}: Agentic Video Generation},
  author        = {Huang, Lingxuan and He, Sizhe and Zhou, Hengji and Nie, Liqiang and Xia, Lianghao and Huang, Chao},
  year          = {2026},
  eprint        = {2606.07649},
  archivePrefix = {arXiv},
  url           = {https://arxiv.org/abs/2606.07649}
}

@misc{wei2026videoweaver,
  title         = {{VideoWeaver}: Evaluating and Evolving Skills for Agentic Long Video Generation},
  author        = {Wei, Jianhui and Zhang, Yan and Tan, Jie and Zhu, Hengchuan and Zhang, Xiaotian and Chen, Ziyi and Zhang, Daoan and Xu, Wei and Jin, Yeying and Liu, Zuozhu},
  year          = {2026},
  eprint        = {2606.08091},
  archivePrefix = {arXiv},
  url           = {https://arxiv.org/abs/2606.08091}
}

@misc{huang2026agenticvisual,
  title         = {Agentic Visual Generation: From Generative Models to Agentic Control},
  author        = {Huang, Yinming and Tu, Shuyuan and Yan, Xi and Zhan, Jiahao and Yang, Zihan and Xing, Zhen and Zhang, Hui and Zhang, Tiehua and Jiang, Yu-Gang and Wu, Zuxuan},
  year          = {2026},
  eprint        = {2609.06758},
  archivePrefix = {arXiv},
  url           = {https://arxiv.org/abs/2609.06758}
}

@article{li2026videoposttraining,
  title   = {Video Generation Models: A Survey of Post-Training and Alignment},
  author  = {Li, Chaoyu and Gu, Xiaoyi and Kulkarni, Yogesh and Im, Eun Woo and Honarmand, Mohammadmahdi and Wang, Zeyu and Song, Juntong and Du, Fei and Jiang, Xilin and Zheng, Kexin and Li, Tianzhi and Tao, Fei and Fazli, Pooyan},
  journal = {Transactions on Machine Learning Research},
  year    = {2026},
  url     = {https://openreview.net/forum?id=YlUEWLESIu},
  note    = {Coverage assessed from the author version, arXiv:2610.00812v1}
}

@misc{chu2026agenticworldmodel,
  title         = {Agentic World Modeling: Foundations, Capabilities, Laws, and Beyond},
  author        = {Chu, Meng and Zhang, Xuan Billy and Lin, Kevin Qinghong and Kong, Lingdong and Zhang, Jize and Tu, Teng and Ma, Weijian and Huang, Ziqi and Yang, Senqiao and others},
  year          = {2026},
  eprint        = {2604.22748},
  archivePrefix = {arXiv},
  url           = {https://arxiv.org/abs/2604.22748}
}

@inproceedings{rombach2021latentdiffusion,
  title         = {High-Resolution Image Synthesis with Latent Diffusion Models},
  author        = {Rombach, Robin and Blattmann, Andreas and Lorenz, Dominik and Esser, Patrick and Ommer, Bj{\"o}rn},
  year          = {2022},
  eprint        = {2112.10752},
  archivePrefix = {arXiv},
  url           = {https://arxiv.org/abs/2112.10752},
  booktitle     = {Proceedings of the IEEE/CVF Conference on Computer Vision and Pattern Recognition (CVPR)},
  pages         = {10674--10685},
  doi           = {10.1109/CVPR52688.2022.01042}
}

@inproceedings{ho2022videodiffusion,
  title         = {Video Diffusion Models},
  author        = {Ho, Jonathan and Salimans, Tim and Gritsenko, Alexey and Chan, William and Norouzi, Mohammad and Fleet, David J.},
  year          = {2022},
  eprint        = {2204.03458},
  archivePrefix = {arXiv},
  url           = {https://arxiv.org/abs/2204.03458},
  booktitle     = {Advances in Neural Information Processing Systems},
  volume        = {35},
  pages         = {8633--8646},
  doi           = {10.52202/068431-0628}
}

@inproceedings{villegas2022phenaki,
  title         = {{Phenaki}: Variable Length Video Generation From Open Domain Textual Descriptions},
  author        = {Villegas, Ruben and Babaeizadeh, Mohammad and Kindermans, Pieter-Jan and Moraldo, Hernan and Zhang, Han and Saffar, Mohammad Taghi and Castro, Santiago and Kunze, Julius and Erhan, Dumitru},
  year          = {2023},
  eprint        = {2210.02399},
  archivePrefix = {arXiv},
  url           = {https://openreview.net/forum?id=vOEXS39nOF},
  booktitle     = {International Conference on Learning Representations}
}

@inproceedings{lipman2022flowmatching,
  title         = {Flow Matching for Generative Modeling},
  author        = {Lipman, Yaron and Chen, Ricky T. Q. and Ben-Hamu, Heli and Nickel, Maximilian and Le, Matt},
  year          = {2023},
  eprint        = {2210.02747},
  archivePrefix = {arXiv},
  url           = {https://arxiv.org/abs/2210.02747},
  booktitle     = {International Conference on Learning Representations}
}

@inproceedings{peebles2022dit,
  title         = {Scalable Diffusion Models with Transformers},
  author        = {Peebles, William and Xie, Saining},
  year          = {2023},
  eprint        = {2212.09748},
  archivePrefix = {arXiv},
  url           = {https://arxiv.org/abs/2212.09748},
  booktitle     = {Proceedings of the IEEE/CVF International Conference on Computer Vision (ICCV)},
  pages         = {4172--4182},
  doi           = {10.1109/ICCV51070.2023.00387}
}

@inproceedings{shinn2023reflexion,
  title         = {Reflexion: Language Agents with Verbal Reinforcement Learning},
  author        = {Shinn, Noah and Cassano, Federico and Gopinath, Ashwin and Narasimhan, Karthik and Yao, Shunyu},
  year          = {2023},
  eprint        = {2303.11366},
  archivePrefix = {arXiv},
  url           = {https://proceedings.neurips.cc/paper_files/paper/2023/hash/1b44b878bb782e6954cd888628510e90-Abstract-Conference.html},
  booktitle     = {Advances in Neural Information Processing Systems},
  volume        = {36},
  pages         = {8634--8652}
}

@inproceedings{kondratyuk2023videopoet,
  title         = {{VideoPoet}: A Large Language Model for Zero-Shot Video Generation},
  author        = {Kondratyuk, Dan and Yu, Lijun and Gu, Xiuye and Lezama, Jos{\'e} and Huang, Jonathan and Schindler, Grant and Hornung, Rachel and Birodkar, Vighnesh and Yan, Jimmy and Chiu, Ming-Chang and others},
  year          = {2024},
  eprint        = {2312.14125},
  archivePrefix = {arXiv},
  url           = {https://arxiv.org/abs/2312.14125},
  booktitle     = {Proceedings of the 41st International Conference on Machine Learning},
  series        = {Proceedings of Machine Learning Research},
  volume        = {235},
  pages         = {25105--25124}
}

@inproceedings{bartal2024lumiere,
  title         = {Lumiere: A Space-Time Diffusion Model for Video Generation},
  author        = {Bar-Tal, Omer and Chefer, Hila and Tov, Omer and Herrmann, Charles and Paiss, Roni and Zada, Shiran and Ephrat, Ariel and Hur, Junhwa and Liu, Guanghui and Raj, Amit and Li, Yuanzhen and Rubinstein, Michael and Michaeli, Tomer and Wang, Oliver and Sun, Deqing and Dekel, Tali and Mosseri, Inbar},
  year          = {2024},
  eprint        = {2401.12945},
  archivePrefix = {arXiv},
  url           = {https://arxiv.org/abs/2401.12945},
  booktitle     = {SIGGRAPH Asia 2024 Conference Papers},
  doi           = {10.1145/3680528.3687614}
}

@misc{kong2024hunyuanvideo,
  title         = {{HunyuanVideo}: A Systematic Framework For Large Video Generative Models},
  author        = {Kong, Weijie and Tian, Qi and Zhang, Zijian and Min, Rox and Dai, Zuozhuo and Zhou, Jin and Xiong, Jiangfeng and Li, Xin and Wu, Bo and Zhang, Jianwei and others},
  year          = {2024},
  eprint        = {2412.03603},
  archivePrefix = {arXiv},
  url           = {https://arxiv.org/abs/2412.03603}
}

@misc{ning2026code,
  title         = {Code as Agent Harness},
  author        = {Ning, Xuying and Tieu, Katherine and Fu, Dongqi and Wei, Tianxin and Li, Zihao and Bei, Yuanchen and Zou, Jiaru and Ai, Mengting and Liu, Zhining and Li, Ting-Wei and others},
  year          = {2026},
  eprint        = {2605.18747},
  archivePrefix = {arXiv},
  url           = {https://arxiv.org/abs/2605.18747}
}

@misc{mmstoryagent2025,
  title = {{MM-StoryAgent: Immersive Narrated Storybook Video Generation with a Multi-Agent Paradigm across Text, Image and Audio}},
  author = {Xu, Xuenan and Mei, Jiahao and Li, Chenliang and Wu, Yuning and Yan, Ming and Lai, Shaopeng and Zhang, Ji and Wu, Mengyue},
  year = {2025},
  howpublished = {Preprint},
  url = {https://arxiv.org/abs/2503.05242},
  eprint = {2503.05242},
  archivePrefix = {arXiv},
}

@inproceedings{filmaster2025,
  title = {{FilMaster: Bridging Cinematic Principles and Generative AI for Automated Film Generation}},
  author = {Huang, Kaiyi and Huang, Yukun and WANG, Xintao and Lin, Zinan and Ning, Xuefei and Wan, Pengfei and ZHANG, Di and Wang, Yu and Liu, Xihui},
  year = {2026},
  booktitle = {ICLR 2026},
  url = {https://proceedings.iclr.cc/paper_files/paper/2026/hash/fb8e52adcd9b59bad73f109c53afc43a-Abstract-Conference.html},
  pages = {155439--155482},
}

@misc{cameraartist2026,
  title = {{Camera Artist: A Multi-Agent Framework for Cinematic Language Storytelling Video Generation}},
  author = {Hu, Haobo and Mao, Qi and Li, Yuanhang and Jin, Libiao},
  year = {2026},
  howpublished = {Preprint},
  url = {https://arxiv.org/abs/2604.09195},
  eprint = {2604.09195},
  archivePrefix = {arXiv},
  note = {Authors list ICME 2026; archival publication not independently verified},
}

@misc{cineagents2026,
  title = {{A Benchmark and Multi-Agent System for Instruction-driven Cinematic Video Compilation}},
  author = {Zhang, Peixuan and Zhou, Chang and Zhang, Ziyuan and Liu, Hualuo and Zhang, Chunjie and Liu, Jingqi and Zhou, Xiaohui and Chen, Xi and Weng, Shuchen and Li, Si and Shi, Boxin},
  year = {2026},
  howpublished = {Preprint},
  url = {https://arxiv.org/abs/2604.10456},
  eprint = {2604.10456},
  archivePrefix = {arXiv},
}

@misc{frameworkers2026,
  title = {{FRAMEWORKERS: A Dynamic Multi-Agent Framework for AI-Generated Video Production}},
  author = {Li, Zhendong and Sun, Lei and Shi, Letian and Zhang, Deheng and Ming, Ruibo and Hu, Mengshun and Xu, Dannong and Wang, Jian and Paudel, Danda and Van Gool, Luc and Gu, Jinjin},
  year = {2026},
  howpublished = {Preprint},
  url = {https://arxiv.org/abs/2608.29814},
  eprint = {2608.29814},
  archivePrefix = {arXiv},
}

@article{lvmark2024,
  title   = {{LVMark}: Robust Watermark for Latent Video Diffusion Models},
  author  = {Jang, Youngdong and Jang, MinHyuk and Lee, JaeHyeok and Yang, Feng and Oh, Gyeongrok and Jeong, Jongheon and Kim, Sangpil},
  journal = {IEEE Transactions on Information Forensics and Security},
  year = {2026},
  volume = {21},
  pages = {4908--4923},
  doi = {10.1109/TIFS.2026.3688194},
  url = {https://doi.org/10.1109/TIFS.2026.3688194},
}

@misc{c2pa2026,
  author       = {{Coalition for Content Provenance and Authenticity}},
  title        = {Content Credentials: {C2PA} Technical Specification},
  howpublished = {\url{https://spec.c2pa.org/specifications/specifications/2.4/specs/C2PA_Specification.html}},
  year         = {2026},
  note         = {Version 2.4, April 2026}
}

@misc{wang2026harnessevaluation,
  title         = {{Rethinking the Evaluation of Harness Evolution for Agents}},
  author        = {Wang, Yike and Zhu, Huaisheng and Hu, Zhengyu and Yuan, Yige and Chen, Zhengyu and Senthil, Shakti and Hajishirzi, Hannaneh and Tsvetkov, Yulia and Smith, Noah A. and Dasigi, Pradeep and Xiao, Teng},
  year          = {2026},
  eprint        = {2607.12227},
  archivePrefix = {arXiv},
  primaryClass  = {cs.AI},
  url           = {https://arxiv.org/abs/2607.12227}
}

@misc{huang2026evobench,
  title         = {{Evo-Bench: Can Language Models Improve Agent Harness?}},
  author        = {Huang, Lisheng and Yang, Chen and Zhou, Hao and Song, Huatong and Chen, Zongchao and Le, Ran and Song, Yang and Zhao, Wayne Xin and Zhang, Tao},
  year          = {2026},
  eprint        = {2608.09096},
  archivePrefix = {arXiv},
  primaryClass  = {cs.CL},
  url           = {https://arxiv.org/abs/2608.09096}
}

@misc{unterthiner2018fvd,
  title = {{Towards Accurate Generative Models of Video: A New Metric \& Challenges}},
  author = {Unterthiner, Thomas and van Steenkiste, Sjoerd and Kurach, Karol and Marinier, Raphael and Michalski, Marcin and Gelly, Sylvain},
  year = {2018},
  eprint = {1812.01717},
  archivePrefix = {arXiv},
  url = {https://arxiv.org/abs/1812.01717}
}

@inproceedings{ge2024contentbias,
  title = {{On the Content Bias in Fr\'echet Video Distance}},
  author = {Ge, Songwei and Mahapatra, Aniruddha and Parmar, Gaurav and Zhu, Jun-Yan and Huang, Jia-Bin},
  year = {2024},
  eprint = {2404.12391},
  archivePrefix = {arXiv},
  url = {https://arxiv.org/abs/2404.12391},
  booktitle = {Proceedings of the IEEE/CVF Conference on Computer Vision and Pattern Recognition (CVPR)},
  pages = {7277--7288},
  doi = {10.1109/CVPR52733.2024.00695}
}

@misc{liu2024fvmd,
  title = {{Fr\'echet Video Motion Distance: A Metric for Evaluating Motion Consistency in Videos}},
  author = {Liu, Jiahe and Qu, Youran and Yan, Qi and Zeng, Xiaohui and Wang, Lele and Liao, Renjie},
  year = {2024},
  eprint = {2407.16124},
  archivePrefix = {arXiv},
  url = {https://arxiv.org/abs/2407.16124}
}

@inproceedings{huang2024vbench,
  title = {{VBench: Comprehensive Benchmark Suite for Video Generative Models}},
  author = {Huang, Ziqi and He, Yinan and Yu, Jiashuo and Zhang, Fan and Si, Chenyang and Jiang, Yuming and Zhang, Yuanhan and Wu, Tianxing and Jin, Qingyang and Chanpaisit, Nattapol and Wang, Yaohui and Chen, Xinyuan and Wang, Limin and Lin, Dahua and Qiao, Yu and Liu, Ziwei},
  year = {2024},
  eprint = {2311.17982},
  archivePrefix = {arXiv},
  url = {https://arxiv.org/abs/2311.17982},
  booktitle = {Proceedings of the IEEE/CVF Conference on Computer Vision and Pattern Recognition (CVPR)},
  pages = {21807--21818},
  doi = {10.1109/CVPR52733.2024.02060}
}

@inproceedings{sun2025t2vcompbench,
  title = {{T2V-CompBench: A Comprehensive Benchmark for Compositional Text-to-video Generation}},
  author = {Sun, Kaiyue and Huang, Kaiyi and Liu, Xian and Wu, Yue and Xu, Zihan and Li, Zhenguo and Liu, Xihui},
  year = {2025},
  eprint = {2407.14505},
  archivePrefix = {arXiv},
  url = {https://arxiv.org/abs/2407.14505},
  booktitle = {Proceedings of the IEEE/CVF Conference on Computer Vision and Pattern Recognition (CVPR)},
  pages = {8406--8416},
  doi = {10.1109/CVPR52734.2025.00787}
}

@misc{zheng2025vbench2,
  title = {{VBench-2.0: Advancing Video Generation Benchmark Suite for Intrinsic Faithfulness}},
  author = {Zheng, Dian and Huang, Ziqi and Liu, Hongbo and Zou, Kai and He, Yinan and Zhang, Fan and Gu, Lulu and Zhang, Yuanhan and He, Jingwen and Zheng, Wei-Shi and Qiao, Yu and Liu, Ziwei},
  year = {2025},
  eprint = {2503.21755},
  archivePrefix = {arXiv},
  url = {https://arxiv.org/abs/2503.21755}
}

@inproceedings{bansal2025videophy2,
  title = {{VideoPhy-2: A Challenging Action-Centric Physical Commonsense Evaluation in Video Generation}},
  author = {Bansal, Hritik and Peng, Clark and Bitton, Yonatan and Goldenberg, Roman and Grover, Aditya and Chang, Kai-Wei},
  year = {2026},
  eprint = {2503.06800},
  archivePrefix = {arXiv},
  url = {https://arxiv.org/abs/2503.06800},
  booktitle = {International Conference on Learning Representations}
}

@misc{miao2026statebench,
  title = {{Do Video Generators Track the World Across Segments? A Benchmark and Method for World-State Reasoning in Video Continuation}},
  author = {Miao, Yingmao and Zhang, Pengfei and Xu, Chaoran and Yu, Meng and Tang, Jing and Chu, Xiangxiang and Shen, Chao and Lin, Chenhao},
  year = {2026},
  eprint = {2609.03673},
  archivePrefix = {arXiv},
  url = {https://arxiv.org/abs/2609.03673}
}

@misc{zeng2026videoargus,
  title = {{VideoArgus: Agentic Rubric-Grounded Unified Evaluation for Video Generation and Editing}},
  author = {Zeng, Ziyun and Wang, Zixuan and Yu, Yongsheng and Hua, Hang and Luo, Jiebo},
  year = {2026},
  eprint = {2608.05485},
  archivePrefix = {arXiv},
  url = {https://arxiv.org/abs/2608.05485}
}

@misc{liu2026v2vbench,
  title = {{V2V-Bench: A Comprehensive Benchmark for Video-to-Video Generation Evaluation}},
  author = {Liu, Tao and Krishna, Leela and Kumar, Gouti Pavan and K, Sreeja and Garg, Vishav},
  year = {2026},
  eprint = {2606.05665},
  archivePrefix = {arXiv},
  url = {https://arxiv.org/abs/2606.05665}
}

@misc{zhou2026avgenbench,
  title = {{AVGen-Bench: A Task-Driven Benchmark for Multi-Granular Evaluation of Text-to-Audio-Video Generation}},
  author = {Zhou, Ziwei and Lai, Zeyuan and Wang, Rui and Yang, Yifan and Xing, Zhen and Yang, Yuqing and Dai, Qi and Qiu, Lili and Luo, Chong},
  year = {2026},
  eprint = {2604.08540},
  archivePrefix = {arXiv},
  url = {https://arxiv.org/abs/2604.08540}
}

@misc{yang2026evalverse,
  title = {{EvalVerse: Pipeline-Aware and Expert-Calibrated Benchmarking for Professional Cinematic Video Generation}},
  author = {Yang, Songlin and Zhong, Haobin and Zhang, Ruilin and Zhao, Xiaotong and Li, Shuai and Zheng, Kai and Yang, Xuyi and Wang, Zhe and Tang, Zhenchen and Li, Yang and Gu, Bohai and Peng, Zhengwei and Huang, Yidan and Luo, Mengzhou and Bo, Yihang and Feng, Dalu and Zhang, Yujia and Ma, Juntao and Wang, Ruiqi and Zhang, Lvmin and Guo, Yuwei and Guan, Frank and Agrawala, Maneesh and Fu, Hongbo and Zhao, Alan and Rao, Anyi},
  year = {2026},
  eprint = {2605.23271},
  archivePrefix = {arXiv},
  url = {https://arxiv.org/abs/2605.23271}
}

@inproceedings{dai2024safesora,
  title = {{SafeSora: Towards Safety Alignment of Text2Video Generation via a Human Preference Dataset}},
  author = {Dai, Josef and Chen, Tianle and Wang, Xuyao and Yang, Ziran and Chen, Taiye and Ji, Jiaming and Yang, Yaodong},
  year = {2024},
  eprint = {2406.14477},
  archivePrefix = {arXiv},
  url = {https://arxiv.org/abs/2406.14477},
  booktitle = {Advances in Neural Information Processing Systems},
  volume = {37},
  pages = {17161--17214},
  doi = {10.52202/079017-0546}
}

@inproceedings{he2024videoscore,
  title = {{VideoScore: Building Automatic Metrics to Simulate Fine-grained Human Feedback for Video Generation}},
  author = {He, Xuan and Jiang, Dongfu and Zhang, Ge and Ku, Max and Soni, Achint and Siu, Sherman and Chen, Haonan and Chandra, Abhranil and Jiang, Ziyan and Arulraj, Aaran and Wang, Kai and Do, Quy Duc and Ni, Yuansheng and Lyu, Bohan and Narsupalli, Yaswanth and Fan, Rongqi and Lyu, Zhiheng and Lin, Yuchen and Chen, Wenhu},
  year = {2024},
  eprint = {2406.15252},
  archivePrefix = {arXiv},
  url = {https://arxiv.org/abs/2406.15252},
  booktitle = {Proceedings of the 2024 Conference on Empirical Methods in Natural Language Processing},
  pages = {2105--2123},
  doi = {10.18653/v1/2024.emnlp-main.127}
}

@inproceedings{liu2025videoalign,
  title = {{Improving Video Generation with Human Feedback}},
  author = {Liu, Jie and Liu, Gongye and Liang, Jiajun and Yuan, Ziyang and Liu, Xiaokun and Zheng, Mingwu and Wu, Xiele and Wang, Qiulin and Xia, Menghan and Wang, Xintao and Liu, Xiaohong and Yang, Fei and Wan, Pengfei and Zhang, Di and Gai, Kun and Yang, Yujiu and Ouyang, Wanli},
  year = {2025},
  eprint = {2501.13918},
  archivePrefix = {arXiv},
  url = {https://arxiv.org/abs/2501.13918},
  booktitle = {Advances in Neural Information Processing Systems},
  volume = {38},
  doi = {10.52202/085713-2750}
}

@article{he2025videoscore2,
  title = {{VideoScore2: Think before You Score in Generated Video Evaluation}},
  author = {He, Xuan and Jiang, Dongfu and Nie, Ping and Liu, Minghao and Jiang, Zhengxuan and Su, Mingyi and Ma, Wentao and Lin, Junru and Ye, Chun and Lu, Yi and Wu, Keming and Schneider, Benjamin and Do, Quy Duc and Li, Zhuofeng and Jia, Yiming and Zhang, Yuxuan and Cheng, Guo and Wang, Haozhe and Zhou, Wangchunshu and Lin, Qunshu and Zhang, Yuanxing and Zhang, Ge and Huang, Wenhao and Chen, Wenhu},
  year = {2026},
  eprint = {2509.22799},
  archivePrefix = {arXiv},
  url = {https://openreview.net/forum?id=MpkVh4jH44},
  journal = {Transactions on Machine Learning Research}
}

@inproceedings{han2025videobench,
  title = {{Video-Bench: Human-Aligned Video Generation Benchmark}},
  author = {Han, Hui and Li, Siyuan and Chen, Jiaqi and Yuan, Yiwen and Wu, Yuling and Deng, Yufan and Leong, Chak Tou and Du, Hanwen and Fu, Junchen and Li, Youhua and Zhang, Jie and Zhang, Chi and Li, Li-jia and Ni, Yongxin},
  year = {2025},
  eprint = {2504.04907},
  archivePrefix = {arXiv},
  url = {https://arxiv.org/abs/2504.04907},
  booktitle = {Proceedings of the IEEE/CVF Conference on Computer Vision and Pattern Recognition (CVPR)},
  pages = {18858--18868},
  doi = {10.1109/CVPR52734.2025.01757}
}

@misc{chinchure2025spotlight,
  title = {{Spotlight: Identifying and Localizing Video Generation Errors Using VLMs}},
  author = {Chinchure, Aditya and Ravi, Sahithya and Shukla, Pushkar and Shwartz, Vered and Sigal, Leonid},
  year = {2025},
  eprint = {2511.18102},
  archivePrefix = {arXiv},
  url = {https://arxiv.org/abs/2511.18102}
}

@misc{tang2026rewardverse,
  title = {{RewardVerse: Rubric-Guided Policy Optimization for Video Reward Modeling}},
  author = {Tang, Zhenchen and Li, Yang and Yang, Songlin and Peng, Bo and Zhao, Xiaotong and Li, Shuai and Fan, Haotian and Zhao, Alan and Dong, Jing},
  year = {2026},
  eprint = {2609.22947},
  archivePrefix = {arXiv},
  url = {https://arxiv.org/abs/2609.22947}
}

@misc{yang2025videogeneval,
  title = {{VideoGen-Eval: Agent-based System for Video Generation Evaluation}},
  author = {Yang, Yuhang and Fan, Ke and Sun, Shangkun and Li, Hongxiang and Zeng, Ailing and Han, FeiLin and Zhai, Wei and Liu, Wei and Cao, Yang and Zha, Zheng-Jun},
  year = {2025},
  eprint = {2503.23452},
  archivePrefix = {arXiv},
  url = {https://arxiv.org/abs/2503.23452}
}

@inproceedings{zhang2025evaluationagent,
  title = {Evaluation Agent: Efficient and Promptable Evaluation Framework for Visual Generative Models},
  author = {Zhang, Fan and Tian, Shulin and Huang, Ziqi and Qiao, Yu and Liu, Ziwei},
  booktitle = {Proceedings of the 63rd Annual Meeting of the Association for Computational Linguistics (Volume 1: Long Papers)},
  year = {2025},
  publisher = {Association for Computational Linguistics},
  pages = {7561--7582},
  doi = {10.18653/v1/2025.acl-long.374},
  url = {https://aclanthology.org/2025.acl-long.374/},
}

@misc{tian2026openevaluationagent,
  title = {{Open Evaluation Agent: Efficient and Promptable Evaluation of Visual Generative Models}},
  author = {Tian, Shulin and Huang, Ziqi and Zhang, Fan and Zhu, Hongyuan and Qiao, Yu and Liu, Ziwei},
  year = {2026},
  eprint = {2608.09666},
  archivePrefix = {arXiv},
  url = {https://arxiv.org/abs/2608.09666}
}

@misc{xu2026dontreadlog,
  title = {{Don't Read the Log: Execution Traces Contaminate Verifiers in Video-Generation Agents}},
  author = {Xu, Jian},
  year = {2026},
  eprint = {2609.28564},
  archivePrefix = {arXiv},
  url = {https://arxiv.org/abs/2609.28564}
}

@inproceedings{liu2025uve,
  title = {{UVE: Are MLLMs Unified Evaluators for AI-Generated Videos?}},
  author = {Liu, Yuanxin and Zhu, Rui and Ren, Shuhuai and Wang, Jiacong and Guo, Haoyuan and Sun, Xu and Jiang, Lu},
  year = {2025},
  eprint = {2503.09949},
  archivePrefix = {arXiv},
  url = {https://arxiv.org/abs/2503.09949},
  booktitle = {Advances in Neural Information Processing Systems},
  volume = {38},
  doi = {10.52202/085713-0564}
}

@inproceedings{matsuda2026slvmeval,
  title = {{SLVMEval: Synthetic Meta Evaluation Benchmark for Text-to-Long Video Generation}},
  author = {Matsuda, Ryosuke and Kudo, Keito and Yoshida, Haruto and Shimizu, Nobuyuki and Suzuki, Jun},
  year = {2026},
  eprint = {2603.29186},
  archivePrefix = {arXiv},
  url = {https://arxiv.org/abs/2603.29186},
  booktitle = {Proceedings of the IEEE/CVF Conference on Computer Vision and Pattern Recognition (CVPR)},
  pages = {7784--7794}
}

@inproceedings{yang2026dualipo,
  title = {{Dual-IPO: Dual-Iterative Preference Optimization for Text-to-Video Generation}},
  author = {Yang, Xiaomeng and Yang, Mengping and Gong, Jia and Qin, Luozheng and Tan, Zhiyu and Li, Hao},
  year = {2026},
  eprint = {2502.02088},
  archivePrefix = {arXiv},
  url = {https://arxiv.org/abs/2502.02088},
  booktitle = {International Conference on Learning Representations}
}

@inproceedings{bansal2024videophy,
  title = {{VideoPhy: Evaluating Physical Commonsense for Video Generation}},
  author = {Bansal, Hritik and Lin, Zongyu and Xie, Tianyi and Zong, Zeshun and Yarom, Michal and Bitton, Yonatan and Jiang, Chenfanfu and Sun, Yizhou and Chang, Kai-Wei and Grover, Aditya},
  year = {2025},
  eprint = {2406.03520},
  archivePrefix = {arXiv},
  url = {https://arxiv.org/abs/2406.03520},
  booktitle = {International Conference on Learning Representations}
}

@misc{lin2026phyground,
  title = {{PhyGround: Benchmarking Physical Reasoning in Generative World Models}},
  author = {Lin, Juyi and Akbari, Arash and He, Yumei and Zhao, Lin and Zhang, Haichao and Akbari, Arman and Xu, Xingchen and Lu, Zoe Y. and Nan, Enfu and Deng, Hokin and Yeh, Edmund and Ostadabbas, Sarah and Fu, Yun and Dy, Jennifer and Zhao, Pu and Wang, Yanzhi},
  year = {2026},
  eprint = {2605.10806},
  archivePrefix = {arXiv},
  url = {https://arxiv.org/abs/2605.10806}
}

@inproceedings{he2026tear,
  title = {{TEAR: Temporal-aware Automated Red-teaming for Text-to-Video Models}},
  author = {He, Jiaming and Hou, Guanyu and Li, Hongwei and Huang, Zhicong and Chen, Kangjie and Yu, Yi and Jiang, Wenbo and Xu, Guowen and Zhang, Tianwei},
  year = {2026},
  eprint = {2511.21145},
  archivePrefix = {arXiv},
  url = {https://arxiv.org/abs/2511.21145},
  booktitle = {Proceedings of the IEEE/CVF Conference on Computer Vision and Pattern Recognition (CVPR)},
  pages = {41--50}
}

@inproceedings{lee2026scenesplit,
  title = {{Jailbreaking on Text-to-Video Models via Scene Splitting Strategy}},
  author = {Lee, Wonjun and Park, Haon and Lee, Doehyeon and Ham, Bumsub and Kim, Suhyun},
  year = {2026},
  eprint = {2509.22292},
  archivePrefix = {arXiv},
  url = {https://arxiv.org/abs/2509.22292},
  booktitle = {International Conference on Learning Representations}
}

@misc{jia2026raveneval,
  title = {{RAVEN-Eval: Rubric-Guided Automatic Evaluation for AI Video Generation Models Based on LMM Preference Judgement}},
  author = {Jia, Ziheng and Qian, Jiaying and Zhang, Zicheng and Zhu, Xiaorong and Gao, Lancheng and Min, Xiongkuo},
  year = {2026},
  eprint = {2608.09111},
  archivePrefix = {arXiv},
  url = {https://arxiv.org/abs/2608.09111}
}

@misc{song2026vqqa,
  title = {{VQQA: An Agentic Approach for Video Evaluation and Quality Improvement}},
  author = {Song, Yiwen and Pfister, Tomas and Song, Yale},
  year = {2026},
  eprint = {2603.12310},
  archivePrefix = {arXiv},
  url = {https://arxiv.org/abs/2603.12310}
}

@misc{liu2025aigvemacs,
  title = {{AIGVE-MACS: Unified Multi-Aspect Commenting and Scoring Model for AI-Generated Video Evaluation}},
  author = {Liu, Xiao and Zhang, Jiawei},
  year = {2025},
  eprint = {2507.01255},
  archivePrefix = {arXiv},
  url = {https://arxiv.org/abs/2507.01255}
}

@inproceedings{hu2024adas,
  title = {{Automated Design of Agentic Systems}},
  author = {Hu, Shengran and Lu, Cong and Clune, Jeff},
  year = {2025},
  eprint = {2408.08435},
  archivePrefix = {arXiv},
  url = {https://arxiv.org/abs/2408.08435},
  booktitle = {International Conference on Learning Representations}
}

@inproceedings{zhang2024aflow,
  title = {{AFlow: Automating Agentic Workflow Generation}},
  author = {Zhang, Jiayi and Xiang, Jinyu and Yu, Zhaoyang and Teng, Fengwei and Chen, Xionghui and Chen, Jiaqi and Zhuge, Mingchen and Cheng, Xin and Hong, Sirui and Wang, Jinlin and Zheng, Bingnan and Liu, Bang and Luo, Yuyu and Wu, Chenglin},
  year = {2025},
  eprint = {2410.10762},
  archivePrefix = {arXiv},
  url = {https://arxiv.org/abs/2410.10762},
  booktitle = {International Conference on Learning Representations}
}

@misc{robeyns2025sica,
  title = {{A Self-Improving Coding Agent}},
  author = {Robeyns, Maxime and Szummer, Martin and Aitchison, Laurence},
  year = {2025},
  eprint = {2504.15228},
  archivePrefix = {arXiv},
  url = {https://arxiv.org/abs/2504.15228}
}

@inproceedings{zhang2025dgm,
  title = {{Darwin G{\"o}del Machine: Open-Ended Evolution of Self-Improving Agents}},
  author = {Zhang, Jenny and Hu, Shengran and Lu, Cong and Lange, Robert and Clune, Jeff},
  year = {2026},
  eprint = {2505.22954},
  archivePrefix = {arXiv},
  url = {https://arxiv.org/abs/2505.22954},
  booktitle = {International Conference on Learning Representations}
}

@inproceedings{agrawal2025gepa,
  title = {{GEPA: Reflective Prompt Evolution Can Outperform Reinforcement Learning}},
  author = {Agrawal, Lakshya A and Tan, Shangyin and Soylu, Dilara and Ziems, Noah and Khare, Rishi and Opsahl-Ong, Krista and Singhvi, Arnav and Shandilya, Herumb and Ryan, Michael J and Jiang, Meng and Potts, Christopher and Sen, Koushik and Dimakis, Alexandros G. and Stoica, Ion and Klein, Dan and Zaharia, Matei and Khattab, Omar},
  year = {2026},
  eprint = {2507.19457},
  archivePrefix = {arXiv},
  url = {https://arxiv.org/abs/2507.19457},
  booktitle = {International Conference on Learning Representations}
}

@inproceedings{zhang2025ace,
  title = {{Agentic Context Engineering: Evolving Contexts for Self-Improving Language Models}},
  author = {Zhang, Qizheng and Hu, Changran and Upasani, Shubhangi and Ma, Boyuan and Hong, Fenglu and Kamanuru, Vamsidhar and Rainton, Jay and Wu, Chen and Ji, Mengmeng and Li, Hanchen and Thakker, Urmish and Zou, James and Olukotun, Kunle},
  year = {2026},
  eprint = {2510.04618},
  archivePrefix = {arXiv},
  url = {https://arxiv.org/abs/2510.04618},
  booktitle = {International Conference on Learning Representations}
}

@misc{lee2026metaharness,
  title = {{Meta-Harness: End-to-End Optimization of Model Harnesses}},
  author = {Lee, Yoonho and Nair, Roshen and Zhang, Qizheng and Lee, Kangwook and Khattab, Omar and Finn, Chelsea},
  year = {2026},
  eprint = {2603.28052},
  archivePrefix = {arXiv},
  url = {https://arxiv.org/abs/2603.28052}
}

@misc{wei2026evoharness,
  title = {{Evo-Harness: Context-to-Harness Skill Compilation for Self-Evolving Agents}},
  author = {Wei, Tianxin and Shi, Zhan and Lin, Minhua and He, Bing and Liu, Zewen and Sang, Yisi and Bei, Yuanchen and Ning, Xuying and Zou, Jiaru and Li, Ting-Wei and Lin, Xiao and Zhao, Yanjun and Wang, Chi and Dumoulin, Benoit and Wang, Dakuo and He, Jingrui and Lu, Hanqing},
  year = {2026},
  eprint = {2608.15071},
  archivePrefix = {arXiv},
  url = {https://arxiv.org/abs/2608.15071}
}

@misc{xia2026rrsi,
  title = {{RRSI: Regularized Recursive Self-Improvement of Agent Harnesses}},
  author = {Xia, Peng and Han, Rujun and Wang, Zifeng and Chen, Yanfei and Zhuang, Yufan and Lee, Yoonho and Huang, Chengsong and Yu, Han and CuiZhu, Zhongying and Ming, Yifei and Yao, Huaxiu and Gokturk, Burak and Pfister, Tomas and Lee, Chen-Yu},
  year = {2026},
  eprint = {2609.24972},
  archivePrefix = {arXiv},
  url = {https://arxiv.org/abs/2609.24972}
}

@misc{ke2026evoharnessbench,
  title = {{EvoHarnessBench: Can Your Agents Keep Pace with an Evolving Harness?}},
  author = {Ke, Zixuan and Patil, Vaidehi and Shi, Haizhou and Li, Yang and Liu, Ye and Shekkizhar, Sarath and Koul, Anurag and Wang, Jiayu and Nguyen, Xuan Phi and Yavuz, Semih and Bansal, Mohit and Joty, Shafiq},
  year = {2026},
  eprint = {2609.04280},
  archivePrefix = {arXiv},
  url = {https://arxiv.org/abs/2609.04280}
}

@inproceedings{liu2024evalcrafter,
  title = {{EvalCrafter: Benchmarking and Evaluating Large Video Generation Models}},
  author = {Liu, Yaofang and Cun, Xiaodong and Liu, Xuebo and Wang, Xintao and Zhang, Yong and Chen, Haoxin and Liu, Yang and Zeng, Tieyong and Chan, Raymond and Shan, Ying},
  booktitle = {Proceedings of the IEEE/CVF Conference on Computer Vision and Pattern Recognition},
  year = {2024},
  pages = {22139--22149},
  url = {https://openaccess.thecvf.com/content/CVPR2024/html/Liu_EvalCrafter_Benchmarking_and_Evaluating_Large_Video_Generation_Models_CVPR_2024_paper.html}
}

@inproceedings{kossen2021activetesting,
  title = {{Active Testing: Sample-Efficient Model Evaluation}},
  author = {Kossen, Jannik and Farquhar, Sebastian and Gal, Yarin and Rainforth, Tom},
  booktitle = {Proceedings of the 38th International Conference on Machine Learning},
  series = {Proceedings of Machine Learning Research},
  volume = {139},
  pages = {5753--5763},
  year = {2021},
  publisher = {PMLR},
  url = {https://proceedings.mlr.press/v139/kossen21a.html}
}

@inproceedings{bain2021frozen,
  title     = {Frozen in Time: A Joint Video and Image Encoder for End-to-End Retrieval},
  author    = {Bain, Max and Nagrani, Arsha and Varol, G{\"u}l and Zisserman, Andrew},
  booktitle = {Proceedings of the IEEE/CVF International Conference on Computer Vision},
  year      = {2021},
  eprint    = {2104.00650},
  archivePrefix= {arXiv},
  url       = {https://arxiv.org/abs/2104.00650},
  pages     = {1708--1718},
  doi       = {10.1109/ICCV48922.2021.00175}
}

@misc{blattmann2023stablevideo,
  title         = {Stable Video Diffusion: Scaling Latent Video Diffusion Models to Large Datasets},
  author        = {Blattmann, Andreas and Dockhorn, Tim and Kulal, Sumith and Mendelevitch, Daniel and Kilian, Maciej and Lorenz, Dominik and Levi, Yam and English, Zion and Voleti, Vikram and Letts, Adam and Jampani, Varun and Rombach, Robin},
  year          = {2023},
  eprint        = {2311.15127},
  archivePrefix = {arXiv},
  url           = {https://arxiv.org/abs/2311.15127}
}

@inproceedings{nan2024openvid,
  title     = {{OpenVid-1M}: A Large-Scale High-Quality Dataset for Text-to-Video Generation},
  author    = {Nan, Kepan and Xie, Rui and Zhou, Penghao and Fan, Tiehan and Yang, Zhenheng and Chen, Zhijie and Li, Xiang and Yang, Jian and Tai, Ying},
  booktitle = {International Conference on Learning Representations},
  year      = {2025},
  eprint    = {2407.02371},
  archivePrefix = {arXiv},
  url       = {https://arxiv.org/abs/2407.02371}
}

@inproceedings{chen2024panda,
  title     = {{Panda-70M}: Captioning 70M Videos with Multiple Cross-Modality Teachers},
  author    = {Chen, Tsai-Shien and Siarohin, Aliaksandr and Menapace, Willi and Deyneka, Ekaterina and Chao, Hsiang-wei and Jeon, Byung Eun and Fang, Yuwei and Lee, Hsin-Ying and Ren, Jian and Yang, Ming-Hsuan and Tulyakov, Sergey},
  booktitle = {Proceedings of the IEEE/CVF Conference on Computer Vision and Pattern Recognition},
  year      = {2024},
  eprint    = {2402.19479},
  archivePrefix= {arXiv},
  url       = {https://arxiv.org/abs/2402.19479},
  pages     = {13320--13331},
  doi       = {10.1109/CVPR52733.2024.01265}
}

@inproceedings{chen2024sharegpt4video,
  title     = {{ShareGPT4Video}: Improving Video Understanding and Generation with Better Captions},
  author    = {Chen, Lin and Wei, Xilin and Li, Jinsong and Dong, Xiaoyi and Zhang, Pan and Zang, Yuhang and Chen, Zehui and Duan, Haodong and Lin, Bin and Tang, Zhenyu and Yuan, Li and Qiao, Yu and Lin, Dahua and Zhao, Feng and Wang, Jiaqi},
  booktitle = {Advances in Neural Information Processing Systems},
  year      = {2024},
  eprint    = {2406.04325},
  archivePrefix = {arXiv},
  url       = {https://arxiv.org/abs/2406.04325}
}

@inproceedings{wang2025videoufo,
  title         = {{VideoUFO}: A Million-Scale User-Focused Dataset for Text-to-Video Generation},
  author        = {Wang, Wenhao and Yang, Yi},
  year          = {2025},
  eprint        = {2503.01739},
  archivePrefix = {arXiv},
  url           = {https://arxiv.org/abs/2503.01739},
  booktitle     = {Advances in Neural Information Processing Systems},
  volume        = {38},
  doi           = {10.52202/085713-0713}
}

@inproceedings{wang2025koala,
  title         = {{Koala-36M}: A Large-scale Video Dataset Improving Consistency between Fine-grained Conditions and Video Content},
  author        = {Wang, Qiuheng and Shi, Yukai and Ou, Jiarong and Chen, Rui and Lin, Ke and Wang, Jiahao and Jiang, Boyuan and Yang, Haotian and Zheng, Mingwu and Tao, Xin and Yang, Fei and Wan, Pengfei and Zhang, Di},
  year          = {2025},
  booktitle     = {Proceedings of the IEEE/CVF Conference on Computer Vision and Pattern Recognition (CVPR)},
  pages         = {8428--8437},
  month         = jun,
  url           = {https://arxiv.org/abs/2410.08260}
}

@inproceedings{savov2025autoexplore,
  title         = {Exploration-Driven Generative Interactive Environments},
  author        = {Savov, Nedko and Kazemi, Naser and Mahdi, Mohammad and Paudel, Danda Pani and Wang, Xi and Van Gool, Luc},
  booktitle     = {Proceedings of the IEEE/CVF Conference on Computer Vision and Pattern Recognition},
  year          = {2025},
  pages         = {27597--27607},
  eprint        = {2504.02515},
  archivePrefix = {arXiv},
  url           = {https://openaccess.thecvf.com/content/CVPR2025/html/Savov_Exploration-Driven_Generative_Interactive_Environments_CVPR_2025_paper.html},
  doi           = {10.1109/CVPR52734.2025.02570}
}

@inproceedings{bai2025ditto,
  title         = {Scaling Instruction-Based Video Editing with a High-Quality Synthetic Dataset},
  author        = {Bai, Qingyan and Wang, Qiuyu and Ouyang, Hao and Yu, Yue and Wang, Hanlin and Wang, Wen and Cheng, Ka Leong and Ma, Shuailei and Zeng, Yanhong and Liu, Zichen and Xu, Yinghao and Shen, Yujun and Chen, Qifeng},
  year          = {2026},
  eprint        = {2510.15742},
  archivePrefix = {arXiv},
  primaryClass  = {cs.CV},
  url           = {https://arxiv.org/abs/2510.15742},
  booktitle     = {Proceedings of the IEEE/CVF Conference on Computer Vision and Pattern Recognition (CVPR)},
  pages         = {37971--37981}
}

@inproceedings{guo2024animatediff,
  title = {{AnimateDiff: Animate Your Personalized Text-to-Image Diffusion Models without Specific Tuning}},
  author = {Yuwei Guo and Ceyuan Yang and Anyi Rao and Zhengyang Liang and Yaohui Wang and Yu Qiao and Maneesh Agrawala and Dahua Lin and Bo Dai},
  year = {2024},
  booktitle = {International Conference on Learning Representations},
  url = {https://proceedings.iclr.cc/paper_files/paper/2024/hash/94894cf9bab20c00f6eaaff3b7f2be70-Abstract-Conference.html}
}

@inproceedings{yuan2024instructvideo,
  title         = {{InstructVideo}: Instructing Video Diffusion Models with Human Feedback},
  author        = {Yuan, Hangjie and Zhang, Shiwei and Wang, Xiang and Wei, Yujie and Feng, Tao and Pan, Yining and Zhang, Yingya and Liu, Ziwei and Albanie, Samuel and Ni, Dong},
  booktitle     = {Proceedings of the IEEE/CVF Conference on Computer Vision and Pattern Recognition},
  year          = {2024},
  pages         = {6463--6474},
  eprint        = {2312.12490},
  archivePrefix = {arXiv},
  url           = {https://openaccess.thecvf.com/content/CVPR2024/html/Yuan_InstructVideo_Instructing_Video_Diffusion_Models_with_Human_Feedback_CVPR_2024_paper.html}
}

@inproceedings{guo2026vlaw,
  title         = {{VLAW}: Iterative Co-Improvement of Vision-Language-Action Policy and World Model},
  author        = {Guo, Yanjiang and Lee, Tony and Shi, Lucy Xiaoyang and Chen, Jianyu and Liang, Percy and Finn, Chelsea},
  year          = {2026},
  eprint        = {2602.12063},
  archivePrefix = {arXiv},
  primaryClass  = {cs.RO},
  url           = {https://arxiv.org/abs/2602.12063},
  booktitle     = {Proceedings of the 43rd International Conference on Machine Learning},
  series        = {Proceedings of Machine Learning Research},
  volume        = {306},
  pages         = {37998--38011}
}

@inproceedings{li2025t2vturbov2,
  title         = {{T2V-Turbo-v2}: Enhancing Video Model Post-Training through Data, Reward, and Conditional Guidance Design},
  author        = {Li, Jiachen and Long, Qian and Zheng, Jian and Gao, Xiaofeng and Piramuthu, Robinson and Chen, Wenhu and Wang, William Yang},
  booktitle     = {International Conference on Learning Representations (ICLR)},
  year          = {2025},
  eprint        = {2410.05677},
  archivePrefix = {arXiv},
  url           = {https://proceedings.iclr.cc/paper_files/paper/2025/hash/e68af7d8a44bc1964f6be4de464e38f9-Abstract-Conference.html}
}

@inproceedings{huang2025selfforcing,
  title         = {Self Forcing: Bridging the Train-Test Gap in Autoregressive Video Diffusion},
  author        = {Huang, Xun and Li, Zhengqi and He, Guande and Zhou, Mingyuan and Shechtman, Eli},
  booktitle     = {Advances in Neural Information Processing Systems},
  year          = {2025},
  eprint        = {2506.08009},
  archivePrefix = {arXiv},
  url           = {https://arxiv.org/abs/2506.08009}
}

@misc{feng2025followyourinstruction,
  title         = {{Follow-Your-Instruction}: A Comprehensive {MLLM} Agent for World Data Synthesis},
  author        = {Feng, Kunyu and Ma, Yue and Zhang, Xinhua and Liu, Boshi and Yuluo, Yikuang and Zhang, Yinhan and Liu, Runtao and Liu, Hongyu and Qin, Zhiyuan and Mo, Shanhui and Chen, Qifeng and Wang, Zeyu},
  year          = {2025},
  eprint        = {2508.05580},
  archivePrefix = {arXiv},
  primaryClass  = {cs.CV},
  url           = {https://arxiv.org/abs/2508.05580}
}

@inproceedings{liu2025videodpo,
  title         = {{VideoDPO}: Omni-Preference Alignment for Video Diffusion Generation},
  author        = {Liu, Runtao and Wu, Haoyu and Zheng, Ziqiang and Wei, Chen and He, Yingqing and Pi, Renjie and Chen, Qifeng},
  booktitle     = {Proceedings of the IEEE/CVF Conference on Computer Vision and Pattern Recognition},
  year          = {2025},
  pages         = {8009--8019},
  eprint        = {2412.14167},
  archivePrefix = {arXiv},
  url           = {https://openaccess.thecvf.com/content/CVPR2025/html/Liu_VideoDPO_Omni-Preference_Alignment_for_Video_Diffusion_Generation_CVPR_2025_paper.html}
}

@inproceedings{li2024t2vturbo,
  title         = {{T2V-Turbo}: Breaking the Quality Bottleneck of Video Consistency Model with Mixed Reward Feedback},
  author        = {Li, Jiachen and Feng, Weixi and Fu, Tsu-Jui and Wang, Xinyi and Basu, Sugato and Chen, Wenhu and Wang, William Yang},
  booktitle     = {Advances in Neural Information Processing Systems},
  year          = {2024},
  volume        = {37},
  eprint        = {2405.18750},
  archivePrefix = {arXiv},
  url           = {https://proceedings.neurips.cc/paper_files/paper/2024/hash/8a57aa8e8b57e64a42e95f7dceb0adb9-Abstract-Conference.html},
  pages         = {75692--75726},
  doi           = {10.52202/079017-2410}
}

@inproceedings{bruce2024genie,
  title     = {Genie: Generative Interactive Environments},
  author    = {Bruce, Jake and Dennis, Michael D. and Edwards, Ashley and Parker-Holder, Jack and Shi, Yuge and Hughes, Edward and Lai, Matthew and Mavalankar, Aditi and Steigerwald, Richie and Apps, Chris and Aytar, Yusuf and Bechtle, Sarah Maria Elisabeth and Behbahani, Feryal and Chan, Stephanie C. Y. and Heess, Nicolas and Gonzalez, Lucy and Osindero, Simon and Ozair, Sherjil and Reed, Scott and Zhang, Jingwei and Zolna, Konrad and Clune, Jeff and de Freitas, Nando and Singh, Satinder and Rockt{\"a}schel, Tim},
  booktitle = {Proceedings of the 41st International Conference on Machine Learning},
  volume    = {235},
  pages     = {4603--4623},
  year      = {2024},
  url       = {https://proceedings.mlr.press/v235/bruce24a.html}
}

@inproceedings{yang2024unisim,
  title     = {Learning Interactive Real-World Simulators},
  author    = {Yang, Sherry and Du, Yilun and Ghasemipour, Seyed Kamyar Seyed and Tompson, Jonathan and Kaelbling, Leslie Pack and Schuurmans, Dale and Abbeel, Pieter},
  booktitle = {International Conference on Learning Representations},
  year      = {2024},
  url       = {https://openreview.net/forum?id=sFyTZEqmUY}
}

@inproceedings{valevski2025gamengen,
  title     = {Diffusion Models Are Real-Time Game Engines},
  author    = {Valevski, Dani and Leviathan, Yaniv and Arar, Moab and Fruchter, Shlomi},
  booktitle = {International Conference on Learning Representations},
  year      = {2025},
  url       = {https://arxiv.org/abs/2408.14837}
}

@inproceedings{che2025gamegenx,
  title     = {{GameGen-X}: Interactive Open-World Game Video Generation},
  author    = {Che, Haoxuan and He, Xuanhua and Liu, Quande and Jin, Cheng and Chen, Hao},
  booktitle = {International Conference on Learning Representations},
  year      = {2025},
  url       = {https://arxiv.org/abs/2411.00769}
}

@article{nvidia2025cosmos,
  title   = {Cosmos World Foundation Model Platform for Physical {AI}},
  author  = {{NVIDIA} and Agarwal, Niket and Ali, Arslan and Bala, Maciej and Balaji, Yogesh and Barker, Erik and Cai, Tiffany and others},
  journal = {arXiv preprint arXiv:2501.03575},
  year    = {2025},
  url     = {https://arxiv.org/abs/2501.03575}
}

@inproceedings{gao2025adaworld,
  title     = {{AdaWorld}: Learning Adaptable World Models with Latent Actions},
  author    = {Gao, Shenyuan and Zhou, Siyuan and Du, Yilun and Zhang, Jun and Gan, Chuang},
  booktitle = {Proceedings of the 42nd International Conference on Machine Learning},
  volume    = {267},
  pages     = {18744--18771},
  year      = {2025},
  url       = {https://proceedings.mlr.press/v267/gao25u.html}
}

@inproceedings{chi2025eva,
  title     = {Empowering World Models with Reflection for Embodied Video Prediction},
  author    = {Chi, Xiaowei and Fan, Chun-Kai and Zhang, Hengyuan and Qi, Xingqun and Zhang, Rongyu and Chen, Anthony and Chan, Chi-Min and Xue, Wei and Liu, Qifeng and Zhang, Shanghang and Guo, Yike},
  booktitle = {Proceedings of the 42nd International Conference on Machine Learning},
  volume    = {267},
  pages     = {10383--10410},
  year      = {2025},
  url       = {https://proceedings.mlr.press/v267/chi25b.html}
}

@article{he2025matrixgame2,
  title   = {{Matrix-Game 2.0}: An Open-Source, Real-Time, and Streaming Interactive World Model},
  author  = {He, Xianglong and Peng, Chunli and Liu, Zexiang and Wang, Boyang and Zhang, Yifan and Cui, Qi and Kang, Fei and Jiang, Biao and An, Mengyin and Ren, Yangyang and Xu, Baixin and Guo, Hao-Xiang and Gong, Kaixiong and Wu, Size and Li, Wei and Song, Xuchen and Liu, Yang and Li, Yangguang and Zhou, Yahui},
  journal = {arXiv preprint arXiv:2508.13009},
  year    = {2025},
  url     = {https://arxiv.org/abs/2508.13009}
}

@inproceedings{sun2026worldplay,
  title     = {{WorldPlay}: Towards Long-Term Geometric Consistency for Real-Time Interactive World Modeling},
  author    = {Sun, Wenqiang and Zhang, Haiyu and Wang, Haoyuan and Wu, Junta and Wang, Zehan and Wang, Zhenwei and Wang, Yunhong and Zhang, Jun and Wang, Tengfei and Guo, Chunchao},
  booktitle = {Proceedings of the 43rd International Conference on Machine Learning},
  volume    = {306},
  year      = {2026},
  url       = {https://arxiv.org/abs/2512.14614}
}

@article{sima2025sima2,
  title   = {{SIMA 2}: A Generalist Embodied Agent for Virtual Worlds},
  author  = {{SIMA Team}},
  journal = {arXiv preprint arXiv:2512.04797},
  year    = {2025},
  url     = {https://arxiv.org/abs/2512.04797}
}

@article{ye2026dreamzero,
  title   = {World Action Models Are Zero-Shot Policies},
  author  = {Ye, Seonghyeon and Ge, Yunhao and Zheng, Kaiyuan and Gao, Shenyuan and Yu, Sihyun and Kurian, George and Indupuru, Suneel and Tan, You Liang and Zhu, Chuning and Xiang, Jiannan and others},
  journal = {arXiv preprint arXiv:2602.15922},
  year    = {2026},
  url     = {https://arxiv.org/abs/2602.15922}
}

@article{jiang2026abotworld,
  title   = {{ABot-World-0}: Infinite Interactive World Rollout on a Single Desktop {GPU}},
  author  = {Jiang, Fan and Sun, Zhaoxu and Wang, Mengchao and Zhu, Ziyu and Wang, Chiyu and Zhang, Yunpeng and Liu, Wenlin and Wang, Yun and Zheng, Xue and Sun, Rui and Ni, Junfeng and Pan, Hongyu and others},
  journal = {arXiv preprint arXiv:2607.19191},
  year    = {2026},
  url     = {https://arxiv.org/abs/2607.19191}
}

@article{ding2026playworld,
  title   = {{PlayWorld}: Benchmarking World Models with Agent Players over Long-Horizon Objectives},
  author  = {Ding, Kaixin and Chen, Xi and Cai, Minghong and Xu, Zhiyuan and Wang, Yiyang and Lu, Yuxiang and Li, Junyi and Chen, Shuyang and Gao, Yuan and Tao, Xin and Wan, Pengfei and Zhao, Hengshuang},
  journal = {arXiv preprint arXiv:2608.13552},
  year    = {2026},
  url     = {https://arxiv.org/abs/2608.13552}
}

@inproceedings{anime2025,
  title = {{AniME: Adaptive Multi-Agent Planning for Long Animation Generation}},
  author = {Zhang, Lisai and Xu, Baohan and Yang, Siqian and Yin, Mingyu and Liu, Jing and Xu, Chao and Wang, Siqi and Wu, Yidi and Hong, Yuxin and Zhang, Zihao and Liang, Yanzhang and Jiang, Yudong},
  year = {2025},
  booktitle = {SIGGRAPH Asia 2025: Posters},
  url = {https://doi.org/10.1145/3757374.3771455},
  doi = {10.1145/3757374.3771455},
}

@misc{scriptallneed2026,
  title = {{The Script is All You Need: An Agentic Framework for Long-Horizon Dialogue-to-Cinematic Video Generation}},
  author = {Mu, Chenyu and He, Xin and Yang, Qu and Chen, Wanshun and Yao, Jiadi and Liu, Huang and Yi, Zihao and Zhao, Bo and Chen, Xingyu and Ma, Ruotian and Ye, Fanghua and Yang, Erkun and Deng, Cheng and Tu, Zhaopeng and Li, Xiaolong and Linus},
  year = {2026},
  howpublished = {Preprint},
  url = {https://arxiv.org/abs/2601.17737},
  eprint = {2601.17737},
  archivePrefix = {arXiv},
}

@inproceedings{lee2026videorepair,
  title = {Self-Correcting Text-to-Video Generation with Misalignment Detection and Localized Refinement},
  author = {Lee, Daeun and Yoon, Jaehong and Cho, Jaemin and Bansal, Mohit},
  year = {2026},
  url = {https://arxiv.org/abs/2411.15115},
  booktitle = {Findings of the Association for Computational Linguistics: ACL 2026},
  pages = {36464--36489},
  doi = {10.18653/v1/2026.findings-acl.1817}
}

@article{frolov2021surveyn01,
  title = {{Adversarial Text-to-Image Synthesis: A Review}},
  author = {Frolov, Stanislav and Hinz, Tobias and Raue, Federico and Hees, Jörn and Dengel, Andreas},
  journal = {arXiv preprint arXiv:2101.09983},
  year = {2021},
  url = {https://arxiv.org/abs/2101.09983v2}
}

@article{yang2022surveyn08,
  title = {{Diffusion Models: A Comprehensive Survey of Methods and Applications}},
  author = {Yang, Ling and Zhang, Zhilong and Song, Yang and Hong, Shenda and Xu, Runsheng and Zhao, Yue and Zhang, Wentao and Cui, Bin and Yang, Ming-Hsuan},
  journal = {arXiv preprint arXiv:2209.00796},
  year = {2022},
  url = {https://arxiv.org/abs/2209.00796v15}
}

@article{bie2023surveyn09,
  title = {{RenAIssance: A Survey into AI Text-to-Image Generation in the Era of Large Model}},
  author = {Bie, Fengxiang and Yang, Yibo and Zhou, Zhongzhu and Ghanem, Adam and Zhang, Minjia and Yao, Zhewei and Wu, Xiaoxia and Holmes, Connor and Golnari, Pareesa and Clifton, David A. and He, Yuxiong and Tao, Dacheng and Song, Shuaiwen Leon},
  journal = {arXiv preprint arXiv:2309.00810},
  year = {2023},
  url = {https://arxiv.org/abs/2309.00810v1}
}

@article{zhang2023surveyn02,
  title = {{Text-to-image Diffusion Models in Generative AI: A Survey}},
  author = {Zhang, Chenshuang and Zhang, Chaoning and Zhang, Mengchun and Kweon, In So and Kim, Junmo},
  journal = {arXiv preprint arXiv:2303.07909},
  year = {2023},
  url = {https://arxiv.org/abs/2303.07909v3}
}

@misc{sun2024sorasurvey,
  title={From Sora What We Can See: A Survey of Text-to-Video Generation},
  author={Rui Sun and Yumin Zhang and Tejal Shah and Jiahao Sun and Shuoying Zhang and Wenqi Li and Haoran Duan and Bo Wei and Rajiv Ranjan},
  year={2024},
  eprint={2405.10674},
  archivePrefix={arXiv},
  url={https://arxiv.org/html/2405.10674v1},
  note={Coverage inspected in v1; 2024-05-17}
}

@article{xiong2024surveyn06,
  title = {{Autoregressive Models in Vision: A Survey}},
  author = {Xiong, Jing and Liu, Gongye and Huang, Lun and Wu, Chengyue and Wu, Taiqiang and Mu, Yao and Yao, Yuan and Shen, Hui and Wan, Zhongwei and Huang, Jinfa and Tao, Chaofan and Yan, Shen and Yao, Huaxiu and Kong, Lingpeng and Yang, Hongxia and Zhang, Mi and Sapiro, Guillermo and Luo, Jiebo and Luo, Ping and Wong, Ngai},
  journal = {arXiv preprint arXiv:2411.05902},
  year = {2024},
  url = {https://arxiv.org/abs/2411.05902v2}
}

@article{melnik2024videodiffusionsurvey,
  title={Video Diffusion Models: A Survey},
  author={Andrew Melnik and Michal Ljubljanac and Cong Lu and Qi Yan and Weiming Ren and Helge Ritter},
  year={2024},
  eprint={2405.03150},
  archivePrefix={arXiv},
  url={https://openreview.net/forum?id=rJSHjhEYJx},
  note={Coverage assessed from the author version, arXiv:2405.03150v2},
  journal={Transactions on Machine Learning Research}
}

@article{huang2024surveyn03,
  title = {{Diffusion Model-Based Image Editing: A Survey}},
  author = {Huang, Yi and Huang, Jiancheng and Liu, Yifan and Yan, Mingfu and Lv, Jiaxi and Liu, Jianzhuang and Xiong, Wei and Zhang, He and Cao, Liangliang and Chen, Shifeng},
  journal = {arXiv preprint arXiv:2402.17525},
  year = {2024},
  url = {https://arxiv.org/abs/2402.17525v3}
}

@misc{sun2024videoeditingsurvey,
  title={Diffusion Model-Based Video Editing: A Survey},
  author={Wenhao Sun and Rong-Cheng Tu and Jingyi Liao and Dacheng Tao},
  year={2024},
  eprint={2407.07111},
  archivePrefix={arXiv},
  url={https://arxiv.org/html/2407.07111v1},
  note={Coverage inspected in v1; 2024-06-26}
}

@article{cao2024surveyn04,
  title = {{Controllable Generation with Text-to-Image Diffusion Models: A Survey}},
  author = {Cao, Pu and Zhou, Feng and Song, Qing and Yang, Lu},
  journal = {arXiv preprint arXiv:2403.04279},
  year = {2024},
  url = {https://arxiv.org/abs/2403.04279v2}
}

@misc{li2024longvideosurvey,
  title={A Survey on Long Video Generation: Challenges, Methods, and Prospects},
  author={Chengxuan Li and Di Huang and Zeyu Lu and Yang Xiao and Qingqi Pei and Lei Bai},
  year={2024},
  eprint={2403.16407},
  archivePrefix={arXiv},
  url={https://arxiv.org/html/2403.16407v1},
  note={Coverage inspected in v1; 2024-03-25}
}

@article{waseem2024surveyn07,
  title = {{Video Is Worth a Thousand Images: Exploring the Latest Trends in Long Video Generation}},
  author = {Waseem, Faraz and Shahzad, Muhammad},
  journal = {arXiv preprint arXiv:2412.18688},
  year = {2024},
  url = {https://arxiv.org/abs/2412.18688v2}
}

@article{chen2026surveyn14,
  title = {{The Past Frames the Future: Memory for Autoregressive Video Generation}},
  author = {Chen, Harold Haodong and Guo, Rongjin and Lan, Disen and Shu, Wen-Jie and Zhang, Hongfei and Hu, Hanzhe and Yao, Shengtao and Zhang, Zixin and Zhang, Guibin and Rao, Zhefan and Liu, Jinxiu and Liu, Yexin and Peng, Rui and Liu, Yuhao and Ren, Bin and Yang, Shuai and Chen, Yukang and Khan, Salman and Chen, Ying-Cong and Lim, Ser-Nam and Lau, Rynson W. H. and Sebe, Nicu and Cheng, Yu and Yang, Ming-Hsuan and Chen, Qifeng},
  journal = {arXiv preprint arXiv:2609.28466},
  year = {2026},
  url = {https://arxiv.org/abs/2609.28466v1}
}

@misc{lei2024humanvideosurvey,
  title={A Comprehensive Survey on Human Video Generation: Challenges, Methods, and Insights},
  author={Wentao Lei and Jinting Wang and Fengji Ma and Guanjie Huang and Li Liu},
  year={2024},
  eprint={2407.08428},
  archivePrefix={arXiv},
  url={https://arxiv.org/html/2407.08428v1},
  note={Coverage inspected in v1; 2024-07-11}
}

@misc{ma2025controllablevideo,
  title={Controllable Video Generation: A Survey},
  author={Yue Ma and Kunyu Feng and Zhongyuan Hu and Xinyu Wang and Yucheng Wang and Mingzhe Zheng and Bingyuan Wang and Qinghe Wang and Xuanhua He and Hongfa Wang and Chenyang Zhu and Hongyu Liu and Yingqing He and Zeyu Wang and Zhifeng Li and Xiu Li and Sirui Han and Yike Guo and Wei Liu and Dan Xu and Linfeng Zhang and Qifeng Chen},
  year={2025},
  eprint={2507.16869},
  archivePrefix={arXiv},
  url={https://arxiv.org/html/2507.16869v3},
  note={Coverage inspected in v3; 2026-01-19}
}

@misc{lin2025physicscognitionsurvey,
  title={Exploring the Evolution of Physics Cognition in Video Generation: A Survey},
  author={Minghui Lin and Xiang Wang and Yishan Wang and Shu Wang and Fengqi Dai and Pengxiang Ding and Cunxiang Wang and Zhengrong Zuo and Nong Sang and Siteng Huang and Donglin Wang},
  year={2025},
  eprint={2503.21765},
  archivePrefix={arXiv},
  url={https://arxiv.org/html/2503.21765v1},
  note={Coverage inspected in v1; 2025-03-27}
}

@article{liu2024surveyn11,
  title = {{Alignment of Diffusion Models: Fundamentals, Challenges, and Future}},
  author = {Liu, Buhua and Shao, Shitong and Li, Bao and Bai, Lichen and Xu, Zhiqiang and Xiong, Haoyi and Kwok, James and Helal, Sumi and Xie, Zeke},
  journal = {arXiv preprint arXiv:2409.07253},
  year = {2024},
  url = {https://arxiv.org/abs/2409.07253v4}
}

@article{wu2025surveyn05,
  title = {{Preference Alignment on Diffusion Model: A Comprehensive Survey for Image Generation and Editing}},
  author = {Wu, Sihao and Si, Xiaonan and Xing, Chi and Wang, Jianhong and Jin, Gaojie and Cheng, Guangliang and Zhang, Lijun and Huang, Xiaowei},
  journal = {arXiv preprint arXiv:2502.07829},
  year = {2025},
  url = {https://arxiv.org/abs/2502.07829v1}
}

@misc{liu2021persuasivevideosurvey,
  title={AI-Empowered Persuasive Video Generation: A Survey},
  author={Chang Liu and Han Yu},
  year={2021},
  eprint={2112.09401},
  archivePrefix={arXiv},
  url={https://arxiv.org/html/2112.09401v1},
  note={Coverage inspected in v1; 2021-12-17}
}

@misc{he2024llmsgenerationediting,
  title={LLMs Meet Multimodal Generation and Editing: A Survey},
  author={Yingqing He and Zhaoyang Liu and Jingye Chen and Zeyue Tian and Hongyu Liu and Xiaowei Chi and Runtao Liu and Ruibin Yuan and Yazhou Xing and Wenhai Wang and Jifeng Dai and Yong Zhang and Wei Xue and Qifeng Liu and Yike Guo and Qifeng Chen},
  year={2024},
  eprint={2405.19334},
  archivePrefix={arXiv},
  url={https://arxiv.org/html/2405.19334v2},
  note={Coverage inspected in v2; 2024-06-09}
}

@misc{mei2026beyond,
  title        = {{Beyond Visual Generation: Frontiers, Challenges, and Future Directions in Agentic Visual Creation}},
  author       = {Mei, Hefei and
                  He, Hangzhou and
                  Qiu, Haiyi and
                  Yang, Longrong and
                  Duan, Lunhao and
                  Zhao, Shanshan and
                  Zhan, Pengxin and
                  Chen, Qing-Guo and
                  Xu, Zhao and
                  Luo, Weihua},
  year         = {2026},
  month        = sep,
  url          = {https://www.preprints.org/manuscript/202609.0682}
}

@misc{xing2026agenticvisualsurvey,
  title={Agentic Visual Generation: A Survey},
  author={Xing, Zihan and Wang, Jin and Xia, Rong and Ye, Keming and Qu, Zhan and Yi, Biao and Liu, Tianqi and Chen, Junhao and Yang, Jie and Zhu, Zhibo and Shen, Zhouzhou and Sheng, Honghui and Chen, Yurun and Zhang, Yuqing and Zhu, Shuanghe and Wang, Wenkai and Xiong, Tao and Lin, Kuncheng and Yu, Qihang and Chen, Kui and Xiong, Yufan and Li, Juncheng and Zhang, Wenqiao and Lv, Zheqi and Zang, Chengxi and Chen, Long and Zhao, Zhou and Zhang, Shengyu and Wu, Fei},
  year={2026},
  howpublished={Preprints.org},
  url={https://www.preprints.org/manuscript/202609.0173/v1},
  note={Version 1, posted 2 September 2026; not peer-reviewed}
}

@misc{puspitasari2024soraworldsurvey,
  title={Sora as a World Model? A Complete Survey on Text-to-Video Generation},
  author={Fachrina Dewi Puspitasari and Chaoning Zhang and Joseph Cho and Adnan Haider and Noor Ul Eman and Omer Amin and Alexis Mankowski and Muhammad Umair and Jingyao Zheng and Sheng Zheng and Lik-Hang Lee and Caiyan Qin and Tae-Ho Kim and Choong Seon Hong and Yang Yang and Heng Tao Shen},
  year={2024},
  eprint={2403.05131},
  archivePrefix={arXiv},
  url={https://arxiv.org/html/2403.05131v3},
  note={Coverage inspected in v3; 2026-01-21}
}

@misc{ding2024worldmodels,
  title={Understanding World or Predicting Future? A Comprehensive Survey of World Models},
  author={Ding, Jingtao and Zhang, Yunke and Shang, Yu and Feng, Jie and Zhang, Yuheng and Zong, Zefang and Yuan, Yuan and Su, Hongyuan and Li, Nian and Piao, Jinghua and Deng, Yucheng and Sukiennik, Nicholas and Gao, Chen and Xu, Fengli and Li, Yong},
  year={2024},
  eprint={2411.14499},
  archivePrefix={arXiv},
  url={https://arxiv.org/html/2411.14499v4},
  note={Coverage inspected in v4; 10 Dec 2025}
}

@article{yue2025surveyn13,
  title = {{Simulating the Visual World with Artificial Intelligence: A Roadmap}},
  author = {Yue, Jingtong and Huang, Ziqi and Chen, Zhaoxi and Wang, Xintao and Wan, Pengfei and Liu, Ziwei},
  journal = {arXiv preprint arXiv:2511.08585},
  year = {2025},
  url = {https://arxiv.org/abs/2511.08585v2}
}

@article{yu2025surveyn12,
  title = {{A Survey of Interactive Generative Video}},
  author = {Yu, Jiwen and Qin, Yiran and Che, Haoxuan and Liu, Quande and Wang, Xintao and Wan, Pengfei and Zhang, Di and Gai, Kun and Chen, Hao and Liu, Xihui},
  journal = {arXiv preprint arXiv:2504.21853},
  year = {2025},
  url = {https://arxiv.org/abs/2504.21853v1}
}

@misc{li2025embodiedworldmodels,
  title={A Comprehensive Survey on World Models for Embodied AI},
  author={Li, Xinqing and He, Xin and Zhang, Le and Wu, Min and Li, Xiaoli and Liu, Yun},
  year={2025},
  eprint={2510.16732},
  archivePrefix={arXiv},
  url={https://arxiv.org/html/2510.16732v3},
  note={Coverage inspected in v3; 25 Jun 2026}
}

@article{zhang2025surveyn18,
  title = {{A Step Toward World Models: A Survey on Robotic Manipulation}},
  author = {Zhang, Peng-Fei and Cheng, Ying and Sun, Xiaofan and Wang, Shijie and Li, Fengling and Zhu, Lei and Shen, Heng Tao},
  journal = {arXiv preprint arXiv:2511.02097},
  year = {2025},
  url = {https://arxiv.org/abs/2511.02097v2}
}

@article{mei2026surveyn19,
  title = {{Robotic Video World Models: A Survey of Applications, Research Challenges, Future Directions}},
  author = {Mei, Zhiting and Yin, Tenny and Shorinwa, Ola and Badithela, Apurva and Zheng, Zhonghe and Bruno, Joseph and Bland, Madison and Zha, Lihan and Hancock, Asher and Fisac, Jaime Fernández and Dames, Philip and Majumdar, Anirudha},
  journal = {arXiv preprint arXiv:2601.07823},
  year = {2026},
  url = {https://arxiv.org/abs/2601.07823v2}
}

@article{wang2026surveyn20,
  title = {{World Models for Robotic Manipulation: A Survey}},
  author = {Wang, Fangyuan and Wang, Ziyuan and Pei, Guorui and Zhang, Mengshi and Liang, Canxi and Hu, Jun and Li, Zhongxuan and Wu, Jinsong and Han, Ning and Zhang, Zeqing and Qi, Jiaming and Wu, Hongmin and Zhang, Shiyao and Zheng, Pai and Pan, Jia and Navarro-Alarcon, David and Liu, Sichao and Zhou, Peng},
  journal = {arXiv preprint arXiv:2606.00113},
  year = {2026},
  url = {https://arxiv.org/abs/2606.00113v1}
}

@article{lu2026surveyn17,
  title = {{World-Action Models for Robot Learning and Control: A Survey}},
  author = {Lu, Zuxing and Zhai, Hongjia and Wang, Guanzhi and Zeng, Huajian and Yang, Jiaqi and Liu, Jingyu and Cheng, Lei and Zhang, Yuantai and Qiu, Yuheng and Cheng, Zezhou and Laptev, Ivan and Xu, Danfei and Riviere, Benjamin and Loianno, Giuseppe and Xing, Eric and Zuo, Xingxing},
  journal = {arXiv preprint arXiv:2609.16074},
  year = {2026},
  url = {https://arxiv.org/abs/2609.16074v1}
}

@misc{hou2026robotworldmodels,
  title={World Model for Robot Learning: A Comprehensive Survey},
  author={Hou, Bohan and Li, Gen and Jia, Jindou and An, Tuo and Guo, Xinying and Leng, Sicong and Geng, Haoran and Ze, Yanjie and Harada, Tatsuya and Torr, Philip and Mees, Oier and Pollefeys, Marc and Liu, Zhuang and Wu, Jiajun and Abbeel, Pieter and Malik, Jitendra and Du, Yilun and Yang, Jianfei},
  year={2026},
  eprint={2605.00080},
  archivePrefix={arXiv},
  url={https://arxiv.org/html/2605.00080v1},
  note={Coverage inspected in v1; 30 Apr 2026}
}

@article{yao2026surveyn16,
  title = {{World Models for Embodied Intelligence: From Plausible to Controllable to Actionable}},
  author = {Yao, Nanjie and Wang, Hao and Cheng, Chong and Chen, Zhikang and Li, Wenzhe and Lyu, Jiafei and Shen, Li and Zhao, Peilin and Lu, Zongqing and Huang, Gao and Hoi, Steven and Tao, Dacheng and Ye, Deheng},
  journal = {arXiv preprint arXiv:2609.16697},
  year = {2026},
  url = {https://arxiv.org/abs/2609.16697v1}
}

@article{tao2024surveyr01,
  title = {{A Survey on Self-Evolution of Large Language Models}},
  author = {Tao, Zhengwei and Lin, Ting-En and Chen, Xiancai and Li, Hangyu and Wu, Yuchuan and Li, Yongbin and Jin, Zhi and Huang, Fei and Tao, Dacheng and Zhou, Jingren},
  journal = {arXiv preprint arXiv:2404.14387},
  year = {2024},
  url = {https://arxiv.org/abs/2404.14387v2}
}

@article{gao2025surveyr02,
  title = {{A Survey of Self-Evolving Agents: What, When, How, and Where to Evolve on the Path to Artificial Super Intelligence}},
  author = {Gao, Huan-ang and Geng, Jiayi and Hua, Wenyue and Hu, Mengkang and Juan, Xinzhe and Liu, Hongzhang and Liu, Shilong and Qiu, Jiahao and Qi, Xuan and Wu, Yiran and Wang, Hongru and Xiao, Han and Zhou, Yuhang and Zhang, Shaokun and Zhang, Jiayi and Xiang, Jinyu and Fang, Yixiong and Zhao, Qiwen and Liu, Dongrui and Ren, Qihan and Qian, Cheng and Wang, Zhenhailong and Hu, Minda and Wang, Huazheng and Wu, Qingyun and Ji, Heng and Wang, Mengdi},
  journal = {arXiv preprint arXiv:2507.21046},
  year = {2025},
  url = {https://arxiv.org/pdf/2507.21046v4}
}

@article{chen2026surveyr04,
  title = {{Recursive Self-Improvement in AI: From Bounded Self-Refinement to Autonomous Research Loops}},
  author = {Chen, Mingguang and Wang, Licheng and Qu, Bo},
  journal = {arXiv preprint arXiv:2607.07663},
  year = {2026},
  url = {https://arxiv.org/abs/2607.07663v2}
}

@article{ren2026surveyr03,
  title = {{Self-Improvements in Modern Agentic Systems: A Survey}},
  author = {Ren, Zhe and Chen, Yimeng and Guo, Dandan and Rong, Guowei and Li, Tonghui and Xiong, R. B. and Lan, Qingfeng and Wang, Wenyi and Nanbo, Li and Yang, Yibo and Zhuge, Mingchen and Schmidhuber, Jürgen},
  journal = {arXiv preprint arXiv:2607.13104},
  year = {2026},
  url = {https://arxiv.org/pdf/2607.13104v1}
}

@article{duan2026surveyn21,
  title = {{The Last AI Built by Humans: Toward Genuine Recursive Self-Improvement}},
  author = {Duan, Yi and Liu, Ying and Tang, Zirui and Chen, Haodong and Zhou, Jun and Liu, Yumou and Xu, Bangrui and Wu, Yukai and Chen, Sidi and Zhou, Yuhan and Wang, Haoyu and Yu, Xiaoyou and Han, Shaokun and Zhu, Xuzhou and Zhou, Le and Lu, Bolin and Zhou, Wei and Liu, Jiachen and Fang, Nuozhou and Tian, Jiaxin and Chen, Ruoyu and Li, Yuxuan and Zuo, Kai and Zhang, Kaiyan and Yang, Qianyu and Wang, Zijie and Qiu, Jiantao and He, Conghui and Li, Guoliang and Zhou, Bowen and Liu, Zhiyuan and Wen, Zhoufutu and Kang, Jihua and Zhou, Xuanhe and Wu, Fan},
  journal = {arXiv preprint arXiv:2609.11873},
  year = {2026},
  url = {https://arxiv.org/abs/2609.11873v3}
}

@misc{chen2026javedit,
  title         = {{JAVEDIT}: Joint Audio-Visual Instruction-Guided Video Editing with Agentic Data Curation},
  author        = {Yinan Chen and Chuming Lin and Zhennan Chen and Yuxiang Zeng and Junwei Zhu and Yali Bi and Xijie Huang and Chengming Xu and Donghao Luo and Zhucun Xue and Xiaobin Hu and Chengjie Wang and Yong Liu and Jiangning Zhang and Shuicheng Yan},
  year          = {2026},
  eprint        = {2606.03168},
  archivePrefix = {arXiv},
  primaryClass  = {cs.CV},
  url           = {https://arxiv.org/abs/2606.03168}
}

@misc{li2026avaencoder,
  title         = {{AVA-Encoder}: Towards Agent-Native Video Representation Learning},
  author        = {Chuyue Li and Jinpeng Yu and Haozhe Wang and Tian Xueyun and Zhijing Zhang and Bingnan Li and Shuqi Gu and Kan Ren and Jiaming Liu and Ruihua Huang},
  year          = {2026},
  eprint        = {2608.12313},
  archivePrefix = {arXiv},
  primaryClass  = {cs.CV},
  url           = {https://arxiv.org/abs/2608.12313v2}
}

@inproceedings{phyt2v2025,
  title = {{PhyT2V: LLM-Guided Iterative Self-Refinement for Physics-Grounded Text-to-Video Generation}},
  author = {Xue, Qiyao and Yin, Xiangyu and Yang, Boyuan and Gao, Wei},
  year = {2025},
  booktitle = {Proceedings of the IEEE/CVF Conference on Computer Vision and Pattern Recognition},
  url = {https://openaccess.thecvf.com/content/CVPR2025/html/Xue_PhyT2V_LLM-Guided_Iterative_Self-Refinement_for_Physics-Grounded_Text-to-Video_Generation_CVPR_2025_paper.html},
  pages = {18826--18836},
}

@inproceedings{videoshield2025,
  title = {{VideoShield}: Regulating Diffusion-based Video Generation Models via Watermarking},
  author = {Hu, Runyi and Zhang, Jie and Li, Yiming and Li, Jiwei and Guo, Qing and Qiu, Han and Zhang, Tianwei},
  booktitle = {International Conference on Learning Representations},
  year = {2025},
  url = {https://proceedings.iclr.cc/paper_files/paper/2025/hash/8227285e32f70e07fa3a247f3a48006d-Abstract-Conference.html}
}

@inproceedings{sigmark2026,
  title = {{SIGMark}: Scalable In-Generation Watermark with Blind Extraction for Video Diffusion},
  author = {Zhu, Xinjie and Zhao, Zijing and Jin, Hui and Guo, Qingxiao and Ma, Yilong and Wang, Yunhao and Guo, Xiaobing and Zhang, Weifeng},
  booktitle = {International Conference on Learning Representations},
  year = {2026},
  url = {https://proceedings.iclr.cc/paper_files/paper/2026/hash/f3f6f1739b646e0bd20111261ce23adb-Abstract-Conference.html}
}

@inproceedings{filmagent2024,
  title = {{FilmAgent: Automating Virtual Film Production Through a Multi-Agent Collaborative Framework}},
  author = {Zhenran Xu and Jifang Wang and Longyue Wang and Zhouyi Li and Senbao Shi and Baotian Hu and Min Zhang},
  year = {2024},
  booktitle = {SIGGRAPH Asia 2024: Technical Communications},
  url = {https://doi.org/10.1145/3681758.3698014},
  doi = {10.1145/3681758.3698014}
}

@InProceedings{holocine2026,
    author    = {Meng, Yihao and Ouyang, Hao and Yu, Yue and Wang, Qiuyu and Wang, Wen and Cheng, Ka Leong and Wang, Hanlin and Ma, Shuailei and Li, Yixuan and Chen, Cheng and Zeng, Yanhong and Zhu, Xing and Shen, Yujun and Qu, Huamin},
    title     = {HoloCine: Holistic Generation of Cinematic Multi-Shot Long Video Narratives},
    booktitle = {Proceedings of the IEEE/CVF Conference on Computer Vision and Pattern Recognition (CVPR)},
    month     = {June},
    year      = {2026},
    pages     = {461-471},
    url = {https://openaccess.thecvf.com/content/CVPR2026/html/Meng_HoloCine_Holistic_Generation_of_Cinematic_Multi-Shot_Long_Video_Narratives_CVPR_2026_paper.html}
}

@InProceedings{vace2025,
    author    = {Jiang, Zeyinzi and Han, Zhen and Mao, Chaojie and Zhang, Jingfeng and Pan, Yulin and Liu, Yu},
    title     = {VACE: All-in-One Video Creation and Editing},
    booktitle = {Proceedings of the IEEE/CVF International Conference on Computer Vision (ICCV)},
    month     = {October},
    year      = {2025},
    pages     = {17191-17202},
    url = {https://openaccess.thecvf.com/content/ICCV2025/html/Jiang_VACE_All-in-One_Video_Creation_and_Editing_ICCV_2025_paper.html}
}

@InProceedings{spdmark2026,
    author    = {Fares, Samar and Tastan, Nurbek and Nandakumar, Karthik},
    title     = {SPDMark: Selective Parameter Displacement for Robust Video Watermarking},
    booktitle = {Proceedings of the IEEE/CVF Conference on Computer Vision and Pattern Recognition (CVPR)},
    month     = {June},
    year      = {2026},
    pages     = {10303-10312},
    url = {https://openaccess.thecvf.com/content/CVPR2026/html/Fares_SPDMark_Selective_Parameter_Displacement_for_Robust_Video_Watermarking_CVPR_2026_paper.html}
}

@article{rbmark2025,
  title = {RBMark: Robust and blind video watermark in DT CWT domain},
  author = {Huang, I.-Chun and Wu, Ji-Yan and Ooi, Wei Tsang},
  journal = {Journal of Visual Communication and Image Representation},
  year = {2025},
  url = {https://doi.org/10.1016/j.jvcir.2025.104438}
}

@inproceedings{sepvamark2025,
  title = {SepVAMark: Deep Separable Visual-Audio Fusion Watermarking for Source Tracing and Deepfake Detection},
  author = {Zhang, Chuan and Li, Zihan and Xu, Zihao and Ren, Xuhao and Zhu, Liehuang},
  booktitle = {Proceedings of the 33rd ACM International Conference on Multimedia},
  year = {2025},
  url = {https://doi.org/10.1145/3746027.3755783},
  pages = {8910--8919},
  doi = {10.1145/3746027.3755783}
}

@misc{mindofdirector2026,
  title = {{Mind-of-Director: Multi-modal Agent-Driven Film Previsualization via Collaborative Decision-Making}},
  author = {Nan, Shufeng and Li, Mengtian and Zheng, Sixiao and Lu, Yuwei and Zhang, Han and Fu, Yanwei},
  year = {2026},
  howpublished = {Preprint},
  url = {https://arxiv.org/abs/2603.14790},
  eprint = {2603.14790},
  archivePrefix = {arXiv},
}

@misc{muse2026,
  title = {{MUSE: A Multi-agent Framework for Unconstrained Story Envisioning via Closed-Loop Cognitive Orchestration}},
  author = {Sun, Wenzhang and Wang, Zhenyu and Hu, Zhangchi and Wang, Chunfeng and Li, Hao and Chen, Wei},
  year = {2026},
  howpublished = {Preprint},
  url = {https://arxiv.org/abs/2602.03028},
  eprint = {2602.03028},
  archivePrefix = {arXiv},
}

@inproceedings{code2video2026,
  title = {{Code2Video: A Code-centric Paradigm for Educational Video Creation}},
  author = {Chen, Yanzhe and Lin, Kevin Qinghong and Shou, Mike Zheng},
  year = {2026},
  booktitle = {ICML 2026},
  url = {https://icml.cc/virtual/2026/poster/65050},
}

@inproceedings{evopresent2026,
  title = {{Presenting a Paper is an Art: Self-Improvement Aesthetic Agents for Academic Presentations}},
  author = {Liu, Chengzhi and Yang, Yuzhe and Zhou, Kaiwen and Zhang, Zhen and Fan, Yue and Xie, Yanan and Qi, Peng and Wang, Xin Eric},
  year = {2026},
  booktitle = {ICLR 2026},
  url = {https://proceedings.iclr.cc/paper_files/paper/2026/hash/d848cb2c84f0bba7f1f73cf232734c40-Abstract-Conference.html},
}

@inproceedings{fantasyhsi2026,
  title = {{FantasyHSI: Video-Generation-Centric 4D Human Synthesis in Any Scene Through a Graph-Based Multi-Agent Framework}},
  author = {Lingzhou Mu and Wang Qiang and Fan Jiang and Mengchao Wang and Mu Xu and Kai Zhang},
  year = {2026},
  booktitle = {AAAI 2026},
  url = {https://ojs.aaai.org/index.php/AAAI/article/view/37758},
  doi = {10.1609/aaai.v40i10.37758},
  pages = {8116--8124},
}

@inproceedings{liver2026,
  title = {{Lighting-grounded Video Generation with Renderer-based Agent Reasoning}},
  author = {Cai, Ziqi and Yang, Taoyu and Chang, Zheng and Li, Si and Jiang, Han and Weng, Shuchen and Shi, Boxin},
  year = {2026},
  booktitle = {Proceedings of the IEEE/CVF Conference on Computer Vision and Pattern Recognition},
  url = {https://openaccess.thecvf.com/content/CVPR2026/html/Cai_Lighting-grounded_Video_Generation_with_Renderer-based_Agent_Reasoning_CVPR_2026_paper.html},
  pages = {20055--20065},
}

@inproceedings{mpresenter2026,
  title = {{mPresenter: An Agentic Framework for Generating Multilingual Presentation Videos from Scientific Papers}},
  author = {Wenhan Han and Xiao Xiao and Mykola Pechenizkiy and Meng Fang},
  year = {2026},
  booktitle = {Findings of the Association for Computational Linguistics: ACL 2026},
  url = {https://aclanthology.org/2026.findings-acl.805/},
  doi = {10.18653/v1/2026.findings-acl.805},
  pages = {16358--16371},
}

@inproceedings{theoremexplainagent2025,
  title = {{TheoremExplainAgent: Towards Video-based Multimodal Explanations for LLM Theorem Understanding}},
  author = {Max Ku and Cheuk Hei Chong and Jonathan Leung and Krish Shah and Alvin Yu and Wenhu Chen},
  year = {2025},
  booktitle = {Proceedings of the 63rd Annual Meeting of the Association for Computational Linguistics (Volume 1: Long Papers)},
  url = {https://aclanthology.org/2025.acl-long.332/},
  doi = {10.18653/v1/2025.acl-long.332},
  pages = {6663--6684},
}

@misc{cutsceneagent2026,
  title = {{Cutscene Agent: An LLM Agent Framework for Automated 3D Cutscene Generation}},
  author = {He, Lanshan and Pang, Haozhou and Gan, Qi and Shen, Xin and Zhang, Ziwei and Liu, Yibo and Fang, Gang and Liu, Bo and Sheng, Kai and Zeng, Shengfeng and Li, Chaofan and Hui, Zhen and Zhou, Keer and Zhou, Lan and Dai, Shujun},
  year = {2026},
  howpublished = {Preprint},
  url = {https://arxiv.org/abs/2604.25318},
  eprint = {2604.25318},
  archivePrefix = {arXiv},
}

@misc{gestagent2026,
  title = {{Authoring for Living Worlds: Tool-Constrained LLM Agents for Executable Multi-Actor Scenarios}},
  author = {Cudlenco, Nicolae and Masala, Mihai and Leordeanu, Marius},
  year = {2026},
  howpublished = {Preprint},
  url = {https://arxiv.org/abs/2604.10383},
  eprint = {2604.10383},
  archivePrefix = {arXiv},
}

@inproceedings{mavfr2026,
  title = {{Evolutionary Multi-Agent Collaboration for Real-World Video Face Restoration}},
  author = {Tang, Bowen and Wang, Tao and Zhang, Miao and Yu, Xin and Chen, Jinwei and Li, Bo and Zhang, Kaihao},
  year = {2026},
  booktitle = {Proceedings of the IEEE/CVF Conference on Computer Vision and Pattern Recognition Findings},
  url = {https://openaccess.thecvf.com/content/CVPR2026F/html/Tang_Evolutionary_Multi-Agent_Collaboration_for_Real-World_Video_Face_Restoration_CVPRF_2026_paper.html},
  pages = {8175--8184},
}

@misc{aurora2026,
  title = {{Aurora: Unified Video Editing with a Tool-Using Agent}},
  author = {Yu, Yongsheng and Zeng, Ziyun and Xiao, Zhiyuan and Zhou, Zhenghong and Hua, Hang and Xiong, Wei and Luo, Jiebo},
  year = {2026},
  howpublished = {Preprint},
  url = {https://arxiv.org/abs/2605.18748},
  eprint = {2605.18748},
  archivePrefix = {arXiv},
}

@misc{crayotter2026,
  title = {Crayotter: Traceable Multi-Agent Workflows for Long-Form Video Editing},
  author = {Yan, Lecheng and Zhang, Yichong and Xu, Xiantao and Lin, Jianze and Pan, Ben and Zheng, Xiaoyu and Qian, Jiawei and Wu, Anqi and Geng, Jiahui and Li, Ruizhe and Cai, Fengyu and Niu, Jingcheng and Li, Raymond and Li, Wenxi and Lyu, Chenyang},
  year = {2026},
  howpublished = {Preprint},
  url = {https://arxiv.org/abs/2606.07636},
  eprint = {2606.07636},
  archivePrefix = {arXiv},
}

@misc{direct2026,
  title = {{DIRECT: Video Mashup Creation via Hierarchical Multi-Agent Planning and Intent-Guided Editing}},
  author = {Li, Ke and Li, Maoliang and Chen, Jialiang and Chen, Jiayu and Zheng, Zihao and Wang, Shaoqi and Chen, Xiang},
  year = {2026},
  howpublished = {Preprint},
  url = {https://arxiv.org/abs/2604.04875},
  eprint = {2604.04875},
  archivePrefix = {arXiv},
}

@misc{mmlveagent2026,
  title = {{Thinking on Shots: Consistent Multi-Shot Video Editing with Agentic Reasoning}},
  author = {Wu, Chenyang and Long, Fuchen and Huang, Binyuan and Sun, Xinlong and Chen, Xi and Guo, Chun-Le and Li, Chongyi},
  year = {2026},
  howpublished = {Preprint},
  url = {https://arxiv.org/abs/2608.26809},
  eprint = {2608.26809},
  archivePrefix = {arXiv},
}

@misc{zhang2025vcagent,
  title = {{VC-Agent: An Interactive Agent for Customized Video Dataset Collection}},
  author = {Zhang, Yidan and Xu, Mutian and Hao, Yiming and Zhou, Kun and Chang, Jiahao and Liu,
    Xiaoqiang and Wan, Pengfei and Fu, Hongbo and Han, Xiaoguang},
  year = {2025},
  howpublished = {Preprint},
  eprint = {2509.21291},
  archivePrefix = {arXiv},
  url = {https://arxiv.org/abs/2509.21291}
}

@misc{zhang2026vidus2,
  title = {{Vidu S2: Real-Time Interactive, Editable, and Spatial Video Generation}},
  author = {Zhang, Jintao and Jiang, Kai and Chen, Jintao and Wang, Xu and Liu, Deyuan and Li,
    Jungang and Chen, Dechuang and Lin, Ming and Zhou, Jingjiang and Jin, Haopeng and Jia, Qi and
    Wang, Xiaohang and Wang, Yaole and Zhang, Zhanqiang and Li, Ran and Huang, Zhengkun and Xiong,
    Shuyue and Wang, Yuji and Dai, Zikun and He, Hui and Luo, Yang and Ning, Mang and Feng, Weiqi
    and Ye, Chengyang and Lin, Xinyue and Zhao, Min and Zhu, Hongzhou and Tan, Hengkai and Wang,
    Zeyuan and Xiang, Chendong and Zheng, Kaiwen and Deng, Zhijie and Bao, Fan and Chen, Jianfei and
    Zhu, Jun},
  year = {2026},
  howpublished = {Preprint},
  eprint = {2609.11638},
  archivePrefix = {arXiv},
  url = {https://arxiv.org/abs/2609.11638}
}

@inproceedings{wang2025vrthinker,
  title = {{VR-Thinker: Boosting Video Reward Models through Think-with-Image Reasoning}},
  author = {Wang, Qunzhong and Liu, Jie and Liang, Jiajun and Zhang, Yuanxing and Jiang, Yilei and Zheng, Yaozhi and Wang, Xintao and Wan, Pengfei and Yue, Xiangyu and Liu, Jiaheng},
  year = {2026},
  eprint = {2510.10518},
  archivePrefix = {arXiv},
  url = {https://proceedings.mlr.press/v306/wang26ep.html},
  booktitle = {Proceedings of the 43rd International Conference on Machine Learning},
  series = {Proceedings of Machine Learning Research},
  volume = {306},
  pages = {128327--128350}
}

@misc{skillpe2026,
  title = {{SkillPE: Creativity-Oriented Cinematic Skill Evolution for Text-to-Video Prompt Engineering}},
  author = {Huang, Yanwei and Zhu, Mingxuan and Li, Shujie and Liu, Shiyuan and Zhang, Yuanxing and Narechania, Arpit},
  year = {2026},
  eprint = {2609.34335},
  archivePrefix = {arXiv},
  howpublished = {Preprint},
  url = {https://arxiv.org/abs/2609.34335}
}

@misc{multimediaagent2026,
  title = {{A Versatile Multimodal Agent for Multimedia Content Generation}},
  author = {Zhang, Daoan and Yao, Wenlin and Wang, Xiaoyang and Hu, Yebowen and Luo, Jiebo and Yu, Dong},
  year = {2026},
  eprint = {2601.03250},
  archivePrefix = {arXiv},
  howpublished = {Preprint},
  url = {https://arxiv.org/abs/2601.03250}
}

@article{spagent2024,
  title = {{SPAgent: Adaptive Task Decomposition and Model Selection for General Video Generation and Editing}},
  author = {Tu, Rong-Cheng and Sun, Wenhao and Jin, Zhao and Liao, Jingyi and Huang, Jiaxing and Tao, Dacheng},
  year = {2026},
  eprint = {2411.18983},
  archivePrefix = {arXiv},
  url = {https://arxiv.org/abs/2411.18983},
  journal = {IEEE Transactions on Image Processing},
  volume = {35},
  pages = {3085--3098},
  doi = {10.1109/TIP.2026.3673949}
}

@misc{cogenpvg2026,
  title = {{CogenPVG: Cognitive-Enhanced Reflective Multi-Agent Framework for Persuasive Video Generation}},
  author = {Xiao, Yuntian and Zhang, Shoulong and Song, Wenfeng and Wang, Yan and Chen, Yi and Li, Shuai},
  year = {2026},
  eprint = {2609.25821},
  archivePrefix = {arXiv},
  howpublished = {Preprint},
  url = {https://arxiv.org/abs/2609.25821}
}

@misc{phyprompt2026,
  title = {{PhyPrompt: RL-based Prompt Refinement for Physically Plausible Text-to-Video Generation}},
  author = {Wu, Shang and Xu, Chenwei and Xia, Zhuofan and Li, Weijian and Lu, Lie and Maneriker, Pranav and Du, Fan and Li, Manling and Liu, Han},
  year = {2026},
  eprint = {2603.03505},
  archivePrefix = {arXiv},
  howpublished = {Preprint},
  url = {https://arxiv.org/abs/2603.03505}
}

@inproceedings{dppgrpo2025,
  title = {{Diverse Video Generation with Determinantal Point Process-Guided Policy Optimization}},
  author = {Kazimi, Tahira and Dunlop, Connor and Yanardag, Pinar},
  year = {2026},
  eprint = {2511.20647},
  archivePrefix = {arXiv},
  url = {https://arxiv.org/abs/2511.20647},
  booktitle = {Proceedings of the IEEE/CVF Conference on Computer Vision and Pattern Recognition (CVPR)},
  pages = {12839--12848}
}

@inproceedings{automvrealestate2025,
  title = {{AutoMV: An Autonomous Agent Framework for Real Estate Marketing Video Generation}},
  author = {Kuizong Wu and Shaozu Yuan and Chang Shen and Long Xu and Meng Chen},
  year = {2025},
  booktitle = {AAAI 2025: Demonstration Track},
  url = {https://ojs.aaai.org/index.php/AAAI/article/view/35377},
  doi = {10.1609/aaai.v39i28.35377},
  pages = {29715--29717},
}

@inproceedings{animdirector2024,
  title = {{Anim-Director: A Large Multimodal Model Powered Agent for Controllable Animation Video Generation}},
  author = {Li, Yunxin and Shi, Haoyuan and Hu, Baotian and Wang, Longyue and Zhu, Jiashun and Xu, Jinyi and Zhao, Zhen and Zhang, Min},
  year = {2024},
  booktitle = {SIGGRAPH Asia 2024: Conference Papers},
  url = {https://doi.org/10.1145/3680528.3687688},
  doi = {10.1145/3680528.3687688},
}

@inproceedings{jeon2026atvg,
  title = {{ATVG: Agentic System for Factually Grounded Travel Advertisement Video Generation}},
  author = {Byung Eun Jeon and Xiao Bai and Wen Zhang and Jinchao Li},
  year = {2026},
  booktitle = {Proceedings of the 49th International ACM SIGIR Conference on Research and Development in Information Retrieval},
  pages = {5166--5170},
  doi = {10.1145/3805712.3808387},
  url = {https://doi.org/10.1145/3805712.3808387}
}

@inproceedings{li2025agentdropout,
  title = {{AgentDropout: Dynamic Agent Elimination for Token-Efficient and High-Performance LLM-Based Multi-Agent Collaboration}},
  author = {Zhexuan Wang and Yutong Wang and Xuebo Liu and Liang Ding and Miao Zhang and Jie Liu and Min Zhang},
  year = {2025},
  booktitle = {Proceedings of the 63rd Annual Meeting of the Association for Computational Linguistics (Volume 1: Long Papers)},
  doi = {10.18653/v1/2025.acl-long.1170},
  pages = {24013-24035},
  url = {https://doi.org/10.18653/v1/2025.acl-long.1170}
}

@inproceedings{yue2025masrouter,
  title = {{MasRouter: Learning to Route LLMs for Multi-Agent Systems}},
  author = {Yanwei Yue and Guibin Zhang and Boyang Liu and Guancheng Wan and Kun Wang and Dawei Cheng and Yiyan Qi},
  year = {2025},
  booktitle = {Proceedings of the 63rd Annual Meeting of the Association for Computational Linguistics (Volume 1: Long Papers)},
  doi = {10.18653/v1/2025.acl-long.757},
  pages = {15549-15572},
  url = {https://doi.org/10.18653/v1/2025.acl-long.757}
}

@article{li2024moreagents,
  title = {{More Agents Is All You Need}},
  author = {Li, Junyou and Zhang, Qin and Yu, Yangbin and Fu, Qiang and Ye, Deheng},
  year = {2024},
  journal = {TMLR},
  url = {https://arxiv.org/abs/2402.05120}
}

@inproceedings{chen2025phantomdata,
  title = {{Phantom-Data}: Towards a General Subject-Consistent Video Generation Dataset},
  author = {Chen, Zhuowei and Li, Bingchuan and Ma, Tianxiang and Liu, Lijie and Liu, Mingcong and Jiang, Yunsheng and Li, Li and Li, Xinghui and Chen, Liyang and Zhou, Siyu and He, Qian and Wu, Xinglong},
  year = {2026},
  eprint = {2506.18851},
  archivePrefix = {arXiv},
  url = {https://proceedings.iclr.cc/paper_files/paper/2026/hash/76c6f9f2475b275b92d03a83ea270af4-Abstract-Conference.html},
  booktitle = {International Conference on Learning Representations}
}

@inproceedings{yang2024cogvideox,
  title = {{CogVideoX}: Text-to-Video Diffusion Models with An Expert Transformer},
  author = {Yang, Zhuoyi and Teng, Jiayan and Zheng, Wendi and Ding, Ming and Huang, Shiyu and Xu, Jiazheng and Yang, Yuanming and Hong, Wenyi and Zhang, Xiaohan and Feng, Guanyu and Yin, Da and Zhang, Yuxuan and Wang, Weihan and Cheng, Yean and Xu, Bin and Gu, Xiaotao and Dong, Yuxiao and Tang, Jie},
  year = {2025},
  eprint = {2408.06072},
  archivePrefix = {arXiv},
  url = {https://arxiv.org/abs/2408.06072},
  booktitle = {International Conference on Learning Representations}
}

@inproceedings{chen2024videocrafter2,
  title = {{VideoCrafter2}: Overcoming Data Limitations for High-Quality Video Diffusion Models},
  author = {Chen, Haoxin and Zhang, Yong and Cun, Xiaodong and Xia, Menghan and Wang, Xintao and Weng, Chao and Shan, Ying},
  year = {2024},
  eprint = {2401.09047},
  archivePrefix = {arXiv},
  url = {https://arxiv.org/abs/2401.09047},
  booktitle = {Proceedings of the IEEE/CVF Conference on Computer Vision and Pattern Recognition (CVPR)},
  pages = {7310--7320},
  doi = {10.1109/CVPR52733.2024.00698}
}

@inproceedings{zhao2023motiondirector,
  title = {{MotionDirector}: Motion Customization of Text-to-Video Diffusion Models},
  author = {Zhao, Rui and Gu, Yuchao and Wu, Jay Zhangjie and Zhang, David Junhao and Liu, Jiawei and Wu, Weijia and Keppo, Jussi and Shou, Mike Zheng},
  year = {2024},
  eprint = {2310.08465},
  archivePrefix = {arXiv},
  url = {https://arxiv.org/abs/2310.08465},
  booktitle = {Computer Vision -- ECCV 2024},
  series = {Lecture Notes in Computer Science},
  pages = {273--290},
  publisher = {Springer},
  doi = {10.1007/978-3-031-72992-8_16}
}

@misc{xue2025dancegrpo,
  title = {{DanceGRPO}: Unleashing {GRPO} on Visual Generation},
  author = {Xue, Zeyue and Wu, Jie and Gao, Yu and Kong, Fangyuan and Zhu, Lingting and Chen, Mengzhao and Liu, Zhiheng and Liu, Wei and Guo, Qiushan and Huang, Weilin and Luo, Ping},
  year = {2025},
  eprint = {2505.07818},
  archivePrefix = {arXiv},
  url = {https://arxiv.org/abs/2505.07818}
}

@inproceedings{yin2025causvid,
  title = {From Slow Bidirectional to Fast Autoregressive Video Diffusion Models},
  author = {Yin, Tianwei and Zhang, Qiang and Zhang, Richard and Freeman, William T. and Durand, Fredo and Shechtman, Eli and Huang, Xun},
  year = {2025},
  eprint = {2412.07772},
  archivePrefix = {arXiv},
  url = {https://arxiv.org/abs/2412.07772},
  booktitle = {Proceedings of the IEEE/CVF Conference on Computer Vision and Pattern Recognition (CVPR)},
  pages = {22963--22974},
  doi = {10.1109/CVPR52734.2025.02138}
}

@article{wu2025rewarddance,
  title={Rewarddance: Reward scaling in visual generation},
  author={Wu, Jie and Gao, Yu and Ye, Zilyu and Li, Ming and Li, Liang and Guo, Hanzhong and Liu, Jie and Xue, Zeyue and Hou, Xiaoxia and Liu, Wei and others},
  journal={arXiv preprint arXiv:2509.08826},
  year={2025}
}

@misc{deng2026agenticvideounderstanding,
  title = {Agentic Video Understanding: A Survey},
  author = {Deng, Xinyu and Luo, Siwen and Liu, Daochang},
  year = {2026},
  eprint = {2609.31713},
  archivePrefix = {arXiv},
  primaryClass = {cs.CV},
  url = {https://arxiv.org/abs/2609.31713v1}
}

@misc{wu2026rsisurvey,
  title = {Recursive Self-Improvement in {AI}: A Survey},
  author = {Wu, Hao and Xu, Fan and Zhang, Fan and Xu, Zhipeng and Zhang, Yongheng and Li, Yinghui and Zhao, Yizhou and Zhao, Penghao and Yao, Weiran and Li, Zengxiang and Gao, Yuan and Gao, Zhangyang and Wang, Hanbin and Tao, Ruihan and Wang, Wenjie and Li, Qi and Fang, Junfeng and Ma, Xingjun and Li, Qiankun and Luo, Xiao and Zheng, Yefeng and Liu, Zhiwei and Wang, Stephen and Wang, Kun and Liu, Yang and Jiang, Yugang and Wang, Liang and Liu, Nan and Yu, Philip S. and Xiong, Caiming and Yang, Qiang and Wen, Qingsong},
  year = {2026},
  howpublished = {Preprints.org},
  url = {https://www.preprints.org/manuscript/202609.2680/v1},
  note = {Version 1, posted 30 September 2026}
}

@misc{gu2026anyflow,
  title = {{AnyFlow}: Any-Step Video Diffusion Model with On-Policy Flow Map Distillation},
  author = {Gu, Yuchao and Fang, Guian and Jiang, Yuxin and Mao, Weijia and Han, Song and Cai, Han and Shou, Mike Zheng},
  year = {2026},
  eprint = {2605.13724},
  archivePrefix = {arXiv},
  primaryClass = {cs.CV},
  url = {https://arxiv.org/abs/2605.13724}
}

@misc{shaul2026pdd,
  title = {Parallel Decoding Distillation for Fast Image and Video Generation},
  author = {Shaul, Neta and Liu, Chao and Vahdat, Arash and Berner, Julius},
  year = {2026},
  eprint = {2607.26004},
  archivePrefix = {arXiv},
  primaryClass = {cs.CV},
  url = {https://arxiv.org/abs/2607.26004}
}

@misc{luo2026videorsi,
  title = {{Video-RSI}: Recursive Self-Improvement of Video Understanding Agents via Harness Evolution},
  author = {Luo, Bingjun and Guo, Jialin and Li, Siqi},
  year = {2026},
  eprint = {2609.37950},
  archivePrefix = {arXiv},
  primaryClass = {cs.AI},
  url = {https://arxiv.org/abs/2609.37950v1}
}

@inproceedings{singer2022make,
  title={Make-a-video: Text-to-video generation without text-video data},
  author={Singer, Uriel and Polyak, Adam and Hayes, Thomas and Yin, Xi and An, Jie and Zhang, Songyang and Hu, Qiyuan and Yang, Harry and Ashual, Oron and Gafni, Oran and others},
  year={2023},
  booktitle={International Conference on Learning Representations},
  eprint={2209.14792},
  archivePrefix={arXiv},
  url={https://arxiv.org/abs/2209.14792}
}

@article{wan2025wan,
  title={Wan: Open and advanced large-scale video generative models},
  author={Wan, Team and Wang, Ang and Ai, Baole and Wen, Bin and Mao, Chaojie and Xie, Chen-Wei and Chen, Di and Yu, Feiwu and Zhao, Haiming and Yang, Jianxiao and others},
  journal={arXiv preprint arXiv:2503.20314},
  year={2025}
}

@inproceedings{wu2023tune,
  title={Tune-a-video: One-shot tuning of image diffusion models for text-to-video generation},
  author={Wu, Jay Zhangjie and Ge, Yixiao and Wang, Xintao and Lei, Stan Weixian and Gu, Yuchao and Shi, Yufei and Hsu, Wynne and Shan, Ying and Qie, Xiaohu and Shou, Mike Zheng},
  booktitle={2023 IEEE/CVF International Conference on Computer Vision (ICCV)},
  pages={7589--7599},
  year={2023},
  organization={IEEE},
  doi={10.1109/ICCV51070.2023.00701}
}

@inproceedings{zhai2024motion,
  title = {Motion Consistency Model: Accelerating Video Diffusion with Disentangled Motion-Appearance Distillation},
  author = {Zhai, Yuanhao and Lin, Kevin and Yang, Zhengyuan and Li, Linjie and Wang, Jianfeng and Lin, Chung-Ching and Doermann, David and Yuan, Junsong and Wang, Lijuan},
  booktitle = {Advances in Neural Information Processing Systems},
  year = {2024},
  url = {https://arxiv.org/abs/2406.06890}
}

@inproceedings{nie2026transition,
  title = {Transition Matching Distillation for Fast Video Generation},
  author = {Nie, Weili and Berner, Julius and Ma, Nanye and Liu, Chao and Xie, Saining and Vahdat, Arash},
  booktitle = {Proceedings of the IEEE/CVF Conference on Computer Vision and Pattern Recognition (CVPR)},
  pages = {4645--4655},
  month = jun,
  year = {2026},
  url = {https://openaccess.thecvf.com/content/CVPR2026/html/Nie_Transition_Matching_Distillation_for_Fast_Video_Generation_CVPR_2026_paper.html}
}

@inproceedings{zhu2026causal,
  title = {{Causal Forcing}: Autoregressive Diffusion Distillation Done Right for High-Quality Real-Time Interactive Video Generation},
  author = {Zhu, Hongzhou and Zhao, Min and He, Guande and Su, Hang and Li, Chongxuan and Zhu, Jun},
  year = {2026},
  eprint = {2602.02214},
  archivePrefix = {arXiv},
  primaryClass = {cs.CV},
  url = {https://arxiv.org/abs/2602.02214},
  booktitle = {Proceedings of the 43rd International Conference on Machine Learning},
  series = {Proceedings of Machine Learning Research},
  volume = {306},
  pages = {168030--168049}
}

@article{wang2023videofactory,
  title = {Swap Attention in Spatiotemporal Diffusions for Text-to-Video Generation},
  author = {Wang, Wenjing and Yang, Huan and Tuo, Zixi and He, Huiguo and Zhu, Junchen and Fu, Jianlong and Liu, Jiaying},
  year = {2025},
  eprint = {2305.10874},
  archivePrefix = {arXiv},
  primaryClass = {cs.CV},
  url = {https://doi.org/10.1007/s11263-025-02349-y},
  journal = {International Journal of Computer Vision},
  volume = {133},
  number = {7},
  pages = {4177--4195},
  doi = {10.1007/s11263-025-02349-y}
}

@inproceedings{zhou2026physinone,
  title = {{PhysInOne}: Visual Physics Learning and Reasoning in One Suite},
  author = {Zhou, Siyuan and Wang, Hejun and Cheng, Hu and Li, Jinxi and Wang, Dongsheng and Jiang, Junwei and Jin, Yixiao and Huang, Jiayue and Mao, Shiwei and Liu, Shangjia and Yang, Yafei and Song, Hongkang and Wei, Shenxing and Zhang, Zihui and Wang, Bing and Wang, Zhihua and Zou, Chuhang and Yang, Bo},
  booktitle = {Proceedings of the IEEE/CVF Conference on Computer Vision and Pattern Recognition (CVPR)},
  month = jun,
  year = {2026},
  pages = {33131--33142},
  url = {https://openaccess.thecvf.com/content/CVPR2026/html/Zhou_PhysInOne_Visual_Physics_Learning_and_Reasoning_in_One_Suite_CVPR_2026_paper.html}
}

@inproceedings{alemohammad2024mad,
  title = {{Self-Consuming Generative Models Go MAD}},
  author = {Alemohammad, Sina and Casco-Rodriguez, Josue and Luzi, Lorenzo and Humayun, Ahmed Imtiaz and Babaei, Hossein and LeJeune, Daniel and Siahkoohi, Ali and Baraniuk, Richard G.},
  booktitle = {International Conference on Learning Representations (ICLR)},
  year = {2024},
  eprint = {2307.01850},
  archivePrefix = {arXiv},
  url = {https://arxiv.org/abs/2307.01850}
}

@misc{yang2026quovadis,
  title = {{Quo Vadis, World Modeling?}},
  author = {Yang, Yu and Yang, Xuemeng and Wen, Licheng and Kong, Lingdong and Hu, Xiaobin and Lu, Dongyue and Chow, Wei and Huang, Xiyan and Feng, Yuxiang and Liao, Yue and Mei, Jianbiao and Fu, Daocheng and Wu, Rong and Cai, Pinlong and Yi, Ran and Tai, Ying and Zhang, Jiangning and Shi, Botian and Liu, Yong and Yan, Shuicheng},
  year = {2026},
  eprint = {2608.02713},
  archivePrefix = {arXiv},
  url = {https://arxiv.org/abs/2608.02713}
}

@misc{wu2026visualnewera,
  title = {{Visual Generation in the New Era: An Evolution from Atomic Mapping to Agentic World Modeling}},
  author = {Wu, Keming and Yang, Zuhao and Zhang, Kaichen and Wang, Shizun and Zhu, Haowei and Leng, Sicong and Yang, Zhongyu and Wang, Qijie and Wang, Sudong and Wang, Ziting and Wang, Zili and Zhang, Hui and Wang, Haonan and Zhou, Hang and Pu, Yifan and Li, Xingxuan and Zhan, Fangneng and Li, Bo and Bing, Lidong and Song, Yuxin and Liu, Ziwei and Chen, Wenhu and Wang, Jingdong and Wang, Xinchao and Qi, Xiaojuan and Lu, Shijian and Wang, Bin},
  year = {2026},
  eprint = {2604.28185},
  archivePrefix = {arXiv},
  url = {https://arxiv.org/abs/2604.28185}
}

@misc{xu2026ags,
  title = {{Agentic Generative Systems: A Survey on Autonomous Multimodal Content Creation}},
  author = {Xu, Xinli and Guo, Litao and Zhang, Yehang and Bai, Haotian and Lin, Jiantao and Feng, Jiawei and Wang, Luozhou and Yang, Zhen and Chen, Man and Zhang, Zixin and Chen, Harold Haodong and Min, Chen and Chen, Ying-Cong},
  year = {2026},
  howpublished = {Preprints.org},
  doi = {10.20944/preprints202607.1997.v1},
  url = {https://doi.org/10.20944/preprints202607.1997.v1}
}

@misc{wang2026artifactcreation,
  title = {{Agentic Artifact Creation: Systems, Evaluation, Principles, and Opportunities}},
  author = {Wang, Tianfu and Hao, Zhezheng and Xia, Xilin and Liu, Lixin and Hu, Mengkang and Liu, Hongzhang and Chen, Xi and Liu, Ziyan and Lin, Xiankun and Zhang, Weijia and Yuan, Nicholas Jing and Xiong, Hui},
  year = {2026},
  eprint = {2608.28122},
  archivePrefix = {arXiv},
  url = {https://arxiv.org/abs/2608.28122}
}

@misc{guo2026harnessdesign,
  title = {{From Question Answering to Task Completion: A Survey on Agent System and Harness Design}},
  author = {Guo, Jianyuan and Hao, Zhiwei and Wang, Chengcheng and Fan, Cheng and Luo, Tingzhang and Li, Hongguang and Gao, Ying and Mei, Hefei and Peng, Jiankun and Xu, Rongjian and Dong, Minjing and Wu, Han and Zheng, Mengyu and Han, Kai and Wang, Shiqi and Xu, Chang and Wang, Yunhe},
  year = {2026},
  eprint = {2606.20683},
  archivePrefix = {arXiv},
  url = {https://arxiv.org/abs/2606.20683}
}

@misc{lu2026orbia,
  title = {{Orbia: Do Generated Videos Form a Consistent 3D World?}},
  author = {Lu, Dongyue and Fang, Guian and Lim, Hur and Wu, Yonggang and Wang, Yanrong and Chen, Hongshan and Qian, Shengju and Liao, Yue and Chow, Wei and Kong, Lingdong and Huang, Tianxin and Yang, Yu and Yang, Hongji and Huang, Junchao and Liang, Ao and Wang, Yihao and Wang, Zihan and Li, Rong and Chen, Hanlin and Wang, Xin and Ooi, Wei Tsang and Shou, Mike Zheng and Yan, Shuicheng},
  year = {2026},
  eprint = {2610.20693},
  archivePrefix = {arXiv},
  url = {https://arxiv.org/abs/2610.20693}
}

@misc{huang2026solarwm,
  title = {{SolarWM: Open Data and Scalable Training for Long-Horizon Video World Models}},
  author = {Huang, Junchao and Fang, Guian and Qian, Shengju and Kong, Xianghao and Zhao, Zhuoran and Huang, Wei and Du, Yihua and Zhang, Zixin and Cui, Justin and Gu, Yuchao and Chen, Yukang and Hu, Xinting and He, Tianyu and Shi, Shaoshuai and Tian, Zhuotao and Wang, Xin and Shou, Mike Zheng and Jiang, Li},
  year = {2026},
  eprint = {2609.02886},
  archivePrefix = {arXiv},
  url = {https://arxiv.org/abs/2609.02886}
}

@inproceedings{gao2023overoptimization,
  title = {{Scaling Laws for Reward Model Overoptimization}},
  author = {Gao, Leo and Schulman, John and Hilton, Jacob},
  booktitle = {Proceedings of the 40th International Conference on Machine Learning},
  year = {2023},
  eprint = {2210.10760},
  archivePrefix = {arXiv},
  url = {https://arxiv.org/abs/2210.10760},
  series = {Proceedings of Machine Learning Research},
  volume = {202},
  pages = {10835--10866}
}

@article{shumailov2023curse,
  title = {{The Curse of Recursion: Training on Generated Data Makes Models Forget}},
  author = {Shumailov, Ilia and Shumaylov, Zakhar and Zhao, Yiren and Gal, Yarin and Papernot, Nicolas and Anderson, Ross},
  journal = {arXiv preprint arXiv:2305.17493},
  year = {2023},
  url = {https://arxiv.org/abs/2305.17493}
}

@misc{parkerholder2025genie3,
  title = {{Genie 3: A New Frontier for World Models}},
  author = {Parker-Holder, Jack and Fruchter, Shlomi},
  year = {2025},
  howpublished = {Google DeepMind blog, August 5, 2025},
  url = {https://deepmind.google/blog/genie-3-a-new-frontier-for-world-models/}
}

@inproceedings{guo2025lct,
  title = {{Long Context Tuning for Video Generation}},
  author = {Guo, Yuwei and Yang, Ceyuan and Yang, Ziyan and Ma, Zhibei and Lin, Zhijie and Yang, Zhenheng and Lin, Dahua and Jiang, Lu},
  year = {2025},
  eprint = {2503.10589},
  archivePrefix = {arXiv},
  url = {https://arxiv.org/abs/2503.10589},
  booktitle = {Proceedings of the IEEE/CVF International Conference on Computer Vision (ICCV)},
  pages = {17281--17291},
  doi = {10.1109/ICCV51701.2025.01605}
}

@misc{yan2026grpb,
  title = {{Crayotter}: Learning Long-Horizon Video Editing Agents via Group-Relative Preference Backpropagation},
  author = {Yan, Lecheng and Lin, Jianze and Zhang, Yichong and Pan, Ben and Li, Wenxi and Lyu, Chenyang and Zhou, Liting and Gurrin, Cathal},
  year = {2026},
  eprint = {2608.02694},
  archivePrefix = {arXiv},
  url = {https://arxiv.org/abs/2608.02694}
}

@article{gao2026lingbotworld,
  title = {Advancing Open-source World Models},
  author = {{Robbyant Team} and Gao, Zelin and Wang, Qiuyu and Zeng, Yanhong and Zhu, Jiapeng and Cheng, Ka Leong and Li, Yixuan and Wang, Hanlin and Xu, Yinghao and Ma, Shuailei and Chen, Yihang and Liu, Jie and Cheng, Yansong and Yao, Yao and Zhu, Jiayi and Meng, Yihao and Zheng, Kecheng and Bai, Qingyan and Chen, Jingye and Shen, Zehong and Yu, Yue and Zhu, Xing and Shen, Yujun and Ouyang, Hao},
  journal = {arXiv preprint arXiv:2601.20540},
  year = {2026},
  url = {https://arxiv.org/abs/2601.20540}
}

@article{gao2026lingbotworld2,
  title = {Infinite Worlds with Versatile Interactions},
  author = {Gao, Zelin and Wang, Qiuyu and Zhu, Jiapeng and Chen, Jingye and Liu, Zichen and Bai, Qingyan and Wang, Jiahao and Yuan, Yufeng and Wang, Hanlin and Lu, Yichong and Cheng, Ka Leong and Zhang, Haojie and Gao, Jian and Feng, Tianrui and Liu, Yuzheng and Yao, Yao and Xu, Yinghao and Zhu, Xing and Shen, Yujun and Ouyang, Hao},
  journal = {arXiv preprint arXiv:2607.07534},
  year = {2026},
  url = {https://arxiv.org/abs/2607.07534}
}

@article{jiang2026scope,
  title = {{SCOPE}: Score-Isolated Agentic Optimization for Video World Models},
  author = {Jiang, Yuhua and Wang, Jiaming and Liu, Qingbin and Gao, Feifei},
  journal = {arXiv preprint arXiv:2608.15043},
  year = {2026},
  url = {https://arxiv.org/abs/2608.15043}
}

@misc{weng2023agents,
  author = {Weng, Lilian},
  title = {{LLM} Powered Autonomous Agents},
  year = {2023},
  howpublished = {Lil'Log},
  url = {https://lilianweng.github.io/posts/2023-06-23-agent/}
}

@misc{weng2026harness,
  author = {Weng, Lilian},
  title = {Harness Engineering for Self-Improvement},
  year = {2026},
  howpublished = {Lil'Log},
  url = {https://lilianweng.github.io/posts/2026-07-04-harness/}
}

@misc{dubois2025llmagents,
  author = {Dubois, Yann},
  title = {Introduction to Training {LLMs} for {AI} Agents},
  year = {2025},
  howpublished = {LLM Agents MOOC, lecture slides},
  note = {September 15, 2025},
  url = {https://rdi.berkeley.edu/agentic-ai/slides/lecture1.pdf}
}

@misc{lu2024aiscientist,
  author = {Lu, Chris and Lu, Cong and Lange, Robert Tjarko and Foerster, Jakob and Clune, Jeff and Ha, David},
  title = {The {AI} Scientist: Towards Fully Automated Open-Ended Scientific Discovery},
  year = {2024},
  eprint = {2408.06292},
  archivePrefix = {arXiv},
  primaryClass = {cs.AI},
  url = {https://arxiv.org/abs/2408.06292}
}

@misc{karpathy2026autoresearch,
  author = {Karpathy, Andrej},
  title = {autoresearch},
  year = {2026},
  howpublished = {GitHub repository},
  note = {Accessed October 12, 2026},
  url = {https://github.com/karpathy/autoresearch}
}

@article{shang2026worldarena,
  author = {Shang, Yu and Li, Zhuohang and Ma, Yiding and Su, Weikang and others},
  title = {{WorldArena}: A Unified Benchmark for Evaluating Perception and Functional Utility of Embodied World Models},
  journal = {arXiv preprint arXiv:2602.08971},
  year = {2026},
  url = {https://arxiv.org/abs/2602.08971}
}

@inproceedings{duan2025worldscore,
  author = {Duan, Haoyi and Yu, Hong-Xing and Chen, Sirui and Fei-Fei, Li and Wu, Jiajun},
  title = {{WorldScore}: A Unified Evaluation Benchmark for World Generation},
  booktitle = {Proceedings of the IEEE/CVF International Conference on Computer Vision},
  year = {2025},
  url = {https://openaccess.thecvf.com/content/ICCV2025/papers/Duan_WorldScore_A_Unified_Evaluation_Benchmark_for_World_Generation_ICCV_2025_paper.pdf},
  pages = {27713--27724},
  doi = {10.1109/ICCV51701.2025.02573}
}

@misc{zhan2026magpie,
  title = {{Magpie: Real-Time World Renderer for Interactive Games}},
  author = {Zhan, Xiaoyu and Wang, Xinyu and Zhang, Xiaohong and Zhu, Huanjie and Sun, Tengjiao and Fang, Pengcheng and Yu, Jiaxing and Guo, Yanwen and Fu, Dongjie},
  year = {2026},
  eprint = {2608.27168},
  archivePrefix = {arXiv},
  url = {https://arxiv.org/abs/2608.27168}
}

@misc{gomeznogales2026coarsetoreal,
  title = {{Coarse-to-Real: Generative Rendering for Populated Dynamic Scenes}},
  author = {Gomez-Nogales, Gonzalo and Hong, Yicong and Ge, Chongjian and Zhuang, Peiye and Comino-Trinidad, Marc and Casas, Dan and Zhou, Yi},
  year = {2026},
  eprint = {2601.22301},
  archivePrefix = {arXiv},
  url = {https://arxiv.org/abs/2601.22301}
}

@misc{nvidia2026streamrender,
  year = {2026},
  title = {{StreamRender-H3: Interactive Coding Worlds with Real-time Streaming Diffusion Render}},
  author = {Lu, Yanzuo and Ye, Tian and Li, Yitong and Xue, Shuchen and Liu, Haozhe and Han, Song and Xie, Enze},
  howpublished = {\url{https://nvlabs.github.io/Sana/Sol-Engine/StreamRender-H3/}},
  note = {Project page}
}

@misc{chi2026agentgarten,
  title = {{AgentGarten: Code Worlds for Evolving Agents}},
  author = {Chi, Jiawei and Miao, Shangchen and Shi, Zhiyuan and Wu, Kailu and Wang, Hanyang and Chen, Weiliang and Dai, Qiyu and Li, Chengyang and Ren, Jinshan and Gao, Jun and Long, Mingsheng and Duan, Yueqi and Lyu, Jiangran and Wu, Jialong and Liu, Fangfu},
  year = {2026},
  eprint = {2610.12374},
  archivePrefix = {arXiv},
  url = {https://arxiv.org/abs/2610.12374}
}

@misc{huang2026programmable,
  title = {{Programmable World Model}},
  author = {Huang, Zheng-Hui and Lin, Guixu and Lin, Jiacheng and Huang, Yi-Chuan and Yu, Ruihan and Niu, Muyao and Yang, Siqi and Liu, Yu-Lun and Chuang, Yung-Yu and Zhang, Kaipeng and Wang, Zhixiang},
  year = {2026},
  eprint = {2609.10540},
  archivePrefix = {arXiv},
  url = {https://arxiv.org/abs/2609.10540}
}

@misc{chen2026cwm,
  title = {{Code World Model: Coding Agent as World Brain}},
  author = {Chen, Yiwen and Lin, Guosheng and Zhang, Chi},
  year = {2026},
  eprint = {2608.25927},
  archivePrefix = {arXiv},
  url = {https://arxiv.org/abs/2608.25927}
}

@misc{liao2026streamingcwm,
  title = {{Streaming Code World Model}},
  author = {Liao, Tingting and Liu, Guangyi},
  year = {2026},
  howpublished = {GitHub repository},
  url = {https://github.com/TingtingLiao/streaming-code-world-model}
}

@misc{zeng2026rgbx,
  title = {{RGBX-Next: Towards Realistic Generative Rendering from G-Buffers}},
  author = {Zeng, Zheng and Salvi, Marco and Wu, Lifan and Novák, Jan and Lin, Daqi and Hadadan, Saeed and Sheng, Yichen and Pottorff, Robert and Liu, Shiqiu and Ramamoorthi, Ravi and Yan, Ling-Qi and Hašan, Miloš},
  year = {2026},
  eprint = {2608.13929},
  archivePrefix = {arXiv},
  url = {https://arxiv.org/abs/2608.13929}
}

@misc{alaya2026renderer,
  title = {{Generative World Renderer at the Speed of Play}},
  author = {Lin, Guixu and Huang, Zheng-Hui and Yang, Siqi and Yang, Ming-Hsuan and Zhang, Kaipeng and Wang, Zhixiang},
  year = {2026},
  eprint = {2607.18703},
  archivePrefix = {arXiv},
  url = {https://arxiv.org/abs/2607.18703}
}

@misc{huang2026worldrenderer,
  title = {{Generative World Renderer}},
  author = {Huang, Zheng-Hui and Wang, Zhixiang and Tan, Jiaming and Yu, Ruihan and Zhang, Yidan and Zheng, Bo and Liu, Yu-Lun and Chuang, Yung-Yu and Zhang, Kaipeng},
  year = {2026},
  eprint = {2604.02329},
  archivePrefix = {arXiv},
  url = {https://arxiv.org/abs/2604.02329}
}

@inproceedings{liang2025diffusionrenderer,
  title = {{Diffusion Renderer: Neural Inverse and Forward Rendering with Video Diffusion Models}},
  author = {Liang, Ruofan and Gojcic, Zan and Ling, Huan and Munkberg, Jacob and Hasselgren, Jon and Lin, Chih-Hao and Gao, Jun and Keller, Alexander and Vijaykumar, Nandita and Fidler, Sanja and Wang, Zian},
  booktitle = {Proceedings of the IEEE/CVF Conference on Computer Vision and Pattern Recognition (CVPR)},
  year = {2025},
  eprint = {2501.18590},
  archivePrefix = {arXiv},
  url = {https://arxiv.org/abs/2501.18590},
  doi = {10.1109/CVPR52734.2025.02428},
  pages = {26069--26080}
}

@misc{xu2026proxy2world,
  title = {{Proxy2World: Learning to Generate Worlds From Lightweight Proxies without Seeing Them}},
  author = {Xu, Hongli and Yan, Weilong and Wang, Anbang and Zou, Chunyu and Hong, Siyu and Huang, Jingwei},
  year = {2026},
  eprint = {2609.35023},
  archivePrefix = {arXiv},
  url = {https://arxiv.org/abs/2609.35023}
}

@misc{liang2026codevideomodel,
  title = {{Code Video Model}},
  author = {Liang, Zhiyang and Zhang, Jiawei and Qian, Guocheng and Chen, Dongdong and Wan, Ziyu},
  year = {2026},
  howpublished = {Technical report, \url{https://code-video-model.github.io/tech_report.pdf}},
  doi = {10.13140/RG.2.2.34014.80961},
  url = {https://code-video-model.github.io/}
}

@misc{vast2026edenlive,
  title = {{EdenLive}: Real-Time Generative Rendering from Any Visual Reference},
  author = {Huang, Zehuan and Xu, Tian-Xing and Peng, Chunli and Zhang, Cheng and Zheng, Zeyi and Song, Yifei and Gu, Zekai and Yu, Zhipeng and Wang, Peng and Cao, Yan-Pei},
  year = {2026},
  howpublished = {Technical report, VAST},
  url = {https://assets.vast-eden.ai/assets/edenlive/EdenLive-tech-report.pdf}
}

@misc{fang2026solarrenderer,
  title = {{SolarRenderer: Few-Step Generative Reference Rendering}},
  author = {Fang, Guian and Qian, Shengju and Lu, Dongyue and Huang, Junchao and Jiang, Lutao and Wang, Xin and Song, Yiren and Jiang, Li and Shou, Mike Zheng},
  year = {2026},
  eprint = {2610.28630},
  archivePrefix = {arXiv},
  url = {https://arxiv.org/abs/2610.28630}
}

@misc{zhang2026matrixgame4,
  title = {{Matrix-Game 4.0: Harnessing Video Models for Real-Time Playable Game Worlds}},
  author = {Zhang, Jihai and Qian, Runjia and Kong, Kaiqin and Jiang, Biao and Wang, Puyi and Chen, Siru and Li, Tianle and Chen, Guanjie and Wu, Junyi and Chen, Chubin and Li, Yuanhang and Huang, Kaichen and Zhang, Haobo and Li, Yangguang and Yu, Wei and Chen, Guibin and Cheng, Yu and Zhou, Yahui},
  year = {2026},
  eprint = {2610.27999},
  archivePrefix = {arXiv},
  url = {https://arxiv.org/abs/2610.27999}
}

@misc{hadsell2026genie4,
  title = {{Genie 4}: A new generation of world simulation},
  author = {Hadsell, Raia and Rivas, Diego and Qi, Hang},
  year = {2026},
  howpublished = {Google DeepMind blog, October 27, 2026},
  url = {https://deepmind.google/blog/genie-4-a-new-generation-of-world-simulation/}
}

@misc{jiang2026olafworld,
  title = {{Olaf-World}: Orienting Latent Actions for Video World Modeling},
  author = {Jiang, Yuxin and Gu, Yuchao and Tsang, Ivor W. and Shou, Mike Zheng},
  year = {2026},
  eprint = {2602.10104},
  archivePrefix = {arXiv},
  url = {https://arxiv.org/abs/2602.10104}
}

@misc{wu2025movieagent,
  title = {{Automated Movie Generation via Multi-Agent CoT Planning}},
  author = {Wu, Weijia and Zhu, Zeyu and Shou, Mike Zheng},
  year = {2025},
  howpublished = {Preprint},
  url = {https://arxiv.org/abs/2503.07314},
  eprint = {2503.07314},
  archivePrefix = {arXiv},
}

@misc{filmworld2026,
  title = {{FilmWorld: Agentic Novel-to-Film Generation through Dynamic Cinematic World Modeling}},
  author = {Zuo, Jialong and Zuo, Haotong and Zhang, Shiwei and Wang, Xiang and Li, Chen and Sang, Nong and Gao, Changxin and Bai, Xiang},
  year = {2026},
  eprint = {2607.19038},
  archivePrefix = {arXiv},
  primaryClass = {cs.CV},
  note = {arXiv preprint},
  url = {https://arxiv.org/abs/2607.19038},
}

@inproceedings{chen2026cinecrew,
  title = {{Better Call CineCrew: Consistent Ultra-Long Narrative-to-Film Generation}},
  author = {Chen, Jiaben and Dong, Sixun and Zhou, Qinhong and Ma, Raine and Dou, Zhiyang and Matusik, Wojciech and Gan, Chuang},
  year = {2026},
  booktitle = {ECCV},
  url = {https://arxiv.org/abs/2609.07720},
  eprint = {2609.07720},
  archivePrefix = {arXiv},
}

@inproceedings{vista2025,
  title = {{VISTA: A Test-Time Self-Improving Video Generation Agent}},
  author = {Long, Do Xuan and Wan, Xingchen and Nakhost, Hootan and Lee, Chen-Yu and Pfister, Tomas and Arik, Sercan Ö.},
  year = {2026},
  booktitle = {Proceedings of the IEEE/CVF Conference on Computer Vision and Pattern Recognition},
  url = {https://openaccess.thecvf.com/content/CVPR2026/html/Long_VISTA_A_Test-Time_Self-Improving_Video_Generation_Agent_CVPR_2026_paper.html},
  pages = {6021--6032},
}

@misc{hu2026cutverse,
  title = {{CutVerse: A Compositional GUI Agents Benchmark for Media Post-Production Editing}},
  author = {Hu, Haobo and Guo, Xiangwu and Chen, Zhiheng and Gao, Difei and Liu, Haotian and Jin, Libiao and Mao, Qi},
  year = {2026},
  eprint = {2605.19484},
  archivePrefix = {arXiv},
  primaryClass = {cs.CV},
  note = {arXiv preprint},
  url = {https://arxiv.org/abs/2605.19484},
}

@inproceedings{yao2023react,
  title = {{ReAct: Synergizing Reasoning and Acting in Language Models}},
  author = {Yao, Shunyu and Zhao, Jeffrey and Yu, Dian and Du, Nan and Shafran, Izhak and Narasimhan, Karthik and Cao, Yuan},
  year = {2023},
  booktitle = {ICLR},
  url = {https://arxiv.org/abs/2210.03629},
  eprint = {2210.03629},
  archivePrefix = {arXiv},
}

@inproceedings{park2023generativeagents,
  title = {{Generative Agents: Interactive Simulacra of Human Behavior}},
  author = {Park, Joon Sung and O'Brien, Joseph C. and Cai, Carrie J. and Morris, Meredith Ringel and Liang, Percy and Bernstein, Michael S.},
  year = {2023},
  booktitle = {Proceedings of the 36th Annual ACM Symposium on User Interface Software and Technology},
  doi = {10.1145/3586183.3606763},
  pages = {1-22},
  url = {https://arxiv.org/abs/2304.03442},
  eprint = {2304.03442},
  archivePrefix = {arXiv},
}

@inproceedings{zhou2024lats,
  title = {{Language Agent Tree Search Unifies Reasoning, Acting, and Planning in Language Models}},
  author = {Zhou, Andy and Yan, Kai and Shlapentokh-Rothman, Michal and Wang, Haohan and Wang, Yu-Xiong},
  year = {2024},
  booktitle = {ICML},
  url = {https://arxiv.org/abs/2310.04406},
  eprint = {2310.04406},
  archivePrefix = {arXiv},
}

@article{wang2024voyager,
  title = {{Voyager: An Open-Ended Embodied Agent with Large Language Models}},
  author = {Wang, Guanzhi and Xie, Yuqi and Jiang, Yunfan and Mandlekar, Ajay and Xiao, Chaowei and Zhu, Yuke and Fan, Linxi and Anandkumar, Anima},
  year = {2024},
  journal = {TMLR},
  url = {https://arxiv.org/abs/2305.16291},
  eprint = {2305.16291},
  archivePrefix = {arXiv},
}

@inproceedings{zhao2024expel,
  title = {{ExpeL: LLM Agents Are Experiential Learners}},
  author = {Zhao, Andrew and Huang, Daniel and Xu, Quentin and Lin, Matthieu and Liu, Yong-Jin and Huang, Gao},
  year = {2024},
  booktitle = {Proceedings of the AAAI Conference on Artificial Intelligence},
  volume = {38},
  number = {17},
  pages = {19632-19642},
  doi = {10.1609/aaai.v38i17.29936},
  url = {https://doi.org/10.1609/aaai.v38i17.29936},
}

@inproceedings{wang2025awm,
  title = {{Agent Workflow Memory}},
  author = {Wang, Zora Zhiruo and Mao, Jiayuan and Fried, Daniel and Neubig, Graham},
  year = {2025},
  booktitle = {Proceedings of the 42nd International Conference on Machine Learning},
  series = {Proceedings of Machine Learning Research},
  volume = {267},
  pages = {63897-63911},
  publisher = {PMLR},
  url = {https://proceedings.mlr.press/v267/wang25bx.html},
}

@misc{aesr2026,
  title = {{Identity-Preserving Text-to-Video Generation via Agentic Enhancement and Semantic Repair}},
  author = {Gao, Jiayi and Hua, Changcheng and Tang, Jiaqi and Peng, Yuxin and Liu, Yang},
  year = {2026},
  eprint = {2608.20749},
  archivePrefix = {arXiv},
  primaryClass = {cs.CV},
  note = {arXiv preprint; ACM MM 2026 Challenge system},
  url = {https://arxiv.org/abs/2608.20749}
}

@misc{cineforge2026,
  title = {{CineForge: Self-Improving Agents for Long-Horizon Video Generation}},
  author = {Liu, Junxiang and Wang, Lin and Shi, Haiyu and Ma, Hongxu and Yang, Xiaoyu and Chen, Chunjie and Xu, Xiaoxiao and Zhan, Kaiqiao and Wang, Boao and Shi, Shuizhou and Zhu, Tianyun and Li, Jie and Li, Jiangtong},
  year = {2026},
  eprint = {2608.29621},
  archivePrefix = {arXiv},
  primaryClass = {cs.AI},
  note = {arXiv preprint},
  url = {https://arxiv.org/abs/2608.29621}
}

@inproceedings{mavis2025,
  title = {{MAViS: A Multi-Agent Framework for Long-Sequence Video Storytelling}},
  author = {Qian Wang and Ziqi Huang and Ruoxi Jia and Paul Debevec and Ning Yu},
  year = {2026},
  booktitle = {Proceedings of the 19th Conference of the European Chapter of the Association for Computational Linguistics (Volume 1: Long Papers)},
  url = {https://aclanthology.org/2026.eacl-long.101/},
  doi = {10.18653/v1/2026.eacl-long.101},
  pages = {2273--2295},
}

@inproceedings{cineagi2026,
  title = {{CineAGI: Character-Consistent Movie Creation through LLM-Orchestrated Multi-Modal Generation and Cross-Scene Integration}},
  author = {Xie, Tianyidan and Huang, Zhentao and Wang, Mingjie and Huang, Xin and Zhou, Jun and Gong, Minglun and Yi, Zili},
  year = {2026},
  booktitle = {ICME},
  url = {https://arxiv.org/abs/2604.23579},
  eprint = {2604.23579},
  archivePrefix = {arXiv},
}

@misc{paper2video2025,
  title = {{Paper2Video: Automatic Video Generation from Scientific Papers}},
  author = {Zhu, Zeyu and Lin, Kevin Qinghong and Shou, Mike Zheng},
  year = {2025},
  eprint = {2510.05096},
  archivePrefix = {arXiv},
  primaryClass = {cs.CV},
  note = {arXiv preprint},
  url = {https://arxiv.org/abs/2510.05096},
}

@inproceedings{preacher2025,
  title = {{Preacher: Paper-to-Video Agentic System}},
  author = {Jingwei Liu and Ling Yang and Hao Luo and Fan Wang and Hongyan Li and Mengdi Wang},
  year = {2025},
  booktitle = {2025 IEEE/CVF International Conference on Computer Vision (ICCV)},
  doi = {10.1109/iccv51701.2025.01591},
  pages = {17129-17139},
  url = {https://arxiv.org/abs/2508.09632},
  eprint = {2508.09632},
  archivePrefix = {arXiv},
}

@inproceedings{presentagent2025,
  title = {{PresentAgent: Multimodal Agent for Presentation Video Generation}},
  author = {Jingwei Shi and Zeyu Zhang and Biao Wu and Yanjie Liang and Meng Fang and Ling Chen and Yang Zhao},
  year = {2025},
  booktitle = {Proceedings of the 2025 Conference on Empirical Methods in Natural Language Processing: System Demonstrations},
  doi = {10.18653/v1/2025.emnlp-demos.58},
  pages = {760-773},
  url = {https://doi.org/10.18653/v1/2025.emnlp-demos.58},
}

@misc{automv2025,
  title = {{AutoMV: An Automatic Multi-Agent System for Music Video Generation}},
  author = {Tang, Xiaoxuan and Lei, Xinping and Zhu, Chaoran and Chen, Shiyun and Yuan, Ruibin and Li, Yizhi and Oh, Changjae and Zhang, Ge and Huang, Wenhao and Benetos, Emmanouil and Liu, Yang and Liu, Jiaheng and Ma, Yinghao},
  year = {2025},
  howpublished = {Preprint},
  url = {https://arxiv.org/abs/2512.12196},
  eprint = {2512.12196},
  archivePrefix = {arXiv},
}

@misc{agentmv2026,
  title = {{AgentMV: A State-Guided Multi-Agent Framework for Budget-Aware Music Video Generation}},
  author = {Wang, Huimin and Xia, Chang and Ouyang, Leilei and Kang, Yongqi and Fu, Yu and Ouyang, Yuqi},
  year = {2026},
  howpublished = {Preprint},
  url = {https://arxiv.org/abs/2605.10723},
  eprint = {2605.10723},
  archivePrefix = {arXiv},
  note = {Authors report ECCV 2026 AI4VA Workshop; publication record not verified},
}

@misc{mvweaver2026,
  title = {{MVWeaver: A Hierarchical Music Video Generation Agent with a Learned Song-to-Visual Bridge}},
  author = {Li, Sifei and Luo, Minyan and Li, Xu and Qi, Guodong and Wang, Xincan and Wang, Hanwen and Zhang, Chen and Wan, Pengfei and Deussen, Oliver and Dong, Weiming},
  year = {2026},
  eprint = {2609.06478},
  archivePrefix = {arXiv},
  primaryClass = {cs.MM},
  note = {arXiv preprint},
  url = {https://arxiv.org/abs/2609.06478},
}

@misc{zhang2026interacvid,
  title = {{InteracVid: Building a Real Interactive Audio-Visual Response Dataset from Live-Chat Videos}},
  author = {Chi Zhang and Haoyang Shi and Yueyi Liu and Zhaokun Yan and Yishu Yin and Yuhang Wu and Miao Liu},
  year = {2026},
  eprint = {2608.01157},
  archivePrefix = {arXiv},
  primaryClass = {cs.CV},
  note = {arXiv preprint},
  url = {https://arxiv.org/abs/2608.01157},
}

@misc{peng2025svbench,
  title = {{SVBench: Evaluation of Video Generation Models on Social Reasoning}},
  author = {Wenshuo Peng and Gongxuan Wang and Tianmeng Yang and Chuanhao Li and Xiaojie Xu and Hui He and Kaipeng Zhang},
  year = {2025},
  eprint = {2512.21507},
  archivePrefix = {arXiv},
  primaryClass = {cs.CV},
  note = {arXiv preprint},
  url = {https://arxiv.org/abs/2512.21507},
}

@inproceedings{chief2026,
  title = {{Bridging Creative Intent and Visual Quality: Creator-Driven Recurrent Video Generation with Agentic Feedback Loops}},
  author = {Savytski, Denis and Lei, Aiden and Liu, Heding and Yang, Warren and Liang, Sihan and Liu, Alexander and Zhao, Zhe},
  year = {2026},
  booktitle = {ICML Workshop on Human-AI Co-Creativity},
  url = {https://arxiv.org/abs/2606.18591},
  eprint = {2606.18591},
  archivePrefix = {arXiv},
}

@article{zhang2026vqinsight,
  title = {{VQ-Insight: Teaching VLMs for AI-Generated Video Quality Understanding via Progressive Visual Reinforcement Learning}},
  author = {Xuanyu Zhang and Weiqi Li and Shijie Zhao and Junlin Li and Li Zhang and Jian Zhang},
  year = {2026},
  journal = {Proceedings of the AAAI Conference on Artificial Intelligence},
  volume = {40},
  number = {15},
  pages = {12870--12878},
  doi = {10.1609/aaai.v40i15.38285},
  url = {https://doi.org/10.1609/aaai.v40i15.38285},
}

@misc{chen2026directorbench,
  title = {{DirectorBench: Diagnosing Long-Form Video Generation with Personalized Multi-Agent Evaluation}},
  author = {Jiamin Chen and Qianben Chen and Jiawen Zhang and Yidi Wu and Yuchen Li and Xiaokun Zhang and Wangchunshu Zhou and Chen Ma},
  year = {2026},
  eprint = {2605.30090},
  archivePrefix = {arXiv},
  primaryClass = {cs.CL},
  note = {arXiv preprint},
  url = {https://arxiv.org/abs/2605.30090},
}

@misc{groundshot2026,
  title = {{GroundShot: Visually Consistent Multi-Shot Long Video Generation via Entity-Grounded Shot Scheduling}},
  author = {Lai, Yixuan and Shao, Tianjia and Dou, Weijia and Zhu, Siyu and Wang, Jingdong},
  year = {2026},
  howpublished = {Preprint},
  url = {https://arxiv.org/abs/2606.20799},
  eprint = {2606.20799},
  archivePrefix = {arXiv},
}

@misc{li2026videogenagent,
  title = {{VideoGen-Agent: Reinforcing Video Generation Agents}},
  author = {Li, Binxu and Duan, Haoyi and Zhang, Yuhui and Zhang, Yaohui and Lin, Zihao and Feng, Kaituo and Huang, Suozhi and Li, Xiangyi and Li, Yu and Li, Chunyuan and Liu, Shilong and Wang, Mengdi},
  year = {2026},
  howpublished = {Preprint},
  url = {https://arxiv.org/abs/2609.24997},
  eprint = {2609.24997},
  archivePrefix = {arXiv},
}

@misc{cai2026liver,
  title = {{Lighting-grounded Video Generation with Renderer-based Agent Reasoning}},
  author = {Cai, Ziqi and Yang, Taoyu and Chang, Zheng and Li, Si and Jiang, Han and Weng, Shuchen and Shi, Boxin},
  year = {2026},
  eprint = {2604.07966},
  archivePrefix = {arXiv},
  url = {https://arxiv.org/abs/2604.07966},
}

@misc{kong2026mvagent,
  title = {{MVAgent: Multi-Agent Video Generation via Consistent Condition Construction and Shot-Level Policy Optimization}},
  author = {Kong, Xiangyu and Zhou, Wenjie and Tian, Fengping and Fang, Lihua and Sun, Haoqin and Lyu, Chenyang and Wang, Longyue and Luo, Weihua},
  year = {2026},
  howpublished = {Preprint},
  url = {https://arxiv.org/abs/2609.30609},
  eprint = {2609.30609},
  archivePrefix = {arXiv},
}

@misc{meng2026omnidrive,
  title = {{OmniDrive: An LLM-Choreographed Multi-Agent World Model with Unified Latent Co-Compression for Multi-View Driving Video Generation}},
  author = {Meng, Zijie and Liu, Yufei and Ma, Chengqian and Li, Zhiyu and Liu, Jiyuan and Nie, Wenhua and Wei, Bingcai and Chen, Shuqin and Xu, Weichen and Yuan, Jiquan and Zhang, Miao},
  year = {2026},
  eprint = {2606.17536},
  archivePrefix = {arXiv},
  url = {https://arxiv.org/abs/2606.17536},
}

@misc{chen2026viper,
  title = {{VIPER: Visual In-Context Physics Reasoning for Physically Plausible Video Generation}},
  author = {Chen, Tianxiao and Chen, Hanmo and Chen, Huajin and Li, Bo and Ye, Qi and Jiang, Peng-Tao},
  year = {2026},
  eprint = {2607.23472},
  archivePrefix = {arXiv},
  url = {https://arxiv.org/abs/2607.23472},
}

@misc{bian2025videoasprompt,
  title = {{Video-As-Prompt: Unified Semantic Control for Video Generation}},
  author = {Bian, Yuxuan and Chen, Xin and Li, Zenan and Zhi, Tiancheng and Sang, Shen and Luo, Linjie and Xu, Qiang},
  year = {2025},
  eprint = {2510.20888},
  archivePrefix = {arXiv},
  url = {https://arxiv.org/abs/2510.20888},
}

@misc{dao2022flashattention,
  title = {{FlashAttention: Fast and Memory-Efficient Exact Attention with IO-Awareness}},
  author = {Dao, Tri and Fu, Daniel Y. and Ermon, Stefano and Rudra, Atri and R\'e, Christopher},
  year = {2022},
  eprint = {2205.14135},
  archivePrefix = {arXiv},
  url = {https://arxiv.org/abs/2205.14135},
}

@misc{liu2023lostmiddle,
  title = {{Lost in the Middle: How Language Models Use Long Contexts}},
  author = {Liu, Nelson F. and Lin, Kevin and Hewitt, John and Paranjape, Ashwin and Bevilacqua, Michele and Petroni, Fabio and Liang, Percy},
  year = {2023},
  eprint = {2307.03172},
  archivePrefix = {arXiv},
  url = {https://arxiv.org/abs/2307.03172},
}

@inproceedings{xu2023freebloom,
  title = {{Free-Bloom: Zero-Shot Text-to-Video Generator with LLM Director and LDM Animator}},
  author = {Huang, Hanzhuo and Feng, Yufan and Shi, Cheng and Xu, Lan and Yu, Jingyi and Yang, Sibei},
  year = {2023},
  booktitle = {NeurIPS},
  url = {https://arxiv.org/abs/2309.14494},
  eprint = {2309.14494},
  archivePrefix = {arXiv},
}

@inproceedings{lian2024llmgrounded,
  title = {{LLM-grounded Video Diffusion Models}},
  author = {Lian, Long and Shi, Baifeng and Yala, Adam and Darrell, Trevor and Li, Boyi},
  year = {2024},
  booktitle = {ICLR},
  url = {https://arxiv.org/abs/2309.17444},
  eprint = {2309.17444},
  archivePrefix = {arXiv},
}

@inproceedings{lin2024videodirectorgpt,
  title = {{VideoDirectorGPT: Consistent Multi-scene Video Generation via LLM-Guided Planning}},
  author = {Lin, Han and Zala, Abhay and Cho, Jaemin and Bansal, Mohit},
  year = {2024},
  booktitle = {COLM},
  url = {https://arxiv.org/abs/2309.15091},
  eprint = {2309.15091},
  archivePrefix = {arXiv},
}

@inproceedings{long2024videostudio,
  title = {{VideoStudio: Generating Consistent-Content and Multi-Scene Videos}},
  author = {Long, Fuchen and Qiu, Zhaofan and Yao, Ting and Mei, Tao},
  year = {2024},
  booktitle = {ECCV},
  url = {https://arxiv.org/abs/2401.01256},
  eprint = {2401.01256},
  archivePrefix = {arXiv},
}

@inproceedings{motionagent2025,
  title = {{MotionAgent: Fine-grained Controllable Video Generation via Motion Field Agent}},
  author = {Liao, Xinyao and Zeng, Xianfang and Wang, Liao and Yu, Gang and Lin, Guosheng and Zhang, Chi},
  year = {2025},
  booktitle = {Proceedings of the IEEE/CVF International Conference on Computer Vision},
  url = {https://openaccess.thecvf.com/content/ICCV2025/html/Liao_MotionAgent_Fine-grained_Controllable_Video_Generation_via_Motion_Field_Agent_ICCV_2025_paper.html},
  pages = {11305--11316},
}

@misc{hu2024storyagent,
  title = {{StoryAgent: Customized Storytelling Video Generation via Multi-Agent Collaboration}},
  author = {Hu, Panwen and Jiang, Jin and Chen, Jianqi and Han, Mingfei and Liao, Shengcai and Chang, Xiaojun and Liang, Xiaodan},
  year = {2024},
  eprint = {2411.04925},
  archivePrefix = {arXiv},
  primaryClass = {cs.AI},
  note = {arXiv preprint},
  url = {https://arxiv.org/abs/2411.04925},
}

@inproceedings{campilot2026,
  title = {{CamPilot: A Multi-Agent Cinematic Assistant for Camera-Controlled Movie Generation}},
  author = {Wu, Yang and Petrangeli, Stefano and Dasgupta, Ishita and Shen, Yu},
  year = {2026},
  booktitle = {EMNLP REALM Workshop},
  url = {https://arxiv.org/abs/2609.10943},
  eprint = {2609.10943},
  archivePrefix = {arXiv},
}

@misc{zhou2026dramadirector,
  title = {{DramaDirector: Geometry-Guided Short Drama Generation}},
  author = {Zhou, Hengji and Liu, Sijie and Chen, Jianrun and Zou, Xingchen and Xia, Lianghao and Nie, Liqiang},
  year = {2026},
  eprint = {2606.24107},
  archivePrefix = {arXiv},
  primaryClass = {cs.CV},
  note = {arXiv preprint},
  url = {https://arxiv.org/abs/2606.24107},
}

@misc{zeng2026cutcraft,
  title = {{Beyond Coherence: Benchmarking Professional Editing-Technique Execution in Multi-Shot Audio-Video Generation}},
  author = {Zeng, Tianyi and Liao, Junchao and Wei, Yujie and Zhang, Ziying and Li, Litao and Wang, Tianyi and Wei, Zhichao and Xu, Shuyao and Qiang, Wenwen and Zhu, Siyu and Zhang, Zhenghao and Qin, Long},
  year = {2026},
  eprint = {2609.08275},
  archivePrefix = {arXiv},
  primaryClass = {cs.AI},
  note = {arXiv preprint},
  url = {https://arxiv.org/abs/2609.08275}
}

@misc{yang2026spiral,
  title = {{SPIRAL: Self-Evolving Action-Conditioned Video Generation via Reflective Planning Agents}},
  author = {Yang, Yu and Liao, Yue and Mei, Jianbiao and Wang, Baisen and Yang, Xuemeng and Wen, Licheng and Zhang, Jiangning and Li, Xiangtai and Lv, Liang and Chen, Hanlin and Shi, Botian and Liu, Yong and Yan, Shuicheng and Lee, Gim Hee},
  year = {2026},
  eprint = {2603.08403},
  archivePrefix = {arXiv},
  primaryClass = {cs.CV},
  note = {arXiv preprint},
  url = {https://arxiv.org/abs/2603.08403},
}

@misc{liu2026reca,
  title = {{ReCA: Multi-Shot Long Video Extrapolation via Recursive Context Allocation}},
  author = {Liu, Akide and Xing, Jinbo and Mao, Chaojie and Li, Ye and Zhang, Zeyu and He, Yefei and Wang, Weijie and Wang, Zihan and Liu, Yu and Haffari, Gholamreza and Zhuang, Bohan},
  year = {2026},
  eprint = {2605.26525},
  archivePrefix = {arXiv},
  primaryClass = {cs.CV},
  note = {arXiv preprint},
  url = {https://arxiv.org/abs/2605.26525},
}

@misc{chen2026wnm,
  title = {{World Narrative Model for Highly Controllable Video Generation: A Paradigm Shift from Pixel Sampling to Physical World Orchestration}},
  author = {Chen, Ye and Chen, Xuanhong and Zhu, Yupeng and Tan, Liming and Wan, Zhewen and Xiong, Yuxuan and Wang, Tielong and Liu, Jinfan and Zhang, Wuze and Zhang, Xiongzhen and Li, Feifei and Luo, Xianglin and Zhao, Zhehan and Zhang, Zhifan and Kou, Laisheng and Liang, Zhujin and Chen, Yugang and Chen, Muchun and Miao, Xu and Zhang, Yijing and Sheng, Xiaojie and Hu, Qiang and Chen, Jialiang and Zhang, Weimin and Zhang, Wenjun and Ni, Bingbing},
  year = {2026},
  eprint = {2606.31946},
  archivePrefix = {arXiv},
  primaryClass = {cs.CV},
  note = {arXiv preprint},
  url = {https://arxiv.org/abs/2606.31946},
}

@misc{feng2026newton,
  title = {{NEWTON: Agentic Planning for Physically Grounded Video Generation}},
  author = {Feng, Yuxiang and Wang, Juncheng and Xu, Chao and Qian, Yijie and Wang, Huihan and Hou, Wenlong and Liu, Yang and Sun, Baigui and Liu, Yong and Wang, Shujun},
  year = {2026},
  eprint = {2605.18396},
  archivePrefix = {arXiv},
  primaryClass = {cs.CV},
  note = {arXiv preprint},
  url = {https://arxiv.org/abs/2605.18396}
}

@misc{nguyen2026physplan,
  title = {{PhysPlan: Grounded Physical State Reasoning and Graph-Guided Optimization for Physically Plausible Video Generation}},
  author = {Nguyen, Minh-Loi and Le, Xuan-Vu and Le, Trung-Nghia and Nguyen, Tam V. and Tran, Minh-Triet and Do, Thanh-Toan},
  year = {2026},
  eprint = {2609.13006},
  archivePrefix = {arXiv},
  primaryClass = {cs.CV},
  note = {arXiv preprint},
  url = {https://arxiv.org/abs/2609.13006},
}

@inproceedings{animaker2025,
  title = {{AniMaker: Multi-Agent Animated Storytelling with MCTS-Driven Clip Generation}},
  author = {Shi, Haoyuan and Li, Yunxin and Chen, Xinyu and Wang, Longyue and Hu, Baotian and Zhang, Min},
  year = {2025},
  booktitle = {SIGGRAPH Asia 2025: Conference Papers},
  url = {https://animaker-dev.github.io/},
}

@misc{codirector2026,
  title = {{Co-Director: Agentic Generative Video Storytelling}},
  author = {Song, Yale and Song, Yiwen and Losier, Nick and Hodson, Nathan and Jin, Ye and Zhu, Rhyard and Xu, Yan and Vlasic, Daniel and Claassen, Carina and Leon, Jasmine and LeViet, Khanh G. and Chomyn, Zack and Timmons, Joe and Slatkin, Brett and Penberthy, Scott and Pfister, Tomas},
  year = {2026},
  howpublished = {Preprint},
  url = {https://arxiv.org/abs/2604.24842},
  eprint = {2604.24842},
  archivePrefix = {arXiv},
  note = {Authors report acceptance at COLM 2026; conservatively counted as preprint},
}

@inproceedings{scmapr2026,
  title = {{SCMAPR: Self-Correcting Multi-Agent Prompt Refinement for Complex-Scenario Text-to-Video Generation}},
  author = {Chengyi Yang and Pengzhen Li and Jiayin Qi and Aimin Zhou and Ji Wu and Ji Liu},
  year = {2026},
  booktitle = {Findings of the Association for Computational Linguistics: ACL 2026},
  url = {https://aclanthology.org/2026.findings-acl.345/},
  doi = {10.18653/v1/2026.findings-acl.345},
  pages = {6942--6973},
}

@inproceedings{huang2026genmac,
  title = {{GenMAC: Compositional Text-to-Video Generation with Multi-Agent Collaboration}},
  author = {Kaiyi Huang and Yukun Huang and Xuefei Ning and Zinan Lin and Yu Wang and Xihui Liu},
  year = {2026},
  booktitle = {Proceedings of the AAAI Conference on Artificial Intelligence},
  doi = {10.1609/aaai.v40i7.37418},
  pages = {5049-5057},
  volume = {40},
  number = {7},
  url = {https://doi.org/10.1609/aaai.v40i7.37418}
}

@misc{soap2soap2026,
  title = {{Soap2Soap: Long Cinematic Video Remaking via Multi-Agent Collaboration}},
  author = {Song, Yiren and Zhong, Huilin and Lin, Kevin Qinghong and Wang, Haofan and Shou, Mike Zheng},
  year = {2026},
  eprint = {2605.17423},
  archivePrefix = {arXiv},
  primaryClass = {cs.AI},
  note = {arXiv preprint},
  url = {https://arxiv.org/abs/2605.17423},
}

@misc{beyondtrial2026,
  title = {{Beyond Trial-and-Error: Agentic Optimization for Image-to-Video Adherence}},
  author = {Tyagi, Aman and Boinpally, Hemanth and Chen, Jonathan and Gebert, Douglas and Hickson, Steven},
  year = {2026},
  eprint = {2608.12290},
  archivePrefix = {arXiv},
  primaryClass = {cs.AI},
  note = {arXiv preprint},
  url = {https://arxiv.org/abs/2608.12290}
}
